\documentclass[11pt]{book}

\usepackage{amsmath}
\usepackage{amsthm}
\usepackage{endnotes}
\usepackage{amssymb}
\usepackage{url}
\usepackage{hyperref}
\usepackage{algorithm}
\usepackage{xspace}
\usepackage{color}
\usepackage{mathtools}
\usepackage{enumerate}
\usepackage{algorithmicx}
\usepackage{algpseudocode}
\usepackage{natbib}
\usepackage{fancyhdr}

\newtheorem{theorem}{Theorem}[chapter]
\newtheorem{definition}[theorem]{Definition}
\newtheorem{example}[theorem]{Example}
\newtheorem{proposition}[theorem]{Proposition}
\newtheorem{corollary}[theorem]{Corollary}
\newtheorem{lemma}[theorem]{Lemma}

\let\footnote=\endnote

\title{ Introduction to Online Convex Optimization \\ {\small Second Edition} } 
\date{}
\author{Elad Hazan}

\DeclareMathOperator*{\argmin}{\arg\min}
\DeclareMathOperator*{\argmax}{\arg\max}

\def\C{{\mathcal C}}
\def\W{{\mathcal W}}

\def\X{{\mathcal X}}
\def\H{{\mathcal H}}
\def\Y{{\mathcal Y}}
\def\D{{\mathcal D}}
\def\L{{\mathcal L}}
\def\F{{\mathcal F}}
\def\mP{{\mathcal P}}
\def\reals{{\mathbb R}}
\def\R{{\mathcal R}}
\def\fhat{{\ensuremath{\hat{f}}}}

\newcommand{\proj}{\mathop{\Pi}}

\newcommand{\err}{\mathop{\mbox{\rm error}}}
\newcommand{\rank}{\mathop{\mbox{\rm rank}}}
\newcommand{\ellipsoid}{{\mathcal E}}

\newcommand{\sign}{\mathop{\mbox{\rm sign}}}
\newcommand{\poly}{\mathop{\mbox{\rm poly}}}

\newcommand{\ignore}[1]{}
\newcommand{\equaldef}{\stackrel{\text{\tiny def}}{=}}
\newcommand{\equaltri}{\equaldef}

\def\dist{{\bf Dist}}

\def\reals{{\mathbb R}}
\newcommand{\E}{\mathop{\mbox{\bf E}}}
\newcommand\ball{\mathbb{B}}
\newcommand\var{\mbox{Var}}
\newcommand\cone{\mbox{cone}}
\def\bzero{\mathbf{0}}

\def\mP{{\mathcal P}}

\def\mA{{\mathcal A}}

\def\mO{{\mathcal O}}

\def\lhat{\hat{\ell}}

\def\bold0{\mathbf{0}}

\newcommand\mycases[4] {{
\left\{
\begin{array}{ll}
    {#1}, & {#2} \\\\
    {#3}, & {#4}
\end{array}
\right. }}
\newcommand\mythreecases[6] {{
\left\{
\begin{array}{ll}
    {#1}, & {#2} \\\\
    {#3}, & {#4} \\\\
    {#5}, & {#6}
\end{array}
\right. }}

\def\be{\mathbf{e}}

\def\bx{\mathbf{x}}
\def\w{\mathbf{w}}
\def\by{\mathbf{y}}

\def\bp{\mathbf{p}}

\def\br{\mathbf{r}}

\def\bv{\mathbf{v}}

\def\ba{\mathbf{a}}
\def\bA{\mathbf{A}}

\def\bI{\mathbf{I}}

\def\bone{\mathbf{1}}

\def\ch{\mathbf{CH}}

\def\xhat{\hat{\mathbf{x}}}
\def\xbar{\bar{\mathbf{x}}}
\def\vol{\mbox{vol}}
\def\trace{{\bf Tr}}

\newcommand{\eps}{\varepsilon}

\def\bone{\mathbf{1}}
\newcommand{\sphere}{\ensuremath{\mathbb {S}}}

\newcommand{\diag}{{\bf diag}}
\newcommand{\K}{\ensuremath{\mathcal K}}

\def\mA{{\mathcal A}}
\newcommand{\x}{\ensuremath{\mathbf x}}
\newcommand{\vn}{\ensuremath{\mathbf n}}
\newcommand{\rv}[1][t]{\ensuremath{\mathbf r_{#1}}}
\newcommand{\y}{\ensuremath{\mathbf y}}
\newcommand{\z}{\ensuremath{\mathbf z}}
\newcommand{\h}{\ensuremath{\mathbf h}}
\newcommand{\xtil}[1][t]{\ensuremath{\mathbf {\tilde{x}}_{#1}}}
\newcommand{\xv}[1][t]{\ensuremath{\mathbf x_{#1}}}

\newcommand{\yv}[1][t]{\ensuremath{\mathbf y_{#1}}}

\newcommand{\fv}[1][t]{\ensuremath{\mathbf f_{#1}}}
\newcommand{\gv}[1][t]{\ensuremath{\mathbf g_{#1}}}

\newcommand{\ev}[1][i]{\ensuremath{\mathbf e_{#1}}}
\newcommand{\uv}{\ensuremath{\mathbf u}}
\newcommand{\vv}{\ensuremath{\mathbf v}}
\newcommand{\A}[1][t]{\ensuremath{\mathbf A_{#1}}}

\newcommand{\ftil}{\tilde{f}}

\def\regret{\ensuremath{\mathrm{{Regret}}}}
\def\dregreteqn{\ensuremath{\mathrm{{DynamicRegret}}}}

\def\newregreteqn{\ensuremath{\mathrm{{AdaptiveRegret}}}}
\def\newregret{{adaptive regret }}

\def\be{\mathbf{e}}

\def\bx{\mathbf{x}}
\def\by{\mathbf{y}}

\def\bp{\mathbf{p}}

\def\br{\mathbf{r}}
\def\bv{\mathbf{v}}
\def\ba{\mathbf{a}}
\def\bA{{A}}
\def\bI{\mathbf{I}}

\def\eps{\varepsilon}
\def\epsilon{\varepsilon}
\def\tsum{{\textstyle \sum}}

\def\ogd{{online gradient descent}\xspace}
\def\R{\ensuremath{\mathcal R}}

\newcounter{exercise}
\newcounter{subexercise}

\long\def\exer#1{\vskip12pt \stepcounter{exercise}
\setcounter{subexercise}{0}
{%\leftskip=1pt
\noindent {\bf \arabic{exercise}. } #1 %  \hfill}\ignorespaces#1
}
}

\long\def\subexer#1{\vskip12pt \stepcounter{subexercise}
{%\leftskip=27.5pt %
\noindent {\bf(\alph{subexercise}) } #1 } %\hfill}\ignorespaces#1\vskip1pt}
}

\begin{document}

\frontmatter  % title page, contents, catalog information
\maketitle

\begin{figure}[ht]
\begin{center}
\includegraphics[width=5in]{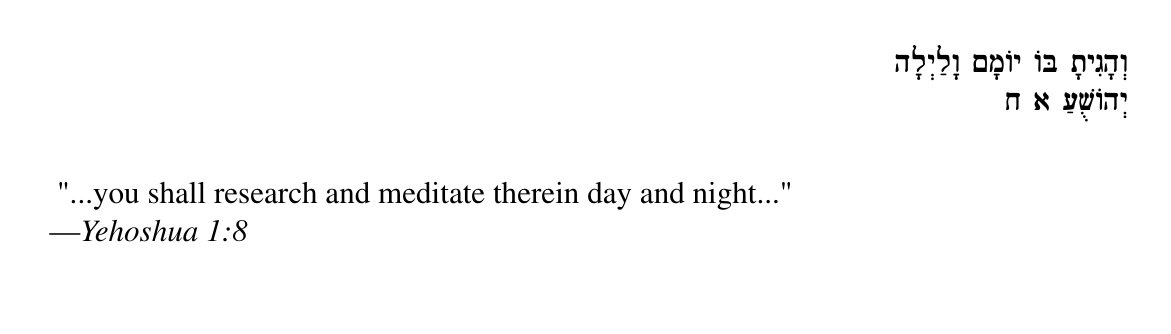}
\end{center}
\end{figure}

\newpage 
\begin{center}
To my family: Dana, Hadar, Yoav, Oded and Deluca, \\ 
---EH \\
\end{center}
\hspace*{\fill}
\vfill
\textcopyright\ [2021] Elad Hazan

\tableofcontents

%%%%%%%%%%%
% Preface %
%%%%%%%%%%%
\chapter*{Preface}
     \addcontentsline{toc}{chapter}{Preface}
     \markboth{\sffamily\slshape Preface}
       {\sffamily\slshape Preface}
%\chapter*{Preface}
%\addcontentsline{toc}{fmbm}{Preface}
%\addtocontents{toc}{\protect\enlargethispage{1\baselineskip}}

This book serves as an introduction to the expanding theory of online convex optimization (OCO). It was written as an advanced textbook to serve as a basis for a graduate course, and/or as a reference to researchers diving into this fascinating world at the intersection of optimization and machine learning. 

Such a course was given at the Technion in 2010--2014, with slight variations from year to year, and later at Princeton University in 2015--2020. The core material in these courses is fully covered in this book, along with exercises that allow students to complete parts of proofs, or that were found illuminating and thought-provoking by those  taking the course. Most of the material is given with examples of applications, which are interlaced throughout various topics. These include prediction from expert advice, portfolio selection, matrix completion and recommendation systems, and support vector machine training. 

Our hope is that this compendium of material and exercises will be useful to you; the researcher and/or educator.

\section*{Placing this Book in the Machine Learning Library}

The broad field of machine learning, as in the sub-disciplines of online learning, boosting, regret minimization in games, universal prediction, and other related topics, have seen a plethora of introductory books in recent years. With this note, we can hardly do justice to all of these, but perhaps point to the most related books on the topics of machine learning, learning in games, and optimization, whose intersection is our main focus. 

The most closely related book, which served as an inspiration to the current, and indeed an inspiration to the entire field of learning in games, is the wonderful text of \citet{CesaBianchiLugosi06book}. From the literature on mathematical optimization theory, there are the numerous introductory essays to convex optimization and convex analysis, to name only a few \citep{boyd.convex,NesterovBook,NY83,Nemirovski04lectures,borwein2006convex,rockafellar1997convex}. The author fondly recommends the text from which he has learned about mathematical optimization theory \citep{Nemirovski04lectures}.  The more broad texts on machine learning are too numerous to state here.

The primary purpose of this  is to serve as an educational textbook for a {\it dedicated} course on OCO and the convex optimization approach to machine learning.  Online convex optimization has already had enough impact to appear in several surveys and introductory texts \citep{HazanSurvey,shalev2011online,SashaLec1,SashaLec2}. We hope this compilation of material and exercises will further enrich the literature.

\section*{Book's tructure}

This book is intended to serve as a reference for a self-contained course for graduate students in computer science/electrical engineering/operations research/statistics and related fields. As such, its organization follows the structure of the course  ``Decision Analysis'',  taught at the Technion, and later ``Theoretical Machine Learning", taught at Princeton University.  

Each chapter should take one or two weeks of classes, depending on the depth and breadth of the intended course. Chapter \ref{chap:intro} is designed to be a teaser for the field, and it is less rigorous than the rest of the book. 

Roughly speaking, the book can be conceived as three units. The first, from chapter \ref{chap:opt} through \ref{chap:second order-methods}, contains the basic definitions, framework and core algorithms for OCO.   
Chapters \ref{chap:regularization} to  \ref{chap:FW} contain more advanced algorithms and in-depth analysis of the framework and its extensions to other computational and information access models. 
The rest of the book deals with more advanced algorithms, more difficult settings, and relationships to well-known machine learning paradigms. 

This book can assist  educators in  designing a complete course on the topic of online convex optimization, or it can serve as a component in a comprehensive course on machine learning. A accompanying manual of solutions to selected exercises given in the book is available for educators only. 

\newpage
\subsection*{New in the Second Edition}

The main additions to the second edition of this book include the following:
\begin{itemize}
    \item Expanded coverage of optimization in chapter \ref{chap:opt}, with a unified gradient descent analysis of the Polyak stepsize.
    
    \item
    Expanded coverage of learning theory in chapter \ref{chap:online2batch}, with an introduction to compression and its use in generalization theory.
    
    \item An expanded chapter \ref{chap:second order-methods}, with addition of the exponential weighted optimizer for exp-concave loss functions. 

    \item A revised chapter \ref{chap:regularization}, with the addition of mirror descent analysis, as well as a revised section on adaptive gradient methods.

    \item
    New chapter \ref{chap:adaptive} on the notion of adaptive regret and algorithms for OCO with near-optimal adaptive regret bounds.
    
    \item 
    New chapter \ref{chap:boosting} on boosting and its relationship to OCO. Derivation of boosting algorithms from regret minimization.
    
    \item
    New chapter \ref{chap:ocoboost} on online boosting. 

    \item 
    New chapter \ref{chap:approach} on Blackwell approachability and its strong connection to OCO.

\end{itemize}

In addition, numerous typos are fixed, exercises are corrected, and solutions to several questions have been made available in a separate manual for educators.

%%%%%%%%%%%%%%%%%%%%%%%%%%%%%%%%%%%%
% Give credit where credit is due. %
% Say thanks!                      %
%%%%%%%%%%%%%%%%%%%%%%%%%%%%%%%%%%%%
\chapter*{Acknowledgements}
     \addcontentsline{toc}{chapter}{Acknowledgements}
     \markboth{\sffamily\slshape Acknowledgements}
       {\sffamily\slshape Acknowledgements}

\section*{The First Version}

First of all, I gratefully acknowledge the numerous contributions and insight of the students of the course ``decision analysis'' given at the Technion during 2010-2014, as well as the students of ``theoretical machine learning'' taught at Princeton University during 2015-2016. 

I would like to  thank the friends, colleagues and students that have contributed many suggestions and corrections. A partial list includes:  Sanjeev Arora, Shai Shalev-Shwartz, Aleksander Madry, Yoram Singer, Satyen Kale, Alon Gonen, Roi Livni, Gal Lavee, Maayan Harel, Daniel Khasabi, Shuang Liu, Jason Altschuler, Haipeng Luo, Zeyuan Allen-Zhu, Mehrdad Mahdavi, Jaehyun Park, Baris Ungun, Maria Gregori,  Tengyu Ma, Kayla McCue, Esther Rolf, Jeremy Cohen, Daniel Suo, Lydia Liu, Fermi Ma, Mert Al, Amir Reza Asadi,  Carl Gabel,  Nati Srebro, Abbas Mehrabian, Chris Liaw, Nikhil Bansal, Naman Agarwal, Raunak Kumar, Zhize Li, Sheng Zhang, Swati Gupta, Xinyi Chen, Liang Zeng and Kunal Mittal.

I thank Udi Aharoni for his artwork and illustrations depicting algorithms in this book. 

I am forever indebted to my teacher and mentor, Sanjeev Arora, without him this book would not be possible. 

Finally, I am grateful for the love and support of my wife and children: Dana, Hadar, Yoav, and Oded.

\section*{The Second Version}

I am most thankful to my students and colleagues that have collaborated with me on research, some of which appears in the second version of this book. Notably my collaborators on boosting methods including Nataly Brukhim, Xinyi Chen, Shay Moran, Naman Agarwal and Karan Singh. 

Thanks to my students that helped me proof check new and existing sections: Edgar Minasyan, Paula Gradu, Karan Singh, Nataly Brukhim, Xinyi Chen, Naman Agarwal and Udaya Ghai. 

I am thankful to Shay Moran for explaining compression schemes and how they simplify generalization for boosting. 

I gratefully acknowledges the help of Ahmed Farah, Charlie Cowen-Breen and the students of ``COS 597C: Computational Control Theory" for many helpful suggestions, corrections and solutions to many of the problem sets. 

I am very thankful to Wouter Koolen-Wijkstra for a helpful suggestion in the analysis of the Online Newton Step algorithm. 

Thanks to Shiyun Lin for finding typos and improving the presentation of the randomized regularization section. 

I thank the extremely helpful and rigorous reviewers of this book found by MIT press who gave fantastic suggestions and improve the final manuscript. 

As in the first version and even more so, I am grateful for the love and support of my wife and children: Dana, Hadar, Yoav, and Oded.

\bigskip

\noindent Elad Hazan \\
Princeton University

\listoffigures
\chapter*{List of Symbols}
\addcontentsline{toc}{chapter}{List of Symbols}

\subsection*{General}

\begingroup
\renewcommand{\arraystretch}{1.5}
\setlength{\tabcolsep}{10pt} % Default value: 6pt
\begin{tabular}{ll}
$\equaldef$ & definition \\
$\argmin\{ \}$ &  the argument minimizing the expression in braces \\
$[n]$ & the set of integers $\{1,2,\ldots,n\}$
\end{tabular}  
\endgroup

\subsection*{Geometry and Calculus}

\begingroup
\renewcommand{\arraystretch}{1.5}
\setlength{\tabcolsep}{10pt} % Default value: 6pt

\begin{tabular}{ll}
$\reals^d$ & $d$ dimensional Euclidean space \\
$\Delta_d$ & $d$ dimensional simplex, $\{ \sum_i \x_i=1, \x_i \geq 0\}$    \\
$\sphere$ & $d$ dimensional sphere, $\{ \|\x\| =1\}$  \\
$\ball$ & $d$ dimensional ball, $\{ \|\x\| \leq 1 \}$  \\
$\reals$ & real numbers \\
$\mathbb{C}$ & complex numbers \\
$|A|$ & determinant of matrix $A$ 
\end{tabular}  
\endgroup

\subsection*{Learning Theory}

\begingroup
\renewcommand{\arraystretch}{1.5}
\setlength{\tabcolsep}{10pt} % Default value: 6pt
\begin{tabular}{ll}
 $\X,\Y$  & feature/label sets \\
 $\D$  & distribution over examples $(\x,y)$ \\
 $\H$  & hypothesis class in $\X \mapsto Y$\\
 $h$  & single hypothesis  $h \in \H$ \\
 $m$  &  training set size \\
 $\err(h)$  & generalization error of hypothesis $h \in \H$ 
\end{tabular}
\endgroup

\subsection*{Optimization}

\begingroup
\renewcommand{\arraystretch}{1.5}
\setlength{\tabcolsep}{10pt} % Default value: 6pt
\begin{tabular}{ll}
$\x$ & vectors in the decision set \\
 $\K$ & decision set \\
 $\nabla^k f $ & the $k$'th differential of  $f$; note $\nabla^k f \in \reals^{d^k}$\\
 $\nabla^{-2} f $ & the inverse Hessian of  $f$ \\
 $\nabla f $ & the gradient of $f$ \\
 $\nabla_t$ & the gradient of $f$ at point $\x_t$ \\
 $\x^\star$ & the global or local optima of objective $f$ \\
 $h_t$ & objective value distance to optimality, $h_t = f(\x_t) - f(\x^\star)$ \\
 $d_t$ & Euclidean distance to optimality $d_t =  \|\x_t - \x^\star\| $ \\
 $G$ & upper bound on norm of subgradients \\
 $D$ & upper bound on Euclidean diameter \\
 $D_p,G_p$ & upper bound on the $p$-norm of the subgradients/diameter 
\end{tabular}
\endgroup

\subsection*{Regularization}

\begingroup
\renewcommand{\arraystretch}{1.5}
\setlength{\tabcolsep}{10pt} % Default value: 6pt
\begin{tabular}{ll}
$R$  & strongly convex and smooth regularization function \\
 $B_R(\x || \y)$  & $R$-Bregman-divergence  $R(\x) - R(\y) - \nabla R(\y)^\top (\x-\y)$ \\
 $G_R$  & upper bound on norm of (sub)gradients \\
 $D_R^2$  & squared $R$ diameter $\max_{\x,\y \in \K} \{ R(\x) - R(\y) \}$ \\
 $ \| \x \|_A^2 $  & squared matrix norm $\x^\top A \x$ \\
 $ \| \x \|_\y^2$  & local norm according to local regularization $\x^\top \nabla^2 R(\y) \x$ \\
 $\| \x \|^*$  & dual norm to $\| \x \| $ 
\end{tabular}
\endgroup

\mainmatter

%%%%%%%%%%%%%%%%
% NEW CHAPTER! %
%%%%%%%%%%%%%%%%
\chapter{Introduction}  \label{chap:intro}

This book considers {\it optimization as a process}.  In many practical applications, the environment is so complex that it is not feasible  to lay out a comprehensive theoretical model and use classical algorithmic theory and mathematical optimization. It is necessary, as well as beneficial, to take a robust approach, by applying an optimization method that learns as more aspects of the problem are observed. 
This view  of optimization as a process has become prominent in various fields, which has led to spectacular successes in modeling  and systems that are now part of our daily lives. 

The growing body of literature of machine learning, statistics, decision science, and mathematical optimization blurs the classical distinctions   between deterministic modeling, stochastic modeling, and optimization methodology.  We continue this trend in this book, studying a prominent optimization framework whose precise location in the mathematical sciences is unclear: the framework of {\it online convex optimization} (OCO), which was first defined in the machine learning literature (see section \ref{sec:bib-of-sec-1}, later in this chapter). The metric of success is borrowed from game theory, and the framework is closely tied to statistical learning theory and convex optimization. 

We embrace these fruitful connections and, on purpose, do not try to use any particular jargon in the discussion. Rather, this book will start with actual problems that can be modeled and solved via OCO. We will proceed to present rigorous definitions, backgrounds, and algorithms. Throughout, we provide connections to the literature in other fields. It is our hope that you, the reader, will contribute to our understanding of these connections from your domain of expertise, and expand the growing amount of literature on this fascinating subject.

\section{The Online Convex Optimization Setting}
\label{section:formaldef}

In OCO, an online player iteratively makes decisions.  At the time of each decision, the outcome or outcomes associated with it are unknown to the player. 

After committing to a decision, the decision maker suffers a loss: every possible decision incurs a (possibly different) loss. These losses are unknown to the decision maker beforehand. The losses can be adversarially chosen, and even depend on the action taken by the decision maker.

Already at this point, several restrictions are necessary in order for this framework to make any sense at all:
\begin{itemize}
\item
The losses determined by an adversary should not be allowed to be unbounded \endnote{Alternatively and equivalently, any performance metric for this setting should depend on the magnitude of the largest loss. This is the viewpoint taken later in the rest of this book. }. Otherwise, the adversary could keep decreasing the scale of the loss at each step, and never allow the algorithm to recover from the loss of the first step. Thus, we assume that the losses lie in some bounded region.

\item 
The decision set must be somehow bounded and/or structured, though not necessarily finite. 

To see why this is necessary, consider decision making with an infinite set of possible decisions. An adversary can assign high loss to all the strategies chosen by the player indefinitely, while setting apart some strategies with zero loss. This precludes any meaningful performance metric. 

\end{itemize}

Surprisingly, interesting statements and algorithms can be derived with not much more than these two restrictions. 
The online convex optimization (OCO) framework models the decision set as a convex  set in Euclidean space denoted as $\K
\subseteq \reals^n$. The costs are modeled as bounded convex functions over $\K$. 

The OCO framework can be seen as a structured repeated game.  The protocol of this learning framework is as follows. 

At iteration $t$, the online player chooses
$\x_t \in \K$ . After the player has committed to this choice, a convex cost
function $f_t \in \F : \K \mapsto \reals$ is revealed. Here, $\F$ is the bounded family of cost functions available to the adversary. The cost incurred
by the online player is $f_t(\x_t)$, the value of the cost function for the
choice  $\x_t$. Let $T$ denote the total number of game iterations.

What would make an algorithm a good OCO algorithm? As the framework is game-theoretic and adversarial in nature, the appropriate performance metric  also comes from game theory: 
define the {\it regret} of the decision maker to be the difference between the total cost she has incurred and that of the best fixed decision in hindsight. In OCO, we are usually interested in an upper bound on the worst-case regret of an algorithm. 

Let $\mA$ be an algorithm for OCO, which maps a certain game history to a decision in the decision set: 
$$ \x_t^\mA = \mA(f_1,...,f_{t-1}) \in \K . $$
We formally define the regret of $\mA$ after $T$ iterations as: 
\begin{equation} \label{eqn:regret-defn}
\regret_T(\mA) = \sup_{\{f_1,...,f_T\} \subseteq \F}  \left\{   \sum_{t=1}^T f_t(\x_t^\mA) -\min_{\x \in \K} \sum_{t=1}^T f_t(\x) \right\} .
\end{equation}

If the algorithm is clear from the context, we henceforth omit the superscript and denote the algorithm's decision at time $t$ simply as $\x_t$. 
Intuitively, an algorithm performs well if its
regret is sublinear as a function of $T$ (i.e.
$\regret_T(\mA) = o(T)$), since this implies that on 
average, the algorithm performs as well as the best fixed strategy
in hindsight.

The running time of an algorithm for OCO  is
defined to be the worst-case expected time to produce $\x_t$, for
an iteration $t \in [T]$\endnote{Here and henceforth, we denote
as $[n]$ the set of integers $\{1,...,n\}$.} in a $T$-iteration
repeated game. Typically, the running time will depend on $n$  (the dimensionality of the decision set $\mathcal{K}$), $T$ (the total number of game iterations), and the parameters of the cost functions and underlying convex set.

\section{Examples of Problems That Can Be Modeled via Online Convex Optimization } \label{subsec:OCOexamples} \sectionmark{Examples of OCO}

Perhaps the main reason that OCO has become a leading online learning framework in recent years is its powerful modeling capability: problems from diverse domains such as online routing, ad selection for search engines, and spam filtering can all be modeled as special cases. In this section, we briefly survey a few special cases and how they fit into the OCO framework. 

\subsection{Prediction from expert advice}
Perhaps the most well known problem in prediction theory is the {\it experts problem}. The decision maker has to choose among the advice of $n$ given experts. After making her choice, a loss between zero and $1$ is incurred. This scenario is repeated iteratively, and at each iteration, the costs of the various experts are arbitrary (and possibly even adversarial,  trying to mislead  the decision maker). The goal of the decision maker is to do as well as the best expert in hindsight. 

The OCO setting captures this as a special case: the set of decisions is the set of all distributions over $n$ elements (experts); that is, the $n$-dimensional simplex $\K = \Delta_n = \{ \x \in \reals^n \  , \ \sum_i \x_i = 1 \  ,  \  \x_i \geq 0\}$. Let the cost of the $i$th expert at iteration $t$ be  $\gv(i)$, and let $\gv$ be the cost vector of all $n$ experts. 
Then the cost function is the expected cost of choosing an expert according to distribution $\x$, and it is given by the linear function $f_t(\x) = \gv^\top \x$.

Thus, prediction from expert advice is a special case of OCO, in which the decision set is the simplex and the cost functions are linear and bounded, in the $\ell_\infty$ norm, to be at most $1$. The bound on the cost functions is derived from the bound on the elements of the cost vector $\gv$.  

The fundamental importance of the experts problem in machine learning warrants special attention, and we shall return to it and analyze it in detail at the end of this chapter.

\subsection{Online spam filtering}

Consider an online spam-filtering system.  Repeatedly, emails arrive in the system and are classified as spam or valid. Obviously, such a system has to cope with adversarially generated data and dynamically change with the varying input---a hallmark of the OCO model. 

The linear variant of this model is captured by representing the emails as vectors according to the ``bag-of-words'' representation. Each email is represented as a vector $\ba \in \reals^d$, where $d$ is the  number of words in the dictionary. The entries of this vector are all zero, except for those coordinates that correspond to words appearing in the email, which are assigned the value one. 

To predict whether an email is spam, we learn a filter, for example a vector $\x \in \reals^d$. Usually a bound on the Euclidean norm of this vector is decided upon a priori, and is a parameter of great importance in practice. 

Classification of an email $\ba \in \reals^d$ by a filter $\x \in \reals^d$ is given by the sign of the inner product between these two vectors, i.e., $\hat{b} = \sign (  \x^\top \ba)  $ (with, for example, $+1$ meaning valid and $-1$ meaning spam). 

In the OCO model of online spam filtering, the decision set is taken to be the set of all such norm-bounded linear filters,  i.e., the Euclidean ball of a certain radius. The cost functions are determined  according to a stream of incoming emails arriving into the system, and their labels (which may be known by the system, partially known, or not known at all). Let $(\ba,b) $ be an email/label pair. Then the corresponding cost function over filters is given by $f(\x) = \ell( \hat{b},b)$. Here $\hat{b}$ is the classification given by the filter $\x$, $b$ is the true label, and $\ell$ is a convex loss function, for example, the scaled square loss  $\ell (\hat{b},b) = \frac{1}{4}(\hat{b} - b)^2$.  

At this point the reader may wonder - why use a square loss rather than any other function? The most natural choice being perhaps a loss of one if $b = \hat{b}$ and zero otherwise. 

To answer this, notice first that if both $b$ and $\hat{b}$ are binary and in $\{-1,1\}$, then the square loss is indeed one or zero. However, moving to a continuous function allows us much more flexibility in the decision making process. We can allow, for example, the algorithm to return a number in the interval $[-1,1]$ depending on its confidence. 

Another reason has to do with the algorithmic efficiency of finding a a good solution. This will be the subject of future chapters.

\subsection{Online shortest paths}

In the online shortest path problem,
the decision maker is given a directed graph $G=(V,E)$ and a source-sink pair $u,v \in V$. At each iteration $t \in [T]$,
the decision maker chooses a path $p_t \in \mP_{u,v}$, where $\mP_{u,v} \subseteq E^{|V|} $ is the set of all $u$-$v$-paths in the graph. The adversary independently chooses weights (lengths) on the edges of the graph, given by a function from the edges to the real numbers $\w_t: E \mapsto \reals$, which can be represented as a vector $\w_t \in \reals^m$, where  $m=|E|$. The decision maker suffers and observes a loss, which is the weighted length of the chosen path $\sum_{e \in p_t} \w_t(e) $.

The discrete description of this problem as an experts problem, where we have an expert for each path, presents an efficiency challenge. There  are potentially exponentially many paths in terms of the graph representation size. 

Alternatively, the online shortest path problem can be cast in the online convex optimization framework as follows.
Recall the standard description of the set of all distributions over paths (flows) in a graph as a convex set in $\reals^{m}$, with $O(m+|V|)$ constraints (figure \ref{flow polytope}). Denote this flow polytope by $\K$. The expected cost of a given flow $\x \in \K$ (distribution over paths) is then a linear function, given by $f_t(\x) = \w_t^\top \x$, where, as defined above, $\w_t(e)$ is the length of the edge $e \in E$. This inherently succinct formulation leads to computationally efficient algorithms.

\begin{figure}[ht]
\begin{align*}
&  \sum_{ e = (u,w) , w\in V} \x_{e} = 1 = \sum_{ e = (w,v), w \in V } \x_{e}  & \mbox{ flow value is one} \\
& \forall w \in V \setminus \{u,v\}  \ \   \sum_{e = (w,x) \in E } \x_{e} = \sum_{e = (x,w) \in E } \x_{e}  & \mbox{ flow conservation} \\
& \forall e \in E   \ \  0 \leq \x_{e} \leq 1  &  \mbox{ capacity constraints} 
\end{align*}
\caption{Linear equalities and inequalities that define the flow polytope, which is the convex hull of all $u$-$v$  paths \label{flow polytope}}
\end{figure}

\subsection{Portfolio selection} \label{section:portfolios}

In this section we consider a portfolio selection model that does not make any statistical assumptions about the stock market (as opposed to the standard geometric Brownian motion model for stock prices), and is called the ``universal portfolio selection'' model. 

At each iteration $t\in[T]$,
the decision maker chooses a distribution of her wealth over $n$ assets $\xv \in \Delta_n$. The adversary independently chooses market returns for the assets, i.e., a vector $\rv \in \reals^n$ with strictly positive entries such that each coordinate $\rv(i)$ is the price ratio for the $i$'th asset between the iterations $t$ and $t+1$. The ratio between the wealth of the investor at iterations $t+1$ and $t$ is $\rv^\top \xv$, and hence the gain in this setting is defined to be the logarithm of this change ratio in wealth $\log (\rv^\top \xv)$. Notice that since $\xv$ is the distribution of the investor's wealth, even if $\xv[t+1] =\xv$, the investor may still need to trade to adjust for price changes.

The goal of regret minimization, which in this case corresponds to minimizing the difference
$ \max_{\x^\star \in \Delta_n} \tsum_{t=1}^T \log(\rv^\top \x^\star) -  \tsum_{t=1}^T \log(\rv^\top \x_t)$, has an intuitive interpretation. The first term is the logarithm of the wealth accumulated by the best possible in-hindsight distribution $\x^\star$. Since this distribution is fixed, it corresponds to a strategy of rebalancing the position after every trading period, and hence, is called a {\it constant rebalanced portfolio}. The second term is the logarithm of the wealth accumulated by the online decision maker. Hence regret minimization corresponds to maximizing the ratio of the investor's wealth to the wealth  of the best benchmark from a pool of investing strategies.

A {\it universal} portfolio selection algorithm is defined to be one that, in this setting, attains regret converging to zero. Such an algorithm, albeit requiring exponential time,  was first described by Cover (see bibliographic notes at the end of this chapter). The online convex optimization framework has given rise to much more efficient algorithms based on Newton's method. We  shall return to study these in detail in chapter \ref{chap:second order-methods}.

\subsection{Matrix completion and recommendation systems} 

The prevalence of large-scale media delivery systems such as the Netflix online video library, Spotify music service and many others, give rise to very large scale recommendation systems. One of the most popular and successful models for automated recommendation is the matrix completion model. 

In this mathematical model, recommendations are thought of as composing a matrix.  The customers are represented by the rows, the different media are the columns, and at the entry corresponding to a particular user/media pair we have a value scoring the preference of the user for that particular media. 

For example, for the case of binary recommendations for music,  we have a matrix $X \in \{0,1\}^{n \times m}$  where $n$ is the number of persons considered, $m$ is the number of songs in our library, and $0/1$ signifies dislike/like respectively:
$$ X_{ij} = \mycases {0}{\mbox{person $i$ dislikes song $j$}}{1}{\mbox{person $i$ likes song $j$}} .$$ 

In the online setting, for each iteration the decision maker outputs a preference matrix  $X_t \in \K$, where $\K \subseteq \{0,1\}^{n \times m}$ is a subset of all possible zero/one matrices. An adversary then chooses a user/song pair $(i_t,j_t)$ along with a ``real'' preference for this pair $y_t \in \{0,1\}$. Thus, the loss experienced by  the decision maker can be described by the convex loss function, 
$$ f_t(X) = ( X_{i_t,j_t} - y_t)^2 .$$ 

The natural comparator in this scenario is a low-rank matrix, which corresponds to the intuitive assumption that preference is determined by few unknown factors. Regret with respect to this comparator means performing, on the average, as few preference-prediction errors as the best low-rank matrix. 

We return to this problem and explore efficient algorithms for it in chapter \ref{chap:FW}.

\sectionmark{Learning from Expert Advice}
\section{A Gentle Start:  Learning from Expert Advice} \label{sec:experts}
\sectionmark{Learning from Expert Advice}

Consider the following fundamental iterative decision making problem:

At each time step $t=1,2,\ldots,T$, the decision maker faces a choice between two actions  $A$ or $B$ (i.e.,  buy or sell a certain stock). The decision maker has assistance in the form of  $N$  ``experts'' that offer their advice. After a choice between the two actions has been made, the decision maker receives feedback in the form of a loss associated with each decision.  For simplicity one of the actions receives a loss of zero (i.e., the ``correct'' decision) and the other a loss of one. 

We make the following elementary observations:
\begin{enumerate} 
\item
A decision maker that chooses an  action uniformly at random each iteration,  trivially attains a loss of $\frac{T}{2} $ and is ``correct''  $50\%$ of the time.
\item
In terms of the number of mistakes, no algorithm can do better in the worst case! In a later  exercise, we will  devise  a randomized setting in which the expected number of mistakes of any algorithm is at least $\frac{T}{2}$. 
\end{enumerate}

We are thus motivated to consider a {\it relative performance metric}: can the decision maker make as few mistakes as the best expert in hindsight? 
The next theorem shows that the answer in the worst case  is negative for a deterministic decision maker.
\begin{theorem}
Let $L \leq \frac{T} {2} $ denote the number of mistakes made by the best expert in hindsight. Then there does not exist a deterministic algorithm that can guarantee less than $2L$ mistakes.
\end{theorem}

\begin{proof}

Assume that there are only two experts and one always chooses option $A$ while the other always chooses option $B$.
Consider the setting in which an adversary always chooses the opposite of our prediction (she can do so, since our algorithm is deterministic).
Then, the total number of mistakes the algorithm makes is $T$.
However, the best expert makes no more than $\frac{T}{2}$ mistakes (at every iteration exactly one of the two experts is mistaken).
Therefore, there is no algorithm that can always guarantee less than $2L$ mistakes.

\end{proof}

This observation motivates the design of random decision making algorithms, and indeed, the OCO framework  gracefully models decisions on a continuous probability space. Henceforth we prove Lemmas \ref{lem:wm} and \ref{lem:rwm} that show the following: 
 
\begin{theorem}
Let $\eps \in (0,\frac{1}{2} )$. Suppose the best expert makes $L$  mistakes. Then:
\begin{enumerate}
\item
There is an efficient deterministic algorithm that can guarantee less than $2(1+\epsilon)L + \frac{2\log N}{\epsilon}$  mistakes;
\item
There is an efficient randomized algorithm for which the expected number of mistakes is at most $(1+\epsilon)L  + \frac{\log N}{\epsilon}$.
\end{enumerate}
\end{theorem}

\subsection {The weighted majority algorithm}

The weighted majority (WM) algorithm is intuitive to describe: 
each expert $i$ is assigned a weight $W_t(i)$ at every  iteration $t$.
Initially, we set $W_1(i) = 1$ for all experts $i \in [N]$.
For all $t \in [T]$ let $S_t(A),S_t(B) \subseteq [N] $ be the set of experts that choose $A$ (and respectively $B$) at time $t$.  Define,
\[
W_t(A) = \smashoperator[r]{\sum_{i \in S_t(A)}} W_t(i) \qquad  \qquad W_t(B) = \smashoperator[r]{\sum_{i \in S_t(B)}} W_t(i) 
\]
and predict according to
\begin{equation*}
a_t =
\begin{cases}
A & \text{if $W_t(A) \ge W_t(B)$}\\
B & \text{otherwise.}
\end{cases}
\end{equation*}
Next,  update the weights $W_t(i)$ as follows:
\begin{equation*}
W_{t+1}(i) =
\begin{cases}
W_t(i) & \text{if expert $i$ was correct}\\
W_t(i)  (1-\eps) & \text{if expert $i$ was wrong}
\end{cases}
,
\end{equation*}
where $\eps$ is a parameter of the algorithm that will affect its performance. This concludes the description of the WM algorithm. We proceed to bound the number of  mistakes it makes. 
\begin{lemma} \label{lem:wm}
Denote by $M_t$ the number of mistakes the algorithm makes until time $t$, and by $M_t(i)$ the number of mistakes made by expert $i$ until time $t$.
Then, for any expert $i \in [N]$ we have
\[
M_T \le 2(1+\epsilon)M_T(i) + \frac{2\log N}{\epsilon} .
\]
\end{lemma}

\noindent We can optimize $\epsilon$ to minimize the above bound.
The expression on the right hand side is of the form $f(x)=ax+b/x$, that reaches its minimum at $x=\sqrt{b/a}$.
Therefore the bound is minimized at $\epsilon^\star = \sqrt{\log N/M_T(i)}$.
Using this optimal value of $\epsilon$, we get that for the best expert $i^\star$ 
\[
M_T \le 2M_T(i^\star) + O\left(\sqrt {M_T(i^\star)\log N}\right).
\]
Of course, this value of  $\epsilon^\star$ cannot be used in advance since we do not know which expert is the best one ahead of time (and therefore we do not know the value of $M_T(i^\star)$). However, we shall see later on that the same asymptotic bound can be obtained even without this prior knowledge. 

Let us now prove Lemma \ref{lem:wm}.

\begin{proof}

Let $\Phi_t= \sum_{i=1}^N W_t(i)$ for all $t \in [T]$, and note that $\Phi_1=N$.

Notice that $\Phi_{t+1} \le \Phi_t$. However, on iterations in which the WM algorithm erred, we have 
$$\Phi_{t+1} \le \Phi_t(1-\frac{\epsilon}{2}) ,$$
the reason being that experts with at least half of total weight were wrong (else WM would not have erred), and therefore
\[
\Phi_{t+1} \le  \frac{1}{2} \Phi_t(1-\epsilon) + \frac {1} {2} \Phi_t =\Phi_t(1-\frac {\epsilon}{2}) .
\]
From both observations,
\[
\Phi_{t} \le \Phi_1 (1-\frac{\epsilon}{2})^{M_t} = N (1-\frac{\epsilon}{2})^{M_t} .
\]
On the other hand, by definition we have for any expert $i$ that
\[
W_T(i) = (1-\epsilon)^{M_T(i)} .
\]
Since the value of $W_T(i)$ is always less than the sum of all weights $\Phi_T$, we conclude that
\[
(1-\epsilon)^{M_T(i)} = W_T(i) \le \Phi_T \le N(1-\frac{\epsilon}{2})^{M_T}.
\]
Taking the logarithm of both sides we get
\[
M_T(i)\log(1-\epsilon) \le \log{N} + M_T\log{(1-\frac{\epsilon}{2})}  .
\]
Next, we use the approximations
\[
-x-x^2 \le \log{(1-x)} \le -x  \qquad  \quad 0 < x < \frac{1}{2},
\]
which follow from the Taylor series of the logarithm function, to obtain that
\[
-M_T(i)(\epsilon+\epsilon^2) \le \log{N} - M_T\frac {\epsilon}{2} ,
\]
and the lemma follows.
\end{proof}

\subsection{Randomized weighted majority}
In the randomized version of the WM algorithm, denoted RWM, we choose expert $i$ w.p. $p_t(i) = W_t(i) /  \sum_{j=1}^N W_t(j)$ at time $t$.
\begin{lemma} \label{lem:rwm}
Let $M_t$ denote the number of mistakes made by RWM until iteration $t$. Then, for any expert $i \in [N]$ we have
\[
\E[ M_T]  \le (1+\epsilon)M_T(i) + \frac{\log N}{\epsilon} .
\]
\end{lemma}
\noindent The proof of this lemma is very similar to the previous one, where the factor of two is saved by the use of randomness:
\begin{proof}

As before, let $\Phi_t= \sum_{i=1}^N W_t(i)$ for all $t \in [T]$, and note that $\Phi_1=N$. Let $\tilde{m}_t = M_t - M_{t-1}$ be the indicator variable that equals one if the RWM algorithm makes a mistake on iteration $t$. Let $m_t(i)$  equal one if the $i$'th expert makes a mistake on iteration $t$ and zero otherwise. 
Inspecting the sum of the weights: 
\begin{align*}
\Phi_{t+1} & = \sum_i W_t(i) (1 - \eps m_t(i)) \\
& = \Phi_t (1 - \epsilon  \sum_i p_t(i) m_t(i)) & \mbox{ $p_t(i) = \frac{W_t(i)}{\sum_j W_t(j) }$} \\
& = \Phi_t ( 1 - \epsilon \E[\tilde{m}_t ]) \\
& \leq \Phi_t e^{-\eps \E[\tilde{m}_t] }. & \mbox{ $1  + x \leq e^x $}  
\end{align*}
On the other hand, by definition we have for any expert $i$ that
\[
W_T(i) = (1-\epsilon)^{M_T(i)} 
\]
Since the value of $W_T(i)$ is always less than the sum of all weights $\Phi_T$, we conclude that
\[
(1-\epsilon)^{M_T(i)} = W_T(i) \le \Phi_T \le N e^{-\eps \E [M_T]}.
\]
Taking the logarithm of both sides we get
\[
M_T(i)\log(1-\epsilon) \le \log{N} - \eps \E[ M_T] 
\]
Next, we use the approximation
\[
-x-x^2 \le \log{(1-x)} \le -x \qquad , \quad 0 < x <  \frac{1}{2}
\]
to obtain
\[
-M_T(i)(\epsilon+\epsilon^2) \le \log{N} - \eps \E[M_T] ,
\]
and the lemma follows.
\end{proof}

\subsection{Hedge}

The RWM algorithm is in fact more general: instead of considering a discrete number of mistakes, we can consider measuring the performance of an expert by a non-negative real number $\ell_t(i)$, which we refer to as the {\it loss} of the expert $i$ at iteration $t$. The randomized weighted majority algorithm guarantees that a decision maker following its advice will incur an average expected loss approaching that of the best expert in hindsight. 

Historically, this was observed by a different and closely related algorithm called Hedge, whose total loss bound will be of interest to us later on in the book. 

\begin{algorithm}[ht]
	\caption{Hedge}
	\label{alg:Hedge}
	\begin{algorithmic}[1]
		\State Initialize: $\forall i\in [N], \ W_1(i) = 1$ 
		\For {$t=1$ to $T$}
		\State Pick $i_t \sim_R W_t$, i.e., $i_t = i$ with probability $\x_t(i) = \frac{W_t(i) } {\sum_j W_t(j) }$
		\State Incur loss $\ell_t(i_t)$. 
		\State Update weights $ W_{t+1}(i) = W_{t}(i) e^{-\eps \ell_t(i)}$
		\EndFor
	\end{algorithmic}
\end{algorithm}

Henceforth, denote in vector notation the expected loss of the algorithm by
$$ \E [ \ell_t(i_t) ] = \sum_{i=1}^N \x_t(i) \ell_t(i) = \x_t^\top \ell_t  $$
\begin{theorem} \label{lem:hedge}
Let $\ell_t^2$ denote the $N$-dimensional vector of square losses, i.e., $\ell_t^2(i) = \ell_t(i)^2$,  let $\eps > 0$, and assume all losses to be non-negative.  
The Hedge algorithm satisfies for any expert $i^\star \in [N]$:
\[
  \sum_{t=1}^T  \x_t^\top \ell_t   \le \sum_{t=1}^T \ell_t(i^\star) + \epsilon \sum_{t=1}^T  \x_t^\top \ell_t^2   + \frac{\log N}{\epsilon} 
\]
\end{theorem}

\begin{proof}
	
As before, let $\Phi_t= \sum_{i=1}^N W_t(i)$ for all $t \in [T]$, and note that $\Phi_1=N$. 
	
Inspecting the sum of weights: 
\begin{eqnarray*}
	\Phi_{t+1} & = \sum_i W_t(i) e^{- \eps \ell_t(i)}  \\
	& = \Phi_t  \sum_i \x_t(i) e^{- \eps \ell_t(i)}  & \mbox{ $\x_t(i) = \frac{W_t(i)}{\sum_j W_t(j) }$} \\
	& \leq \Phi_t \sum_i \x_t(i) ( 1 - \epsilon \ell_t(i) + \eps^2 \ell_t(i)^2 ) )  & \mbox{ for $x \geq 0$, }    \\
	&   & \mbox{  $e^{-x} \leq 1 - x + x^2 $}   \\
	& = \Phi_t (  1 - \epsilon \x_t^\top \ell_t  + \eps^2 \x_t^\top \ell_t^2   )  \\
		& \leq \Phi_t e^{-\eps \x_t^\top \ell_t + \epsilon^2 \x_t^\top \ell_t^2  }. & \mbox{ $1  + x \leq e^x $}  
\end{eqnarray*}
On the other hand, by definition, for  expert $i^\star$ we have that
	\[
	W_{T+1}(i^\star) = e^{ -\epsilon \sum_{t=1}^{T} \ell_t(i^\star) } 
	\]
Since the value of $W_T(i^\star)$ is always less than the sum of all weights $\Phi_t$, we conclude that
	\[
	 W_{T+1}(i^\star) \le \Phi_{T+1} \le N e^{-\eps \sum_t  \x_t^\top \ell_{t} + \epsilon^2 \sum_{t} \x_t^\top \ell_t^2 }.
	\]
	Taking the logarithm of both sides we get
	\[
	-\epsilon \sum_{t=1}^T \ell_t(i^\star)  \le \log{N} - \eps \sum_{t=1}^T  \x_t^\top \ell_t + \epsilon^2 \sum_{t=1}^T \x_t^\top \ell_t^2
	\]
	and the theorem follows by simplifying.
\end{proof}

\newpage
\section{Bibliographic Remarks} \label{sec:bib-of-sec-1}

The OCO model was first defined by \citet{Zinkevich03} and has since become widely influential in the learning community and significantly extended since (see thesis and surveys  \citep{HazanThesis,HazanSurvey,shalev2011online}).

The problem of prediction from expert advice and the Weighted Majority algorithm were devised in \citep{WarmuthLittlestone89,LitWar94}. This seminal work was one of the  first uses of the multiplicative updates method---a ubiquitous meta-algorithm in computation and learning,  see the survey \citep{AHK-MW} for more details. The Hedge algorithm was introduced by \citet{FreundSch1997}.

The Universal Portfolios model was put forth in \citep{cover}, and is one of the first examples of a worst-case online learning model. Cover gave an optimal-regret algorithm for universal portfolio selection that runs in exponential time. A polynomial time algorithm was given in \citep{KalaiVempalaPortfolios}, which was further sped up in \citep{AgarwalHKS06,HAK07}. Numerous extensions to the model also appeared in the literature, including addition of transaction costs \citep{BlumKalaiPortfolios} and relation to the Geometric Brownian Motion model for stock prices \citep{HazanKNips09}.

In their influential paper,   \citet{AweKle08} put forth the application of online convex optimization to online routing. A great deal of work has been  devoted since then to improve the initial bounds, and generalize it into a complete framework for decision making with limited feedback. This framework is an extension of OCO, called Bandit Convex Optimization (BCO). We defer further bibliographic remarks to chapter \ref{chap:bandits} which is devoted to the BCO framework.

\newpage

\begin{exercises}

%\begin{enumerate}
%	\item
\exer{ (Attributed  to Claude Shannon) \\ Construct market returns over two stocks for which the wealth accumulated over any single stock decreases exponentially, whereas the best constant rebalanced portfolio increases wealth exponentially.  More precisely, construct two sequences of numbers in the range $(0,\infty)$, that represent returns, such that:}
%	 \begin{enumerate}
%\item
\subexer{Investing in any of the individual stocks results in exponential decrease in wealth. This means that the product of the prefix of numbers in each of these sequences decreases exponentially.}
%\item
\subexer{Investing  evenly on the two assets and rebalancing after every iteration increases wealth exponentially. }
% \end{enumerate}

%	\item
%	\begin{enumerate}
%		\item

\exer{}  \subexer{
Consider the experts problem in which the losses are between zero and a positive real number $G > 0$. Give an algorithm that attains expected loss upper bounded by:
$$   \sum_{t=1}^T \E[  \ell_t (i_t) ]  \le \min_{i^\star \in [N]} \sum_{t=1}^T \ell_t(i^\star)  + c \sqrt{ T \log N }  $$
for the best constant $c$ you can (the constant $c$ should be independent of the number of game iterations $T$,  and the number of experts  $N$. Assume that $T$ is known in advance).
}

\subexer{ %\item 
Suppose the upper bound $G$ is not known in advance. Give an algorithm whose performance is asymptotically as good as your algorithm in part (a), up to an additive and/or multiplicative constant which is independent of $T,N,G$. Prove your claim.
}
%\end{enumerate}

%\item
\exer{Consider the experts problem in which the losses can be negative and are real numbers in the range $[-1,1]$. Give an algorithm with regret guarantee of $O(\sqrt{T\log N})$ and prove your claim. 
	}	
	
%\end{enumerate}

\end{exercises}

%!TEX root = OCObook.tex

%%%%%%%%%%%%%%%%%%%%%%%%%%%%%%%%%%%%%%%%%%%%%%%%%%%%%%%%%%%%
%%%%%%%%%%%%%%%%%%%%%%%%%%%%%%%%%%%%%%%%%%%%%%%%%%%%%%%%%%%%
%  Offline	 Convex Optimization
%%%%%%%%%%%%%%%%%%%%%%%%%%%%%%%%%%%%%%%%%%%%%%%%%%%%%%%%%%%%
%%%%%%%%%%%%%%%%%%%%%%%%%%%%%%%%%%%%%%%%%%%%%%%%%%%%%%%%%%%%
\chapter{Basic Concepts in Convex Optimization}  \label{chap:opt}
\chaptermark{Convex Optimization} 

In this chapter we give a gentle introduction to convex optimization and present some basic algorithms for solving convex mathematical programs. Although offline convex optimization is not our main topic, 
it is useful to recall the basic definitions and results before we move on to OCO. This will help in assessing the advantages and limitations of OCO. Furthermore, we describe some tools that will be our bread-and-butter later on. 

The material in this chapter is far from being new.  A broad and significantly more detailed literature exists, and the reader is deferred to the bibliography at the end of this chapter for references. We give here only the most elementary analysis, and focus on the techniques that will be of use to us later on. 

\section{Basic Definitions and Setup}  \label{sec:optdefs}

The goal in this chapter is to minimize a continuous and convex function over a convex subset of Euclidean space. Henceforth,  let $\K \subseteq \reals^d$ be a bounded convex and closed set in Euclidean space. We denote by $D$ an upper bound on the diameter of $\K$:
$$ \forall \x,\y \in \K , \ \|\x-\y\| \leq D.$$
A set $\K$ is convex if for any  $\x,\y \in \K$, all the points on the line segment connecting $\x$ and $\y$ also belong to $\K$, i.e., 
$$ \forall \alpha \in [0,1]  , \ \alpha \x + (1-\alpha)\y \in \K.$$
A function $f: \K \mapsto \reals$ is convex if for  any $\x,\y \in \K$  
$$\forall \alpha \in [0,1] , \  f( (1 - \alpha) \x + \alpha \y) \leq (1- \alpha) f(\x) + \alpha f(\y).$$
This inequality, and generalizations thereof, is also known as Jensen's inequality. 
Equivalently, if $f$ is differentiable, that is, its gradient $\nabla f(\x)$ exists for all $\x \in\K$, then it is convex if and only if $\forall \x,\y \in \K$
$$  f(\y) \geq f(\x) + \nabla f(\x)^\top (\y-\x).$$
For convex and non-differentiable functions $f$, the subgradient  at  $\x$ is {\it defined} to be any member of the set of vectors $\{ \nabla f(\x) \}$ that satisfies the above for all $\y \in \K$.   

We denote by $G > 0$ an upper bound on the norm of the subgradients of $f$ over $\K$, i.e., $\|\nabla f(\x)\| \leq G$ for all $\x \in \K$. Such an upper bound implies that the function  $f$ is Lipschitz continuous with parameter $G$, that is, for all $\x,\y \in \K$
$$ |f(\x) - f(\y)| \leq G \|\x-\y\|.$$

The optimization and machine learning literature studies special types of convex functions that admit useful properties, which in turn allow for more efficient optimization. Notably, we say that a function is $\alpha$-strongly convex if
$$  f(  \y) \geq  f(\x) + \nabla f(\x)^\top (\y-\x)  + \frac{\alpha}{2} \|\y-\x\|^2.  $$
A function is $\beta$-smooth if
$$  f(  \y) \leq  f(\x) + \nabla f(\x)^\top (\y-\x)  + \frac{\beta}{2} \|\y-\x\|^2.  $$
The latter condition is equivalent to a Lipschitz condition over the gradients, i.e., 
$$ \| \nabla f(\x) - \nabla f(\y) \| \leq {\beta} \|\x-\y\|.$$

If the function is twice differentiable and admits a second derivative, known as a Hessian for a function of several variables, the above conditions are equivalent to the following condition on the Hessian, denoted $\nabla^2 f(\x)$:
$$ \alpha I \preccurlyeq  \nabla^2 f(\x) \preccurlyeq \beta I, $$
where $A\preccurlyeq B$ if the matrix $B-A$ is positive semidefinite.

When the function $f$ is both $\alpha$-strongly convex and $\beta$-smooth, we say that it is $\gamma$-well-conditioned where $\gamma$ is the ratio between  strong convexity and smoothness, also called the {\it condition number} of $f$
$$ \gamma = \frac{\alpha}{\beta} \leq 1$$

\subsection{Projections onto convex sets}  \label{sec:projections}
In the following algorithms we shall make use of a projection operation onto a convex set, which is defined as the closest point in terms of Euclidean distance \endnote{ We will discuss projections with respect to other distance notions in chapter \ref{chap:regularization}.} inside the convex set to a given point. Formally,
$$ \proj_\K (\y) \equaltri \argmin_{\x \in \K} \| \x - \y \|.$$
When clear from the context, we shall remove the $\K$ subscript. It is left as an exercise to the reader to prove that the projection of a given point over a closed, bounded and convex set exists and is unique.

The computational complexity of projections is a subtle issue that depends much on the characterization of $\K$ itself. Most generally, $\K$ can be represented by a membership oracle---an efficient procedure that is capable of deciding whether a given $\x$ belongs to $\K$ or not. In this case, projections can be computed in polynomial time. In certain special cases, projections can be computed very efficiently in near-linear time. The computational cost of projections, as well as optimization algorithms that avoid them altogether, is discussed in chapter \ref{chap:FW}.  

A crucial property of projections that we shall make extensive use of  is the Pythagorean theorem, which we state here for completeness:
\begin{figure}[ht]
\begin{center}
\includegraphics[width=3.5in]{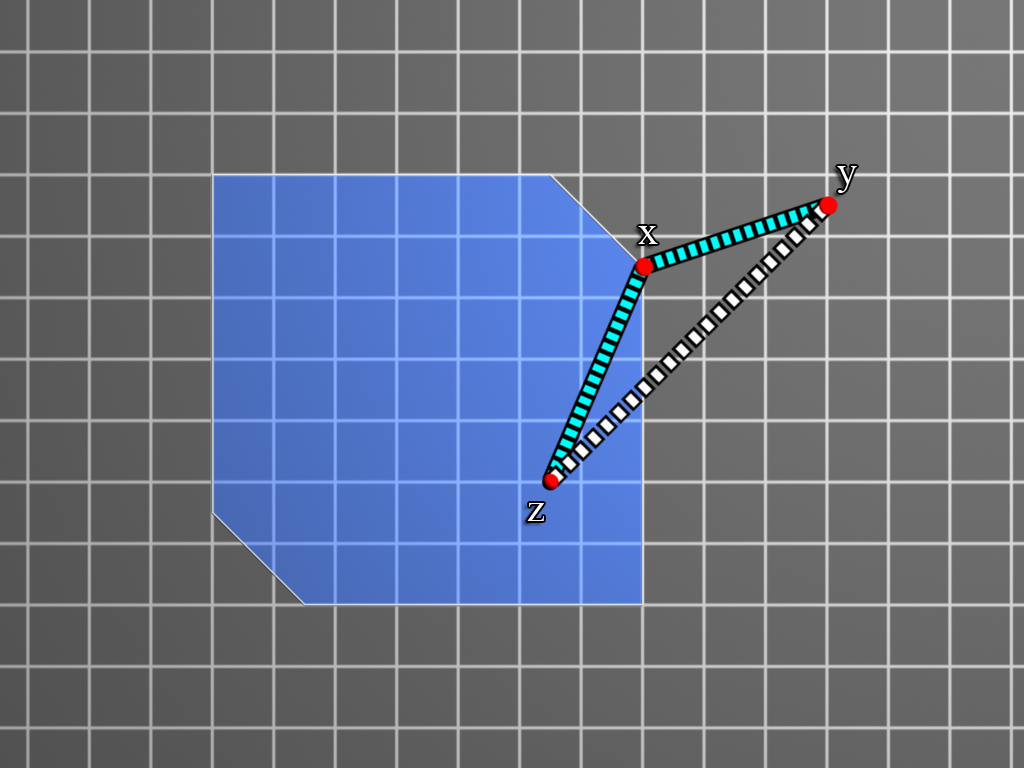}%projection}
\end{center}
\caption{Pythagorean theorem}
\end{figure}
\begin{theorem}[Pythagoras, 	circa 500 BC] \label{thm:pythagoras}
	Let $\K \subseteq \reals^d$ be a convex set, $\y \in \reals^d$ and $\x = \proj_\K(\y)$. Then for any $\z \in \K$ we have
	$$ \| \y - \z \| \geq \| \x - \z \|.$$
\end{theorem}

We note that there exists a more general version of the Pythagorean theorem. The above theorem and the definition of projections are true and valid not only for Euclidean norms, but for projections according to other distances that are not norms. In particular, an analogue of the Pythagorean theorem remains valid with respect to Bregman divergences (see chapter \ref{chap:regularization}). 

\subsection{Introduction to optimality conditions} \label{subsec:optimality-conditions}

The standard curriculum of high school mathematics contains the basic facts concerning when a function (usually in one dimension) attains a local optimum or saddle point. The generalization of these conditions to more than one dimension is called the KKT (Karush-Kuhn-Tucker) conditions, and the reader is referred to the bibliographic material at the end of this chapter for an in-depth rigorous discussion of optimality conditions in general mathematical programming. 

For our purposes, we describe only briefly and intuitively the main facts that we will require henceforth. Naturally, we restrict ourselves to convex programming, and thus a local minimum of a convex function is also a global minimum  (see exercises at the end of this chapter). 
In general there can be many points in which a function is minimized, and thus we refer to the {\it set} of minima of a given objective function, denoted as $\argmin_{\x \in \reals^n} \{ f(\x)\} $ \endnote{this notation stands for the arguments that minimize the expression inside the brakets, which are a subset in $\reals^d$.}.

The generalization of the  fact that a minimum of a convex differentiable function on $\reals$ is a point in which its derivative is equal to zero, is given by the multi-dimensional analogue that its gradient is zero:
$$ \nabla f(\x ) =  0  \ \ \Longleftrightarrow  \ \ \x \in \argmin_{\x \in \reals^n} \{ f(\x) \}.$$

\begin{figure}[ht]
\begin{center}
\includegraphics[width=4in]{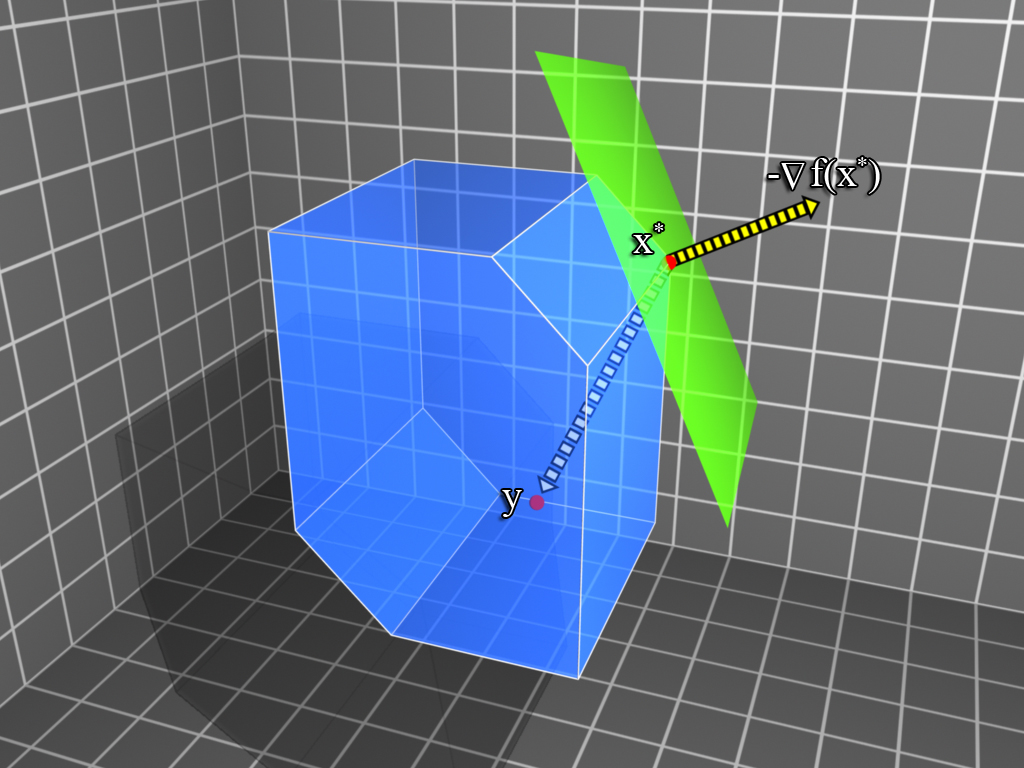}
\end{center}
\caption{Optimality conditions: negative subgradient pointing outwards \label{fig:optimality}}
\end{figure}

We will require a slightly more general, but equally intuitive, fact for constrained optimization: at a minimum point of a constrained convex function, the inner product between the negative gradient and direction towards the interior of $\K$ is non-positive. This is depicted in figure \ref{fig:optimality}, which shows that $-\nabla f(\x^\star)$ defines a supporting hyperplane to $\K$. The intuition is that if the inner product were positive, one could improve the objective by moving in the direction of the projected negative gradient. This fact is stated formally in the following theorem.
\begin{theorem}[Karush-Kuhn-Tucker]  \label{thm:optim-conditions}
Let $\K \subseteq \reals^d$ be a convex set, $\x^\star \in \argmin_{\x \in  \K} f(\x)$.  Then for any $\y \in \K$ we have
$$ \nabla f(\x^\star) ^\top ( \y - \x^\star ) \geq 0.  $$
\end{theorem}

\section{Gradient Descent}

Gradient descent (GD) is the simplest and oldest of optimization methods. It is an {\it iterative method}---the optimization procedure proceeds in iterations, each improving the objective value. The basic method  amounts to iteratively moving the current point in the direction of the gradient, which is a linear time operation if the gradient is given explicitly (indeed, for many functions computing the gradient at a certain point is a simple linear-time operation). 

\begin{figure}[ht] 
	\begin{center}
		\includegraphics[width=2.3in]{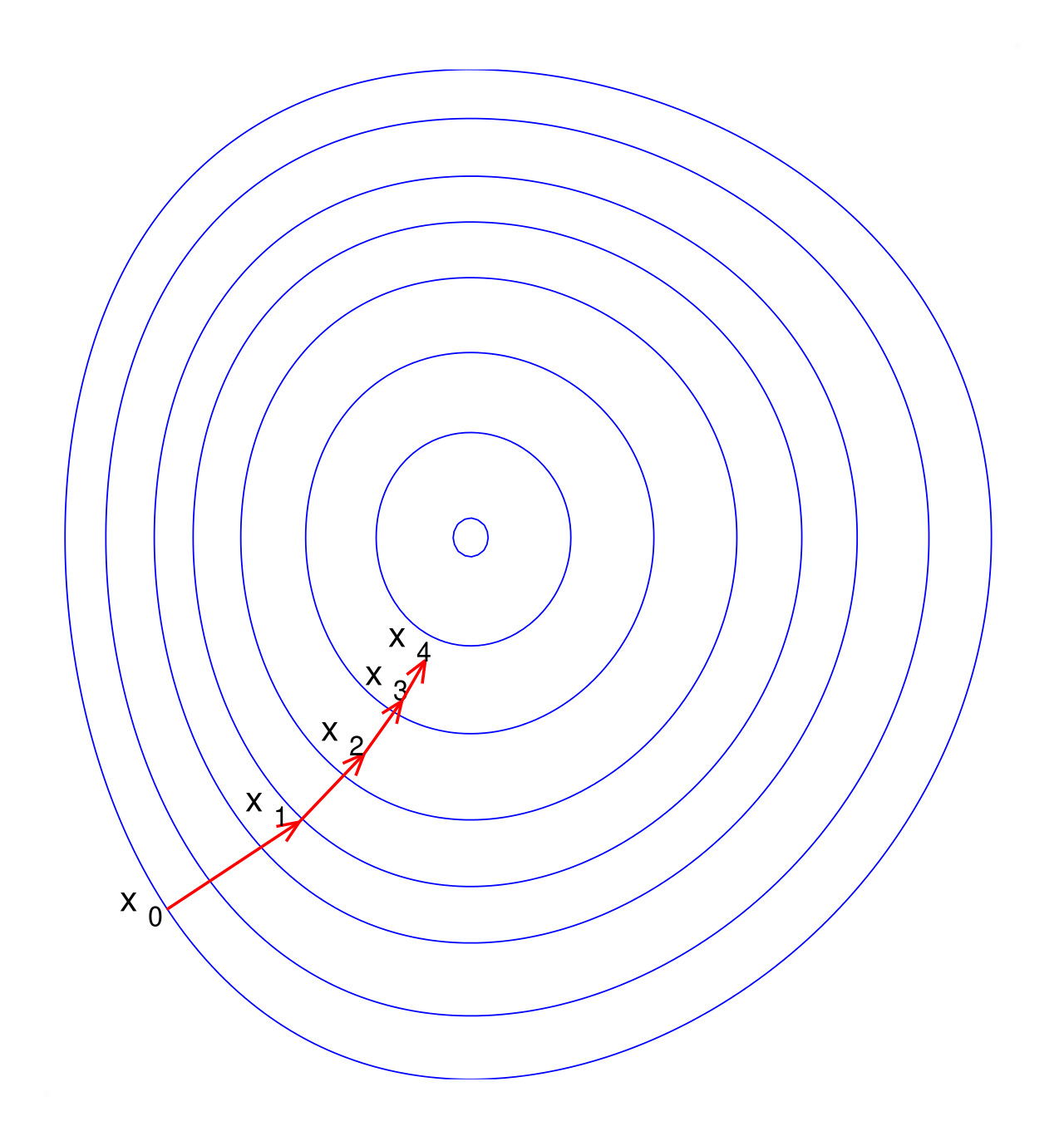}
	\end{center}
	\caption{Iterates of the GD algorithm \label{fig:gradient_descent}}
\end{figure}

The basic template algorithm, for unconstrained optimization, is given in \ref{alg:basic}, and a depiction of the iterates it produced in figure \ref{fig:gradient_descent}.

\begin{algorithm}[ht]
\caption{Gradient Descent}
\label{alg:basic}
\begin{algorithmic}[1]
\State Input: time horizon $T$, initial point $x_0$, step sizes $\{\eta_t\}$
\For{$t = 0, \ldots, T-1$}
%\State Set $\eta_t =  \frac{h_t}{\|\nabla_t\|^2} $
\State  $ \x_{t+1}  =   \x_t - \eta_t \nabla_t $
\EndFor
\State \Return $\xbar = \argmin_{\x_t} \{ f(\x_t) \}$
\end{algorithmic}
\end{algorithm}

For a convex function there always exists a choice of step sizes that will cause GD to converge to the optimal solution. The rates of convergence, however, differ greatly and depend on the smoothness and strong convexity properties of the objective function. 
The following table summarises the convergence rates of GD variants for convex functions with different convexity parameters. The rates described omit the (usually small) constants in the bounds---we focus on asymptotic rates. 

\begin{table}[ht]
\begin{center} 
\begin{tabular}{c|c|c|c|c}
%\hline 
 & general & $\alpha$-strongly & $\beta$-smooth & $\gamma$-well\\ 
 &  		&  convex    		& 			 & conditioned\\ 
 \hline 
Gradient descent  & $\frac{1}{\sqrt{T}}$ & $\frac{1}{\alpha T}$ & $\frac{\beta}{ T}$ & $ e^{- \gamma T } $\\ 
%\hline
%Accelerated GD  & $\frac{d}{{T}}$ & $\frac{d}{\sqrt{\alpha} T^2}$ & $\frac{\sqrt{\beta}}{ T^2}$ & $ e^{- \sqrt{\gamma}T} $\\ 
Accelerated GD  & --- & --- & $\frac{{\beta}}{ T^2}$ & $ e^{- \sqrt{\gamma} \ T} $\\ 
%\hline
\end{tabular}
\end{center}
\caption{Rates of convergence of first order (gradient-based) methods as a function of the number of iterations and the smoothness and strong-convexity of the objective. Dependence on other parameters and constants, namely the Lipchitz constant, diameter of constraint set and initial distance to the objective is omitted. Acceleration for non-smooth functions is not possible in general.} \label{table:GD}
\label{table:offline}
\end{table}%

In this section we address only the first row of Table \ref{table:GD}. For accelerated methods and their analysis see references at the bibliographic section.

\subsection{The Polyak stepsize} 

Luckily, there exists a simple choice of step sizes that yields the optimal convergence rate, called the Polyak stepsize. It has a huge  advantage of not depending on the strong convexity and/or smoothness parameters of the objective function. 

However, it does depend on the distance in function value to optimality and gradient norm. While the latter can be efficiently estimated, the distance to optimality is not always available if $f(\x^*)$ is not known ahead of time. This can be remedied, as referred to in the bibliography.

We henceforth denote:
\begin{enumerate}
    \item Distance to optimality in value: $h_t = h(\x_t) = f(\x_t) - f(\x^*) $
    \item Euclidean distance to optimality: $d_t = \| \x_t - \x^*\| $
    \item Current gradient norm $ \| \nabla_t\| = \|\nabla f(\x_t)\|$
\end{enumerate}

With these notations we can describe the algorithm precisely in Algorithm \ref{alg:basicpolyak}: 

\begin{algorithm}[ht]
\caption{Gradient Descent with Polyak stepsize}
\label{alg:basicpolyak}
\begin{algorithmic}[1]
\State Input: time horizon $T$, $x_0$
\For{$t = 0, \ldots, T-1$}
\State Set $\eta_t =  \frac{h_t}{\|\nabla_t\|^2} $
\State  $ \x_{t+1}  =   \x_t - \eta_t \nabla_t $
\EndFor
\State Return $\xbar = \argmin_{\x_t} \{ f(\x_t) \}$
\end{algorithmic}
\end{algorithm}

To prove precise convergence bounds, assume $\|\nabla_t\| \leq G$, and define:
\begin{eqnarray*}
  B_T    &=&  \min\left\{
  \frac{G d_0}{\sqrt{ T}},
  \frac {2 \beta d_0^2}{ T },
  \frac{3 G^2}{  \alpha  T } ,
   \beta d_0^2\left(1-\frac{\gamma}{4}\right)^T
 \right\}
\end{eqnarray*}

We can now state the main guarantee of GD with the Polyak stepsize: 

\begin{theorem} \label{thm:simple}
(GD with the Polyak Step Size) 
Algorithm \ref{alg:basicpolyak} guarantees the following after $T$ steps: 
\begin{eqnarray*}
f(\xbar)- f(\x^\star) \leq \min_{ 0 \leq t \leq  T} \{ h_t \} \leq B_T % \\
\end{eqnarray*}
\end{theorem}

\subsection{Measuring distance to optimality}

When analyzing convergence of gradient methods, it is useful to use potential functions in lieu of function distance to optimality, such as gradient norm and/or Euclidean distance. The following relationships hold between these quantities. 

\begin{lemma} \label{lem:elementary_properties}
The following properties hold for $\alpha$-strongly-convex functions and/or $\beta$-smooth functions over Euclidean space $\reals^d$. % over a constraint set $\K$: 
\begin{enumerate}
    \item $\frac{\alpha}{2} d_t^2 \leq h_t$
    \item $ h_t \leq \frac{\beta}{2} d_t^2$
    \item $\frac{1}{2 \beta} \|\nabla_t\|^2 \leq h_t$
    \item $ h_t \leq \frac{1}{2 \alpha} \|\nabla_t\|^2 $
\end{enumerate}
\end{lemma}

\begin{proof}
\begin{enumerate}
    \item  $h_t \geq \frac{\alpha}{2} d_t^2 $: %% case 1

    By strong convexity, we have 
\begin{eqnarray*}
h_t & =  f(\x_t) - f(\x^{\star}) \\
& \geq  \nabla f(\x^{\star})^\top (\x_t - \x^{\star}) + \frac{\alpha}{2} \|\x_t - \x^{\star}\|^2  \\
 & = \frac{\alpha}{2} \|\x_t - \x^{\star}\|^2 
\end{eqnarray*}
where the last inequality follows since the gradient at the global optimum is zero. 
    
    \item  $h_t \leq \frac{\beta}{2} d_t^2 $:  %% case 2
    
By smoothness, 
\begin{eqnarray*}
h_t & =  f(\x_t) - f(\x^{\star}) \\
& \leq  \nabla f(\x^{\star})^\top (\x_t - \x^{\star}) + \frac{\beta}{2} \|\x_t - \x^{\star}\|^2  \\
 & = \frac{\beta}{2} \|\x_t - \x^{\star}\|^2 
\end{eqnarray*}
where the last inequality follows since the gradient at the global optimum is zero.

\item  $h_t \geq \frac{1}{2\beta} \|\nabla_t\|^2 $: % case 3
Using smoothness, and let $\x_{t+1} = \x_t - \eta \nabla_t$ for $\eta = \frac{1}{\beta}$, 
\begin{eqnarray*}
h_t =  & f(\x_t) - f(\x^{\star}) \\
& \geq  f(\x_t) - f(\x_{t+1})   \\
 & \geq   \nabla f(\x_t)^\top (\x_{t} - \x_{t+1}) - \frac{\beta}{2} \|\x_t - \x_{t+1} \|^2   \\
 & = \eta \|\nabla_t\|^2  - \frac{\beta}{2} \eta^2 \|\nabla_t\|^2 \\
 & = \frac{1}{2\beta} \|\nabla_t\|^2  .
\end{eqnarray*}

\item   $h_t \leq \frac{1}{2\alpha} \|\nabla_t\|^2 $:  %% case 3
    
We have for any pair $\x,\y \in \reals^d$:
\begin{align*}
f(\y)  & \ge  f(\x) +   \nabla f(\x)^\top  (\y - \x ) + \frac{\alpha}{2}  \|\x - \y\|^2  \\
&\ge  \min_{\z \in \reals^d } \left\{ f(\x) +   \nabla f(\x)^\top  (\z - \x ) + \frac{\alpha}{2}  \|\x - \z\|^2 \right\} \\
& =  f(\x) - \frac{1}{2  \alpha} \| \nabla f(\x)\|^2. \\
& \text{ by taking $\z = \x - \frac{1}{ \alpha} \nabla f(\x) $ }
\end{align*}
In particular, taking $\x = \x_t \ , \ \y = \x^\star$, we get
\begin{equation} \label{eqn:gradlowerbound}
 h_t =  f(\x_t) - f(\x^\star)  \leq \frac{1}{2 \alpha} \|\nabla_t\|^2  .
\end{equation}

\end{enumerate}
\end{proof}

\subsection{Analysis of the Polyak stepsize} 

We are now ready to prove Theorem~\ref{thm:simple}, which directly follows from the following lemma.  
\ignore{
Let $0\leq \gamma \leq 1 $,   define $R_{T,\gamma}$ as follows:
\[
  B_{T,\gamma} = \min\left\{
  \frac{G d_0}{\sqrt{\gamma T}},
  \frac {2 \beta d_0^2}{\gamma T },
  \frac{ 3 G^2}{{\gamma}  \alpha  T } ,
   \beta d_0^2\left(1-\gamma\frac{\alpha}{4 \beta}\right)^T
 \right\}
\, .
\]
}
\begin{lemma} \label{lemma:shalom2}
Suppose that a sequence $\x_0,
\ldots \x_t$ satisfies:
\begin{equation} \label{eqn:shalom3}
d_{t+1}^2 \leq d_t^2 -   \frac{ h_t^2}{\|\nabla_t\|^2}
\end{equation}
then for $\xbar$ as defined in the algorithm, %$ = \argmin_{\x_t} \{ f(\x_t) \}$
we have:
\[
f(\xbar) - f(\x^\star)  \leq \frac{1}{T} \sum_t h_t \leq   B_{T}\, .
\]
\end{lemma}

\begin{proof}
The proof analyzes different cases:
\begin{enumerate}
\item
For convex functions with gradient bounded by $G$, 
\begin{eqnarray*}
d_{t+1}^2 -  d_t^2 & \leq - \frac{ h_t^2}{\|\nabla_t\|^2} \leq -
                     \frac{ h_t^2}{G^2}  
\end{eqnarray*}
Summing up over $T$ iterations, and using Cauchy-Schwartz on the $T$-dimensional vectors of $\frac{1}{T} \bone$ and $(h_1,...,h_T)$, we have
\begin{eqnarray*}
\frac{1}{T} \sum_t h_t 
& \leq&  \frac{1}{\sqrt{T}} \sqrt{\sum_t h_t^2} \\
& \leq& \frac{ G}{\sqrt{ T}} \sqrt{\sum_t (d_{t}^2 - d_{t+1}^2)} \leq
\frac{ G d_0 }{\sqrt{ T}} \, .
\end{eqnarray*}

\item
For smooth functions whose gradient is bounded by $G$,  Lemma~\ref{lem:elementary_properties} implies:
\[
d_{t+1}^2 - d_t^2 \leq - \frac{ h_t^2}{\|\nabla_t\|^2} \leq -
\frac{ h_t}{2 \beta} \, .
\]
This implies
\[
\frac{1}{T} \sum_t h_t  \leq \frac{2 \beta d_0^2}{ T}\, .
\]

\item
For strongly convex functions, Lemma~\ref{lem:elementary_properties} implies:
\[ d_{t+1}^2 - d_t^2
  \leq -  \frac{h_t^2}{\|\nabla_t\|^2}
  \leq -  \frac{h_t^2}{G^2}
  \leq  -   \frac{\alpha^2 d_t^4 }{4 G^2} \, .
\]
In other words,
$d_{t+1}^2  \leq  d_t^2 ( 1-  \frac{\alpha^2 d_t^2}{4 G^2} ) \, .$ 
Defining $a_t := \frac{\alpha^2 d_t^2}{4 G^2}$, we have:
\[
a_{t+1}  \leq  a_t (1-a_t) \, .
\]
This implies that $a_t \leq \frac{1}{t+1}$, which can be seen by
induction\endnote{That $a_0\leq 1$ follows from Lemma
  \ref{lem:elementary_properties}. For $t=1$, $a_1\leq \frac{1}{2}$ since
  $a_1  \leq  a_0 (1-a_0)$ and $0\leq a_0\leq 1$.
  For the induction step, $
  a_t  \leq  a_{t-1} (1-a_{t-1}) \leq
  \frac{1}{t}(1-\frac{1}{t})
  =\frac{t-1}{t^2}=\frac{1}{t+1}(\frac{t^2-1}{t^2})
  \leq \frac{1}{t+1}$.}. The proof is completed as follows\endnote{This assumes $T$ is even. $T$ odd
    leads to the same constants.} :  
\begin{eqnarray*}
\frac{1}{ T/2 } \sum_{t= T/2 }^T h_t^2 &
\leq& \frac{2G^2}{  T  }\sum_{t= T/2 }^T ( d_t^2 -
                                   d_{t+1}^2)  \\
  &=&\frac{2 G^2}{  T } ( d _{ T/2 }^2 - d_T^2)  \\
  &=&\frac{8 G^4}{  \alpha^2   T} ( a
    _{ T/2 } - a_T)  \\ 
   & \leq &\frac{9 G^4}{ \alpha^2 T ^2} 
  \, .
\end{eqnarray*}
Thus, there exists a $t$ for which $h_t^2 \leq \frac{ 9 G^4}{ \alpha^2  T^2} $. Taking the square root completes the claim.

\item
For both strongly convex and smooth functions: 
\[ d_{t+1}^2 - d_t^2 \leq -  \frac{h_t^2}{\|\nabla_t\|^2} \leq
 - \frac{ h_t}{2 \beta} \leq
  -  \frac{\alpha}{4\beta} d_t^2
  \]
  Thus,
  \[
    h_{T} \leq \beta d_{T}^2 \leq \beta d_0^2
    \left(1-\frac{\alpha}{4 \beta}\right)^T = \beta d_0^2
    \left(1-\frac{\gamma}{4}\right)^T \, .
    \]
  \end{enumerate}
This completes the proof of all cases.
\end{proof}

\section{Constrained Gradient/Subgradient Descent} 
\sectionmark{Constrained GD}

The vast majority of the problems considered in this text include constraints. Consider the examples given in section \ref{subsec:OCOexamples}: a path is a point in the flow polytope, a portfolio is a point in the simplex and so on. In the language of optimization, we require $\x$ not only to minimize a certain objective function, but also to belong to a convex set $\K$. 

In this section we describe and analyze constrained gradient descent. Algorithmically, the change from the previous section is small: after updating the current point in the direction of the gradient, one may need to project back to the decision set. However, the analysis is somewhat more involved, and instructive for the later parts of this text.

\subsection{Basic gradient descent---linear convergence} 

Algorithmic box  \ref{alg:BasicGD} describes a  template for  gradient descent over a constrained set. It is a template since the sequence of step sizes $\{\eta_t\}$ is left as an input parameter, and the several variants of the algorithm differ on its choice. 

\begin{algorithm}[ht]
\caption{Basic gradient descent}
\label{alg:BasicGD}
\begin{algorithmic}[1]
\State Input: $f$, $T$, initial point $\x_1 \in \K$, sequence of step sizes $\{\eta_t\}$
\For{$t=1$ to $T$}
\State Let $ \y_{t+1} = \x_{t}-\eta_t {\nabla f}(\x_t) , \  \x_{t+1}= \proj_{\K} \left( \y_{t+1}  \right) $
%\State Let  $x_{t+1}= \min_{x\in\mathcal{K}} |x-y_{t+1}|$
\EndFor
\State \Return ${\x}_{T+1} $ 
\end{algorithmic}
\end{algorithm}

As opposed to the unconstrained setting, here we require a precise setting of the learning rate to obtain the optimal convergence rate. 
\begin{theorem} \label{thm:basicGD}
For constrained minimization of $\gamma$-well-conditioned functions and $\eta_t = \frac{1}{\beta} $,  Algorithm \ref{alg:BasicGD} converges as
$$ h_{t+1} \leq  h_1 \cdot e^{-\frac{\gamma t}{ 4}} $$ 
\end{theorem}

\begin{proof}
By strong convexity we have for every $\x,\x_t \in \K$ (where we denote $\nabla_t = \nabla f(\x_t)$ as before): 
\begin{equation} \label{eqn:tempGD}
   \nabla_t^\top (\x - \x_t) \leq f(\x) - f(\x_t) - \frac{\alpha}{2} \|\x - \x_t\|^2.
\end{equation}
Next, appealing to the algorithm's definition and the choice $\eta_t = \frac{1}{\beta}$, we have
\begin{eqnarray}
\x_{t+1} 
& = \argmin_{\x \in \K}  \left\{  \nabla_t^\top ( \x-\x_t) + \frac{\beta}{2 } \|\x - \x_t\|^2 \right\} \label{eqn:alg_defn_GD} .
\end{eqnarray}
To see this, notice that 
\begin{align*} %\label{eqn:quad-solution}
&  \proj_\K ( \x_t - \eta_t \nabla_t ) \\
& = \argmin_{\x \in \K} \left\{ \|\x - (\x_t - \eta_t \nabla_t) \|^2 \right\} & \mbox{ definition of projection}  \\
& =  \argmin_{\x \in \K}  \left\{  \nabla_t^\top ( \x-\x_t) + \frac{1}{2 \eta_t} \|\x - \x_t\|^2 \right\} . & \mbox{ see exercise 6} 
\end{align*}
Thus, we have 
\begin{align*}
h_{t+1} - h_{t} & = f(\x_{t+1}) - f(\x_t)  \\
& \leq \nabla_t^\top ( \x_{t+1} - \x_t) + \frac{\beta}{2} \|\x_{t+1} - \x_{t} \|^2 & \mbox{ smoothness}  \\
& \leq \min_{\x \in \K}  \left\{  \nabla_t^\top ( \x-\x_t) + \frac{\beta}{2} \|\x - \x_t\|^2 \right\} & \mbox{ by \eqref{eqn:alg_defn_GD} } \\
& \leq  \min_{\x \in \K}  \left\{ f(\x) - f(\x_t)  + \frac{\beta - \alpha}{2} \|\x - \x_t\|^2 \right\}. & \mbox{by } \eqref{eqn:tempGD} \\
\end{align*}
The minimum can only grow if we take it over a subset of $\K$. Thus we can restrict our attention to all points that are convex combination of $\x_t$ and $\x^\star$, which we denote by the interval $[\x_t,\x^\star] = \{ (1 - \mu )\x_t + \mu \x^\star , \mu \in [0,1]\}$, and write  
\begin{align} \label{eqn:shalom22}
h_{t+1} - h_{t}  & \leq  \min_{\x \in [\x_t,\x^\star] }  \left\{ f(\x) - f(\x_t)  + \frac{\beta - \alpha}{2} \|\x - \x_t\|^2 \right\} \notag \\
& =   f( (1-\mu) \x_t + \mu \x^\star ) - f(\x_t)  + \frac{\beta - \alpha}{2}\mu^2 \|\x^\star  - \x_t\|^2 \notag \\ %& \mbox{ $ \x = (1-\mu) \x_t + \mu \x^\star $ } \\
& \le   (1-\mu) f( \x_t)  + \mu f(\x^\star ) - f(\x_t)  + \frac{\beta - \alpha}{2}\mu^2 \|\x^\star  - \x_t\|^2  & \mbox{convexity}   \notag \\
& =  -  \mu h_t + \frac{\beta - \alpha}{2} \mu^2 \|\x^\star  - \x_t\|^2 . 
\end{align}
Where the equality is by writing  $\x$ as $ \x = (1-\mu) \x_t + \mu \x^\star $.
Using strong convexity, we have for any  $\x_t$ and the minimizer $\x^\star$:
\begin{align*}
h_t & = f(\x_t) - f(\x^\star ) \\
 &\ge    \nabla f(\x^\star)^\top  (\x_t - \x^\star ) + \frac{\alpha}{2}  \|\x^\star - \x_t\|^2 & \text{ $\alpha$-strong convexity} \\
& \ge   \frac{\alpha}{2} \| \x^\star - \x_t \|^2. & \text{  optimality Thm \ref{thm:optim-conditions} } 
\end{align*}
Thus, plugging this into equation \eqref{eqn:shalom22}, we get
\begin{align*} 
h_{t+1} - h_t & \leq ( - \mu  + \frac{\beta - \alpha}{\alpha} \mu^2 ) h_t \\
& \leq  - \frac{\alpha}{4 (\beta- \alpha)} h_t. & \mbox{ optimal choice of $\mu$}
\end{align*}
Thus, 
$$ h_{t+1} \leq h_t (1 - \frac{\alpha}{4(\beta - \alpha)}) \leq h_t( 1 - \frac{\alpha}{4 \beta} ) \leq h_t e^{ -\gamma/4}. $$
This gives the theorem statement by induction. 
\end{proof}

\section{Reductions to Non-smooth and Non-strongly Convex Functions} \label{sec:gd-reductions}

The previous section dealt with $\gamma$-well-conditioned functions, which may seem like a significant restriction over vanilla convexity. Indeed, many interesting convex functions are not strongly convex nor smooth, and as we have seen, the convergence rate of gradient descent greatly differs for these functions. We have completed the picture for unconstrained optimization, and in this section we complete it for a bounded set. 

The literature on first order methods is abundant with specialized analyses that explore the convergence rate of gradient descent for more general functions. 
In this manuscript we take a different approach: instead of analyzing variants of GD from scratch, we use reductions to derive near-optimal convergence rates for smooth functions that are not strongly convex, or strongly convex functions that are not smooth, or general convex functions without any further restrictions. 

While attaining sub-optimal convergence bounds (by logarithmic factors), the advantage of this approach is two-fold: first, the reduction method is very simple to state and analyze, and its analysis is  significantly shorter than analyzing GD from scratch.  Second, the reduction method is generic, and thus extends to the analysis of accelerated gradient descent (or any other first order method) along the same lines. 
We turn to these reductions next. 

\subsection{Reduction to smooth, non strongly convex functions}
\sectionmark{Reductions}

Our first reduction applies the GD algorithm to functions that are  $\beta$-smooth but not strongly convex.

The idea is to add a controlled amount of strong convexity to the function $f$, and then apply the algorithm \ref{alg:BasicGD} to optimize the new function. The solution is distorted by the added strong convexity, but a tradeoff guarantees a meaningful convergence rate.

\begin{algorithm}[ht]
\caption{Gradient descent, reduction to $\beta$-smooth functions}
\label{alg:non-strongly convex-GD}
\begin{algorithmic}[1]
\State Input:  $f$, $T$, $\x_1 \in \K$, parameter $\tilde{\alpha}$.
\State Let $g(\x) = f(\x) + \frac{\tilde{\alpha}}{2} \|\x - \x_1 \|^2 $
\State Apply Algorithm \ref{alg:BasicGD}  with parameters $g,T, \{\eta_t = \frac{1}{\beta}\},\x_1$, return $\x_T$.
\end{algorithmic}
\end{algorithm}

\begin{lemma} \label{thm:smoothGDreduction}
For $\beta$-smooth convex functions,  Algorithm \ref{alg:non-strongly convex-GD} with parameter $\tilde{\alpha} = \frac{ \beta \log t }{D^2 t}$ converges  as
$$ h_{t+1} = O \left(  \frac{\beta \log t } {t} \right)  $$
\end{lemma}
\begin{proof}
The function $g$ is $\tilde{\alpha}$-strongly convex and $(\beta+ \tilde{\alpha})$-smooth (see exercises). Thus, it is  $\gamma = \frac{\tilde{\alpha}}{\tilde{\alpha} + \beta}$-well-conditioned. 
Notice that
\begin{align*}
h_t & = f(\x_t) - f(\x^\star) \\
& = g(\x_t) - g(\x^\star) + \frac{\tilde{\alpha}}{2} (\|\x^\star - \x_1\|^2 - \|\x_t -\x_1\|^2) \\
& \le  h_t^g + \tilde{\alpha}D^2. & \mbox{ def of $D$, \S \ref{sec:optdefs}}
\end{align*}
Here, we denote $h^g_t = g(\x_t) - g(\x^\star)$. Since  $g(\x)$ is   $\frac{\tilde{\alpha}}{\tilde{\alpha} + \beta}$-well-conditioned, 
\begin{align*}
 h_{t+1} & \le h_{t+1}^g  + \tilde{\alpha} D^2 \\
& \leq   h_1^g e^{-\frac{\tilde{\alpha} t}{  4( \tilde{\alpha}+\beta)}} + \tilde{\alpha} D^2    & \mbox{ Theorem \ref{thm:basicGD}} \\
& = O ( \frac{ \beta \log t}{ t} ),   & \mbox { choosing $\tilde{\alpha} = \frac{ \beta \log t }{D^2 t}$} 
\end{align*}
where we ignore constants and terms depending on  $D$ and $h_1^g$.
\end{proof}

Stronger convergence rates of $O(\frac{\beta}{t})$ can be obtained by analyzing GD from scratch, and these are known to be tight. Thus, our reduction is suboptimal by a factor of $O(\log T)$, which we tolerate for the reasons stated at the beginning of this section.

\subsection{Reduction to strongly convex, non-smooth functions}

Our reduction from non-smooth functions to $\gamma$-well-conditioned functions is similar in spirit to the one of the previous subsection. However, whereas for strong convexity the obtained rates were off by a factor of $\log T$, in this section we will also be off by factor of $d$, the dimension of the decision variable $\x$, as compared to the standard analyses in convex optimization. 
For tight bounds, the reader is referred to the excellent reference books and surveys listed in the bibliography section \ref{sec:bib_of_optimization}. 

\begin{algorithm}[ht]
\begin{algorithmic}[1]
\caption{Gradient descent, reduction to non-smooth functions \label{alg:reduction2}}
\State Input: $f,\mathbf{x}_1,T,\delta$
\State Let $ \hat{f}_\delta (\x) =  \E  _{\vv \sim \ball} \left[ f ( \x+ \delta \vv) \right]$
\State Apply Algorithm \ref{alg:BasicGD} on $\hat{f}_\delta,\x_1,T,\{\eta_t = {\delta}\} $, return $\x_T$
\end{algorithmic}
\end{algorithm} 

We apply the GD algorithm to a smoothed variant of the objective function. In contrast to the previous reduction, smoothing cannot be obtained by simple addition of a smooth (or any other) function. Instead, we need a smoothing operation. The one we describe is particularly simple and amounts to taking a local integral of the function. More sophisticated, but less general, smoothing operators exist that are based on the Moreau-Yoshida regularization, see bibliographic section for more details. 

Let $f$ be $G$-Lipschitz continuous and $\alpha$-strongly convex. Define for any $\delta > 0$,
$$ S_\delta[f] : \reals^d \mapsto \reals \  \ , \ \  S_\delta[f](\x) =  \E  _{\vv \sim  \ball} \left[ f ( \x+ \delta \vv) \right] , $$
where $\ball = \{ \x \in \mathbb{R} ^d : \|\x\|  \leq 1 \}$ is the Euclidean ball and $\vv \sim \ball$ denotes a random variable drawn from the uniform distribution over  $\ball$. When the function $f$ is clear from the context, we henceforth use the simpler notation $\hat{f}_\delta = S_\delta[f]$. 

We will prove that the function $\fhat_\delta = S_\delta[f] $ is a smooth approximation to $f: \reals^d \mapsto \reals$, i.e., it is both smooth and close in value to $f$, as given in the following lemma. 

\begin{lemma}
\label{lem:SmoothingLemma}
$\hat{f}_\delta$ has the following properties:
\begin{enumerate}
\item If $f$ is $\alpha$-strongly convex, then so is $\fhat_\delta$ 
\item $\hat{f}_\delta$ is $\frac{d G}{\delta}$-smooth 
\item $ |\hat{f} _\delta (\x) - f(\x) | \le \delta G$ for all $ \x \in \mathcal{K}$ .
\end{enumerate}
\end{lemma}

Before proving this lemma, let us first complete the reduction. 
Using Lemma \ref{lem:SmoothingLemma} and the convergence for $\gamma$-well-conditioned functions the following approximation bound is obtained.

\begin{lemma}
For $\delta = \frac{dG}{\alpha} \frac{\log{t}}{t}$ Algorithm \ref{alg:reduction2}  converges as 
$$h_t=O\left( \frac{G^2 d  \log{t}}{\alpha t}\right).$$
\end{lemma}
Before proving this lemma, notice that the gradient descent method is applied with gradients of the smoothed function $\fhat_\delta$ rather than gradients of the original objective $f$. In this section we ignore the computational cost of computing such gradients given only access to gradients of $f$, which may be significant. Techniques for estimating these gradients are further explored in chapter \ref{chap:bandits}.

\begin{proof}
Note that by Lemma \ref{lem:SmoothingLemma} the function $\fhat_\delta$ is $\gamma$-well-conditioned for 
$ \gamma = \frac{\alpha \delta}{d G}.  $ 

\begin{align*}
 h_{t+1} & = f(\x_{t+1})-f(\x^\star) \\
&\le \hat{f}_\delta(\x_{t+1})-\hat{f}_\delta(\x^\star) + 2\delta G &\mbox{Lemma \ref{lem:SmoothingLemma}} \\
&\le h_1 e^{-\frac{\gamma t}{4}}+2\delta G & \mbox{Theorem \ref{thm:basicGD}}\\
&= h_1 e^{-\frac{\alpha t \delta}{4 dG}}+2\delta G& \mbox{$\gamma = \frac{\alpha \delta}{d G}$ by Lemma \ref{lem:SmoothingLemma}}\\
&= O \left( \frac{dG^2 \log t }{\alpha t} \right). &\mbox{$\delta = \frac{dG}{\alpha} \frac{\log{t}}{t}$}\\
\end{align*}
\end{proof}

We proceed to prove that $\fhat_\delta$ is indeed a good approximation to the original function. 

\begin{proof} [Proof of Lemma \ref{lem:SmoothingLemma}]
First, since $\hat{f}_\delta$ is an average of $\alpha$-strongly convex functions, it is also $\alpha$-strongly convex.
In order to prove smoothness, we will use Stokes' theorem from calculus:
For all $\x \in \reals^d$ and for a  vector random variable $\vv$ which is uniformly distributed over the Euclidean sphere $\sphere = \{ \y \in \mathbb{R} ^d : \|\y\|  = 1 \}$, 
\begin{equation}
\label{lem:stokes_application}
 \E _{\vv  \sim \sphere} [ f(\x + \delta \vv) \vv ] = \frac{\delta}{d} \nabla\hat{f}_\delta (\x).
\end{equation}

Recall that a function $f$ is $\beta$-smooth if and only if  for all $  \x,\y \in \K$, it holds that $ \| \nabla f(\x) -\nabla f(\y) \| \le \beta \|\x-\y\|$. Now,
\begin{align*}
& \| \nabla \hat{f}_\delta (\x) -\nabla \hat{f}_\delta (\y) \|  =  \\
& = \frac{d}{\delta} \| \E _{\vv \sim \sphere} \left[ f(\x + \delta \vv) \vv \right] -\E _{\vv \sim  \sphere} \left[ f(\y + \delta \vv) \vv \right]\| & \mbox{by \eqref{lem:stokes_application}}  \\
& = \frac{d}{\delta} \| \E _{\vv \sim \sphere} \left[ f(\x + \delta \vv) \vv - f(\y + \delta \vv) \vv \right] \|  &\mbox{linearity of expectation} \\
& \le \frac{d}{\delta} \E_{\vv \sim \sphere}   \| f(\x + \delta \vv) \vv - f(\y + \delta \vv) \vv \|    & \mbox{Jensen's inequality}\\
& \le  \frac{dG}{\delta} \| \x- \y\|   \E _{\vv \sim \sphere} \left[ \|\vv\| \right] & \mbox{Lipschitz continuity}\\
&= \frac{dG}{\delta} \| \x- \y\|.   & \mbox{$ \vv \in \sphere$}
\end{align*}
This proves the second property of Lemma \ref{lem:SmoothingLemma}. We proceed to show the third property, namely that  $\fhat_\delta$ is a good approximation to $f$. 

\begin{align*}
& |\hat{f}_\delta (\x)-f(\x)| 
 = \left|\E_{\vv \sim  \ball} \left[  f(\x+ \delta \vv)\right] - f(\x) \right| &\mbox{definition of $\hat{f}_\delta$}\\
& \leq \E_{\vv \sim \ball} \left[  |f(\x+ \delta \vv) - f(\x) |\right]  &\mbox{ Jensen's inequality} \\
& \le \E_{\vv \sim \ball}\left[ G\| \delta \vv \| \right] & \mbox{$f$ is $G$-Lipschitz}\\
& \leq G \delta. & \mbox{ $\vv \in \ball$}
 \end{align*}
\end{proof}

We note that GD variants for $\alpha$-strongly convex functions, even without the smoothing approach used in our reduction, are known to converge quickly and without dependence on the dimension. We state the known algorithm and result here without proof (see bibliography for references). 
\begin{theorem} \label{thm:strongly convex-GD-bubeck}
Let $f$ be $\alpha$-strongly convex, and let $\x_1,...,\x_t$ be the iterates of  Algorithm \ref{alg:BasicGD} applied to $f$ with $\eta_t = \frac{2}{\alpha (t+1)}$. Then 
$$ f\left( \frac{1}{t} \sum_{s=1}^t \frac{2 s }{t+1} \x_s \right) - f(\x^\star) \leq \frac{2 G^2}{\alpha (t+1)}  . $$
\end{theorem}

\subsection{Reduction to general convex functions}

One can apply both reductions simultaneously to obtain a rate of  $\tilde{O}(\frac{d}{\sqrt{t}})$. While near-optimal in terms of the number of iterations, the weakness of this bound lies in its dependence on the dimension. In the next chapter  we shall show a rate of $O(\frac{1}{\sqrt{t}})$ as a direct consequence of a more general online convex optimization algorithm.

\section{Example: Support Vector Machine  Training} \label{sec:svmexample}
\sectionmark{Example: SVM}

To illustrate the usefulness of the gradient descent method, let us describe an  optimization problem that has gained much attention in machine learning and can be solved efficiently using the methods we have just analyzed. 

A very basic and  successful learning paradigm is  the linear classification model.  In this model, the learner is presented with  positive and negative examples of a concept. Each example, denoted by $\ba_i$, is represented in Euclidean space
by a $d$ dimensional feature vector. For example, a common representation for emails in the spam-classification problem  are binary vectors in Euclidean space, where the dimension of the space is the number of words in the language. The $i$'th email is a vector $\ba_i$ whose entries are given as ones for coordinates corresponding to words that appear in the email, and zero otherwise\endnote{Such a representation may seem na\"ive at first as it completely ignores the words' order of appearance and their context.  Extensions to capture these features are indeed studied in the Natural Language Processing literature.}. In addition, each example has a label  $b_i \in \{-1,+1\}$, corresponding to whether the email has been labeled spam/not spam. The goal is to find a hyperplane separating the two classes of vectors: those with positive labels and those with negative labels. If such a hyperplane, which completely  separates the training set according to the labels, does not exist, then the goal is to find a 
hyperplane that achieves a separation of the training set with the smallest number of mistakes. 

Mathematically speaking, given a set of $n$ examples to train on, we seek $\x \in \reals^d$ that minimizes the number of incorrectly classified examples, i.e.
\begin{equation} \label{eqn:linear-classification}
\min_{\x \in \reals^d} \sum_{i \in [n]}  \delta( \sign(\x^\top \ba_i ) \neq b_i)  
\end{equation}
where $\sign(x) \in \{-1,+1\}$ is the sign function, and $\delta(z) \in \{0,1\}$ is the indicator function that takes the value $1$ if the condition $z$ is satisfied and zero otherwise.

This optimization problem, which is at the heart of the linear classification formulation, is NP-hard, and in fact NP-hard to even approximate non-trivially \endnote{see bibliography for references to these results.}. However, in the special case that a linear classifier (a hyperplane $\x$) that classifies all of the examples correctly exists, the problem is solvable in polynomial time via linear programming. 

Various relaxations have been proposed to solve the more general case, when no perfect linear classifier exists. One of the most successful in practice is the Support Vector Machine (SVM)  formulation.  

\begin{figure}[t!]
	\begin{center}
		\includegraphics[width=2.3in]{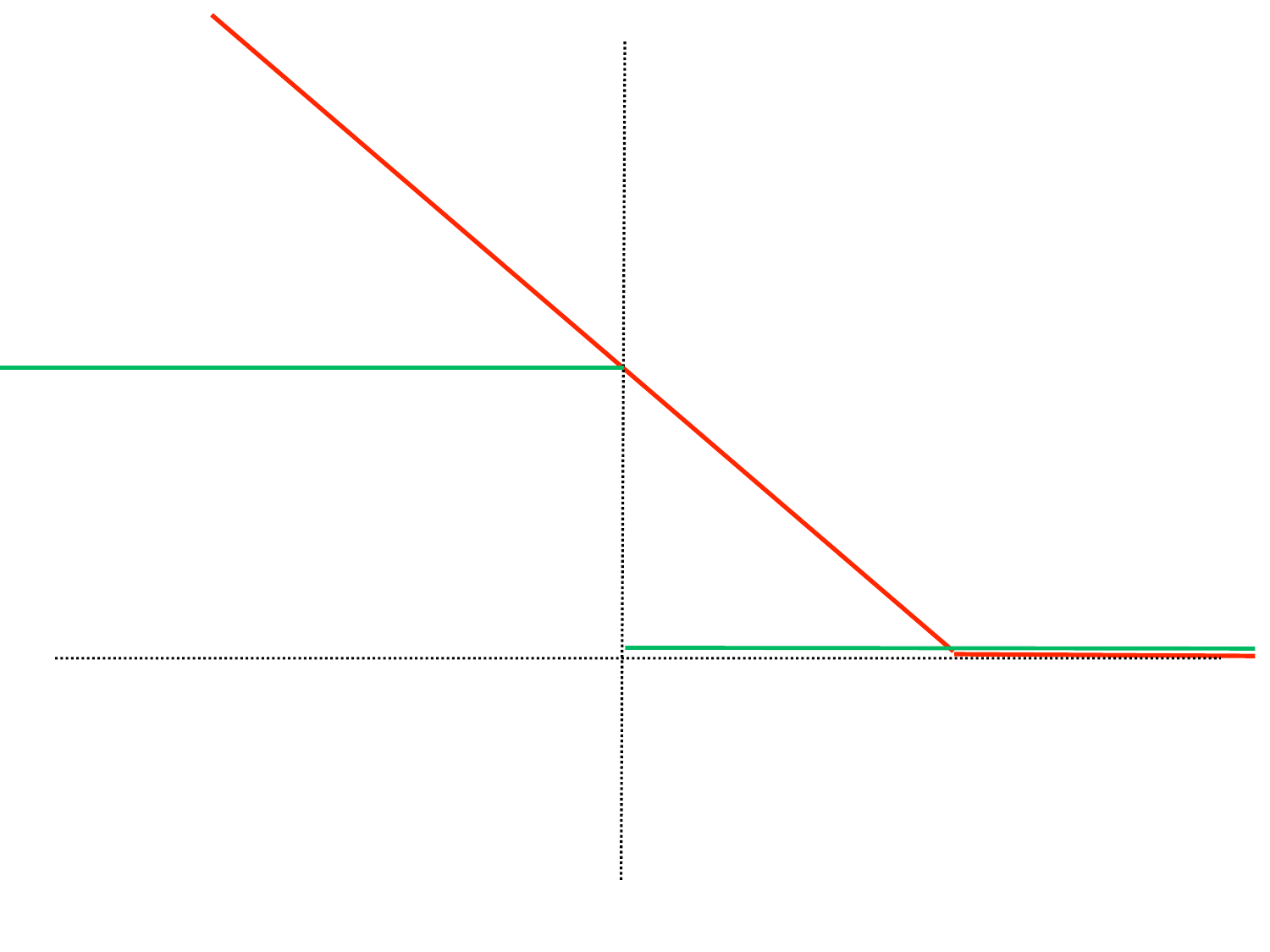}
	\end{center}
	\caption{The hinge loss function versus the 0/1 loss function \label{fig:hinge}}
\end{figure}

The soft margin SVM relaxation replaces the $0/1$ loss  in \eqref{eqn:linear-classification}  with a convex loss function, called the hinge-loss, given by
$$ \ell_{\ba,b}(\x) = \text{hinge}(b \cdot \x^\top \ba )  = \max\{0, 1 - b \cdot \x^\top \ba \}.$$
In figure \ref{fig:hinge} we depict how the hinge loss is a convex relaxation for the non-convex $0/1$ loss.
Further, the SVM formulation adds to the loss minimization objective a term that regularizes the size of the elements in $\x$. The reason and meaning of this additional term   shall be addressed in later sections. For now, let us consider the SVM convex program: 
\begin{equation} \label{eqn:soft-margin}
\min_{\x \in \reals^d} \left \{ \lambda \frac{1}{n} \sum_{i \in [n]}  \ell_{\ba_i,b_i}(\x)  + \frac{1}{2} \| \x \|^2  \right \} 
\end{equation}

\begin{algorithm}[ht]
	\caption{SVM training via subgradient descent}
	\label{alg:BasicGDSVM}
	\begin{algorithmic}[1]
			\State Input: training set of $n$ examples $\{(\ba_i,b_i) \}$,  $T$, learning rates $\{\eta_t\}$, initial $\x_1 = 0$.
			\For {$t=1$ to $T$}
			\State Let ${\nabla_t} = \lambda \frac{1}{n} \sum_{i=1}^n \nabla \ell_{\ba_i,b_i} (\x_t)  + \x_t $ where 
			$$ \nabla \ell_{\ba_i,b_i}(\x) = \mycases {0}  { b_i \x^\top \ba_i > 1 } { - b_i \ba_i}{ \text{otherwise}}$$
			\State  ${\x}_{t+1}  = \x_{t}-\eta_t {\nabla_t}$ for $\eta_t = \frac{2}{t+1}$
			\EndFor
			\State \Return $\bar{\x}_T =  \frac{1}{T} \sum_{t=1}^T \frac{2 t }{T+1} \x_t $ 
	\end{algorithmic}
\end{algorithm}

\bigskip
This is an unconstrained non-smooth and strongly convex program. It follows from  Theorems \ref{thm:simple} and \ref{thm:strongly convex-GD-bubeck} that ${O}(\frac{1}{\varepsilon})$ iterations suffice to attain an $\eps$-approximate solution. We spell out the details of applying the subgradient descent algorithm to this formulation  in Algorithm  \ref{alg:BasicGDSVM}. 

Notice that the learning rates are left unspecified, even though they can be explicitly set as in Theorem  \ref{thm:strongly convex-GD-bubeck}, or using the Polyak rate. The Polyak rate requires knowing the function value at optimality, although this can be relaxed (see bibliography).  

A caveat of using gradient descent for SVM is the requirement to compute the full gradient, which may require a full pass over the data for each iteration. We will see a significantly more efficient algorithm in the next chapter! 

%that the above algorithm returns an $\eps$-approximate solution after $T = O(\frac{1}{\epsilon})$ iterations. 

\newpage
\section{Bibliographic Remarks} \label{sec:bib_of_optimization}

The reader is referred to dedicated books on convex optimization for much more in-depth treatment of the topics surveyed in this background chapter.  For background in convex analysis see the texts  \citep{borwein2006convex,rockafellar1997convex}. The classic textbook of \citet{boyd.convex} gives a broad introduction to convex optimization with numerous applications, see also \citep{BoydNotes}. For detailed rigorous convergence proofs and in depth analysis of first order methods, see lecture notes by \citet{NesterovBook} and books by \citet{NY83,Nemirovski04lectures}, as well as more recent lecture notes and texts \citep{bubeckOPT,hazan2019lecture}.   
Theorem \ref{thm:strongly convex-GD-bubeck} is taken from \citep{bubeckOPT} Theorem 3.9. 

The logarithmic overhead in the reductions of section \ref{sec:gd-reductions} can be removed with a more careful reduction and analysis, for details see \citep{ZeyuanH06}. A more sophisticated smoothing operator is the Moreau-Yoshida regularization: it avoids the dimension factor loss. However, it is sometimes  less computationally efficient to work with \citep{parikh2014proximal}. 

The Polyak learning rate is detailed in \citep{polyak}. A recent exposition allows obtaining the same optimal rate without knowledge of the optimal function value \citep{hazan2019revisiting}.

Using linear separators and halfspaces to learn and separate data was considered in the very early days of AI \citep{rosenblatt1958perceptron,minsky69perceptrons}.  Notable the Perceptron algorithm was one of the first learning algorithms, and closely related to gradient descent. 
Support vector machines were introduced in \citep{CortesV95,Boser92}, see also the book of \citet{ScSm02}. 

Learning halfspaces with the zero-one loss is computationally hard, and hard to even approximate non-trivially \citep{daniely2016complexity}. Proving that a problem is hard to approximate is at the forefront of computational complexity, and based on novel characterizations of the complexity class NP \citep{AroraBarakbook}.

\newpage

%\section{Exercises}
\begin{exercises}

%\begin{enumerate}
%\item
\exer{Prove that a differentiable function $f(x) : \mathbb{R} \rightarrow \mathbb{R}$ is convex if and only if for any $x,y\in\mathbb{R}$ it holds that $f(x)-f(y) \leq (x-y)f'(x)$.
}

%\item
\exer{Recall that we say that a function $f:\mathbb{R}^n\rightarrow\mathbb{R}$ has a condition number $\gamma = \alpha/\beta$ over $
K \subseteq \reals^d$ if the following two inequalities hold for all $\x,\y \in \K$:
%\begin{enumerate}
\subexer{$  f(\y) \geq f(\x) + (\y-\x)^{\top}\nabla{}f(\x) + \frac{\alpha}{2}\Vert{\x-\y}\Vert^2$}
\subexer{$  f(\y) \leq f(\x) + (\y-\x)^{\top}\nabla{}f(\x) + \frac{\beta}{2}\Vert{\x-\y}\Vert^2$}
%\end{enumerate}
%For matrices $A,B \in \reals^{n \times n}$ we denote $A \succcurlyeq B$ if $A-B$ is positive semidefinite. 
Prove that if $f$ is twice differentiable and it holds that $\beta\textbf{I} \succcurlyeq \nabla^2f(\x) \succcurlyeq \alpha\textbf{I}$ for any $\x\in \K$, then the condition number of $f$ over $\K$  is $\alpha/\beta$.
}

%\item
\exer{Prove:} 
%\begin{enumerate}
%	\item
\subexer{The sum of convex functions is convex. }
%	\item
\subexer{Let $f$ be $\alpha_1$-strongly convex and $g$ be $\alpha_2$-strongly convex. Then $f+g$ is $(\alpha_1+\alpha_2)$-strongly convex. }
%	\item
\subexer{Let $f$ be $\beta_1$-smooth and $g$ be $\beta_2$-smooth. Then $f+g$ is $(\beta_1+\beta_2)$-smooth. }
	
%\end{enumerate}
%}

%\item
\exer{Let $\K \subseteq \reals^d$ be bounded and closed. Prove that convexity of $\K$ is a necessary and sufficient condition for all $\x \in \reals^d$ for $\proj_K(\x)$ to be a singleton, that is for $|\proj_K(\x)| = 1$. To prove that this is a necessary condition, it is enough to provide a counterexample. 
}

%\item
\exer{Consider the $n$-dimensional simplex 
$$\Delta_n = \lbrace{\x\in\mathbb{R}^n \, | \, \sum_{i=1}^n \x_i = 1, \, \x_i \geq 0 \ ,  \ \forall{i\in[n]}}\rbrace .$$
Give an algorithm for computing the projection of a point $\x\in\mathbb{R}^n$ onto the set $\Delta_n$ (a near-linear time algorithm exists).
}

%\item
\exer{
Prove the following identity:
\begin{align*}
 & \argmin_{\x \in \K}  \left\{  \nabla_t^\top ( \x-\x_t) + \frac{1}{2 \eta_t} \|\x - \x_t\|^2 \right\} \notag \\
  = &  \argmin_{\x \in \K} \left\{ \|\x - (\x_t - \eta_t \nabla_t) \|^2 \right\} .
\end{align*}
}

%\item
\exer{Let $f(\x):\mathbb{R}^n\rightarrow\mathbb{R}$ be a convex differentiable function and $\K\subseteq \mathbb{R}^n$ be a convex set. Prove that $\x^\star\in \K$ is a minimizer of $f$ over $\K$ if and only if for any $\y\in \K$ it holds that $(\y-\x^\star)^{\top}\nabla f(\x^\star) \geq 0$.
}

%\item $^*$
\exer{ * Extending Nesterov's accelerated GD algorithm: \\
Assume a black-box access to Nesterov's algorithm that attains the rate of $e^{-\sqrt{\gamma} \ T}$ for $\gamma$-well-conditioned functions, as in Table \ref{table:offline}. Apply a reduction to obtain the $\frac{\beta}{T^2}$ rate for $\beta$-smooth functions, as in Table \ref{table:offline}, up to logarithmic factors. }

%\end{enumerate}
\end{exercises}

%%%%%%%%%%%%%%%%%%%%%%%%%%%%%%%%%%%%%%%%%%%%%%%%%%%%%%%%%%%%
%%%%%%%%%%%%%%%%%%%%%%%%%%%%%%%%%%%%%%%%%%%%%%%%%%%%%%%%%%%%
%  Online 	 Convex Optimization
%%%%%%%%%%%%%%%%%%%%%%%%%%%%%%%%%%%%%%%%%%%%%%%%%%%%%%%%%%%%
%%%%%%%%%%%%%%%%%%%%%%%%%%%%%%%%%%%%%%%%%%%%%%%%%%%%%%%%%%%%

\chapter{First-Order Algorithms for Online Convex Optimization} \label{chap:first order}
\chaptermark{First-Order Algorithms}

In this chapter we describe and analyze the most simple and basic algorithms for online convex optimization (recall the definition of the model as introduced in chapter \ref{chap:intro}), which are also surprisingly useful and applicable in practice. 
 We use the same notation introduced in \S \ref{sec:optdefs}. However, in contrast to the previous chapter, the goal of the algorithms introduced in this chapter is to minimize {\it regret}, rather than the optimization error (which is ill-defined in an online setting). 
 
Recall the definition of regret in an OCO setting, as given in equation \eqref{eqn:regret-defn}, with subscripts, superscripts and the supremum over the function class omitted when they are clear from the context:
$$ \regret_T = \sum_{t=1}^{T} f_t(\bx_t) -\min_{\bx \in \K}\sum_{t=1}^{T} f_t(\bx) . $$

Table \ref{table:regret-rates} details known upper and lower bounds on the regret for different types of convex functions as it depends on the number of prediction iterations. 

\begin{table}[ht]
\begin{center}
\begin{tabular}{c|c|c|c }
%\hline 
 &   $\alpha$-strongly convex & $\beta$-smooth & $\delta$-exp-concave \\ 
 \hline 
Upper bound  & $\frac{1}{ \alpha} \log T $ & $\sqrt{T}$ &   $\frac{n}{ \delta} \log T $  \\ 
\hline
Lower bound  & $\frac{1}{ \alpha} \log T $ & $\sqrt{T}$ &   $\frac{n}{ \delta } \log T $  \\
\hline 
Average regret  & $\frac{\log T}{ \alpha T}  $ & $\frac{1}{\sqrt{T}}$ &   $\frac{n \log T}{ \delta T} $  \\ 
%\hline
\end{tabular}
\end{center}
\caption{Attainable asymptotic regret bounds for loss function classes.} \label{table:regret-rates}
\label{default}
\end{table}%

In order to compare regret to optimization error it is useful to consider the average regret, or ${\regret}/{T} $. Let  $\bar{\x}_T = \frac{1}{T} \sum_{t=1}^T \x_t$ be the average decision. If the functions $f_t$ are all equal to a single function $f : \K\mapsto \reals$, then Jensen's inequality implies that $f( \bar{\x}_T)$ converges to $f(\x^\star)$ at a rate at most the average regret, since
$$ f(\bar{\x}_T) - f(\x^\star ) \leq \frac{1}{T} \sum_{t=1} ^T [f(\x_t)  - f(\x^\star) ] = \frac{\regret_T}{T} .$$

The reader may recall Table \ref{table:GD} describing offline convergence of first order methods: as opposed to offline optimization,  smoothness does not improve asymptotic regret rates.  However, exp-concavity, a weaker property than strong convexity, comes into play and gives improved regret rates.

This chapter will present algorithms and lower bounds that realize the above known results for OCO. The property of exp-concavity and its applications, as well as  logarithmic regret algorithms for exp-concave functions are deferred to the next chapter.

\section{Online Gradient Descent} \label{section:ogd}

Perhaps the simplest algorithm that applies to the most general
setting of online convex optimization is online gradient descent.
This algorithm, which is based on  standard gradient descent
 from offline optimization, was introduced in its online form by Zinkevich (see bibliography at the end of this section).

\begin{algorithm}[ht]
		\caption{\label{alg:ogd} \ogd }
		\begin{algorithmic}[1]
			\State Input: convex set $\K$, $T$, $\x_1 \in \mathcal{K}$, step sizes $\{ \eta_t \}$
			\For {$t=1$ to $T$}
			\State Play $\x_t$ and observe cost $f_t(\x_t)$. 
			\State Update and project:
			\begin{align*}
			& \y_{t+1} = \bx_{t}-\eta_{t} \nabla f_{t}(\bx_{t}) \\
			& \bx_{t+1} = \proj_\K(\y_{t+1})
			\end{align*}
			\EndFor
		\end{algorithmic}
\end{algorithm}

\begin{figure}[ht] 
	\begin{center}
		\includegraphics[width=4in]{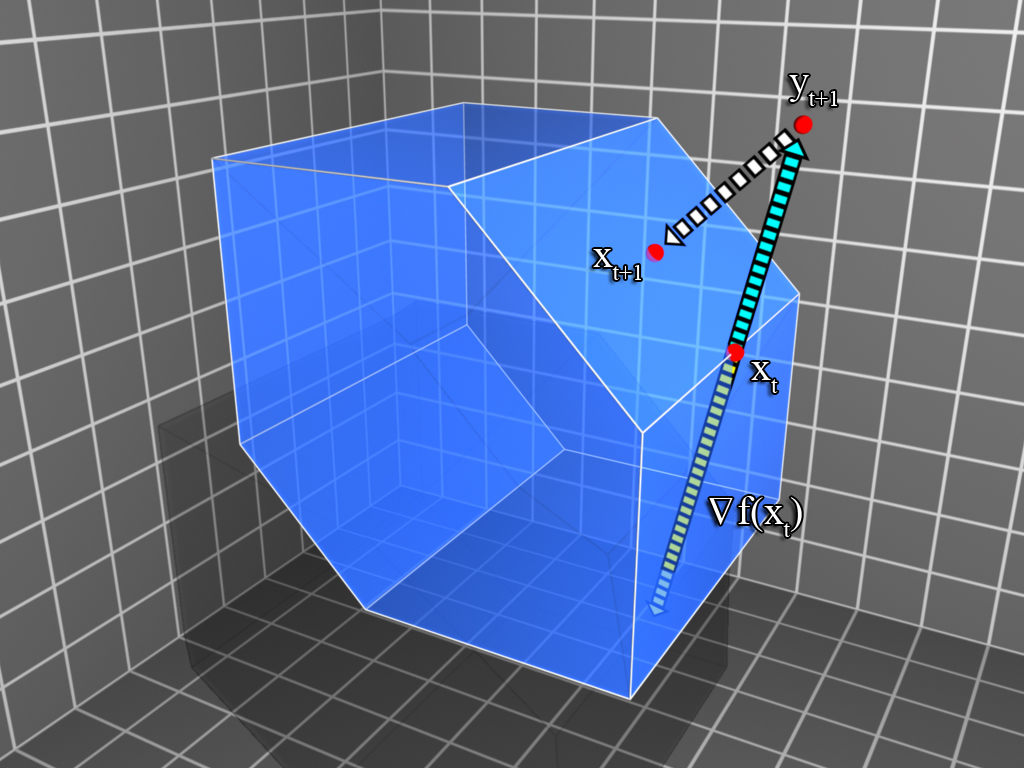}
	\end{center}
	\caption{OGD: the iterate $\x_{t+1}$ is derived by advancing $\x_t$ in the direction of the current gradient $\nabla_t$, and projecting back into $\K$ \label{fig:ogd}}
\end{figure}

Pseudo-code for the algorithm is given in Algorithm \ref{alg:ogd}, and a conceptual illustration is given in figure \ref{fig:ogd}.

In each iteration, the algorithm takes a step from the previous point in the direction of the gradient of the previous cost.  This step may  result in a point outside of the underlying convex set. In such cases,
the algorithm projects the point back to the convex set, i.e.
finds its closest  point in the convex set.
Despite the fact that the next cost function may be completely
different than the costs observed thus far, the regret
attained by the algorithm is sublinear. This is formalized in the following
theorem (recall the definition of $G$ and $D$ from the previous chapter).

\begin{theorem}\label{thm:gradient}
Online gradient descent with step sizes $\{\eta_t = \frac{D}{G \sqrt{t}} , \ t \in [T] \}$ guarantees the following for all $T \geq 1$:
$$ \regret_T = \sum_{t=1}^{T} f_t(\bx_t) -\min_{\bx^\star \in \K}\sum_{t=1}^{T}
f_t(\bx^\star)\ \leq  \frac{3}{2} {G D}\sqrt{T} . $$
\end{theorem}

\begin{proof}
Let $\bx^\star \in \argmin_{\bx \in \K} \sum_{t=1}^T f_t(\bx)$.
Define $\nabla_t \equaltri \nabla f_{t}(\bx_{t})$. By convexity
\begin{eqnarray}  \label{eqn:gradient_inequality}
f_t(\bx_t) - f_t(\bx^\star) \leq   \nabla_t^\top (\bx_t - \bx^\star)
\end{eqnarray}
We first  upper-bound $\nabla_t^\top (\bx_t-\bx^\star)$ using the update
rule for $\bx_{t+1}$ and Theorem  \ref{thm:pythagoras} (the Pythagorean theorem):
\begin{equation} \label{eqn:ogdtriangle}
\| \bx_{t+1}-\bx^\star \|^2\ =\  \left\|\proj_\K (\bx_t - \eta_t
\nabla_{t}) -\bx^\star\right\|^2 \leq  \left\|\bx_t - \eta_t \nabla_t-\bx^\star\right\|^2 .
\end{equation}

Hence,
\begin{eqnarray} \label{eqn:ogd_eq2}
\|\bx_{t+1}-\bx^\star\|^2\ &\leq&\ \|\bx_t- \bx^\star\|^2 + \eta_t^2
\|\nabla_t\|^2 -2 \eta_t \nabla_t^\top (\bx_t -\bx^\star)\nonumber\\
2 \nabla_t^\top (\bx_t-\bx^\star)\ &\leq&\ \frac{ \|\bx_t-
\bx^\star\|^2-\|\bx_{t+1}-\bx^\star\|^2}{\eta_t} + \eta_t G^2 .
\end{eqnarray}
Summing \eqref{eqn:gradient_inequality} and \eqref{eqn:ogd_eq2}  from $t= 1$ to $T$, and setting $\eta_t =
\frac{D}{G \sqrt{t}}$ (with $\frac{1}{\eta_0} \equaltri 0$):
\begin{align*}
& 2 \left( \sum_{t=1}^T f_t(\bx_t)-f_t(\bx^\star) \right ) \leq 2\sum_{t=1}^T \nabla_t^\top (\xv - \x^\star) \\
&\leq  \sum_{t=1}^T \frac{ \|\bx_t-
	\bx^\star\|^2-\|\bx_{t+1}-\bx^\star\|^2}{\eta_t} + G^2 \sum_{t=1}^T \eta_t    \\
&\leq  \sum_{t=1}^T \|\bx_t - \bx^\star\|^2 \left(
\frac{1}{\eta_{t}} -
\frac{1}{\eta_{t-1}} \right) + G^2 \sum_{t=1}^T \eta_t & \frac{1}{\eta_0} \equaltri 0, \\
& &  \|\xv[T+1] - \xv[]^* \|^2 \geq 0 \\
&\leq D^2 \sum_{t=1}^T \left(
\frac{1}{\eta_{t}} -
\frac{1}{\eta_{t-1}} \right) + G^2 \sum_{t=1}^T \eta_t \\
& \leq  D^2  \frac{1}{\eta_{T}}  + G^2 \sum_{t=1}^T \eta_t  & \mbox{ telescoping series } \\
& \leq 3 DG \sqrt{T}.
\end{align*}
The last inequality follows since $\eta_t = \frac{D}{G \sqrt{t}}$ and  $\sum_{t=1}^T \frac{1}{\sqrt{t}} \leq 2 \sqrt{T}$.
\end{proof}

The \ogd algorithm is straightforward to implement, and updates
take linear time given the gradient. However, there is a
projection step which may take significantly longer, as discussed in \S \ref{sec:projections} and chapter \ref{chap:FW}.

\section{Lower Bounds} \label{section:lowerbound}

The previous section introduces and analyzes a very simple and natural approach to online convex optimization. Before continuing our venture, it is worthwhile to consider whether the previous bound can be improved? We measure performance of OCO algorithms both by regret and  by computational efficiency. Therefore, we ask ourselves whether even  simpler algorithms that attain tighter regret bounds exist. 

The computational efficiency of \ogd  seemingly leaves little room for improvement, putting aside the projection step it runs in linear time per iteration. What about obtaining better regret? 

Perhaps surprisingly, the answer is negative: \ogd attains, in the worst case, tight regret bounds up to small constant factors!  This is formally given in the following theorem.

\begin{theorem} \label{thm:lowerbound}
Any algorithm for online convex optimization incurs $\Omega(DG
\sqrt{T})$ regret in the worst case. This is true even if  the cost functions are generated from
a fixed stationary distribution.
\end{theorem}

We give a sketch of the proof; filling in all details is left as an exercise at the end of this chapter. 

Consider an instance of OCO where the convex set $\K$ is the
$n$-dimensional hypercube, i.e.
$$ \K = \{ \bx \in \reals^n \ , \  \|\bx\|_\infty \leq 1 \}.$$
There are $2^n$ linear cost functions, one for each vertex $\bv \in \{
\pm 1\}^n$, defined as
$$ \forall \bv \in \{ \pm 1 \}^n \ , \ f_\bv(\bx)  = \bv^\top \bx. $$
%Here, the cost functions are linear:  they are inner products with  vectors which
%are  vertices of the hypercube. 
Notice that both the diameter of $\K$ and
the bound on the norm of the cost function gradients, denoted G, are bounded by
$$  D \leq \sqrt{ \sum_{i=1}^n 2^2 } = 2 \sqrt{n} , \ G \leq \sqrt{ \sum_{i=1}^n (\pm1)^2 } = \sqrt{n}  $$

The cost functions in each iteration are chosen at random, with uniform probability, from the set  $\{ f_\bv , \bv \in \{\pm 1\}^n \}$. Denote by $\bv_t \in \{\pm 1\}^n $ the vertex chosen in
iteration $t$, and denote $f_t = f_{\bv_t}$. By uniformity and independence, for any $t$ and
$\bx_t$ chosen online, $\E_{\bv_t}[f_{t}(\bx_t)]= \E_{\bv_t}[ \bv_t^\top
\bx_t] = 0$. However,
\begin{align*}
\E_{\bv_1,\ldots,\bv_T}\left[\min_{\bx \in \K} \sum_{t=1}^T f_t(\bx)\right] & =
\E \left[\min_{\bx \in \K} \sum_{i \in [n]} \sum_{t=1}^T \bv_t(i) \cdot \bx_i \right] \\
& = n \E\left[-\left|\sum_{t=1}^T \bv_t(1) \right|\right] & \mbox{i.i.d. coordinates}\\
& = -\Omega(n \sqrt{T}).
\end{align*}
The last equality is left as an exercise. 

The facts above nearly complete the proof of Theorem \ref{thm:lowerbound}; see the exercises at the end of this chapter.

\section{Logarithmic Regret} 

At this point, the reader may wonder: we have introduced a seemingly sophisticated and obviously general framework for learning and prediction, as well as  a linear-time algorithm for the most general case, complete with tight regret bounds, and done so with elementary proofs! Is this all OCO has to offer? 

The answer to this question is two-fold:
\begin{enumerate}
\item
Simple is good: the philosophy behind OCO treats simplicity as a merit. The main reason OCO has taken the stage in online learning in recent years is the simplicity of its algorithms and their analysis, which allow for numerous variations and tweaks in their host applications. 

\item
A very wide class of settings, which will be the subject of the next sections, admit more efficient algorithms, in terms of both regret and computational complexity. 
\end{enumerate}

In \S \ref{chap:opt} we surveyed optimization algorithms with convergence rates that vary greatly according to the convexity properties of the function to be optimized. Do the regret bounds in online convex optimization vary as much as the convergence bounds in offline convex optimization over different classes of convex cost functions?

Indeed, next we show that for important classes of loss functions significantly better regret bounds are possible.

\subsection{Online gradient descent for strongly convex functions} \label{section:ogdnew}

The first algorithm that achieves regret logarithmic in the number
of iterations is a twist on the online gradient descent algorithm, changing only the step size. The following theorem establishes logarithmic bounds on the regret if the
cost functions are strongly convex.

\begin{theorem}\label{thm:gradient2}
For  $\alpha$-strongly convex loss functions, 
\ogd with step sizes $\eta_t = \frac{1}{\alpha {t}}$ achieves the
following guarantee for all $T \geq 1$
$$\regret_T\ \leq\ \frac{G^2}{2 \alpha}(1 + \log T).$$
\end{theorem}

\begin{proof}
Let $\bx^\star \in \argmin_{\bx \in \K} \sum_{t=1}^T f_t(\bx)$.
Recall the definition of regret 
$$ \regret_T\ = \sum_{t=1}^{T} f_t(\bx_t) - \sum_{t=1}^{T} f_t(\bx^\star). $$

Define $\nabla_t \equaltri \nabla f_t(\bx_t)$. Applying the definition of $\alpha$-strong convexity to the pair of points $\{\x_t$,$\x^*\}$, we have
\begin{eqnarray}
2(f_t(\bx_t)-f_t(\bx^\star)) &\leq& 2\nabla_t^\top (\bx_t-\bx^\star)-\alpha
\|\bx^\star-\bx_t\|^2.\label{eqsz}
\end{eqnarray}
We proceed to upper-bound $\nabla_t^\top
(\bx_t-\bx^\star)$. Using the update rule for $\bx_{t+1}$ and the
Pythagorean theorem \ref{thm:pythagoras}, we
get
$$\| \bx_{t+1}-\bx^\star \|^2\ =\ \|\proj_\K(\bx_t - \eta_{t} \nabla_t)-\bx^\star\|^2 \leq  \|\bx_t - \eta_{t} \nabla_t-\bx^\star\|^2.$$
Hence,
\begin{eqnarray*}
\|\bx_{t+1}-\bx^\star\|^2\ &\leq&\ \|\bx_t- \bx^\star\|^2 + \eta_{t}^2
\|\nabla_t\|^2 -2
\eta_{t} \nabla_t^\top (\bx_t - \bx^\star)\nonumber\\
\end{eqnarray*}
and
\begin{eqnarray}
2 \nabla_t^\top (\bx_t-\bx^\star)\ &\leq&\ \frac{ \|\bx_t-
\bx^\star\|^2-\|\bx_{t+1}-\bx^\star\|^2}{\eta_{t}} + \eta_{t} G^2.
\label{eqer}
\end{eqnarray}
Summing  \eqref{eqer} from $t= 1$ to $T$, setting $\eta_{t} = \frac{1}{\alpha t}$ (define $\frac{1}{\eta_0} \equaltri 0$),
and combining with \eqref{eqsz}, we have:
\begin{eqnarray*}
&  & 2 \sum_{t=1}^T (f_t(\bx_t)-f_t(\bx^\star) ) \\
&\leq &\
 \sum_{t=1}^T \|\bx_t-\bx^\star\|^2
\left(\frac{1}{\eta_{t}}-\frac{1}{\eta_{t-1}}-\alpha\right) +G^2
\sum_{t=1}^{T} \eta_{t}   \\
& & \mbox{ since }  \frac{1}{\eta_0} \equaltri 0, \|\xv[T+1] - \xv[]^* \|^2 \geq 0 \\ \\
&=&\ 0 + G^2 \sum_{t=1}^{T} \frac{1}{\alpha t} \\
& \leq &  \frac{G^2}{\alpha}(1 + \log T )
\end{eqnarray*}
\end{proof}

\section{Application: Stochastic Gradient Descent}  \label{sec:sgd}
\sectionmark{Stochastic Gradient Descent}

A special case of Online Convex Optimization is the well-studied  setting of stochastic optimization. In stochastic optimization, the optimizer attempts to minimize a convex function over a convex domain as given by the mathematical program:
\begin{align*}
\min_{\x \in \K } f(\x). 
\end{align*}
However, unlike standard offline optimization, the optimizer is given access to a noisy gradient oracle, defined by
$$ \mathcal{O}(\x) \equaltri \tilde{\nabla }_\x  \ \mbox{ s.t. } \  \E[\tilde{\nabla }_\x ] = \nabla f(\x) \ , \   \E[ \|\tilde{\nabla }_\x\|^2 ] \leq G^2  $$
That is, given a point in the decision set, a noisy gradient oracle returns a random vector whose expectation is the gradient at the point and whose variance is bounded by $G^2$. 
	
We will show that regret bounds for OCO  translate to convergence rates for stochastic optimization.  As a special case, consider the online gradient descent algorithm whose regret is bounded by 
$$\regret_{T} = O(DG\sqrt{T}) $$
Applying the OGD algorithm over a sequence of linear functions that are defined by the noisy gradient oracle at consecutive points, and finally returning the average of all points along the way, we obtain the stochastic gradient descent algorithm, presented in Algorithm \ref{alg:sgd}.
\begin{algorithm}[ht]
\caption{stochastic  gradient descent}
\label{alg:sgd}
\begin{algorithmic}[1]
\State Input: $\mO$,$\K$, $T$, $\x_1 \in \mathcal{K}$, step sizes $\{ \eta_t \}$
\For {$t=1$ to $T$}
\State \label{alg:sgd-defnft} Let $\tilde{\nabla}_t = \mathcal{O}(\x_t)$ 
\State Update and project:
$$ \y_{t+1} = \bx_{t}-\eta_t \tilde{\nabla}_t$$
$$ \bx_{t+1} = \proj_\K(\y_{t+1})$$
\EndFor
\State \Return $\bar{\x}_T \equaltri \frac{1}{T} \sum_{t=1}^T \x_t$ 
\end{algorithmic}
\end{algorithm}

\begin{theorem} \label{thm:sgd}
Algorithm \ref{alg:sgd} with step sizes $\eta_t = \frac{D}{G \sqrt{t}}$ guarantees
$$ \E[ f(\bar{\x}_T) ]  \leq \min_{\x^\star \in \K } f(\x^\star) + \frac{3 GD }{2\sqrt{T}} . $$
\end{theorem}
\begin{proof}
For the analysis, we define the linear functions $ f_t(\x) \equaltri   \tilde{\nabla}_t^\top  \x  $.
Using the regret guarantee of OGD, we have
\begin{align*}
&  \E [ f(\bar{\x}_T) ] - f(\x^\star) \\
 & \leq \E [ \frac{1}{T} \sum_t  f(\x_t) ]  - f(\x^\star)  & \mbox{ convexity of $f$ (Jensen) }\\
&\leq \frac{1}{T}  \E [ \sum_t   \nabla f(\x_t)^\top(  \x_t - \x^\star)  ] & \mbox{ convexity again }\\
& = \frac{1}{T} \E[ \sum_t  \tilde{\nabla}_t^\top (  \x_t -\x^\star )  ] & \mbox{ noisy gradient estimator }\\
& = \frac{1}{T} \E[ \sum_t  f_t( \x_t)  -f_t(\x^\star) ] & \mbox{ Algorithm \ref{alg:sgd}, line \eqref{alg:sgd-defnft} }\\
& \leq  \frac{\regret_T }{T} & \mbox{ definition }\\
& \leq  \frac{3GD }{2\sqrt{T}} & \mbox{ theorem \ref{thm:gradient}}
\end{align*}
\end{proof}
	
It is important to note that in the proof above, we have used the fact that the regret bounds of \ogd hold against an adaptive adversary. This need arises since the cost functions $f_t$ defined in Algorithm \ref{alg:sgd} depend on the choice of decision $\x_t \in \K$. 
	
In addition, the careful reader may notice that by plugging in different step sizes (also called learning rates) and applying SGD to strongly convex functions, one can attain $\tilde{O}({1}/{T})$ convergence rates, where the $\tilde{O}$ notation hides logarithmic factors in $T$. Details of this derivation are left as an exercise. 
	
\subsection{Example: stochastic gradient descent for SVM training}
\sectionmark{SGD for SVM}
	
Recall our example of Support Vector Machine training from \S \ref{sec:svmexample}. The task of training an SVM over a given data set amounts to solving the following convex program (equation \eqref{eqn:soft-margin}):
\begin{align*} 
& f(\x)  = \min_{\x \in \reals^d} \left \{ \lambda \frac{1}{n} \sum_{i \in [n]}  \ell_{\ba_i,b_i}(\x)  + \frac{1}{2} \| \x \|^2  \right \} \\
& \ell_{\ba,b}(\x) = \max\{0, 1 - b \cdot \x^\top \ba \} .
\end{align*}

\begin{algorithm}[ht]
\caption{SGD for SVM training}
\label{alg:sgd4svm}
\begin{algorithmic}[1]
\State Input: training set of $n$ examples $\{(\ba_i,b_i) \}$,  $T$. Set $\x_1 = 0$
\For {$t=1$ to $T$}
	\State Pick an example uniformly at random $t \in [n]$. 
	\State Let $\tilde{\nabla}_t = \lambda  \nabla \ell_{\ba_t,b_t} (\x_t)  + \x_t $ where 
	$$ \nabla \ell_{{\ba_t},b_t}(\x_t) = \mycases {0}  { b_t \x_t^\top \ba_t > 1 } { - b_t \ba_t}{ \mbox{otherwise}}$$
	\State  ${\x}_{t+1}  = \x_{t}-\eta_t \tilde{\nabla}_t$ 
\EndFor
\State \Return $\bar{\x}_T \equaltri \frac{1}{T} \sum_{t=1}^T \x_t$ 
\end{algorithmic}
\end{algorithm}

Using the technique described in this chapter, namely the OGD and SGD algorithms, we can devise a much faster algorithm than the one presented in the previous chapter. The idea is to generate an unbiased estimator for the gradient of the objective using a single example in the dataset, and use it in lieu of the entire gradient. This is given formally in the  SGD algorithm for SVM training presented in  Algorithm \ref{alg:sgd4svm}.

It follows from Theorem \ref{thm:sgd} that this algorithm, with appropriate parameters $\eta_t$, returns an $\eps$-approximate solution after $T = O(\frac{1}{\epsilon^2})$ iterations. Furthermore, with a little more care and using Theorem \ref{thm:gradient2}, a rate of $\tilde{O}(\frac{1}{\epsilon})$ is obtained with parameters $\eta_t = O( \frac{1}{t}) $. 

This matches the convergence rate of standard offline gradient descent. However, observe that each iteration is significantly cheaper---only one example in the data set need be considered! That is the magic of SGD;  we have matched the nearly optimal convergence rate of first order methods using extremely cheap iterations. This makes it the method of choice in numerous applications.

\newpage
\section{Bibliographic Remarks}

The OCO framework was introduced by \citet{Zinkevich03}, where the OGD algorithm was introduced and analyzed. Precursors to this algorithm, albeit for less general settings, were introduced and analyzed in \citep{KivWar97}. 
Logarithmic regret algorithms for Online Convex Optimization were introduced and analyzed in \citep{HAK07}.

The stochastic gradient descent (SGD) algorithm dates back to \citet{robbins1951}, where it was called ``stochastic approximation". The importance of SGD for machine learning was advocated for in \citep{bottou1998online,bottou2008tradeoffs}. 
The literature on SGD is vast and the reader is referred to the text of \citet{bubeckOPT} and paper by \citet{lan2012optimal}. 

Application of SGD to soft-margin SVM training was explored in \citep{Shalev-ShwartzSSC11}. Tight convergence rates of SGD for strongly convex and non-smooth functions were only recently obtained in \citep{hazan:beyond,RSS,SZ}.

\newpage
\begin{exercises}
%\section{Exercises}

%\begin{enumerate}
%\item
\exer{
Prove that SGD for a strongly convex function can, with appropriate parameters $\eta_t$, converge as $\tilde{O}(\frac{1}{T})$. Recall that the $\tilde{O}$ notation hides logarithmic factors in the parameters, including $T$.  You may assume that the gradient estimators have Euclidean norms bounded by the constant $G$.  
}

\exer{$^*$ 
%\item$^*$
In this exercise we show how to remove some a-priory knowledge from the design of online convex optimization algorithms. 
%\begin{enumerate}
   % \item 
   \subexer{Design an OCO algorithm that attains the same asymptotic regret bound as OGD, up to factors logarithmic in $G$ without knowing the parameter $G$ ahead of time. }
      %$ \item 
       \subexer{Do the same for the parameter $D$: design an OCO algorithm that attains the same asymptotic regret bound as OGD, up to factors logarithmic in $D$ without knowing the parameter $D$ ahead of time. This time you may assume $G$ is known. You may assume that it is possible to compute projections onto $\K$ without knowing its diameter. }
%\end{enumerate}
}

%\item
\exer{In this exercise we prove a tight lower bound on the regret of any algorithm for online convex optimization.}
%\begin{enumerate}
%\item
\subexer{For any sequence of $T$ fair coin tosses, let $N_h$ be the number of head outcomes and $N_t$ be the number of tails. Give an asymptotically tight upper and lower bound on $\E[ \left| N_h - N_t \right|]$. That is, give an order of growth of this random variable as a function of $T$, up to multiplicative and additive constants. }
%\item
\subexer{Consider a 2-expert problem, in which the losses are inversely correlated: either expert one incurs a loss of one and the second expert negative one, or vice versa. Use the fact above to design a setting in which any experts algorithm incurs regret asymptotically matching the upper bound.}
%\item
\subexer{Consider the general OCO setting over a convex set $\mathcal{K}$. Design a setting in which the cost functions have gradients whose norm is bounded by $G$, and obtain a lower bound on the regret as a function of $G$, the diameter of $\mathcal{K}$, and the number of game iterations.}
%\end{enumerate}
	
%\item
\exer{Implement the SGD algorithm for SVM training. Apply it on the MNIST dataset. Compare your results to the offline GD algorithm from the previous chapter. }

%\end{enumerate}
\end{exercises}

%!TEX root = OCObook.tex

%%%%%%%%%%%%%%%%%%%%%%%%%%%%%%%%%%%%%%%%%%%%%%%%%%%%%%%%%%%%
%%%%%%%%%%%%%%%%%%%%%%%%%%%%%%%%%%%%%%%%%%%%%%%%%%%%%%%%%%%%
%  Online 	 Convex Optimization
%%%%%%%%%%%%%%%%%%%%%%%%%%%%%%%%%%%%%%%%%%%%%%%%%%%%%%%%%%%%
%%%%%%%%%%%%%%%%%%%%%%%%%%%%%%%%%%%%%%%%%%%%%%%%%%%%%%%%%%%%

\chapter{Second-Order Methods}  \label{chap:second order-methods}

The motivation for this chapter is the application of online portfolio selection, considered in the first chapter of this book.  We begin with a detailed description of this application. We proceed to describe a new class of convex functions that model this problem. This new class of functions is more general than the class of strongly convex functions discussed in the previous chapter. It allows for logarithmic regret algorithms, which are  based on second order methods from convex optimization. In contrast to first order methods, which have been our focus thus far and relied on (sub)gradients, second order methods exploit information about the second derivative of the objective function.

\section{Motivation: Universal Portfolio Selection}

In this subsection we give the formal definition of the universal  portfolio selection problem that was informally described in \S \ref{subsec:OCOexamples}.

\subsection{Mainstream portfolio theory}

Mainstream financial theory models stock prices as a stochastic process known as Geometric Brownian Motion (GBM). This model assumes that the fluctuations in the prices of the stocks behave essentially as a random walk. It is perhaps easier to think about a price of an asset (stock) on time segments, obtained from a  discretization of time into equal segments. Thus, the logarithm of the price at segment $t+1$, denoted $l_{t+1}$, is given by the sum of the logarithm of the price at segment $t$ and a Gaussian random variable with a particular mean and variance, 
$$ l_{t+1} \sim l_t + \mathcal{N}(\mu,\sigma).$$

This is only an informal way of thinking about GBM. The formal model is a continuous time process, similar to the discrete time stochastic process we have just described, obtained as the time intervals, means, and variances approach zero.

The GBM model gives rise to particular algorithms for portfolio selection (as well as more sophisticated applications such as options pricing). Given the means and variances of the stock prices over time of a set of assets, as well as their cross-correlations,  a portfolio with maximal expected gain (mean) for a specific risk (variance) threshold can be formulated. 

The fundamental question is, of course, how does one obtain  the mean and variance parameters, not to mention the cross-correlations, of a given set of stocks?  One accepted solution is to estimate these from historical data, e.g., by taking the recent history of stock prices.

\subsection{Universal portfolio theory}
\sectionmark{Universal portfolio selection}

The theory of universal portfolio selection is very different from the GBM model. The main difference being the lack of statistical assumptions about the stock market.  The idea is to model investing as a repeated decision making scenario, which fits nicely into our OCO framework, and to measure regret as a performance metric. 

Consider the following scenario: at each iteration $t \in [T]$,
the decision maker chooses $\x_t$, a distribution of her wealth over $n$ assets, such that $\xv \in \Delta_n$. Here $\Delta_n = \{ \x \in \reals^n_+ , \sum_i \x_i = 1 \}$ is the $n$-dimensional simplex, i.e., the set of all distributions over $n$ elements. An adversary independently chooses market returns for the assets, i.e., a vector $\rv \in \reals_+^n$ such that each coordinate $\rv(i)$ is the price ratio for the $i$'th asset between the iterations $t$ and $t+1$. For example, if the $i$'th coordinate is the Google ticker symbol GOOG traded on the NASDAQ, then
$$ \rv(i) = \frac{\mbox{price of GOOG at time $t+1$}}{\mbox{price of GOOG at time $t$}} $$ 
How does the decision maker's wealth change? Let $W_t$ be her total wealth at iteration $t$. Then, ignoring transaction costs, we have
$$ W_{t+1} = W_t \cdot \rv^\top \xv$$
Over $T$ iterations, the total wealth of the investor is given by 
$$W_{T} = W_1 \cdot \prod_{t=1}^{T-1} \rv^\top \xv$$
The goal of the decision maker, to maximize the overall wealth gain ${W_T}/{W_0}$, can be attained by maximizing the following more convenient logarithm of this quantity, given by
$$ \log \frac{W_T}{W_1} =  \sum_{t=1}^{T-1} \log \rv^\top \xv$$ 

The above formulation is already very similar to our OCO setting, albeit phrased as a gain maximization rather than a loss minimization setting. Let $$f_t(\x) = \log (\rv^\top \x)$$  
The convex set is the $n$-dimensional simplex $\K = \Delta_n$, and define the regret to be
$$ \regret_T = \max_{\x^\star \in \K } \sum_{t=1}^T f_t(\x^\star) - \sum_{t=1}^T f_t(\x_t)$$
The functions $f_t$ are concave rather than convex, which is perfectly fine as we are framing the problem as a maximization rather than a minimization. Note also that the regret is the negation of the usual regret notion  \eqref{eqn:regret-defn} we have considered for minimization problems. 

Since this is an online convex optimization instance, we can use the \ogd algorithm from the previous chapter to invest, which ensures $O(\sqrt{T})$ regret (see exercises). What guarantee do we attain in terms of investing? To answer this, in the next section we reason about what $\x^\star$  in the above expression may be.

\subsection{Constant rebalancing portfolios}

As $\x^\star \in \K = \Delta_n$ is a point in the $n$-dimensional simplex,  consider the special case of $\x^\star = \ev[1]$, i.e., the first standard basis vector (the vector that has zero in all coordinates except the first, which is set to one). The term $\sum_{t=1}^T f_t(\ev[1])$ becomes $\sum_{t=1}^T \log \rv[t](1)$, or 
$$\log \prod_{t=1}^T  \rv(1) = \log \left(  \frac{\mbox{price of  stock at time $T+1$}} {\mbox{initial price of  stock}} \right)   $$
As $T$ becomes large, any sublinear regret guarantee (e.g., the $O(\sqrt{T})$ regret guarantee  achieved using online gradient descent) achieves an average regret that approaches zero. In this context, this implies that the log-wealth gain achieved (in average over $T$ rounds) is as good as that of the first stock.
%Thus, any regret guarantee that asymptotically grows sublinearly with $T$, gives that on average, the log-wealth gain of the investor is as good as that of the first stock. 
Since $\x^\star$ can be taken to be any vector, sublinear regret guarantees average log-wealth growth as good as any stock!

However, $\x^\star$ can be significantly better, as shown in the following example. Consider a market of two stocks that fluctuate wildly. The first stock increases by $100\%$ every even day and returns to its original price the following (odd)  day. The second stock does exactly the opposite: decreases by $50\%$ on even days and rises back on odd days. Formally, we have
$$ \rv(1)  = (2 \ ,  \ \frac{1}{2} \  , \   2 \ , \ \frac{1}{2} , ... ) $$
$$ \rv(2)  = (\frac{1}{2} \ , \  2 \ ,  \ \frac{1}{2} \  , \   2 \  , ... ) $$
Clearly, any investment in either of the stocks will not gain in the long run. However, the portfolio $\x^\star = (0.5,0.5)$ increases wealth by a factor of $\rv^\top \x^\star = (\frac{1}{2})^2 + 1 = 1.25$ daily! Such a mixed distribution is called a fixed rebalanced portfolio, as it needs to rebalance the proportion of total capital invested in each stock at each iteration to maintain this fixed distribution strategy.
 
Thus, vanishing average regret guarantees long-run growth as the best constant rebalanced portfolio in hindsight. Such a portfolio strategy is called {\it universal}. We have seen that the \ogd algorithm gives  essentially a universal algorithm with regret $O(\sqrt{T})$. Can we get better regret guarantees?

\section{Exp-Concave Functions}

For convenience, we return to considering losses of convex functions, rather than gains of concave functions as in the application for portfolio selection. The two problems are equivalent:  we simply replace the maximization of  the concave $f(\x)= \log(\rv^\top \x)$  with the minimization of the convex $f(\x)=-\log(\rv^\top \x)$.

In the previous chapter we have seen that the OGD algorithm with carefully chosen step sizes can deliver logarithmic regret for strongly convex functions. However, the loss  function for the OCO setting of portfolio selection, $f_t(\x) = -\log (\rv^\top \x),$ is not strongly convex. Instead, the Hessian of this function is given by 
$$ \nabla^2 f_t(\x) =  \frac{\rv \rv^\top }{(\rv^\top \x)^2}$$
which is a rank one matrix.  Recall that  the Hessian of a twice-differentiable strongly convex function is larger than a multiple of identity matrix  and is positive definite and in particular has full rank. Thus, the loss function above is quite far from being strongly convex. 

However, an important observation is that this Hessian is large in the direction of the gradient. This property is called exp-concavity. We proceed to  define this property  rigorously  and  show that it  suffices to attain logarithmic regret. %  The application of portfolio selection falls into this category - its cost functions are exp-concave. 

\begin{definition}
A convex function $f : \reals^n \mapsto \reals$ is defined to be $\alpha$-exp-concave over $\K \subseteq \reals^n$  if the function $g$ is concave, where $g: \K \mapsto \reals $ is defined as
$$   g(\x) = e^{-\alpha f(\x) }$$
\end{definition}

For the following discussion, recall the notation of \S \ref{sec:optdefs}, and in particular our convention over matrices that  $A \succcurlyeq B$  if and only if $A - B $ is positive semidefinite. 
Exp-concavity implies strong-convexity in the direction of the gradient. This reduces to the following property:
\begin{lemma} \label{lem:quadratic_approximation1}
A twice-differentiable function $f : \reals^n \mapsto \reals$ is $\alpha$-exp-concave at $\x$ if and only if 
$$ \nabla^2 f(\x)  \succcurlyeq {\alpha} \nabla f(\x) \nabla f(\x)^\top. $$ 
\end{lemma}

The proof of this lemma is given as a guided exercise at the end of this chapter. We prove a slightly stronger lemma below. 

\begin{lemma} \label{lem:quadratic_approximation2}
Let $f :\K \rightarrow \reals$ be an $\alpha$-exp-concave  function, and $D,G$ denote the diameter of $\K$ and a bound on the (sub)gradients of $f$ respectively.  The
following holds for all $\gamma \leq \frac{1}{2}\min\{\frac{1}{GD},\alpha\}$ and all $\x,\y \in \K$:
$$  f(\bx) \geq f(\by) + \nabla f(\by)^\top (\bx-\by) +
\frac{\gamma}{2} (\bx - \by)^\top \nabla f(\by) \nabla
f(\by)^\top(\bx-\by).$$
\end{lemma}
\begin{proof}
The composition of a concave and non-decreasing function with another concave function is concave \endnote{see exercises.}. Therefore, since $2\gamma \leq \alpha$, the composition of $g(x)= x^{2 \gamma/\alpha} $ with $f(\x) = \exp(-\alpha f(\bx))$ is concave. It follows %from Lemma \ref{lem:quadratic_approximation1} 
that the function $h(\bx) \equaltri \exp(-2\gamma f(\bx))$ is 
concave. Then by the concavity of $h(\bx)$,
\begin{equation*}
h(\bx) \leq h(\by) + \nabla h(\by)^\top(\bx - \by)
\end{equation*}
Plugging in $\nabla h(\by) = -2\gamma \exp(-2\gamma f(\by)) \nabla
f(\by)$ gives
\begin{equation*}
\exp(-2\gamma f(\bx))  \leq \exp(-2\gamma f(\by)) [1 - 2\gamma \nabla f(\by)^\top (\bx - \by)].
\end{equation*}
Simplifying gives
$$f(\bx) \geq f(\by) - \frac{1}{2\gamma} \log \left( 1-2\gamma \nabla f(\by)^\top
(\bx-\by) \right).$$ 
Next, note that $|2\gamma \nabla f(\by)^\top
(\bx-\by)|\ \leq\ 2\gamma GD \leq 1 $ and that using the Taylor approximation, for
$z \geq -1$, it holds that
 $-\log(1-z) \geq z+\frac{1}{4}{z^2}$. Applying the inequality for $z = 2\gamma \nabla f(\by)^\top
(\bx-\by)$ implies the lemma.
\end{proof}

\section{Exponentially Weighted Online Convex Optimization}

Before diving into efficient second order methods, we first describe a simple algorithm based on the multiplicative updates method which gives logarithmic regret for exp-concave losses. Algorithm \eqref{alg:ewoo} below, called EWOO, is a close relative to the Hedge Algorithm \eqref{alg:Hedge}.  Its regret guarantee is robust: it does not include a Lipschitz constant or a diameter bound. In addition, it is particularly simple to describe and analyze. 

The downside of EWOO is its running time. A naive
implementation would run in exponential time of the dimension. It is possible to given a randomized polynomial time  implementation based on random sampling techniques, where the polynomial depends both on the dimension as well as the number of iterations, see bibliographic section for more details.

\begin{algorithm}[ht]
	\caption{\label{alg:ewoo} Exponentially Weighted Online Optimizer }
	\begin{algorithmic}[1]
	\State Input: convex set $\K$, $T$, parameter $\alpha > 0$.
	\For {$t=1$ to $T$}
	\State  Let $w_t(\bx) = e^{-\alpha\tsum_{\tau=1}^{t-1} f_\tau(\bx)} $.
    \State Play $\bx_t$ given by
    $$ \x_t = \frac{\int_\K \bx \ w_t(\bx) d \bx}{\int_\K w_t(\bx)d \bx} . $$
	\EndFor
	\end{algorithmic}
\end{algorithm}

In the analysis below, it can be observed that choosing $\bx_t$ at random with density proportional to $w_t(\bx)$, instead of computing the entire integral, also guarantees our regret bounds on the expectation. This is the basis for the 
polynomial time implementation.
We proceed to give the logarithmic regret bounds.

\begin{theorem} \label{thm:exp}
%Assume that the functions $f_t$ are such that $\exp(-\alpha
%f_t(\bx))$ is concave. Then the \ewoo algorithm achieves the
%following guarantee, for all $T\geq 1$.
$$ \regret_T(EWOO) \ \leq \ \frac{d}{\alpha} \log T + \frac{2}{\alpha} .$$
\end{theorem}

\begin{proof}

Let $h_t(\bx) = e^{- \alpha f_t(\bx)}$.  Since $f_t$ is $\alpha$-exp-concave, we have that $h_t$ is concave and thus
$$h_t(\bx_t) \geq \frac{\int_\K h_t(\bx) \prod_{\tau=1}^{t-1} h_\tau(\bx) ~d \bx}{\int_\K
 \prod_{\tau=1}^{t-1} h_\tau(\bx)  ~ d \bx}.$$
Hence, we have by telescoping product,
\begin{equation} \label{eqn:telescope}
\prod_{\tau=1}^t h_\tau(\bx_\tau) \geq \frac{\int_\K
\prod_{\tau=1}^t h_\tau(\bx) ~ d \bx}{\int_\K 1 ~ d \bx} =
\frac{\int_\K \prod_{\tau=1}^t h_\tau(\bx) ~ d \bx}{\vol(\K)}
\end{equation}

By definition of $\bx^\star$ we have $\bx^\star \in \arg\max_{\bx \in \K}
\prod_{t=1}^T h_t(\bx)$. Denote by  $S_\delta \subset \K$ the translated Minkowski set given by
$$S_\delta  = (1-\delta) \x^\star +  \K_{1-\delta} =  \left\{  \bx = (1-\delta)  \bx^\star + \delta \by \ , \ \by \in
\K \right\}.$$ 
By concavity of $h_t$ and the fact that $h_t$ is
non-negative, we have that,
$$\forall \bx \in S_\delta \ . \  \quad h_t(\bx) \geq (1-\delta) h_t(\bx^\star).$$
Hence,
$$\forall \bx \in S_\delta \quad \prod_{\tau=1}^T h_\tau(\bx)
\geq \left( 1  - \delta\right)^T \prod_{\tau=1}^T h_\tau(\bx^\star)$$
%\geq e^{-\delta T} \prod_{\tau=1}^T h_\tau(\bx^\star)$$ 
Finally, since
$S_\delta = (1-\delta)  \bx^\star + \delta  \K$ is simply a rescaling of $\K$ followed by a translation, and we are in $d$
dimensions, $\vol(S_\delta) = \vol(\K) \times \delta^d$. Putting this together with equation \eqref{eqn:telescope}, we have
$$\prod_{\tau=1}^T h_\tau(\bx_\tau) \geq \frac{\vol{(S_\delta)}}{\vol(\K)} (1-\delta)^T 
\prod_{\tau=1}^T h_\tau(\bx^\star) \geq
{\delta^d}(1-\delta)^T\prod_{\tau=1}^T h_\tau(\bx^\star).$$
We can now simplify by taking logarithms and changing sides, 
\begin{eqnarray*}
\regret_T(EWOO) & = \sum_t f_t(\x_t) - f_t(\x^\star) \\
& = \frac{1}{\alpha} \log \frac { \prod_{\tau=1}^T h_\tau(\bx^\star)} {\prod_{\tau=1}^T h_\tau(\bx_\tau) } \\
& \leq \frac{1}{\alpha} \left( d \log \frac{1}{\delta} + T \log \frac{1}{1-\delta} \right) \leq \frac{d}{\alpha} \log T + \frac{2}{\alpha} ,
\end{eqnarray*}
where the last step is by choosing $\delta = \frac{1}{T}$. 
\end{proof}

\def\ons{{online Newton step}\xspace}

\section{The Online Newton Step Algorithm} \label{section:ons}

Thus far we have only considered first order methods for regret minimization. In this section we consider a quasi-Newton approach,  i.e., an online convex optimization algorithm that approximates the second derivative, or Hessian in more than one dimension. However, strictly speaking, the algorithm we analyze is also first order, in the sense that it only uses gradient information.

The algorithm we introduce and analyze, called \ons, is detailed in Algorithm \ref{alg:ons}. 
At each iteration, this algorithm chooses a vector that is the projection of the sum of the vector chosen at the previous iteration and an additional vector.
Whereas for the \ogd algorithm this added
vector was the gradient of the previous cost function, for \ons this vector is
different: it is reminiscent to the direction in which the Newton-Raphson
method would proceed if it were an offline optimization problem for the
previous cost function. The Newton-Raphson algorithm would move in the
direction of the vector which is the inverse Hessian times  the
gradient. In \ons, this direction is $A_t^{-1} \nabla_t$, where the matrix
$A_t$ is related to the Hessian as will be shown in the analysis.

Since  adding a multiple of the Newton vector $A_t^{-1} \nabla_t$ to the current
vector may result in a point outside the convex set, an additional projection step is required to obtain $\x_t$, the decision at time $t$. This projection is 
different than the standard Euclidean projection used by \ogd in Section
\ref{section:ogd}. It is the projection according to the norm
defined by the matrix $A_t$, rather than the Euclidean norm.

\begin{algorithm}[ht]
		\caption{\label{alg:ons} \ons }
		\begin{algorithmic}[1]
			\State Input: convex set $\K$, $T$, $\x_1 \in \mathcal{K} \subseteq \reals^n$, parameters $\gamma,\varepsilon > 0$, $A_ 0 = \varepsilon \bI_n$ % step sizes $\{ \eta_t \}$
%			\State Let $\gamma = \frac{1}{2} \min\{\frac{1}{4GD},\alpha\} $, $\varepsilon = \frac{1}{\gamma^2 D^2}$, $A_0 = \epsilon \bI_n$
			\For {$t=1$ to $T$}
%			\State Set  $\nabla_\tau = \nabla f_\tau(\bx_\tau)$, $\bA_{t} = \sum_{i=1}^{t} \nabla_{i} \nabla_i^\top  + \varepsilon \bI_n$
			\State Play $\x_t$ and observe cost $f_t(\x_t)$. 
			\State Rank-1 update: $\bA_t = \bA_{t-1} + \nabla_t \nabla_t^\top$
			\State Newton step and generalized projection:
			$$ \y_{t+1} =  \bx_{t} - \frac{1}{\gamma}  \bA_{t}^{-1} \nabla_{t}  $$
			$$ \bx_{t+1} = \proj_\K^{\bA_t} (\y_{t+1}) = \argmin_{\x \in \K} \left\{ \|\y_{t+1} - \x\|^2_{\bA_t} \right\}$$
			\EndFor
		\end{algorithmic}
\end{algorithm}

The advantage of the \ons algorithm is its logarithmic regret guarantee for exp-concave functions, as defined in the previous section. 
The following theorem bounds the regret of \ons.

\begin{theorem}
\label{thm:onsregret} %The regret of the ONS algorithm us bounded by:
Algorithm \ref{alg:ons} with parameters  $\gamma = \frac{1}{2}\min\{\frac{1}{GD},\alpha\}$,  $\varepsilon = \frac{1}{\gamma^2 D^2}$ and $T \geq 4$ guarantees 
$$\regret_T \ \leq\  2 \left(\frac{1}{\alpha} + GD\right) n
\log T.$$
\end{theorem}

As a first step, we prove the following lemma. 

\begin{lemma} \label{lemma:onsbound}
The regret of \ons is bounded by
\begin{equation*}
\regret_T(\text{ONS})\ \leq\ \left(\frac{1}{\alpha}
+ GD\right) \left(\sum_{t=1}^T \nabla_t^\top \bA_t^{-1} \nabla_t + 1\right) 
\end{equation*}
\end{lemma}
\begin{proof}
Let $\bx^\star \in \arg\min_{\bx \in \K} \sum_{t=1}^T f_t(\bx)$ be
the best decision in hindsight. By Lemma
\ref{lem:quadratic_approximation2}, we have for $\gamma = \frac{1}{2}\min\{\frac{1}{GD},\alpha\}$,
\begin{equation*}
f_t(\bx_t ) - f_t(\bx^\star)\ \leq\ R_t  ,  
\end{equation*}
where we define
$$ R_t  \equaltri\ \nabla_t^\top
(\bx_t - \bx^\star) - \frac{\gamma}{2} (\bx^\star - \bx_t)^\top \nabla_t
\nabla_t^\top (\bx^\star - \bx_t) . $$
According to the update rule of the
algorithm $\bx_{t+1} = \proj_{\K}^{\bA_t}(\by_{t+1})$. Now, by the
definition of $\by_{t+1}$:
\begin{equation} \label{eq:update-rule}
\by_{t+1} - \bx^\star = \bx_{t} - \bx^\star - \frac{1}{\gamma} \bA_t^{-1}
\nabla_t, \text{ and}
\end{equation}
\begin{equation} \label{eq:A_t-multiply}
\bA_t (\by_{t+1} - \bx^\star) = \bA_t(\bx_t - \bx^\star) - \frac{1}{\gamma}
\nabla_t.
\end{equation}
Multiplying the transpose of \eqref{eq:update-rule} by
\eqref{eq:A_t-multiply} we get
\begin{gather}
(\by_{t+1} - \bx^\star)^\top \bA_t(\by_{t+1} - \bx^\star) = \notag \\
(\bx_t\! -\! \bx^\star)^\top \bA_t(\bx_t\! -\! \bx^\star) -
\frac{2}{\gamma} \nabla_t^\top (\bx_t\! -\! \bx^\star) +
\frac{1}{\gamma^2} \nabla_t^\top \bA_t^{-1} \nabla_t.
\label{eq:multiplied}
\end{gather}
Since $\bx_{t+1}$ is the projection of $\by_{t+1}$ in the norm induced by
$\bA_t$, we have by the Pythagorean theorem (see \S \ref{sec:projections})
\begin{eqnarray*}
 (\by_{t+1} - \bx^\star)^\top \bA_t(\by_{t+1} - \bx^\star)  & = \| \by_{t+1} - \bx^\star \|_{\bA_t}^2 \\
& \ge  \| \bx_{t+1} - \bx^\star \|_{\bA_t}^2   \\
&  = (\bx_{t+1} - \bx^\star)^\top \bA_t(\bx_{t+1} - \bx^\star ).
\end{eqnarray*}
This inequality is the reason for using generalized projections as
opposed to standard projections, which were used in the analysis
of \ogd (see \S \ref{section:ogd} Equation
\eqref{eqn:ogdtriangle}). This fact together with
\eqref{eq:multiplied} gives
\begin{align*}
\nabla_t^\top (\bx_t \! -\! \bx^\star) &\leq \ \frac{1}{2\gamma}
\nabla_t^\top \bA_t^{-1} \nabla_t + \frac{\gamma}{2} (\bx_t\! -\!
\bx^\star)^\top \bA_t (\bx_t\! -\! \bx^\star) \\
&  - \frac{\gamma}{2}
(\bx_{t+1} - \bx^\star)^\top \bA_t(\bx_{t+1} - \bx^\star).
\end{align*}
Now, summing up over $t=1$ to $T$ we get that
\begin{align*}
&\sum_{t=1}^T \nabla_t^\top (\bx_t - \bx^\star)
 \leq \frac{1}{2\gamma} \sum_{t=1}^T \nabla_t^\top \bA_t^{-1} \nabla_t +
\frac{\gamma}{2}  (\bx_{1} - \bx^\star)^\top \bA_1 (\bx_{1} - \bx^\star) \\
&\quad + \frac{\gamma}{2} \sum_{t=2}^{T} (\bx_t - \bx^\star)^\top
(\bA_t - \bA_{t-1}) (\bx_t - \bx^\star) \\
& \quad - \frac{\gamma}{2} (\bx_{T+1} - \bx^\star)^\top \bA_T (\bx_{T+1} - \bx^\star) \\
&\leq \frac{1}{2\gamma} \sum_{t=1}^T \nabla_t^\top \bA_t^{-1}
\nabla_t+ \frac{\gamma}{2} \sum_{t=1}^{T} (\bx_t\! -\! \bx^\star)^\top
\nabla_t \nabla_t^\top (\bx_t\! -\! \bx^\star) \\
& + \frac{\gamma}{2}
(\bx_{1} - \bx^\star)^\top (\bA_1 - \nabla_1\nabla_1^\top) (\bx_{1} -
\bx^\star).
\end{align*}
In the last inequality we use the fact that $A_t - A_{t-1} =
\nabla_t \nabla_t^\top$, and the fact that the matrix $A_T$ is PSD and hence the last term before the inequality is negative. %The matrix $A_t$ was defined in the first place to give this equation, so now when looking at $\sum_t R_t$ the terms of the form $(\bx_t - \bx^\star)^\top (A_t - A_{t-1}) (\bx_t
% - \bx^\star)$ would cancel out: 
Thus,
\begin{equation*}
\sum_{t=1}^T R_t\ \leq\ \frac{1}{2 \gamma } \sum_{t=1}^T
\nabla_t^\top \bA_t^{-1} \nabla_t + \frac{\gamma}{2} (\bx_{1} -
\bx^\star)^\top (\bA_1 - \nabla_1\nabla_1^\top) (\bx_{1} - \bx^\star).
\end{equation*}

Using the algorithm parameters  $\bA_1 - \nabla_1 \nabla_1^\top = \varepsilon
\bI_n$ , $\varepsilon = \frac{1}{\gamma^2 D^2}$ and our notation for the diameter $\|\bx_1 - \bx^\star\|^2 \leq D^2$ we have
\begin{eqnarray*}
\regret_T(\text{\em ONS})\ & \leq\ &  \sum_{t=1}^T R_t\
\leq\ \frac{1}{2 \gamma } \sum_{t=1}^T \nabla_t^\top \bA_t^{-1}
\nabla_t + \frac{ \gamma }{2}  {D^2}{\varepsilon} \\
& \leq & \frac{1}{2 \gamma } \sum_{t=1}^T \nabla_t^\top \bA_t^{-1}
\nabla_t + \frac{1}{2 \gamma}.
\end{eqnarray*}
Since $\gamma = \frac{1}{2}\min\{\frac{1}{GD},\alpha\}$, we have
$\frac{1}{\gamma} \leq 2( \frac{1}{\alpha} + GD)$. This gives the
lemma.
\end{proof}

We can now prove Theorem \ref{thm:onsregret}.

\begin{proof}[Proof of Theorem \ref{thm:onsregret}]
First we show that the term $\sum_{t=1}^T \nabla_t^\top \bA_t^{-1}
\nabla_t$ is upper bounded by a telescoping sum. Notice that
$$ \nabla_t^\top \bA_t^{-1} \nabla_t = A_t^{-1} \bullet \nabla_t \nabla_t^\top = A_t^{-1} \bullet (A_{t} - A_{t-1})$$
where for matrices $A,B \in \reals^{n \times n}$ we denote by $A
\bullet B = \sum_{i = 1}^n \sum_{j=1}^nA_{ij} B_{ij} = \trace(AB^\top) $, which is equivalent to the inner product of
these matrices as vectors in $\reals^{n^2}$.

For real numbers $a,b \in \reals_+$, the first order Taylor expansion of the logarithm of $b$ at $a$ implies  $a^{-1} (a-b) \leq \log \frac{a}{b}$. An
analogous fact holds for positive semidefinite matrices, i.e., $A^{-1} \bullet (A-B)
\leq \log \frac{|A|}{|B|}$, where $|A|$ denotes the determinant of
the matrix $A$ (this is proved in Lemma \ref{lem:logdet}). Using
this fact  we have 
\begin{eqnarray*}
\sum_{t=1}^T \nabla_t^\top \bA_t^{-1} \nabla_t & = & \sum_{t=1}^T
A_t^{-1} \bullet \nabla_t \nabla_t^\top \\
& = & \sum_{t=1}^T A_t^{-1} \bullet (A_{t} - A_{t-1}) \\
& \leq &  \sum_{t=1}^T \log \frac{ |A_t|} {|A_{t-1}|} = \log
\frac{|A_T|}{|A_0|}.
\end{eqnarray*}

Since $A_T = \sum_{t=1}^T \nabla_t\nabla_t^\top + \varepsilon I_n$
and $\|\nabla_t\| \leq G$, the largest eigenvalue of $A_T$ is at
most $T G^2 + \varepsilon$. Hence the determinant of $A_T$ can be
bounded by $|A_T | \leq (T G^2 + \varepsilon)^n$. Hence recalling
that $\varepsilon = \frac{1}{\gamma^2D^2}$ and $\gamma =
\frac{1}{2}\min\{\frac{1}{GD},\alpha\}$, for  $T > 4$,

\begin{eqnarray*}
\sum_{t=1}^T \nabla_t^\top \bA_t^{-1} \nabla_t\ & \leq\   \log
\left( \frac{T G^2 + \varepsilon}{\varepsilon }\right)^n \leq n
\log (TG^2 \gamma^2 D^2 + 1) \leq n \log T.
\end{eqnarray*}
Plugging into Lemma \ref{lemma:onsbound} we obtain
$$\regret_T(\text{ONS})\ \leq\  \left(\frac{1}{\alpha} +  GD\right) (n \log T + 1),$$
which implies the theorem for $n > 1, \  T \geq 4$.

\end{proof}

It remains to prove the technical lemma for positive semidefinite (PSD)  matrices
used above.
\begin{lemma} \label{lem:logdet}
Let $A \succcurlyeq B \succ 0$ be positive definite matrices. Then
$$ A^{-1} \bullet (A - B) \ \leq\ \log \frac{|A|}{|B|}$$
\end{lemma}
\begin{proof} For any positive definite matrix $C$, denote by
$\lambda_1(C), \ldots, \lambda_n(C)$ its eigenvalues (which are positive).
\begin{align*}
& A^{-1} \bullet (A - B) \   =\  \trace(A^{-1} (A - B)) \\
& =  \trace(A^{-1/2} (A - B) A^{-1/2}) & \trace(XY) = \trace(YX) \\
& =  \trace(I - A^{-1/2} B  A^{-1/2}) \\
& =  \sum_{i=1}^n \left[ 1 - \lambda_i( A^{-1/2} B  A^{-1/2}) \right] &  \trace(C) = \sum_{i=1}^n \lambda_i(C)  \\
& \leq  - \sum_{i=1}^n \log \left[  \lambda_i( A^{-1/2} B
A^{-1/2}) \right] &  1-x \leq -\log(x)  \\
& =  - \log \left[ \prod_{i=1}^n  \lambda_i( A^{-1/2} B
A^{-1/2}) \right] \\
& =  - \log  |  A^{-1/2} B A^{-1/2}|  = \log \frac{|A|}{|B|} &
 |C| = \prod_{i=1}^n \lambda_i(C)
\end{align*}
In the last equality we use the facts $|AB| = |A||B|$ and
$|A^{-1}| = \frac{1}{|A|}$ for positive definite matrices (see exercises).
\end{proof}

\paragraph*{Implementation and running time.}

The \ons algorithm requires $O(n^2)$ space to store the matrix $A_t$. Every
iteration requires the computation of the matrix $A_{t}^{-1}$, the current
gradient, a matrix-vector product, and possibly a projection onto the underlying
convex set $\K$.

A na\"{\i}ve implementation would require computing the inverse of
the matrix $A_t$ on every iteration. However, in the case that $A_t$ is
invertible, the matrix inversion lemma (see bibliography) states that
for invertible matrix $A$ and vector $\bx$,
$$ (A + \bx \bx^\top)^{-1} = A^{-1} - \frac{A^{-1} \bx \bx^\top A^{-1}}{1 + \bx^\top A^{-1} \bx}.$$
Thus, given $A_{t-1}^{-1}$ and $\nabla_t$ one can compute $A_t^{-1}$ in time
$O(n^2)$ using only matrix-vector and vector-vector products.

The \ons algorithm also needs to make projections onto $\K$, but
of a slightly different nature than \ogd and other online convex
optimization algorithms. The required projection, denoted by
$\proj_\K^{A_t}$, is in the vector norm induced by the matrix
$A_t$, viz. $\|\bx\|_{A_t} = \sqrt{\bx^\top A_t \bx}$. It is
equivalent to finding the point $\bx \in \K$ which minimizes
$(\bx - \by)^\top A_t(\bx - \by)$ where $\y$ is the point we are
projecting. This is a convex program which can be solved up to any
degree of accuracy in polynomial time.

Modulo the computation of generalized projections, the \ons algorithm can be
implemented in time and space $O(n^2)$. In addition, the only information
required is the gradient at each step (and the exp-concavity constant $\alpha$ of the
loss functions).

\newpage
\section{Bibliographic Remarks}

The Geometric Brownian Motion model for stock prices was suggested and studied as early as 1900 in the PhD thesis of  Louis Bachelier \citep{bachelier}, see also  \citep{osborne}, and used in the Nobel Prize winning work of Black and Scholes on options pricing \citep{black-scholes}.   
In a strong deviation from standard financial theory, Thomas Cover  \citep{cover} put forth the universal portfolio model, whose algorithmic theory we have historically sketched in chapter \ref{chap:intro}. The EWOO algorithm was essentially given in Cover's paper for the application of portfolio selection and logarithmic loss functions, and extended to exp-concave loss functions in \citep{HazanKKA06}. The randomized extension of Cover's algorithm that runs in polynomial running time is due to \citet{KalaiVempalaPortfolios}, and it naturally extends to EWOO. 

Some bridges between classical portfolio theory and the universal model appear in \citep{AbernethyStoc12}.  Options pricing and its relation to regret minimization was recently also explored in the work of \citep{DKM-options}.

Exp-concave functions have been considered in the context of prediction in \citep{kivinen-warmuth}, see also \citep{CesaBianchiLugosi06book} (chapter 3.3 and bibliography).  A more general condition than exp-concavity called mixability was used by \citet{vovk1990aggregating} to give a general multiplicative update algorithm, see also \citep{foster2018logistic}. For a thorough discussion of various conditions that allow logarithmic regret in online learning see \citep{van2015fast}.

For  the square-loss,  \citep{Azoury} gave a specially tailored and near-optimal prediction algorithm. Logarithmic regret algorithms for online convex optimization and the Online Newton Step algorithm were presented in \citep{HAK07}.

Logarithmic regret algorithms were used to derive $\tilde{O}(\frac{1}{\eps})$-convergent algorithms for non-smooth convex optimization in the context of training support vector machines in \citep{Shalev-ShwartzSSC11}. Building upon these results, tight convergence rates of SGD for strongly convex and non-smooth functions were obtained in \citep{hazan:beyond}.

The Sherman-Morrison formula, a.k.a. the matrix inversion lemma, gives  the form of the inverse of a matrix after a rank-1 update, see \citep{pseudoinverse}.

\newpage
\begin{exercises}
%\section{Exercises}

%\begin{enumerate}
%\item
\exer{For this question, assume all functions are twice differentiable. Prove that exp-concave functions are a larger class than strongly convex and Lipschitz functions. That is, prove that a strongly convex function over a bounded domain that is also $G$-Lipschitz is also exp-concave. Show that the converse is not necessarily true.  }

%\item
\exer{Prove that a twice-differentiable function $f$ is $\alpha$-exp-concave over $\K$ if and only if for all $\x \in \K  $,
$$  \nabla^2 f(\x) \succcurlyeq {\alpha} \nabla f(\x) \nabla f(\x)^\top . $$
Hint: consider the Hessian of the function $e^{-\alpha f(\x)}$, and use the fact that the Hessian of a convex function is always positive semidefinite.  }

%\item
\exer{Write pseudo-code for a portfolio selection algorithm based on online gradient descent. That is, given a set of return vectors, spell out the exact constants and updates based upon the gradients of the reward functions. Derive the regret bound based on Theorem \ref{thm:gradient}. You may assume that the multiplicative change in price for any single asset is bounded, and use this quantity in your regret bound. 
\\
Do the same (pseudo-code and regret bound) for the Online Newton Step algorithm applied to portfolio selection. \\
\\
Note: you are not required to give pseudo-code for projections onto the simplex. }

\exer{ %\item
Download stock prices from your favorite online finance website over a period of at least three years. Create a dataset for testing portfolio selection algorithms by creating price-return vectors. Implement the OGD and ONS algorithms and benchmark them on your data. 
}

\exer{ %\item
Prove that for positive definite matrices, $A,B \succ 0$, the following hold: 
$|AB| = |A||B|$ and
$|A^{-1}| = \frac{1}{|A|}$, where $|A|$ denotes the determinant of $A$. 
}

\exer{ %\item
Let $h(x): \reals \mapsto \reals $ be concave and non-decreasing, and let $g(\x):\K \mapsto \reals $ be concave. Prove that the function 
 $f(\x) = h(g(\x) ) $ is concave. 
}
\end{exercises}
%\end{enumerate}

%!TEX root = OCObook.tex

\chapter{Regularization} \label{chap:regularization}

In the previous chapters we have explored algorithms for OCO that are motivated by convex optimization. 
However, unlike convex optimization, the OCO framework optimizes the Regret performance metric. This distinction motivates a family of algorithms, called ``Regularized Follow The Leader'' (RFTL), which we introduce in this chapter.

In an OCO setting of regret minimization, the most straightforward approach  for the online player is to use at any time the optimal decision (i.e., point in the convex set) in hindsight. Formally, let
$$\x_{t+1} = \argmin_{\x \in \K} \sum_{\tau=1}^{t} f_\tau(\x).$$
This flavor of strategy is known as ``fictitious play'' in economics, and has been named ``Follow the Leader'' (FTL) in machine learning. It is not hard to see that this simple strategy fails miserably in a worst-case sense. That is, this strategy's regret can be linear in the number of iterations, as the following example shows: Consider $\K = [-1,1]$, let $f_1(x) = \frac{1}{2} x $, and let $f_\tau$ for $\tau=2 , \ldots , T$ alternate between $- x $ or $x $. Thus, 
$$ \sum_{\tau=1}^t f_\tau(x)  = \mycases{ \frac{1}{2} x } {t \mbox{  is odd} } {-\frac{1}{2} x } {\text{otherwise}}  $$
The FTL strategy will keep shifting between $x_t = -1$ and $x_t = 1$, always making the wrong choice.

The intuitive FTL strategy fails in the example above because it is unstable. Can we modify the FTL strategy such that it won't change decisions often,  thereby causing it to attain low regret? 

This question motivates the need for a general means of stabilizing the FTL method. Such a means is referred to as ``regularization''.

\section{Regularization Functions}
In this chapter we consider  regularization functions, denoted $R : \K \mapsto \reals $, which are strongly convex and smooth (recall definitions in \S \ref{sec:optdefs}). 

Although it is not strictly necessary, we assume that the regularization functions in this chapter are twice differentiable over $\K$  and, for all points $\x \in \text{int}(\K)$ in the interior of the decision set, have a Hessian $\nabla^2 R(\x)$ that is, by the strong convexity of $R$, positive definite.

We denote the diameter of the set $\K$ relative to the function $R$ as 
$$ D_R = \sqrt{ \max_{\x,\y \in \K} \{ R(\x) - R(\y) \}} . $$ 

Henceforth we make use of general norms and their dual. The dual norm to a norm $\| \cdot \|$ is given by the following definition:
$$ \| \y \|^* \equaltri \sup_{ \| \x \| \leq 1 }  \left\{ \x^\top \y \right\} . $$
A positive definite matrix $A$ gives rise to the matrix norm $\|\x\|_A = \sqrt{\x^\top A \x}$. 
The  dual norm of a matrix norm is $\|\x\|_A^*=\|\x\|_{A^{-1}}$. 

The generalized Cauchy-Schwarz theorem asserts $  \x^\top  \y  \leq \| \x \| \| \y \|^*$ and in particular for matrix norms, $ \x^\top \y  \leq \|\x\|_A \| \y\|_A^*$ (see exercises). 

In our derivations, we usually consider matrix norms with respect to $\nabla^2R(\x)$, the Hessian of the regularization function $R(\x)$, as well as the inverse Hessian denoted $\nabla^{-2} R(\x)$.
In such cases, we use  the notation 
$$\|\x\|_\y \equaltri \|\x\|_{\nabla^2 {R}(\y)} ,$$
and similarly 
$$\|\x\|_\y^* \equaltri \|\x\|_{\nabla^{-2} {R}(\y)} . $$

A crucial quantity in the analysis of OCO algorithms that use regularization is the remainder term of the Taylor approximation of the regularization function, and especially the remainder term of the first order Taylor approximation. The difference between the value of the regularization function at $\x$ and the value of the first order Taylor approximation is known as the Bregman divergence, given by 
\begin{definition}
	Denote by $B_{R}(\x||\y)$ the
	Bregman divergence with respect to the function ${R}$, defined as
	$$ B_{R}(\x||\y) = {R}(\x) - {R}(\y) - \nabla {R}(\y)^\top  (\x-\y) .  $$
\end{definition}

For twice differentiable functions, Taylor expansion and the mean-value theorem assert that the Bregman divergence is equal to the second derivative at an intermediate point, i.e., (see exercises)  
$$ B_{R}(\x||\y) = \frac{1}{2} \|\x - \y\|_\z^2, $$ 
for some point $\z \in [\x,\y]$, meaning there exists some $\alpha \in [0,1]$ such that $\z = \alpha \x + (1-\alpha) \y$. 
Therefore, the Bregman divergence defines a local norm, which has a dual norm. We shall denote this dual norm by 
$$ \| \cdot \|_{\x,\y}^*  \equaltri \| \cdot \|_\z^*.$$ 
With this notation we have
$$ B_{R}(\x||\y) = \frac{1}{2} \|\x - \y\|_{\x,\y} ^2. $$ 
In online convex optimization, we commonly refer to the Bregman divergence between two consecutive decision points $\x_t$ and $\x_{t+1}$. In such cases, we shorthand notation for the norm  defined by the Bregman divergence with respect to  ${R}$ on the intermediate point in $[\x_t,\x_{t+1}]$ as $\| \cdot \|_t \equaltri \| \cdot \|_{\x_t,\x_{t+1}} $. The latter norm is called the local norm at iteration $t$. With this notation, we have $B_{R}(\x_t||\x_{t+1}) = \frac{1}{2} \|\x_t - \x_{t+1}\|_t^2 $. 

Finally, we consider below generalized projections that use the Bregman divergence as a distance instead of a norm. Formally, the projection of a point $\y$ according to the Bregman divergence with respect to  function $R$ is given by
$$\argmin_{\x \in \K} B_{R}(\x||\y) . $$

\section{The RFTL Algorithm and its Analysis}
\sectionmark{The RFTL Algorithm}

Recall the caveat with straightforward use of follow-the-leader: as in the bad example we have considered, the predictions of FTL may vary wildly from one iteration to the next.
This motivates the modification of the basic FTL strategy in order to stabilize the prediction. By adding a  regularization term, we obtain the RFTL (Regularized Follow the Leader) algorithm.

We proceed to formally describe the RFTL algorithmic template and analyze it. The analysis gives asymptotically optimal regret bounds. However,  we do not optimize the constants in the regret bounds in order to  improve clarity of presentation.

Throughout this chapter, recall   the notation of $\nabla_t$ to denote the gradient of the current cost function at the current point, i.e., 
$$ \nabla_t \equaltri \nabla f_t(\x_t) . $$
In the OCO setting, the regret of convex cost functions can be bounded by a linear function via the inequality $f_t(\xv) - f_t(\x^\star) \leq \nabla_t^\top (\xv - \x^\star)$. Thus, the overall regret (recall definition \eqref{eqn:regret-defn}) of an OCO algorithm can be bounded by 
\begin{equation} \label{eqn:rftl-shalom}
\sum_t f_t(\x_t)  - f_t(\x^\star) \leq \sum_t \nabla_t^\top (\x_t - \x^\star).
\end{equation}

\subsection{Meta-algorithm definition}

The generic RFTL meta-algorithm is defined in Algorithm \ref{alg:RFTLmain}. The regularization function ${R}$ is assumed to be strongly convex, smooth, and twice differentiable. 

\begin{algorithm}
	[ht] \caption{Regularized Follow The Leader} \label{alg:RFTLmain} 
	\begin{algorithmic}[1]
		\State Input: $\eta > 0$, regularization function ${R}$, and a bounded, convex and closed set $\K$.
		\State Let $\xv[1]  = \arg\min_{\x \in \K} {\left\{ {R}(\x)\right\} }$.
		\For{$t=1$ to $T$}
		\State Play $\x_t$ and observe cost $f_t(\x_t)$. 
%		\State Observe the loss function $f_t$ and let $\nabla_t = \nabla f_t(\x_t) $.
		\State Update
		\begin{align*}
			\xv[t+1] = \argmin_{\x \in \K} {\left\{\eta\sum_{s=1}^t \nabla_s^\top \x + {R}(\x)\right\}}
		\end{align*}
		\EndFor
	\end{algorithmic}
\end{algorithm}

\subsection{The regret bound} \label{sec:thm1.1}

\begin{theorem} \label{thm:RFTLmain1}
The RFTL  Algorithm \ref{alg:RFTLmain} attains for every $\uv \in \K$ the following bound on the regret:
$$  \regret_T \le    2   \eta  \sum_{t=1}^T \| \nabla_t \|_t^{* 2} + \frac{R(\uv) - R(\x_1)}{\eta }  . $$ %B_{R}(\uv||\x_1)
\end{theorem}
If an upper bound on the local norms is known, i.e., $\| \nabla_t\|_t^* \leq G_R$ for all times $t$, then we  can further optimize over the choice of $\eta$ to obtain
$$ \regret_T \leq  2  D_R G_R \sqrt{ 2T  } .$$

To prove Theorem \ref{thm:RFTLmain1}, we first relate the regret to the ``stability'' in prediction. This is formally captured by the following lemma\endnote{Historically, this lemma is known as the ``FTL-BTL,'' which stands for follow-the-leader vs. be-the-leader. BTL is a hypothetical algorithm that predicts $\x_{t+1}$ at iteration $t$, where $\xv$ is the prediction made by FTL. These terms were coined by Kalai and Vempala \citep{KV-FTL}.}.
\begin{lemma}
	\label{lem:FTL-BTL} Algorithm \ref{alg:RFTLmain}  guarantees the following regret bound
	\begin{eqnarray*}
		\regret_T  \leq  \sum_{t=1}^T \nabla_t^\top
		(\xv-\xv[t+1]) + \frac{1}{\eta} D_R^2 % B_{R}(\uv||\xv[1])
	\end{eqnarray*}
\end{lemma}

\begin{proof}
	For convenience of the derivations, define the functions
	$$g_0(\bx) \equaltri \frac{1}{\eta}R(\bx)  \ , \ g_t(\bx) \equaltri \nabla_t^\top \bx. $$
By equation \eqref{eqn:rftl-shalom}, it suffices to bound $ \sum_{t=1}^T [ g_t(\xv) - g_t (\uv)] $.
As a first step, we prove the following inequality:
	\begin{lemma} \label{prop:ftl-btl}
	For every $\uv\in \K$, 
		$$  \sum_{t=0}^T g_t(\uv)  \geq  \sum_{t=0}^T g_t(\xv[t+1]) . $$
	\end{lemma}
	\begin{proof}
		by induction on $T$:\\
		\noindent {\bf Induction base:} \\
		By definition, we have that $\xv[1] = \argmin _{\x \in \K} \{R(\x)\}$, and thus $g_0(\uv) \ge g_0(\xv[1]) $ for all $\uv$. \\
		{\bf Induction step:}\\
		Assume that for $T$, we have
		\begin{eqnarray*}
			  \sum_{t=0}^{T} g_t(\uv)  \geq \sum_{t=0}^{T} g_t(\xv[t+1] )
		\end{eqnarray*}
		and let us prove the statement for $T+1$. Since $\xv[T+2] = \argmin _{\x \in \K} \{ \sum_{t=0}^{T+1} g_t(\x)\}$ we have:
		\begin{eqnarray*}
			  \sum_{t=0}^{T+1} g_t (\uv) & \geq  &   \sum_{t=0}^{T+1} g_t (\xv[T+2])  \\
			& = &    \sum_{t=0}^{T} g_t (\xv[T+2])  + g_{T+1}(\xv[T+2])  \\
			& \geq &    \sum_{t=0}^{T}   g_t (\xv[t+1])  + g_{T+1}(\xv[T+2])  \\
			& = &  \sum_{t=0}^{T+1}  g_t (\xv[t+1]).
		\end{eqnarray*}
		Where in the third line we used the induction hypothesis for $\uv = \xv[T+2]$. 
	\end{proof}
	We conclude that
	\begin{eqnarray*}
		\sum_{t=1}^T [ g_t(\xv) - g_t (\uv)]  & \leq &  \sum_{t=1}^T [g_t(\xv) - g_t (\xv[t+1])] + \left[  g_0(\uv) - g_0(\xv[1]) \right] \\
		& = & \sum_{t=1}^T g_t(\xv) - g_t (\xv[t+1]) + \frac{1}{\eta} \left[  R(\uv)  - R(\xv[1]) \right]   \\
		& \le & \sum_{t=1}^T g_t(\xv) - g_t (\xv[t+1]) + \frac{1}{\eta} D_R^2 .
	\end{eqnarray*}
	
\end{proof}

%\begin{Cor}
%For any $\uv \in \K$,
%\begin{eqnarray*}
%\sum_{t=1}^T \nabla^\top (\xv-\uv) 		& \leq &  \sum_{t=1}^T \nabla^\top
%(\xv-\xv[t+1]) + \frac{1}{\eta} [B_{R}(\uv,\xv[T+1]) - {R}(\uv,\xv[1]) ]
%\end{eqnarray*}
%\end{Cor}
%\begin{proof}
%This is a simple consequence of the definition of $\xv$ and the Bregman divergence. Notice that
%$$ B_R(\uv,\x) = {R}(\uv) - {R}(\x) - \nabla {R}(\x)(\uv - \xv[1]) $$
%Now by optimality conditions which follow from the definition of $\xv[T+1]$, we have that $\nabla {R}(\xv[T+1]) (\uv - \xv[T+1]) \geq 0$ for any vector $\uv$. Hence,
%$$ B_R(\uv,\x[T+1]) \geq {R}(\uv) - {R}(\x[T+1])  $$
%
%\end{proof}

\begin{proof}[Proof of Theorem \ref{thm:RFTLmain1}]

Recall that ${R}(\x)$ is a convex function and $\K$ is a convex set. Denote:
$$\Phi_t(\x) \equaltri \eta\sum_{s=1}^t \nabla_s^\top \x + {R}(\x) . $$
By the Taylor expansion (with its explicit remainder term via the mean-value theorem) at $\xv[t+1]$, and by the definition of the Bregman divergence, 
%there exists a $\zv \in [\xv[t+1],\xv[t]]$ for which
\begin{eqnarray*}
	\Phi_t(\xv) & = & \Phi_t(\xv[t+1])
	+  (\xv - \xv[t+1])^\top \nabla \Phi_t(\xv[t+1])
	+ B_{\Phi_t}(\x_t||\x_{t+1} ) \\ % \frac{1}{2} \|\xv - \xv[t+1]\|^2_{\zv} \\
	& \geq &
	\Phi_t(\xv[t+1]) + B_{\Phi_t} (\x_t||\x_{t+1} ) \\ %\frac{1}{2}\|\xv - \xv[t+1]\|^2_{\zv}
	& = &
	\Phi_t(\xv[t+1]) + B_{{R}} (\x_t||\x_{t+1} ). %\frac{1}{2}\|\xv - \xv[t+1]\|^2_{\zv}
\end{eqnarray*}
	The inequality holds  since $\xv[t+1]$ is a minimum of
	$\Phi_t$ over $\K$, as in Theorem \ref{thm:optim-conditions}. The last equality holds since the component $\nabla_s^\top \x$ is linear and thus does not affect the Bregman divergence. 
	Thus,
	\begin{eqnarray} \label{eqn:chap5shalom}
		B_{R}(\x_t || \x_{t+1})  & \leq & \,\Phi_t(\xv) - \,\Phi_t(\xv[t+1]) \\
		& = & \,\ (\Phi_{t-1}(\xv) - \Phi_{t-1}(\xv[t+1])) +  \eta \nabla_t^\top (\xv - \xv[t+1]) \notag \\
		& \leq & \,\eta \,\nabla_t^\top (\xv - \xv[t+1]) \quad \mbox{($\x_t$ is the minimizer)} \notag
	\end{eqnarray}
	To proceed, recall the shorthand for the norm induced by the Bregman divergence with respect to  ${R}$ on point $\x_t,\x_{t+1}$ as $\| \cdot \|_t = \| \cdot \|_{\x_t,\x_{t+1}}$. Similarly for the dual local norm $\| \cdot \|^*_t = \| \cdot \|^*_{\x_t,\x_{t+1}}$. With this notation, we have $B_{R}(\x_t||\x_{t+1}) = \frac{1}{2} \|\x_t - \x_{t+1}\|_t^2 $. 
	By the generalized Cauchy-Schwarz inequality,
	\begin{align*}
		\nabla_t^\top (\xv[t]-\xv[t+1]) &\leq \|\nabla_t \|_{t}^* \cdot
		\|\xv - \xv[t+1] \|_{t} & \mbox{ Cauchy-Schwarz} \\
		& = \|\nabla_t \|_{t}^* \cdot
		\sqrt{2 B_{R}(\x_t||\x_{t+1}) } \\
		& \leq \|\nabla_t  \|_{t}^* \cdot \sqrt{2\, \eta\, \nabla_t^\top (\xv -
			\xv[t+1])  }.  & \eqref{eqn:chap5shalom} \nonumber
	\end{align*}
After rearranging we get
	\begin{align*}
		\nabla_t^\top (\xv[t]-\xv[t+1]) &\leq 2\, \eta \, \|\nabla_t  \|^{* 2}_{t}.
	\end{align*}
	Combining this inequality with Lemma \ref{lem:FTL-BTL} we obtain the theorem statement.
\end{proof}

\section{Online Mirror Descent}

In the convex optimization literature, ``Mirror Descent'' refers to a general class of first order methods generalizing gradient descent. Online Mirror descent (OMD) is the online counterpart of this class of methods. This relationship is analogous to the relationship of online gradient descent  to traditional (offline) gradient descent.

OMD is an iterative algorithm that computes the current decision using a simple  gradient update rule and the previous decision, much like OGD. The generality of the method stems from the update being carried out in a ``dual'' space, where the duality notion is defined by the choice of regularization: the gradient of the regularization function defines a mapping from $\reals^n$ onto itself, which is a vector field. The gradient updates are then carried out in this vector field.

For the RFTL algorithm the intuition was straightforward---the regularization was used to ensure stability of the decision. For OMD, regularization has an additional purpose: regularization transforms the space in which gradient updates are performed. This transformation enables better bounds in terms of the geometry of the space.

The OMD algorithm comes in two flavors:  an agile and a lazy version. The lazy version keeps track of a point in Euclidean space and projects onto the convex decision set $\K$ only at decision time. In contrast, the agile version maintains a feasible point at all times, much like OGD.

\begin{algorithm}
	[H] \caption{Online Mirror Descent} \label{alg:flpl}
	\begin{algorithmic}
		[1] \State Input: parameter $\eta > 0$, regularization function ${R}(\x)$.
		\State Let $\yv[1]$ be such that $\nabla {R}(\yv[1]) = \bzero$ and 	$\xv[1] = \arg\min_{\x \in \K} B_{R}(\x||\yv[1])$.
		\For{$t=1$ to $T$}
				\State Play $\xv[t]$.
				\State Observe the loss function $f_t$ and let $\nabla_t = \nabla f_t(\x_t) $.
		\State Update $\y_t$ according to the rule:
		\begin{align*}
			&\text{[Lazy version]} 
			&\nabla {R}(\yv[t+1]) = \nabla {R}(\yv[t]) - \eta\, \nabla_{t}\\
			&\text{[Agile version]}
			&\nabla {R}(\yv[t+1]) = \nabla {R}(\xv[t]) - \eta\, \nabla_{t}
		\end{align*}
Project according to $B_{R}$:
		$$\xv[t+1] = \argmin_{\x \in \K} B_{R}(\x||\yv[t+1])$$
		\EndFor
	\end{algorithmic}
\end{algorithm}

Both versions can be analyzed to give roughly the same regret bounds as the RFTL algorithm. In light of what we will see next, this is not surprising: for linear cost functions, the RFTL and lazy-OMD algorithms are equivalent! 

Thus, we get regret bounds for free for the lazy version. The agile version can be shown to attain similar regret bounds, and is in fact superior in certain settings that require adaptivity. This issue is further explored in chapter \ref{chap:adaptive}.  The analysis of the agile version is of independent interest and we give it below.

\subsection{Equivalence of lazy OMD and RFTL}

The OMD  (lazy version) and RFTL are identical for linear cost functions, as we show next. 

\begin{lemma}
	Let $f_1,...,f_T$ be linear cost functions. The lazy OMD and RFTL algorithms produce identical predictions, i.e.,
	$$ \argmin_{\x \in \K} \left\{ B_{R}(\x ||\yv[t]) \right\} = \argmin_{\x \in \K}
	\left( \eta \sum_{s=1}^{t-1} \nabla_s^\top \x + {R}(\x) \right) . $$
\end{lemma}
\begin{proof}
	First, observe that the unconstrained minimum
	$$ \xv[t]^\star \equaltri \argmin_{\x \in \reals^n}
	\bigg\{\sum_{s=1}^{t-1} \nabla_s^\top \x + \frac{1}{\eta} {R}(\x) \bigg\}  $$
	satisfies
	$$ \nabla {R}(\xv[t]^\star) = - \eta \sum_{s=1}^{t-1} \nabla_s.$$
	By definition, $\yv[t]$ also satisfies the above equation, but
	since ${R}(\x)$ is strictly convex, there is only one solution for the above
	equation and thus $\yv[t]= \x^\star_t$. Hence,
	\begin{align*}
		B_{R}(\x||\yv[t])\ &=\ {R}(\x) - {R}(\yv[t]) - (\nabla {R}(\yv[t]))^\top (\bx-\yv[t])\\
		&=\ {R}(\x) - {R}(\yv[t]) + \eta\,  \sum_{s=1}^{t-1} \nabla_s^\top (\x-\yv[t])~.
	\end{align*}
	Since ${R}(\yv[t])$ and $\sum_{s=1}^{t-1} \nabla_s^\top \yv[t]$ are independent of $\x$,
	it follows that
	$B_{R}(\x||\yv[t])$ is minimized at the point $\x$ that minimizes
	${R}(\x) + \eta\, \sum_{s=1}^{t-1} \nabla_s^\top \x$ over $\K$ which, in turn, implies that
	\begin{align*}
		\argmin_{\x \in \K} B_{R}(\x||\yv[t])\ =\
		\argmin_{\x \in \K} \bigg\{ \sum_{s=1}^{t-1} \nabla_s^\top \x +
		\frac{1}{\eta} {R}(\x) \bigg\}~.
	\end{align*}
\end{proof}

\subsection{Regret bounds for Mirror Descent}

In this subsection we prove regret bounds for the agile version of the RFTL algorithm. The analysis is quite different than the one for the lazy version, and of independent interest. 

\begin{theorem} \label{thm:mirrordescent}
The OMD Algorithm \ref{alg:flpl} attains for every $\uv \in \K$ the following bound on the regret:
$$  \regret_T \le     \frac{\eta}{4}  \sum_{t=1}^T \| \nabla_t \|_t^{* 2} + \frac{R(\uv) - R(\x_1)}{2\eta }  . $$ %B_{R}(\uv||\x_1)
\end{theorem}
If an upper bound on the local norms is known, i.e., $\| \nabla_t\|_t^* \leq G_R$ for all times $t$, then we  can further optimize over the choice of $\eta$ to obtain
$$ \regret_T \leq    D_R G_R \sqrt{ T  } .$$

%\begin{theorem}
%\label{thm:main_primal_dual}
%Suppose that $\R$ is such that $B_\R(\x,\y) \geq
%\frac{1}{2} \|\x-\y\|^2$ for some norm $\|\cdot \|$. Let $\|\nabla \fv(\x_t)\|^* \leq
%G_*$ for all $t$, and $\forall \x \in K \  B_\R(\x,\x_1) \leq D^2$. Applying the primal-dual algorithm (active version) with
%$\eta = \frac{D}{2 G_* \sqrt{T}} $,  we have
%\begin{align*}
%\regret_T \leq DG_* \sqrt{T}
%\end{align*}
%\end{theorem}
\begin{proof}
Since  the functions $\fv$ are convex, for any $\x^* \in K$,
$$ \fv(\x_t) - \fv(\x^*) \leq \nabla \fv(\x_t)^\top (\x_t - \x^*)  .$$
The following property of Bregman divergences follows from the definition: for any vectors $\x,\y,\z$,
$$ (\x - \y)^\top (\nabla \R(\z) - \nabla \R(\y)) = B_\R(\x,\y)-B_\R(\x,\z) +
B_\R(\y,\z). $$
Combining both observations,
\begin{align*}
\fv(\x_t) - \fv(\x^*) & \leq \nabla \fv(\x_t)^\top (\x_t - \x^*)  \\
& =   \frac{1}{\eta}  (\nabla \R(\y_{t+1}) - \nabla \R(\x_{t}))^\top(\x^* - \x_t) \\
& =  \frac{1}{\eta} [B_\R(\x^*,\xv)-B_\R(\x^*,\y_{t+1}) + B_\R(\x_t,\y_{t+1})]   \\
& \leq  \frac{1}{\eta} [B_\R(\x^*,\xv)-B_\R(\x^*,\x_{t+1}) +
B_\R(\x_t,\y_{t+1})]
\end{align*}
where the last inequality follows from the generalized Pythagorean theorem,  as $\x_{t+1}$ is the projection w.r.t the Bregman divergence of $\y_{t+1}$ and $\x^* \in K$ is in the convex set. Summing over all iterations,
\begin{eqnarray} \label{eq:general1}
\regret & \leq & \frac{1}{\eta} [ B_\R(\x^*,\x_1) -  B_\R(\x^*,\x_T) ] + \sum_{t=1}^T \frac{1}{\eta} B_\R(\x_t,\y_{t+1}) \notag \\
& \leq & \frac{1}{\eta} D^2_R  + \sum_{t=1}^T \frac{1}{\eta} B_\R(\x_t,\y_{t+1})
\end{eqnarray}

We proceed to bound $B_\R(\x_t,\y_{t+1})$. By definition of Bregman divergence, and the generalized Cauchy-Schwartz inequality,
\begin{align*}
 B_\R(\x_t,\y_{t+1}) + B_\R(\y_{t+1},\x_t) &= (\nabla \R(\x_t) - \nabla \R(\y_{t+1}))^\top (\x_t - \y_{t+1}) \\
 &=  \eta \nabla \fv(\x_t)^\top(\x_t - \y_{t+1}) \\
 & \leq \eta \| \nabla \fv(\x_t) \|^*_t \| \x_t - \y_{t+1} \|_t \\
 &\leq  \frac{1}{2} \eta^2 G_R^{ 2} + \frac{1}{2} \|\x_t - \y_{t+1}\|^2_t.
\end{align*}
where in the last inequality follows from $(a-b)^2 \geq 0$. 
Thus, we have
$$ B_\R(\x_t,\y_{t+1}) \leq \frac{1}{2} \eta^2 G_R^2 + \frac{1}{2} \|\x_t -
\y_{t+1}\|^2_t - B_\R(\y_{t+1},\x_t) = \frac{1}{2} \eta^2 G^2_R. $$

Plugging back into Equation \eqref{eq:general1}, and by non-negativity of the Bregman divergence, we get
$$ \regret \leq  \frac{1}{2} [\frac{1}{\eta} D^2_R + \frac{1}{2} \eta T G_{R}^{2} ] \leq  D_R G_R \sqrt{T} \ ,  $$
by taking $\eta = \frac{ D_R}{\sqrt{T} G_R}$

\end{proof}

\section{Application and Special Cases}

In this section we illustrate the generality of the regularization technique: we show how to derive the two most important and famous online algorithms---the online gradient descent algorithm and the online exponentiated gradient (based on the multiplicative update method)---from the RFTL meta-algorithm. 

Other important special cases of the RFTL meta-algorithm are derived with matrix-norm regularization---namely, the von Neumann entropy function, and the log-determinant function, as well as  self-concordant barrier regularization---which we shall explore in detail in the next chapter. 

\subsection{Deriving online gradient descent}

To derive the online gradient descent algorithm, we take ${R}(\x) = \frac{1}{2} \|\x - \x_0\|_2^2$ for an arbitrary $\x_0 \in \K$. Projection with respect to  this divergence is the standard Euclidean projection (see exercises), and in addition, $\nabla {R}(\x) = \x - \x_0$. Hence, the update rule for the OMD Algorithm \ref{alg:flpl} becomes:
\begin{align*}
	& \xv = \proj_\K (  \yv)  , \ \yv =  \yv[t-1]  - \eta \nabla_{t-1}  & \mbox{lazy version}  \\
	& \xv = \proj_\K (  \yv)  , \ \yv =  \xv[t-1]  - \eta \nabla_{t-1}  & \mbox{agile version} 
\end{align*}

The latter algorithm is exactly online gradient descent, as described in Algorithm \ref{alg:ogd} in chapter \ref{chap:first order}. However, both variants behave very differently, as explored in chapter \ref{chap:adaptive} (see also exercises). % \ref{exercise:equiv-lazy-agile}). 

Theorem \ref{thm:RFTLmain1} gives us the following bound on the regret (where $D_R, \| \cdot\|_t$ are the diameter and local norm defined with respect to  the regularizer $R$ as defined in the beginning of this chapter, and $D$ is the Euclidean diameter as defined in chapter \ref{chap:opt}) 
$$  \regret_T \le \frac{1}{\eta }  D_R ^2  + 2 \eta \sum_t \| \nabla_t \|_t^{* 2}    \leq \frac{1}{2 \eta} D^2 + 2 \eta  \sum_t \|\nabla_t \|^2 \leq 2GD \sqrt{  T  }, $$
where the second inequality follows since for ${R}(\x) = \frac{1}{2} \|\x - \x_0\|^2 $,  the local norm $\|\cdot\|_t$ reduces to the Euclidean norm. 

\subsection{Deriving multiplicative updates} 

Let   ${R}(\xv[]) =  \xv[] \log \xv[] = \sum_i \x_i \log \x_i$ be the negative entropy function, where $\log \x$ is to be interpreted element-wise. Then $\nabla {R}(\x) = \bone + \log \x$, and hence the update rules for the OMD algorithm become:
\begin{align*}
	& \xv = \argmin_{\x \in \K} B_{R}(\x ||\yv)    , \ \log \yv =  \log \yv[t-1]  - \eta \nabla_{t-1}  & \mbox{lazy version}  \\
	& \xv = \argmin_{\x \in \K} B_{R}(\x ||\yv)    , \ \log \yv =  \log \xv[t-1]  - \eta \nabla_{t-1}  & \mbox{agile version} 
\end{align*}

With this choice of regularizer, a notable special case is the experts problem we encountered in \S \ref{sec:experts}, for which the decision set $\K$ is the $n$-dimensional simplex $ \Delta_n = \{ \x \in \reals^n_+ \ | \ \sum_i \x_i =  1  \}$.
In this special case, the projection according to the negative entropy becomes scaling by the $\ell_1$ norm (see exercises), which implies that both update rules amount to the same algorithm: 
$$ \x_{t+1}(i) = \frac{\x_t(i) \cdot e^{-\eta \nabla_t(i)}}{\sum_{j=1}^n \x_t(j) \cdot e^{-\eta \nabla_t(j)} }, $$
which is exactly the Hedge algorithm from the first chapter! 

Theorem \ref{thm:mirrordescent} gives us the following bound on the regret: 
$$  \regret_T \le 2 \sqrt{  2 D_R^2  \sum_t \| \nabla_t \|_t^{* 2}   }  .$$
If the costs per individual expert are in the range $[0,1]$, it can be shown that 
$$ \|\nabla_t\|_t^* \leq \| \nabla_t \|_\infty \leq 1 = G_R.$$ 
In addition, when $R$ is the negative entropy function, the diameter over the simplex can be shown to be bounded by 
$ D_R^2 \leq \log n$ (see exercises), giving rise to the bound
%In addition, the Bregman divergence corresponding to the negative entropy regularization over the simplex is known as the relative entropy  (see exercises), or  the Kullback-Leibler divergence, given by  
%$$ B_{R}(\x,\y) = \sum_i \x_i \log \frac{\x_i}{\y_i }.$$
%Assuming that $\x_1$ is the uniform distribution, the latter is always bounded by $\log n$, giving rise to the bound
$$  \regret_T \le 2  D_R G_R \sqrt{2 T   }  \leq 2\sqrt{2 T \log n}. $$

For an arbitrary range of costs, we obtain the exponentiated gradient algorithm described in Algorithm \ref{alg:eg}.
\begin{algorithm}
	[H] \caption{Exponentiated Gradient } \label{alg:eg}
	\begin{algorithmic}
		[1] \State Input: parameter $\eta > 0$.
		\State Let $\yv[1] = \bone \ , \ \xv[1] = \frac{\yv[1]}{\|\yv[1]\|_1}$.
		\For{$t=1$ to $T$}
		\State Predict $\x_t$. 
		\State Observe $f_t$, update 
		$  \yv[t+1](i) = \yv[t](i) e^{- \eta\, \nabla_{t}(i)} $ for all $i \in [n]$.
		\State Project: $\xv[t+1] = \frac{\yv[t+1]}{\| \yv[t+1] \|_1 } $
		\EndFor
	\end{algorithmic}
\end{algorithm}
The regret achieved by the exponentiated gradient algorithm can be bounded using the following corollary of Theorem \ref{thm:RFTLmain1}:
\begin{corollary} \label{cor:eg}
	The exponentiated gradient algorithm with gradients bounded by $\|\nabla_t\|_\infty \leq G_\infty$ and  parameter $\eta = \sqrt{ \frac{\log n}{  2 T G_\infty^2 }}  $ has regret bounded by
	$$  \regret_T  \leq 2 G_\infty \sqrt{2  T \log n}. $$
\end{corollary}

\section{Randomized Regularization} \label{sec:randomized-regularization}

The connection between stability in decision making and low regret has motivated our discussion of regularization thus far. However, this stability need not be achieved only using strongly convex regularization functions. An alternative method to achieve stability in decisions is by introducing  randomization into the algorithm. In fact, historically, this method preceded methods based on strongly convex regularization (see bibliography).

In this section we first describe a deterministic algorithm for online convex optimization that is easily amenable to speedup via randomization. We then give an efficient randomized algorithm for the special case of OCO with linear losses.  

\paragraph*{Oblivious vs. adaptive adversaries.}
For simplicity, we consider ourselves in this section with a slightly restricted version of OCO. So far, we have not restricted the cost functions in any way, and they could depend on the choice of decision by the online learner. However, when dealing with randomized algorithms, this issue becomes a bit more subtle: can the cost functions depend on the randomness of the decision making algorithm itself?  Furthermore, when analyzing the regret, which is now a random variable, dependencies across different iterations require probabilistic machinery which adds little to the fundamental understanding of randomized OCO algorithms.  To avoid these complications, we make the following assumption throughout this section: the cost functions $\{\fv\}$ are adversarially chosen ahead of time, and do not depend on the actual decisions of the online learner.  This version of OCO is called the {\it oblivious} setting, to distinguish it from the {\it adaptive} setting.

\subsection{Perturbation for convex losses}
The prediction in Algorithm \ref{alg:FPL} is according to a version of the follow-the-leader algorithm, augmented with an additional component of randomization. 
It is a deterministic algorithm that computes the expected decision according to a random variable. The random variable is the minimizer over the decision set according to the sum of gradients of the cost functions and an additional random vector. 

In practice, the expectation need not be computed exactly. Estimation  (via random sampling) up to a precision that depends linearly on the number of iterations would suffice.  

The algorithm accepts as input a distribution, with the probability density function (PDF) denoted $\D$, over vectors in $n$-dimensional Euclidean space  $\vn \in \reals^n$. For $\sigma, L \in \reals$, we say that a distribution $\D$ is $(\sigma,L)=(\sigma_a,L_a)$ stable with respect to the norm $\| \cdot \|_a$ if 
$$  \E_{\vn \sim \D}  [ \|\vn\|_a^* ] = \sigma_a  , $$
and
$$ \forall \uv,  \  \int_{\vn} \left| \D(\vn) - \D(\vn - \uv) \right| d \vn \leq L_a \| \uv\|_a^* . $$ 
Here $\vn \sim \D$ denotes a vector $\vn \in \reals^n$ sampled according to distribution $\D$,  and $\D(\vn)$ is the value of the probability density function $\D$ over $\vn$. The subscript $a$ is omitted if clear from the context.

The first parameter, $\sigma$, is related to the variance of $\D$, while the second, $L$, is a measure of the sensitivity of the distribution\endnote{In harmonic analysis of Boolean functions, a similar quantity is called ``average sensitivity''.}. For example, if $\D$ is the uniform distribution over the hypercube $[0,1]^n$, then it holds that (see exercises) for the Euclidean norm
$$ \sigma_2 \leq \sqrt{n} \ ,\ L_2 \leq  1. $$
Reusing notation from previous chapters, denote by $D= D_a$ the diameter of the set $\K$ according to the norm $\| \cdot \|_a$, and by $D^* = D_a^*$ the diameter according to its dual norm. % , i.e., 
%$$ D_a = \max_{ \x  ,  \y \in \K} \|\x - \y \|_a \ , \ D_a^* = \max_{ \x  ,  \y \in \K} \|\x - \y \|_a^*  $$
Similarly, denote by $G = G_a$ and $G^* = G_a^*$ an upper bound on the norm (and dual norm) of the gradients. 

\begin{algorithm}
	[ht] \caption{Follow-the-perturbed-leader for convex losses} \label{alg:FPL} 
	\begin{algorithmic}[1]
		\State Input: $\eta > 0$, distribution $\D$ over $\reals^n$, decision  set $\K \subseteq \reals^n$.
		\State Let $\x_1 = \E_{\vn \sim \D} \left[  \argmin_{\x \in \K}
		\left\{  \vn ^\top \x \right\} \right]$. 
		\For{$t=1$ to $T$}
		\State Predict $\xv[t]$.
		\State Observe the loss function $f_t$, suffer loss $f_t(\xv)$ and let $\nabla_t = \nabla f_t(\x_t) $.
		\State Update
		\begin{align} \label{eqn:fpl-oco}
		\xv[t+1] = \E_{\vn \sim \D} \left[  \argmin_{\x \in \K}
		\left\{  \eta\sum_{s=1}^{t} \nabla_s^\top \x +
			\vn ^\top \x \right\} \right]
		\end{align}
		\EndFor
	\end{algorithmic}
\end{algorithm}

\begin{theorem} \label{thm:fpl}
Let the distribution $\D$ be  $(\sigma,L)$-stable with respect to norm $\|\cdot \|_a$.
The FPL algorithm attains  the following bound on the regret:
$$  \regret_T \le\eta D G^{* 2} L T+ \frac{1}{\eta} \sigma D .$$
\end{theorem}
\noindent We can further optimize over the choice of $\eta$ to obtain
$$ \regret_T \leq 2 L D G^* \sqrt{ \sigma  T }. $$

\begin{proof}
Define the random variable $\x_t^\vn =  \argmin_{\x \in \K}
\left\{  \eta\sum_{s=1}^{t} \nabla_s^\top \x +
\vn ^\top \x \right\}$, and the random function $g_0^\vn$  as  
$$g_0^\vn(\bx) \equaltri \frac{1}{\eta} \vn^\top \bx . $$

It follows from Lemma \ref{prop:ftl-btl} applied to the functions $\{g_t(\x) = \nabla_t^\top \x \}$  that 
\begin{eqnarray*} 
\E \left[ \sum_{t=0}^T g_t (\uv) \right] & \geq \E \left[ g_0^\vn(\x_1^\vn) + \sum_{t=1}^T g_t(\x_{t+1}^\vn ) \right] \\
& \geq  \E \left[ g_0^\vn(\x_1^\vn) \right] + \sum_{t=1}^T g_t(\E[ \x_{t+1}^\vn]  ) & \mbox{convexity} \\
& =  \E \left[ g_0^\vn(\x_1^\vn) \right] + \sum_{t=1}^T g_t(\x_{t+1}  ) 
\end{eqnarray*}
and thus, 
\begin{align*} \label{eqn:ftl-shalom1}
& \sum_{t=1}^T \nabla_t(\xv - \bx^\star )  \\
& =  \sum_{t=1}^T g_t (\x_{t}) - \sum_{t=1}^T g_t(\x^\star) \\
& \leq \sum_{t=1}^T g_t (\x_{t}) - \sum_{t=1}^T g_t(\x_{t+1}) + \E [ g_0^\vn(\x^\star) - g_0^\vn(\x_1^\vn) ] \\
& \leq \sum_{t=1}^T \nabla_t(\xv - \xv[t+1] ) + \frac{1}{\eta} \E [ \| \vn \|^* \| \x^\star - \x_1^\vn \| ] & \mbox { Cauchy-Schwarz } \\
& \leq \sum_{t=1}^T \nabla_t(\xv - \xv[t+1] ) + \frac{1}{\eta}\sigma D .
\end{align*}
Hence,
\begin{eqnarray} \label{eqn:ftl-shalom-main}
	&  \sum_{t=1}^T f_t(\x_t) - \sum_{t=1}^T f_t(\x^\star)  \notag \\
	 & \leq   \sum_{t=1}^T \nabla_t^\top (\xv - \xv[]^*)  \notag \\
	& \leq   \sum_{t=1}^T \nabla_t^\top (\xv - \xv[t+1]) + \frac{1}{\eta} \sigma D & \mbox{above}  \notag \\
	& \leq  G^* \sum_{t=1}^T \|\xv - \xv[t+1] \| + \frac{1}{\eta} \sigma D . & \mbox{ Cauchy-Schwarz}
\end{eqnarray}
We now argue that $   \|\xv - \xv[t+1]\|    = O(\eta)$. Let 
$$h_t(\vn) =   \arg \min_{\x \in \K} \left\{ \eta \sum_{s=1}^{t-1} \nabla_s^\top \x + \vn^\top \x \right\}  , $$
and hence $\x_t  = \E_{\vn \sim \D} [h_t(\vn)]$. Recalling that $\D(\vn)$ denotes the value of the probability density function $\D$ over $\vn \in \reals^n$, we can write:
$$ \xv  =  \int\limits_{\vn \in \reals^n } h_t(\vn) \D(\vn) d \vn ,$$
and:
$$ \xv[t+1]  = \int\limits_{\vn \in \reals^n } h_t(\vn + \eta \nabla_t) \D(\vn) d \vn = \int\limits_{\vn \in \reals^n } h_t(\vn ) \D(\vn - \eta \nabla_t ) d \vn .$$
Notice that $\xv,\xv[t+1]$ may  depend on each other. However, by linearity of expectation, we have that
\begin{align*}
&  \| \xv  -  \xv[t+1]\|    \\
 & = \left\| \int\limits_{\vn \in \reals^n } ( h_t(\vn ) - h_t (\vn  + \eta \nabla_t ) ) \D(\vn ) d \vn \right\|  \\
& = \left\| \int\limits_{\vn \in \reals^n }  h_t(\vn ) (\D ( \vn ) - \D ( \vn   - \eta \nabla_t))  d \vn \right\| \\
& = \left\| \int\limits_{\vn \in \reals^n }  (h_t(\vn ) - h_t(\bzero) )  (\D ( \vn ) - \D ( \vn   - \eta \nabla_t))  d \vn \right\| \\
& \leq   \int\limits_{\vn \in \reals^n }  \|h_t(\vn ) - h_t(\bzero) \| |\D ( \vn ) - \D ( \vn   - \eta \nabla_t) |  d \vn  \\
& \leq  D  \int\limits_{\vn \in \reals^n }  \left|  \D(\vn) - \D( \vn - \eta \nabla_t) \right|  d \vn    \mbox{\ \ \ since } \|\x_t - h_t(\bzero)\| \leq  D  \\
& \leq D L \cdot  \eta \|\nabla_t\|^*  \leq \eta D L G^*  .  \mbox{\ \ since $\D$ is $(\sigma,L)$-stable}. 
\end{align*}
Substituting this bound back into \eqref{eqn:ftl-shalom-main} we have
\begin{eqnarray*} 
&  \sum\limits_{t=1}^T f_t(\x_t) - \sum\limits_{t=1}^T f_t(\x^\star)   \leq \eta  L D G^{* 2}  T   + \frac{1}{\eta}  \sigma D.
\end{eqnarray*}
\end{proof}

For the choice of $\D$ as the uniform distribution over the unit hypercube $[0,1]^n$, which has parameters $\sigma_2 \leq \sqrt{n} $ and $ L_2 \leq 1$ for the Euclidean norm, the optimal choice of $\eta$ gives a regret bound of $DG n^{1/4}  \sqrt{ T}$. This is a factor ${n}^{1/4}$ worse than the online gradient descent regret bound of Theorem \ref{thm:gradient}.  For certain decision sets $\K$ a better choice of distribution $\D$ results in near-optimal regret bounds.

\subsection{Perturbation for linear cost functions}

The case of linear cost functions $f_t(\x) = \gv^\top \x$ is of particular interest in the context of randomized regularization. 
Denote 
$$ w_t(\vn) = \arg\min_{\x \in \K} \left\{  \eta\sum_{s=1}^{t} \gv[s]^\top \x + \vn ^\top \x \right\} . $$

By linearity of expectation, we have that
$$ f_t(\x_t) = f_t( \E_{\vn \sim \D} [w_t(\vn) ] ) = \E_{\vn \sim \D} [ f_t(w_t(\vn)) ]. $$
Thus, instead of computing $\x_t$ precisely, we can sample a single vector $\vn_0 \sim \D$, and use it to compute 
$ \xhat_t = w_t(\vn_0) $, as illustrated in Algorithm \ref{alg:FPL-linear}.
\begin{algorithm}
	[ht] \caption{FPL for linear losses} \label{alg:FPL-linear} 
	\begin{algorithmic}[1]
		\State Input: $\eta > 0$, distribution $\D$ over $\reals^n$, decision  set $\K \subseteq \reals^n$.
		\State Sample $\vn_0 \sim \D$.  Let $\xhat_1 \in \argmin_{\x \in \K}  \{ -\vn_0^\top \x \} $.
		\For{$t=1$ to $T$}
		\State Predict $\xhat_t$.
		\State Observe the linear loss function, suffer  loss $\gv[t]^\top\xv $.
		\State Update
		\begin{align*} 
		\xhat_t =   \argmin_{\x \in \K}
		\left\{  \eta\sum_{s=1}^{t-1} \gv[s]^\top \x +
			\vn_0 ^\top \x \right\} 
		\end{align*}
%		\State Observe and suffer linear loss $\gv[t]^\top \xhat_t$
		\EndFor
	\end{algorithmic}
\end{algorithm}

By the above arguments, we have that the expected regret for the random variables $\xhat_t$ is the same as that for $\x_t$. We obtain the following Corollary:
\begin{corollary} \label{cor:fpl-linear}
$$ \E_{\vn_0 \sim \D}  \left[ \sum_{t=1}^T f_t(\xhat_t) - \sum_{t=1}^T f_t(\x^\star) \right] \leq  \eta  L D G^{*  2}  T   + \frac{1}{\eta}  \sigma D  .$$
\end{corollary}

The main advantage of this algorithm is computational:  with a single linear optimization step over the decision set $\K$ (which does not even have to be convex!), we attain near optimal expected regret bounds.

%\ignore{
\subsection{Follow-the-perturbed-leader for expert advice}

An interesting special case (and in fact the first use of perturbation in decision making) is that of non-negative linear cost functions over the unit $n$-dimensional simplex with costs bounded by one, or the problem of prediction of expert advice we have considered in chapter \ref{chap:intro}. 

Algorithm \ref{alg:FPL-linear} applied to the probability simplex and with exponentially distributed noise is known as the follow-the-perturbed-leader for prediction from expert advice method. We spell it out in Algorithm \ref{alg:FPL*}.

\begin{algorithm}
	[ht] \caption{FPL for prediction from expert advice} \label{alg:FPL*} 
	\begin{algorithmic}[1]
		\State Input: $\eta > 0$
		\State Draw $n$  exponentially distributed variables  $\vn(i) \sim  e^{- \eta x}$. 
		\State Let $\xv[1] = \argmin_{\be_i \in \Delta_n}   \{ -\be_i^\top \vn \} $.  
		\For{$t=1$ to $T$}
		\State Predict using expert $i_t$ such that $\xhat_t = \be_{i_t}$
		\State Observe the loss vector  and suffer loss $\gv[t]^\top \xhat_t = \gv(i_t) $
		\State Update (w.l.o.g. choose $\xhat_{t+1}$ to be a vertex) 
		\begin{align*} 
		\xhat_{t+1} =   \argmin_{\x \in \Delta_n}
		\left\{  \sum_{s=1}^{t} \gv[s]^\top \x -
			\vn ^\top \x \right\} 
		\end{align*}
		\EndFor
	\end{algorithmic}
\end{algorithm}

Notice that we take the perturbation to be distributed according to the one-sided negative exponential distribution, i.e., $\vn(i) \sim 	e^{-\eta x} $, or more precisely
$$  \Pr[  \vn(i) \leq x ] = 1 -  e^{-\eta x} \quad \forall x  \geq 0 .  $$ 

Corollary \ref{cor:fpl-linear} gives regret bounds that are suboptimal for this special case, thus we give here an alternative analysis that gives tight bounds up to constants amounting to the following theorem.
\begin{theorem} \label{thm:fpl-experts}
Algorithm \ref{alg:FPL*}  outputs a sequence of predictions $\xhat_1,...,\xhat_T \in \Delta_n$ such that:
$$ (1 - \eta) \E \left[ \sum_t  \gv^\top \xhat_t \right] \leq  \min_{\x^\star\in \Delta_n} \sum_t \gv^\top \x^\star  + \frac{4 \log n}{\eta }  . $$
\end{theorem}

Notice that as a special case of the above theorem, choosing $\eta = \sqrt{\frac{\log n}{T}}$ yields a regret bound of
$$ \regret_T = O ( \sqrt{  T \log n }), $$
which is equivalent up to constant factors to the guarantee given for the Hedge algorithm in Theorem \ref{lem:hedge}. 

\begin{proof}
We start with the same analysis technique throughout this chapter: let $\gv[0] = -\vn $.  It follows from Lemma \ref{prop:ftl-btl} applied to the functions $\{f_t(\x) = \gv^\top \x \}$  that 
\begin{equation*} 
\E \left[ \sum_{t=0}^T \gv ^\top \uv \right] \geq \E \left[ \sum_{t=0}^T \gv^\top \xhat_{t+1} \right]  ,
\end{equation*}
and thus, 
\begin{align} \label{eqn:ftl-shalom3}
\E \left[ \sum_{t=1}^T \gv^\top (\xhat_t - \x^\star ) \right]
& \leq \E \left[ \sum_{t=1}^T \gv ^\top (\xhat_{t}  - \xhat_{t+1}) \right] + \E [ \gv[0]^\top (\x^\star - \x_1) ] \notag \\
& \leq \E \left[ \sum_{t=1}^T \gv^\top (\xhat_t - \xhat_{t+1} ) \right] +  \E [ \| \vn \|_\infty \| \x^\star - \x_1 \|_1 ]  \notag \\ % & \mbox { Cauchy-Schwarz} \notag \\
& \leq   \sum_{t=1}^T \E \left[ \gv^\top (\xhat_t - \xhat_{t+1} ) \ | \ \xhat_t \right] + \frac{4}{\eta} \log n ,
\end{align}
where the second inequality follows by the generalized Cauchy-Schwarz inequality, and the last inequality follows since  (see exercises)
$$  \E_{\vn \sim \D}  [ \|\vn\|_\infty ] \leq \frac{ 2 \log n}{\eta} .$$

We proceed to bound $\E [ \gv^\top (\xhat_t - \xhat_{t+1} ) | \xhat_t ]$, which is naturally bounded by the probability that $\xhat_{t} $ is not equal to $\xhat_{t+1}$ multiplied by the maximum value that $\gv$ can attain (i.e., its $\ell_\infty$ norm):
$$ \E [ \gv^\top (\xhat_t - \xhat_{t+1} ) \ | \ \xhat_t ] \leq \|\gv\|_\infty \cdot \Pr[ \xhat_t \neq \xhat_{t+1} \ |\  \xhat_t ]  \leq \Pr[ \xhat_t \neq \xhat_{t+1} \ |\  \xhat_t ]  .$$
Above we have that $\|\gv\|_\infty \leq 1$ by assumption that the losses are bounded by one. 

To bound the latter, notice that the probability $\xhat_t = \be_{i_t} $ is the leader at time $t$ is the probability that $- \vn({i_t}) > v$ for some value $v$ that depends on the entire loss sequence till now.  On the other hand, given $\xhat_t$, we have that $\xhat_{t+1} = \xhat_t$ remains the leader if $- \vn(i_t) > v + \gv(i_t)$, since it was a leader by a margin of more than the cost it will suffer. Thus,
\begin{eqnarray*} 
\Pr[ \xhat_t \neq \xhat_{t+1} \ |\  \xhat_t ] & = 1 - \Pr[- \vn({i_t}) > v+ \gv(i_t) \ |\  -\vn({i_t}) > v ] \\
& = 1 - \frac{ \int_{v + \gv(i_t) }^\infty \eta e^{-\eta x } } {\int _{v}^\infty \eta e^{-\eta x}} \\
& = 1 - e^{ - \eta \gv(i_t) } \\
& \leq \eta \gv(i_t)  = \eta \gv^\top \xhat_t .
\end{eqnarray*} 
Substituting this bound back into \eqref{eqn:ftl-shalom3} we have
\begin{eqnarray*} 
&  \E [ \sum_{t=1}^T \gv^\top (\xhat_t - \x^\star ) ]  \leq \eta  \sum_t \E_t[ \gv^\top \xhat_t] + \frac{4 \log n}{\eta} ,
\end{eqnarray*}
which simplifies to the Theorem. 
\end{proof}

\section{* Adaptive Gradient Descent} \label{sec:adagrad}

Thus far we have introduced regularization as a general methodology for deriving online convex optimization algorithms. 
The main theorem of this chapter, Theorem \ref{thm:RFTLmain1}, bounds the regret of the RFTL algorithm for any strongly convex regularizer as 
\begin{equation} \label{eqn:general-regret-form}
 \regret_T \leq  \max_{\uv \in \K} \sqrt{ 2 \sum_t \|\nabla_t \|_t^{* 2}  B_{R}( \uv||\x_1) }. 
\end{equation}
In addition, we have seen how to derive the online gradient descent and the multiplicative weights algorithms as special cases of the RFTL methodology. But are there other special cases of interest, besides these two basic algorithms, that warrant such general and abstract treatment? 

There are surprisingly few cases of interest besides the Euclidean and Entropic regularizations and their matrix analogues\footnote{One such example is the self-concordant barrier regularization which we shall explore in the next chapter.}. However, in this chapter we will give some justification of the abstract treatment of regularization. 

Our treatment is motivated by the following question: thus far we have thought of $R$ as a strongly convex function. But which strongly convex function should we choose to minimize regret? 
This is a deep and difficult question which has been considered in the optimization literature since its early developments. Naturally, the optimal regularization should depend on both the convex underlying decision set, as well as the actual cost functions (see exercises for a natural candidate of a regularization function that depends on the convex decision set). 

We shall treat this question no differently than we treat other optimization problems throughout this manuscript itself: we'll learn the optimal regularization online! 
That is, a regularizer that adapts to the sequence of cost functions and is in a sense the ``optimal'' regularization to use in hindsight. 
This gives rise to the AdaGrad (Adaptive subGradient method) algorithm \ref{alg:adagrad}, which explicitely optimizes over the regularization choice in line \eqref{eqn:adagrad1} to minimize the gradient norms, which is the dominant expression in \eqref{eqn:general-regret-form}. % $ G_T \bullet H = \sum_{t=1}^T \nabla^\top H \nabla_t = \sum_t \|\nabla_t\|_H^2 .$

\ignore{
To be more formal, let us consider the set of all strongly convex regularization functions with a fixed and bounded Hessian in the set 
$$\forall \x \in \K \ . \ \nabla^2 R(\x) = \nabla^2 \in \H \equaltri \{ X \in \reals^{n \times n} \ ; \ \trace(X^{-1}) \leq 1 \ , \ X \succcurlyeq 0 \}$$ 

The set $\H$ is a restricted class of regularization functions (which does not include the entropic regularization). However, it is a general enough class to capture online gradient descent along with any rotation of the Euclidean regularization.  
}

\begin{algorithm}
	[H] \caption{AdaGrad} \label{alg:adagrad}
	\begin{algorithmic}
		[1] \State Input: parameters $\eta, \x_1 \in \K$.
		\State Initialize: $G_0 = \bzero $, 
		\For{$t=1$ to $T$}
		\State Predict $\x_t$, suffer loss $f_t(\x_t)$.
		\State \label{eqn:adagrad1} Update $ G_t = G_{t-1} + \nabla_t \nabla_t^\top$ and define
		\begin{align*}
			&\text{[Diagonal version]}
			& H_t =  \argmin_{H \succeq 0,H=\diag(H)} \left\{ G_t \bullet H^{-1} + \trace(H) \right\} = \diag({G_t}^{1/2}) \\
			&\text{[Full matrix version]} 	& H_t =  \argmin_{H \succeq \bzero} \left\{ G_t \bullet H^{-1} + \trace(H) \right\} =  {G_t}^{1/2}
		\end{align*}
        \State Update
		$$ \yv[t+1] = \xv - \eta H_t^{-1} \nabla_t $$ 
		$$ \xv[t+1] = \argmin_{\x \in \K} \| \yv[t+1]  - \x\|^{2}_{H_t} $$ 
				\EndFor
	\end{algorithmic}
\end{algorithm}

AdaGrad comes in two versions: diagonal and full matrix, the first being particularly efficient to implement with negligible computational overhead over online gradient descent. In the algorithm definition and throughout this chapter, the notation $A^{-1}$ refers to the Moore-Penrose pseudoinverse of the matrix $A$.

The computation in line \eqref{eqn:adagrad1} finds the regularization matrix $H$ which minimizes the norm of the gradients from within the positive semi-definite cone, with or without a diagonal constraint. This is closely related, as we shall see, to optimization w.r.t. two natural sets of matrices: 
\begin{enumerate}
    \item $ \H_1 = \{ H = \diag(H) , H \succeq 0 \ , \ \trace(H) \leq 1 \}  $ 
    \item $ \H_2 = \{ H \succeq 0 \ , \ \trace(H) \leq 1 \}  $. 
\end{enumerate}
This results in a regularization matrix that is provably optimal in the following sense,
\begin{lemma} \label{lem:regularzation-optimality-adagrad}
For $\H_i \in \{\H_1,\H_2\}$ with the corresponding $H_T$,
\begin{eqnarray*}
\sqrt{  \min_{H \in \H_i} \sum_{t=1}^T \|\nabla_t \|_H^{* 2} }  & =  \trace(H_T)  .
\end{eqnarray*}
\end{lemma}

Using this lemma, we show the regret of AdaGrad is at most a constant factor larger than the minimum regret of all RFTL algorithm with regularization functions whose Hessian is fixed and belongs to the class $\H_i$. Furthermore, the regret of the diagonal version can be a factor $\sqrt{d}$ smaller than that of online gradient descent for certain gradient geometries. 
The regret bound on AdaGrad is formally stated in the following theorem.  
\begin{theorem}
\label{theorem:adagrad-main}
Let	$\{\x_t\}$ be defined by Algorithm~\ref{alg:adagrad} with parameters $ \eta  = {D}$ (full matrix) or $\eta = D_\infty$ (diagonal). 
Then for any $\x^\star \in \K$,
\begin{align*} \label{eqn:adagrad_regret}
& \regret_{T}(\mbox{AdaGrad-diag}) \le \sqrt{2} D_\infty    \sqrt{  \min_{H \in \H_1} \sum_t \|\nabla_t \|_H^{* 2}  } , \\
& \regret_{T}(\mbox{AdaGrad-full}) \le \sqrt{2} D    \sqrt{  \min_{H \in \H_2} \sum_t \|\nabla_t \|_H^{* 2}  } .
\end{align*}
\end{theorem}

Before proceeding to the analysis, we consider when the regret bounds for AdaGrad improve upon those of Online Gradient Descent. One such case is when $\K$ is the unit cube in $d$-dimensional Euclidean space. This convex set has $D_\infty =1$ and $D = \sqrt{d}$. Lemma \ref{lem:regularzation-optimality-adagrad} and Theorems \ref{theorem:adagrad-main},\ref{thm:RFTLmain1} imply that the regret of diagonal AdaGrad and OGD are bounded by 
\begin{align*}
& \regret_{T}(\mbox{AdaGrad-diag}) \le \sqrt{2} \trace(\diag(G_T)^{1/2})  ,\\
& \regret_{T}(\mbox{OGD}) \le \sqrt{2 d} \sqrt{ \sum_t \|\nabla_t\|^2} = \sqrt{2d \trace(\diag(G_T)) }  .
\end{align*}
The relationship between the two terms depends on the matrix $\diag(G_T)$. If this matrix is sparse, then AdaGrad has a superior bound by at most $\sqrt{d}$ factor. For other convex bodies, such as the Euclidean ball, and when the matrix $G_T$ is dense, the regret of OGD can be a factor $\sqrt{d}$ lower.

\subsection{Analysis of adaptive regularization} 

We proceed with the proof of Theorem \ref{theorem:adagrad-main}. The first component is the following Lemma, which generalizes the RFTL analysis to changing regularization. 

\begin{lemma} \label{lem:adagradlem}
Let $H_{0} = \argmin_{H \succeq 0} \left\{ \trace(H) \right\} = 0$,
$$ \regret_T(\text{GenAdaReg}) \leq \frac{\eta}{2}  ( G_T \bullet H_T^{-1} + \trace(H_T)) 
  +  \frac{1}{2 \eta} \sum_{t=0}^T  \| \bx_t  -  \bx^\star\|^2_{ 
H_t - H_{t-1}}  .$$
\end{lemma}
\begin{proof}
By the definition of $\by_{t+1}$:
\begin{align*} 
& \by_{t+1} - \bx^\star = \bx_{t} - \bx^\star - \eta {H_t}^{-1}
\nabla_t \\
& H_t (\by_{t+1} - \bx^\star) = H_t (\bx_t - \bx^\star) - \eta 
\nabla_t.
\end{align*}
Multiplying the transpose of the first equation by the second we get
\begin{gather}
(\by_{t+1} - \bx^\star)^\top H_t(\by_{t+1} - \bx^\star) = \notag \\
(\bx_t\! -\! \bx^\star)^\top H_t(\bx_t\! -\! \bx^\star) -
2 \eta  \nabla_t^\top (\bx_t\! -\! \bx^\star) +
\eta^2 \nabla_t^\top H_t^{-1} \nabla_t.
\label{eq:multiplied-adagrad}
\end{gather}
Since $\bx_{t+1}$ is the projection of $\by_{t+1}$ in the norm induced by
$H_t$, we have (see \S \ref{sec:projections})
\begin{align*}
 (\by_{t+1} - \bx^\star)^\top H_t(\by_{t+1} - \bx^\star)  & = \| \by_{t+1} - \bx^\star \|_{H_t}^2  \ge  \| \bx_{t+1} - \bx^\star \|_{H_t}^2   .
%&  = (\bx_{t+1} - \bx^\star)^\top G_t(\bx_{t+1} - \bx^\star ).
\end{align*}
This inequality is the reason for using generalized projections as
opposed to standard projections, which were used in the analysis
of \ogd (see \S \ref{section:ogd} Equation
\eqref{eqn:ogdtriangle}). This fact together with
\eqref{eq:multiplied-adagrad} gives
\begin{align*}
\nabla_t^\top (\bx_t \! -\! \bx^\star) &\leq \ \frac{\eta}{2}
\nabla_t^\top H_t^{-1} \nabla_t  +  \frac{1}{2 \eta} \left( \| \bx_{t} - \bx^\star \|_{H_t}^2 - \| \bx_{t+1} - \bx^\star \|_{H_{t}}^2  \right) .
\end{align*}
Now, summing up over $t=1$ to $T$ we get that
\begin{align}  \label{eqn:adagrad-shalom}
&\sum_{t=1}^T \nabla_t^\top (\bx_t - \bx^\star)
 \leq \frac{\eta}{2} \sum_{t=1}^T \nabla_t^\top H_t^{-1} \nabla_t +
\frac{1}{2\eta}  \| \bx_{1} - \bx^\star \|_{H_{0}}^2  \\
& + \frac{1}{2 \eta} \sum_{t=1}^T  \left( \| \bx_{t} - \bx^\star \|_{H_t}^2 - \| \bx_{t} - \bx^\star \|_{H_{t-1}}^2 \right)   - \frac{1}{2 \eta} \| \bx_{T+1}   - \bx^\star \|_{H_{T}}^2  \notag \\
&\leq \frac{\eta}{2} \sum_{t=1}^T \nabla_t^\top H_t^{-1}
\nabla_t + \frac{1}{2\eta} \sum_{t=0}^{T} \| \bx_t\! -\! \bx^\star\|^2_{ 
H_t - H_{t-1}}  . \notag
\end{align}
In the last inequality we use the definition $H_{0} = 0 $. 
We proceed to bound the first term. To this end, define the functions 
$$ \Psi_t(H) = \nabla_t \nabla_t^\top \bullet H^{-1} \ , \ \Psi_0(H) = \trace(H) . $$ 
By definition, $H_t$ is the minimizer of $\sum_{i=0}^{t} \Psi_i$ over $\H$. Therefore, using the BTL Lemma \ref{prop:ftl-btl}, we have that 
\begin{eqnarray*}
\sum_{t=1}^T \nabla_t^\top H_t^{-1} \nabla_t & = \sum_{t=1}^T \Psi_t(H_t) \\
& \leq \sum_{t=1}^T \Psi_t(H_T)  + \Psi_0(H_T) - \Psi_0(H_0) \\
& =  G_T \bullet H_T^{-1} + \trace(H_T) . % = 2 \trace(H_T) ,
\end{eqnarray*}
%where we used the fact that $H_T = G_T^{1/2}$.
\end{proof}

We can now continue with the proof of Theorem \ref{theorem:adagrad-main}.
\begin{proof}[Proof of Theorem \ref{theorem:adagrad-main}]
We bound both parts of Lemma \ref{lem:adagradlem}, with the following two lemmas,
\begin{lemma} \label{lemma:opt-distance-bound-adagrad}
For both the diagonal and full matrix versions of AdaGrad, the following holds
$$ G_T \bullet H_T^{-1} \leq \trace(H_T) . $$ 
\end{lemma}
\begin{lemma} \label{lemma:opt-reg-bound2-adagrad}
Let $D_\infty$ denote the $\ell_\infty$ diameter of $\K$, and $D$ the Euclidean diameter. Then the following bounds hold,
\begin{align*}
& \mbox{Diagonal AdaGrad: } & \sum_{t=1}^{T} \| \bx_t  -  \bx^\star\|_{H_t - H_{t-1}}^2  \leq D^2_\infty \trace(H_T). \\
& \mbox{Full matrix AdaGrad: } &  \sum_{t=1}^{T} \| \bx_t  -  \bx^\star\|_{H_t - H_{t-1}}^2  \leq D^2 \trace(H_T). 
\end{align*}
\end{lemma}
Now combining Lemma \ref{lem:adagradlem} with the above two lemmas, and using $\eta = \frac{D}{\sqrt{2}}$ or $\eta = \frac{D_\infty}{\sqrt{2}}$ appropriately, we obtain the theorem. 
\end{proof}

We proceed to complete the proof of the two lemmas above.

\begin{proof}[Proof of Lemma \ref{lemma:opt-distance-bound-adagrad}]

The optimization problem of choosing $H_t$ in line \eqref{eqn:adagrad1} of Algorithm \ref{alg:adagrad}  has an explicit solution, given in the following proposition (whose proof is left as an exercise). 
\begin{proposition}
\label{proposition:solution-inv-trace}
Consider the following optimization problems, for $A \succcurlyeq 0$:
\begin{align*}
\min_{X \succeq 0 , \trace(X) \leq 1}  \left\{ X^{-1} \bullet A \right\}  \quad \quad \min_{X \succeq 0} \left\{ A \bullet X^{-1} + \trace(X) \right\} .
\end{align*} 
Then the global optimizer to these problems is obtained at $X = \frac{A^{1/2}} { \trace(A^{1/2})}$ and $X = A^{1/2}$ respectively. 
Over the set of diagonal matrices, the global optimizer is obtained at $X = \frac{\diag(A)^{1/2}} { \trace(A^{1/2})}$ and $X = \diag(A)^{1/2}$ respectively. 
\end{proposition}

A direct corollary of this proposition gives Lemma \ref{lem:regularzation-optimality-adagrad} as follows:
\begin{corollary}
\begin{eqnarray*}
\sqrt{  \min_{H \in \H} \sum_t \|\nabla_t \|_H^{* 2} }  & = \sqrt{ \min_{H \in \H } \trace (  H^{-1} \sum_t \nabla_t \nabla_t^\top  ) }  \\
& =  \trace{ \sqrt{ \sum_t \nabla_t \nabla_t^\top } } = \trace(H_T)  % = \trace(\sqrt{S_T - \delta n}) \\ 
%& = \trace(G_T) - \delta n. 
\end{eqnarray*}
\end{corollary}

\end{proof}

The remaining term from Lemma \ref{lem:adagradlem} is the expression $\sum_{t=0}^T  \| \bx_t  -  \bx^\star\|^2_{ H_t - H_{t-1}}$, which we proceed to bound. 

\begin{proof}[Proof of Lemma \ref{lemma:opt-reg-bound2-adagrad}]
By definition $G_t \succcurlyeq G_{t-1}$, and hence using proposition \ref{proposition:solution-inv-trace} and the definition of $H_t$ in line \eqref{eqn:adagrad1}, we have that $H_t = \diag(G_t^{1/2} ) \succcurlyeq  \diag(G_{t-1}^{1/2} ) =  H_{t-1}$. Since for a diagonal matrix $H$ it holds that $\x^\top H \x \leq \|\x\|_\infty^2 \trace(H)$, we have
\begin{align*}
& \sum_{t=1}^{T} (\bx_t\! -\! \bx^\star)^\top (H_t - H_{t-1} ) (\bx_t\! -\! \bx^\star) \\
& \leq \sum_{t=1}^{T} D^2_\infty  \trace( H_t - H_{t-1} ) & \mbox{diagonal structure, $H_t - H_{t-1} \succeq 0$}\\
%& \leq D^2_\infty \sum_{t=1}^{T}  \trace (H_t - H_{t-1})  & A \succcurlyeq 0 \ \Rightarrow \  \lambda_{\max}(A) \leq \trace(A) \\
& = D^2_\infty \sum_{t=1}^{T}  (\trace (H_t ) - \trace( H_{t-1}))  &  \mbox{ linearity of the trace} \\
& \leq D^2_\infty \trace(H_T). 
\end{align*}

Next, we consider the full matrix case.
By definition $G_t \succcurlyeq G_{t-1}$, and hence $H_t \succcurlyeq H_{t-1}$. Thus, 
\begin{align*}
& \sum_{t=1}^{T} (\bx_t\! -\! \bx^\star)^\top (H_t - H_{t-1} ) (\bx_t\! -\! \bx^\star) \\
& \leq \sum_{t=1}^{T} D^2  \lambda_{\max}( H_t - H_{t-1} ) \\
& \leq D^2 \sum_{t=1}^{T}  \trace (H_t - H_{t-1})  & A \succcurlyeq 0 \ \Rightarrow \  \lambda_{\max}(A) \leq \trace(A) \\
& = D^2 \sum_{t=1}^{T}  (\trace (H_t ) - \trace( H_{t-1}))  &  \mbox{ linearity of the trace} \\
& \leq D^2 \trace(H_T). 
\end{align*}
\end{proof}

\ignore{
To prove this theorem, we consider a generic meta-algorithm with adaptive regularization given in Algorithm \eqref{alg:adareg}. This meta-algorithm captures both versions of AdaGrad, as well as other adaptive methods such as the online Newton step algorithm, by changing the regularization set $\H$.  
\begin{algorithm}
	[H] \caption{Generic Adaptive Regularization} \label{alg:adareg}
	\begin{algorithmic}
		[1] \State Input:  $ \x_1 \in \K$, regularization set $\H$. 
		\State Initialize: $ G_0 = \bzero $, 
		\For{$t=1$ to $T$}
		\State Predict $\x_t$, suffer loss $f_t(\x_t)$.
		\State Update: 
		$$G_t = G_{t-1} + \nabla_t \nabla_t^\top $$
		$$ H_t = \argmin_{H \in \H} \left\{ G_t \bullet H \right\} $$ % + \Phi(H) \right\} $$
		$$ \yv[t+1] = \xv - \eta H_t^{-1} \nabla_t $$ 
		$$ \xv[t+1] = \argmin_{\x \in \K} \| \yv[t+1]  - \x\|^2_{H_t} $$ 
				\EndFor
	\end{algorithmic}
\end{algorithm}

The intuition behind meta algorithm \eqref{alg:adareg} is that it iteratively computes $H_t$ in a greedy way: minimizing the sum of gradient norms according to the current regularization. Its regret bound is given by the following lemma, which we later specialize to derive the AdaGrad bound. 
}

\newpage
\section{Bibliographic Remarks}

Regularization in the context of online learning was first studied in \citep{GroveLS01} and \citep{KivinenW01}. The influential paper of \citet{KV-FTL} coined the term ``follow-the-leader'' and introduced many of the techniques that followed in OCO. The latter paper studies random perturbation as a regularization and analyzes the follow-the-perturbed-leader algorithm, following an early development by \citet{Hannan57} that was overlooked in learning for many years.

In the context of OCO, the term follow-the-regularized-leader was coined in \citep{ShwartzS07,ShalevThesis}, and at roughly the same time an essentially identical algorithm was called ``RFTL'' in \citep{AbernethyHR08}. The equivalence of RFTL and Online Mirror Descent was observed by  \citep{DBLP:conf/colt/HazanK08}.
The AdaGrad algorithm was introduced in \citep{DuchiHS10,duchi2011adaptive},  its diagonal  version was also discovered in parallel in \citep{McMahanS10}. The analysis of AdaGrad presented in this chapter is due to \citep{gupta2017unified}.

Adaptive regularization has received significant attention due to its success in training deep neural networks, and notably the development of adaptive algorithms that incorporate momentum and other heuristics, most popular of which are AdaGrad, RMSprop \citep{tieleman2012lecture} and Adam \citep{kingma2014adam}. For a survey of optimization for deep learning, see the comprehensive text of \citet{Goodfellow-et-al-2016}.

There is a strong connection between randomized perturbation and deterministic regularization. For some special cases, adding randomization can be thought of as a special case of deterministic strongly convex regularization, see \citep{abernethy2014online,abernethy16perturbation}.

\newpage

\begin{exercises}

\exer{
\label{exercise:dualnorm}
This exercise concerns the notion of a dual norm. 
\subexer{Show that the dual norm to a matrix norm given by $A \succ 0$ corresponds to the matrix norm of $A^{-1}$.}

\subexer{Prove the generalized Cauchy-Schwarz inequality for any norm, i.e., 
	$$  \x^\top  \y  \leq \|\x \| \|\y \|^* . $$ }
}
	
\exer{% \item
Prove that the Bregman divergence is equal to the local norm at an intermediate point, that is:
$$ B_{R}(\x||\y) = \frac{1}{2} \|\x - \y\|_\z^2, $$ 
where  $\z \in [\x,\y]$, and the interval $[\x,\by]$ is defined as 
$$ [\bx,\by] = \{ \vv  = \alpha \bx + (1-\alpha) \by \ , \ \alpha \in [0,1] \} . $$
}

\exer{% \item
 \label{exercise:bregman-Euclid}
Let ${R}(\x) = \frac{1}{2} \|\x - \x_0\|^2$ be the (shifted) Euclidean regularization function. Prove that the corresponding Bregman divergence is the Euclidean metric. Conclude that projections with respect to  this divergence are standard Euclidean projections.
}

\exer{% \item 
\label{exercise:equiv-lazy-agile}
	 Prove that the agile and lazy versions of the OMD meta-algorithm are different, in the sense that they can produce different predictions over the same setting and cost functions. Show this for the case that the regularization is Euclidean and the decision set is the Euclidean ball. 
}

\exer{% \item 
\label{exercise:bregman-entropy}
For this problem the decision set is the $n$-dimensional simplex.  Let ${R}(\x) = \x \log \x $ be the negative entropy regularization function. Prove that the corresponding Bregman divergence is the relative entropy, and prove that the diameter $D_R$ of the $n$-dimensional simplex with respect to  this function is bounded by $\log n$. Show that projections with respect to  this divergence over the simplex amounts to scaling by the $\ell_1$ norm. 
}

\exer{% \item
Prove that for the uniform distribution $\D$ over the unit hypercube $[0,1]^n$, the parameters $\sigma,L$ defined in \S \ref{sec:randomized-regularization} with respect to the Euclidean norm can be bounded as $\sigma < \sqrt{n} \ , \ L \leq 1$. 
}

\exer{% \item
Let $\D$ be a one-sided multi-dimensional  exponential distribution, such that a vector $\vn \sim \D$ is distributed over each coordinate exponentially:
$$ \ \Pr[ \vn_i \leq x ] = 1 - e^{-x} \quad \forall i \in [n] , \ x \geq 0 . $$
Prove that 
$$ \E_{\vn \sim \D} [ \|\vn\|_\infty ] \leq 2 \log n .$$
(Hint: use the Chernoff bound)
\\
Extra credit: prove that $ \E_{\vn \sim \D} [ \|\vn\|_\infty ] = H_n ,$ where $H_n$ is the $n$-th harmonic number. 
}

\exer{$^*$ % \item $^*$   
A  set $\K \subseteq \reals^d$ is symmetric if $\x \in \K$ implies $-\x \in \K$. Symmetric sets gives rise to a natural definition of a norm. 
Define the function $\| \cdot \|_\K : \reals^d \mapsto \reals$ as
$$ \| \x \|_\K = \arg \min_{\alpha > 0 }  \left \{ \frac{1}{\alpha} \x \in \K \right \} .$$ 
Prove that $\| \cdot \|_\K$ is a norm if and only if $\K$ is convex. 
}

\exer{$^{* *}$ % \item $^{**}$
Prove a lower bound of $\Omega(T)$ on the regret of the  RFTL algorithm with $\| \cdot \|_\K$ as a regularizer. 
}
%\item
%Prove that for positive definite matrices $A \succcurlyeq B \succ 0$ it holds that
%$$ \trace(A - B) = \trace(A ) - \trace(B) $$

\exer{$^*$ %\item $^*$
Prove that for positive definite matrices $A \succcurlyeq B \succ 0$ it holds that
%\begin{enumerate}
%\item
\subexer{$A^{1/2} \succcurlyeq B^{1/2} $}
%\item
\subexer{$ 2 \trace( ({A - B})^{1/2}  )  + \trace( A^{-1/2}B)  \leq 2 \trace( {A}^{1/2} ).  $}
%\end{enumerate}
}

%\item $^*$ 
\exer{$^*$ Consider the  following minimization problem where $A \succ 0$:
\begin{align*}
 \min_X \ \ &  \trace(X^{-1}A) \\
  \text{subject to  } \ \  & X \succ 0 \\
 &  \trace(X) \le 1.
%\label{eqn:inv-trace-prob}
\end{align*}
Prove that its minimizer is given by $X = A^{1/2} / \trace(A^{1/2})$, and the minimum is obtained at $\trace^2(A^{1/2})$.
}

\end{exercises}	
%\end{enumerate}

%!TEX root = OCObook.tex

%\newcommand{\mc}{\ignore}

%%%%%%%%%%%%%%%%%%%%%%%%%%%%%%%%%%%%%%%%%%%%%%%%%%%%%%%%%%%%
%%%%%%%%%%%%%%%%%%%%%%%%%%%%%%%%%%%%%%%%%%%%%%%%%%%%%%%%%%%%
%  Bandit	 Convex Optimization
%%%%%%%%%%%%%%%%%%%%%%%%%%%%%%%%%%%%%%%%%%%%%%%%%%%%%%%%%%%%
%%%%%%%%%%%%%%%%%%%%%%%%%%%%%%%%%%%%%%%%%%%%%%%%%%%%%%%%%%%%
\chapter{Bandit	 Convex Optimization} \label{chap:bandits}

In many real-world scenarios the feedback available to the decision maker is noisy, partial or incomplete. Such is the case in online routing in data networks, in which an online decision maker iteratively chooses a path through a known network, and her loss is measured by the length (in time) of the path chosen.
In data networks, the decision maker can measure the RTD (round trip delay) of a packet through the network, but rarely has access to the congestion pattern of the entire network. 

Another useful example is that of online ad placement in web search. The decision maker iteratively chooses an ordered set of ads from an existing pool. Her reward is measured by the viewer's response---if the user clicks a certain ad, a reward  is generated according to the weight assigned to the particular ad. In this scenario, the search engine can inspect which ads were clicked through, but cannot know whether different ads, had they been chosen to be displayed, would have been clicked through or not. 

The examples above can readily be modeled in the OCO framework, with the underlying sets being the convex hull of decisions. The pitfall of the general OCO model is the feedback; it is unrealistic to expect that the decision maker has access to a gradient oracle at any point in the space for every iteration of the game.

\section{The Bandit Convex Optimization Setting} 

The Bandit Convex Optimization (short: BCO) model is identical to the general OCO model we have explored in previous chapters with the only difference being the feedback available to the decision maker. 

To be more precise, the BCO framework can be seen as a structured repeated game.  The protocol of this learning framework is as follows: 
At iteration $t$, the online player chooses
$\x_t \in \K.$ After committing to this choice, a convex cost
function $f_t \in \F : \K \mapsto \reals$ is revealed. Here $\F$ is the bounded family of cost functions available to the adversary. The cost incurred
to the online player is the value of the cost function at the
point she committed to $f_t(\x_t)$.  As opposed to the OCO model, in which the decision maker has access to a gradient oracle for $f_t$ over $\K$,  in BCO {\bf the loss $f_t(\x_t)$ is the only feedback available to the online player at iteration $t$.} In particular, the decision maker does not know the loss had she chosen a different point $\x \in \K$ at iteration $t$.

As before, let $T$ denote the total number of game iterations (i.e., predictions and their incurred loss). 
Let $\mA$ be an algorithm for BCO, which maps a certain game history to a decision in the decision set. We formally define the regret of $\mA$ that predicted $x_1,...,x_T$ to be 

$$ \regret_T(\mA) = \sup_{\{f_1,...,f_T\} \subseteq \F}  \left\{   \tsum_{t=1}^T f_t(\x_t) -\min_{\x \in \K} \tsum_{t=1}^T f_t(\x) \right\}.$$

\section{The Multiarmed Bandit (MAB) Problem}
\sectionmark{Multiarmed Bandits}

A classical model for decision making under uncertainty is the multiarmed bandit (MAB) model. The term MAB nowadays refers to a multitude of different variants and sub-scenarios that are too large to survey. This section addresses perhaps the simplest variant---the non-stochastic MAB problem---which is defined as follows: 

Iteratively, a decision maker chooses between $n$ different actions $i_t \in \{1,2,...,n\}$, while, at the same time, an adversary assigns each action a loss in the range $[0,1]$. The decision maker receives the loss for $i_t$ and observes this loss, and nothing else. The goal of the decision maker is to minimize her regret. 

The reader undoubtedly observes this setting is identical to the setting of prediction from expert advice,  the only difference being the feedback available to the decision maker: whereas in the expert setting the decision maker can observe the rewards or losses for all experts in retrospect, in the MAB setting, only the losses of the decisions actually chosen are known.

It is instructive to explicitly model this problem as a special case of BCO. Take the decision set to be the set of all distributions over $n$ actions, i.e., $\K = \Delta_n$ is the $n$-dimensional simplex. The loss function is taken to be the linearization of the costs of the individual actions, that is:
$$  f_t(\x) = \ell_t^\top \x =  \sum_{i=1}^n \ell_t(i) \x(i) \quad \forall \x \in \K , $$
where $\ell_t(i)$ is the loss associated with the $i$'th action at the $t$'th iteration. Thus, the cost functions are linear functions in the BCO model.

The MAB problem exhibits an exploration-exploitation tradeoff: an efficient (low regret) algorithm has to explore the value of the different actions in order to make the best decision. On the other hand, having gained sufficient information about the environment, a reasonable algorithm needs to exploit this action by picking the best action.

The simplest way to attain a MAB algorithm would be to separate exploration and exploitation. Such a method would proceed by 
\begin{enumerate}
\item
With some probability, explore the action space (i.e., by choosing an action uniformly at random). Use the feedback to construct an estimate of the actions' losses.
\item
Otherwise, use the estimates to apply a full-information experts algorithm as if the estimates are the true historical costs.  
\end{enumerate}

This simple scheme already gives a sublinear regret algorithm, presented in algorithm  \ref{alg:simpleMAB}.

\begin{algorithm}[ht]
\caption{Simple MAB algorithm}
\label{alg:simpleMAB}
\begin{algorithmic}[1]
\State Input: OCO algorithm $\mA$, parameter $\delta$.
\For{$t = 1$ to $T$ }	
	\State Let $b_t$ be a Bernoulli random variable that equals 1 with probability $\delta$. 
	\If { $b_t = 1$ } 
	\State Choose $i_t \in \{1,2,...,n\}$ uniformly at random and play $i_t$.  \\
	\State Let 
$$ \hat{\ell}_t(i)= \mycases{ \frac{n}{\delta} \cdot \ell_t (i_t)} { i = i_t} {0 } {\text{otherwise}} .$$
	\State Let $\fhat_t(\x)  = \hat{\ell}_t^\top \x$ and update
	$\x_{t+1} = \mA(\fhat_1,...,\fhat_t) $.
	\Else
	\State Choose $i_t \sim \x_t$ and play $i_t$. 
	\State Update $\hat{f}_t = 0, \lhat_t = \bzero$, $\x_{t+1} = \x_t $. %\mA(\fhat_1,...,\fhat_t) $. 
	\EndIf
	\EndFor
\end{algorithmic}
\end{algorithm}

\begin{lemma}
Algorithm \ref{alg:simpleMAB}, with $\mathcal{A}$ being the the online gradient descent algorithm, guarantees the following regret bound:
\begin{equation}
\E\left[\sum_{t=1}^T {\ell_t(i_t)}-\min_i{\sum_{t=1}^T {\ell_t(i)}}\right] \leq O( T^{\frac{2}{3}} n^{\frac{2}{3}} ) \nonumber
\end{equation}
\end{lemma}

\begin{proof}
For the random functions $\{\hat{\ell}_t\}$ defined in algorithm \ref{alg:simpleMAB},  notice that 
\begin{enumerate}
\item
$ \E[ \hat{\ell}_t (i) ]  = \Pr[ b_t = 1] \cdot \Pr[ i_t = i | b_t=1] \cdot \frac{n}{\delta} \ell_t(i) = \ell_t(i)  $.
\item
$ \| \hat{\ell}_t \|_2 \leq \frac{n}{\delta} \cdot |\ell_t(i_t)| \leq \frac{n}{\delta} $.
\end{enumerate}
Therefore the regret of the simple algorithm can be related to that of $\mathcal{A}$ on the estimated functions. 

On the other hand, the simple MAB algorithm does not always play according to the distribution generated by $\mathcal{A}$: with probability $\delta$ it plays uniformly at random, which may lead to a regret of one on these exploration iterations. Let $S_t \subseteq [T]$ be those iterations in which $b_t=1$. 
This is captured by the following lemma:
\begin{lemma} \label{lem:shalom3}
$$\E[ \ell_t(i_t) ] \leq \E [ \lhat_t^\top x_t ] + \delta $$
\end{lemma}
\begin{proof}
\begin{eqnarray*}
&\E [\ell_t(i_t)]\\
&= \Pr[b_t=1] \cdot \E [ \ell_t(i_t)|b_t=1] \\ & + \Pr[b_t = 0] \cdot \E[\ell_t(i_t)|b_t=0] \\
& \le \delta + \Pr[b_t = 0] \cdot \E [\ell_t(i_t)|b_t=0] \\
& = \delta + (1-\delta)  \E [ \ell_t^\top \x_t | b_t = 0 ]  & \mbox{ $b_t = 0 \rightarrow i_t \sim \x_t$,  independent of $l_t$} \\
& \leq  \delta +   \E [ \ell_t^\top \x_t ]  & \mbox{non-negative random variables } \\
& = \delta +   \E [ \lhat_t^\top \x_t ]  & \mbox{$\lhat_t$ is independent of $\x_t$} 
\end{eqnarray*}
\end{proof}
We thus have, 
\begin{eqnarray*}
& \E [ \regret_T ]  \\
& =  \E[ \sum_{t=1}^T{\ell_t(i_t)}-{\sum_{t=1}^T{\ell_t(i^\star)}}] \\
& = \E[ \sum_{t  }{\ell_t(i_t)}-{\sum_{t } {\hat{\ell}_t(i^\star)}}   ] & \mbox{ $i^\star$ is indep. of $\hat{\ell}_t$} \\
& \leq \E[ \sum_{t  }{\hat{\ell}_t(\x_t)}-\min_i{\sum_{t } {\hat{\ell}_t(i)}} ]   + \delta T & \mbox{Lemma \ref{lem:shalom3} } \\
& =  \E[ \regret_{S_T}(\mathcal{A}) ] + \delta \cdot T  \\
& \leq \frac{3}{2} GD  \sqrt{\delta T} + \delta \cdot T & \mbox{ Theorem \ref{thm:gradient}}, \E[ |S_T|] = \delta T  \\
& \leq 3 \frac{n}{ \sqrt{\delta}}  \sqrt{T  } + \delta \cdot T & \mbox{ For $\Delta_n$,  $D \leq 2$ , $\|\lhat_t\|\leq \frac{n}{\delta}  $}  \\
& =  O( T^{\frac{2}{3}} n^{\frac{2}{3}}) . & \delta= n^{\frac{2}{3}}  T^{-\frac{1}{3}} 
\end{eqnarray*}
\end{proof}

\subsection{EXP3: simultaneous exploration and exploitation}

The simple algorithm of the previous section can be improved by combining the exploration and exploitation steps. This gives a near-optimal regret algorithm, called EXP3, presented below. 

\begin{algorithm}[ht]
\caption{EXP3 - simple version}
\label{alg:EXP3}
\begin{algorithmic}[1]
\State Input: parameter $\epsilon > 0$.  Set $\x_1 = ({1}/{n}) \bone$.
\For{$t \in \{1,2,...,T\}$}	
	\State Choose $i_t \sim \x_t$ and play $i_t$. 
	\State Let 
$$ \hat{\ell}_t(i)= \mycases{ \frac{1}{\x_t(i_t)} \cdot \ell_t (i_t)} { i = i_t} {0 } {\text{otherwise}} $$
	\State Update
	%$\y_{t+1} (i)  = \x_t(i) e^{\eps \hat{\ell}_t(i)}  \ , \ \x_{t+1} = (1 - \frac{1}{\sqrt{T}} )\frac{\y_{t+1} }{\|\y_{t+1}\|_1 } + \frac{1}{n\sqrt{T}} \bone $
	$\y_{t+1} (i)  = \x_t(i) e^{-\eps \hat{\ell}_t(i)}  \ , \ \x_{t+1} = \frac{\y_{t+1} }{\|\y_{t+1}\|_1 } $
		\EndFor
\end{algorithmic}
\end{algorithm}

As opposed to the simple multiarmed bandit algorithm, the EXP3 algorithm explores every iteration by always creating an unbiased estimator of the entire loss vector. This results in a possibly large magnitude of the vectors $\hat{\ell}$ and a large gradient bound for use with online gradient descent. However, the large magnitude vectors are created with low probability (proportional to their magnitude),  which allows for a finer analysis. 

Ultimately, the EXP3 algorithm attains a worst case regret bound of $O(\sqrt{T n \log n})$, which is nearly optimal (up to a logarithmic term in the number of actions). 

\begin{lemma} \label{Lemma:exp3regret}
Algorithm \ref{alg:EXP3} with non-negative losses and $\epsilon = \sqrt{\frac{\log n}{T n} }$ guarantees the following regret bound:
\begin{equation}
\E[\sum{\ell_t(i_t)}-\min_i{\sum{\ell_t(i)}}] \leq 2 \sqrt{ T n \log n }  .\nonumber
\end{equation}
\end{lemma}

\begin{proof}
For the random losses $\{\hat{\ell}_t\}$ defined in algorithm \ref{alg:EXP3},  notice that 
	\begin{eqnarray}
		& \E[ \hat{\ell}_t (i) ]  = \Pr[ i_t = i] \cdot  \frac{ \ell_t(i)}{ \x_t(i) }  =  \x_t(i) \cdot  \frac{ \ell_t(i)}{ \x_t(i) }  = \ell_t(i) . \notag \\
		& \E [ \x_t^\top \hat{\ell}_t^2 ]  = \sum_i \Pr[ i_t = i] \cdot  \x_t(i)  \hat{\ell}_t(i)^2 \notag \\
		& =  \sum_i   \x_t(i)^2  \hat{\ell}_t(i)^2 
		 = \sum_i \ell_t(i)^2  \leq  {n} . \label{eqn:shalom1234}
	\end{eqnarray}
Therefore we have $E[\hat{f}_t]=f_t$, and the expected regret with respect to  the functions $\{\hat{f}_t\}$ is equal to that with respect to  the functions $\{f_t\}$.
Thus, the regret with respect to $\hat{\ell}_t$ can be related to that of $\ell_t$. 

The EXP3 algorithm applies Hedge to the losses given by $\hat{\ell}_t$, which are all non-negative and thus satisfy the conditions of Theorem \ref{lem:hedge}. Thus, the expected regret with respect to  $\hat{\ell}_t$, can be bounded by,

	\begin{eqnarray*}
		& \E [ \regret_T ]   =   \E[ \sum_{t=1}^T{\ell_t(i_t)}-\min_i{\sum_{t=1}^T{\ell_t(i)}}] \\
		& =  \E[ \sum_{t=1}^T{\ell_t(i_t)}-{\sum_{t=1}^T{\ell_t(i^\star)}}] \\
		& \leq \E[ \sum_{t=1}^{T} {\hat{\ell}_t(\xv)}-{\sum_{t=1}^{T} {\hat{\ell}_t(i^\star)}}  ] & \mbox{ $i^\star$ is indep. of $\hat{\ell}_t$} \\
		& \leq \E [ \eps \sum_{t=1}^T \sum_{i=1}^n \hat{\ell}_t(i)^2 \x_t(i) + \frac{\log n}{\eps} ]  & \mbox{ Theorem \ref{lem:hedge} }  \\
		& \leq \eps T n + \frac{\log n}{\epsilon}  & \mbox{ equation \eqref{eqn:shalom1234}  }  \\
		& \leq  2 \sqrt{T n \log n }. & \mbox { by choice of $\epsilon$ }
	\end{eqnarray*}
\end{proof}

We proceed to derive an algorithm for the more general setting of bandit convex optimization that attains near-optimal regret.

\sectionmark{From Limited to Full Information} 
\section{A Reduction from Limited Information to Full Information} 
\sectionmark{From Limited to Full Information} 

In this section we derive a low regret algorithm for the general setting of bandit convex optimization. In fact, we shall describe a general technique for designing bandit algorithms, which is composed of two parts:
\begin{enumerate}
\item
A general technique for taking an online convex optimization algorithm that uses only the gradients of the cost functions (formally defined below), and applying it to a family of vector random variables with carefully chosen properties.

\item
Designing the random variables that allow the template reduction to produce meaningful regret guarantees.  
\end{enumerate}

We proceed to describe the two parts of this reduction, and in the remainder of this chapter we describe two examples of using this reduction to design bandit convex optimization algorithms.

\subsection{Part 1: using unbiased estimators} 

The key idea behind many of the efficient algorithms for bandit convex optimization  is the following:  although  we cannot calculate $ \nabla f_t(\x_t)$ explicitly, it is possible to find an \emph{observable} random variable $\gv$ that satisfies $\E[\gv] \approx \nabla f_t (\x_t) = \nabla_t $. Thus, $\gv$  can be seen as an estimator of the gradient. By substituting $\gv$ for $\nabla_t$ in an OCO algorithm, we will show that many times it retains its sublinear regret bound. 

Formally, the family of regret minimization algorithms for which this reduction works is captured in the following definition. 
\begin{definition}(\textbf{first order OCO Algorithm})
Let $\mA$ be an OCO (deterministic) algorithm receiving an arbitrary sequence of  differential  loss functions $f_1,\ldots,f_T$,
and producing decisions $\x_1 \gets \mA (\emptyset), \x_t \gets \mA(f_1,\ldots,f_{t-1})$.  
$\mA$ is called a \emph{first order online algorithm} if the following holds:
\begin{itemize}
\item The family of loss functions $\mathcal{F}$ is closed under addition of linear functions: if $f\in \mathcal{F}$ and $\uv\in \reals^n$ then  $f+ \uv^\top \x  \in \mathcal{F}$.
\item Let $\hat{f}_t$ be the linear function $ \hat{f}_t(\x) = \nabla f_t(\x_t) ^\top \x $, then for every  iteration $t\in[T]$:
$$\mA(f_1,\ldots,f_{t-1}) = \mA(\hat{f}_1,...,\hat{f}_{t-1}) $$
\end{itemize}
\end{definition}

We can now consider a formal reduction from any first order online algorithm to a bandit convex optimization algorithm as follows. 

\begin{algorithm}
	[ht] \caption{Reduction to bandit feedback.} \label{alg:reductionBCO2OCO} \label{BCO2OCO}
	\begin{algorithmic}[1]
		\State Input: convex set $\K \subset \reals^n$,  first order online algorithm $\mA$.
		\State Let  $\x_1 = \mA( \emptyset ) $.
        \For{$t=1$ to $T$}
		\State Generate distribution $\D_t$, sample  $\y_t \sim \D_t$ with $\E[\y_t] = \x_t$. 
		\State Play $\y_t$.
		\State Observe  $f_t(\y_t)$, generate $\gv$ with $\E[\gv] = \nabla f_t (\x_t) $.
		\State Let $\xv[t+1]   = \mA(\gv[1],...,\gv ) $.
		\EndFor
	\end{algorithmic}
\end{algorithm}

Perhaps surprisingly, under very mild conditions the reduction above guarantees the same regret bounds as the original first order algorithm up to the magnitude of the estimated gradients. This is captured in the following lemma. 
\begin{lemma} \label{Lemma:Flaxman_FirstOrderAlgos}
Let $\uv$ be a \emph{fixed} point in $\K$. Let $f_1,\ldots,f_T:\K \to \reals$ be a  sequence of differentiable  functions. 
Let $\mA$ be a first order online algorithm that ensures a regret bound of the form $\regret_T({\mA}) \leq B_{\mA}( \nabla f_1(\x_1),\ldots,\nabla f_T(\x_T))$ in the full information setting.  Define the points  $\{ \x_t \}$ 
as: $\x_1\gets\mA(\emptyset) $, $\x_t \gets \mA(\gv[1],\ldots,\gv[t-1])$ where each $\gv$ is a vector valued random variable such that:
$$\E[\gv\big \vert \x_1,f_1,\ldots, \x_t,f_t]=\nabla f_t(\x_t) . $$
Then the following holds for all $\uv \in \K$:
\begin{align*}
\E[\sum_{t=1}^T f_t(\x_t)] - \sum_{t=1}^T f_t(\uv) \leq  \E[B_{\mA}(\gv[1],\ldots,\gv[T])] .
\end{align*}
\end{lemma}
\begin{proof}
Define the functions  $h_t:\K\to\reals$ as follows:
$$h_t(\x) = f_t(\x) + \boldsymbol\xi_t^\top  \x, \; \text{where } \boldsymbol\xi_t = \gv -\nabla f_t(\x_t).$$
Note that
 $$\nabla h_t(\x_t) =\nabla f_t(\x_t)+  \gv -\nabla f_t(\x_t)=\gv.$$
Therefore, deterministically applying a  first order method $\mA$ on the random functions $h_t$ is equivalent to applying $\mA$ on a stochastic first order approximation of the deterministic functions $f_t$.
Thus by the full-information regret bound of $\mA$ we have:
\begin{align}\label{equation:regretBeforeExpectation}
 \sum_{t=1}^T h_t(\x_t) - \sum_{t=1}^T h_t(\uv) \leq  B_{\mA}(\gv[1],\ldots,\gv[T]).
\end{align}
Also note that:
\begin{align*}
\E[h_t(\x_t)]&=\E[f_t(\x_t)]+\E[\boldsymbol\xi_t^\top  \x_t] \\
& = \E[f_t(\x_t)]+\E[\E[\boldsymbol\xi_t^\top \x_t\big\vert \x_1,f_1,\ldots,\x_t,f_t] ] \\
&= \E[f_t(\x_t)]+\E[\E[\boldsymbol\xi_t \big\vert \x_1,f_1,\ldots,\x_t,f_t] ^\top \x_t] \\
& = \E[f_t(\x_t)].
\end{align*}
where we used $\E[\boldsymbol\xi_t\vert \x_1,f_1,\ldots,\x_t,f_t]=0$. Similarly, since  $\uv\in\K$ is fixed we have that $\E[h_t(\uv)] = f_t(\uv)$. The lemma follows from taking the expectation of Equation \eqref{equation:regretBeforeExpectation}.
\end{proof}

\subsection{Part 2: point-wise gradient estimators}

In the preceding part we have described how to convert a first order algorithm for OCO to one that uses bandit information, using specially tailored random variables. We now describe how to create these vector random variables. 

Although  we cannot calculate $ \nabla f_t(\x_t)$ explicitly, it is possible to find an \emph{observable} random variable $\gv$ that satisfies $\E[\gv] \approx \nabla f_t$, and serves as an estimator of the gradient. % By using $\gv$ instead of $\nabla_t$ in an OCO algorithm, we will show that in many cases the algorithm retains its sublinear regret bound. 

The question is how to find an appropriate $\gv$, and in order to answer it we begin with an example in a 1-dimensional case.

\begin{example}
A 1-dimensional gradient estimate\newline
Recall the definition of the derivative:
\begin{equation}
\label{derivative}
f'(x)=\lim_{\delta \rightarrow 0}{\frac{f(x+\delta)-f(x-\delta)}{2 \delta}}. \nonumber
\end{equation}
The above shows that for a 1-dimensional derivative, two evaluations of $f$ are required. Since in our problem we can perform only one evaluation, let us define $g(x)$ as follows:
\begin{equation}
g(x) = \mycases {\frac{f(x+\delta)}{\delta}} {\text{with probability }  \frac{1}{2}} { - \frac{f(x-\delta)}{\delta}} { \text{with probability }  \frac{1}{2}}.
\label{gt}
\end{equation}
It is clear that
\begin{equation*}
\E[g(x)]={\frac{f(x+\delta)-f(x-\delta)}{2 \delta}}. 
\end{equation*}
Thus, {\bf in expectation}, for small $\delta$,  $g(x)$  approximates $f'(x)$.
\end{example}

\subsubsection{The sphere sampling estimator} 

We will now show how the gradient estimator \eqref{gt} can be extended to the multidimensional case. Let $\x\in \mathbb{R}^n$, and let $B_{\delta}$ and $S_{\delta}$ denote the $n$-dimensional ball and sphere with radius $\delta:$
\begin{equation*}
B_{\delta}=\left\{\x|\left\|\x\right\| \leq \delta \right\},
\end{equation*}
\begin{equation*}
S_{\delta}=\left\{\x|\left\|\x\right\| = \delta \right\}.
\end{equation*}

\noindent We define $\hat{f}(\x)= \hat{f}_\delta(\x)$ to be a $\delta$-smoothed version of $f(\x)$:
\begin{equation}
 \label{fhat}
\hat{f}_\delta \left(\x\right)=\E_{\vv\in \ball }\left[f\left(\x+\delta \vv \right)\right],
\end{equation}
where $\vv$ is drawn from a uniform distribution over the unit ball.
This construction is very similar to the one used in Lemma \ref{lem:SmoothingLemma} in context of convergence analysis for convex optimization. However, our goal here is very  different.

Note that when $f$ is linear, we have $\hat{f}_\delta(\x)=f(\x)$. We shall address the case in which  $f$ is indeed linear as a special case, and show how to estimate the gradient of $\hat{f}(\x)$, which, under the assumption, is also the gradient of $f(\x)$.
The following lemma shows a simple relation between the gradient $\nabla \hat{f}_\delta$ and a uniformly drawn unit vector.
\begin{lemma}
\label{lem_stokes}
Fix $\delta>0$. Let  $\hat{f}_\delta(\x)$  be as defined in \eqref{fhat}, and let $\uv$ be a uniformly drawn unit vector $\uv\sim \sphere$. Then
\[
\E_{\uv\in \sphere}\left[f\left(\x+\delta \uv \right) \uv\right]=\frac{\delta}{n}\nabla\hat{f}_\delta \left( \x\right).
\]
\end{lemma}
\begin{proof}
Using Stokes' theorem from calculus, we have
\begin{equation}
\nabla\underset{B_{\delta}}{\int}f\left(\x+\vv\right)d \vv=\underset{S_{\delta}}{\int}f\left(\x+\uv\right)\frac{\uv}{\left\Vert \uv \right\Vert }d \uv .\label{stokes}
\end{equation}
From \eqref{fhat}, and by definition of expectation, we have
\begin{equation}
\hat{f}_\delta(\x)=\frac{\underset{B_{\delta}}{\int}f\left(\x+ \vv \right)d \vv}{\vol( B_{\delta})} . \label{vol1}
\end{equation}
where $\vol(B_{\delta})$ is the volume of an n-dimensional ball
of radius $\delta$. Similarly,
\begin{equation}
\E_{\uv \in S}\left[f\left(\x+\delta \uv \right)\uv \right]=\frac{\underset{S_{\delta}}{\int}f\left(\x+ \uv \right)\frac{\uv }{\left\Vert \uv \right\Vert }du}{\vol (S_{\delta} ) }  . \label{vol2}
\end{equation}
Combining \eqref{fhat}, \eqref{stokes}, \eqref{vol1}, and \eqref{vol2}, and the fact that the ratio of the volume of a ball in $n$ dimensions and the  sphere of dimension $n-1$ is $\textrm{vol}_{n}B_{\delta}/\textrm{vol}_{n-1}S_{\delta}=\delta/n$
gives the desired result.
\end{proof}
Under the assumption that $f$ is linear, Lemma \ref{lem_stokes} suggests a simple estimator for the gradient $\nabla f$. Draw a random unit vector $\uv$, and let $g\left(\x\right)=\frac{n}{\delta}f\left(\x+\delta \uv \right)\uv$.

\subsubsection{The ellipsoidal sampling estimator} 

The sphere estimator above is at times difficult to use: when the center of the sphere is very close to the boundary of the decision set only a very small sphere can fit completely inside. This results in a gradient estimator with large variance. 

In such cases, it is useful to consider ellipsoids rather than spheres. Luckily, the generalisation to ellipsoidal sampling for gradient estimation is a simple corollary of our derivation above: 

\begin{corollary} \label{Corollary:Gradient_Estimate_SinglePoint}
Consider a continuous function $f:\reals^n\to \reals$, an invertible matrix $A\in \reals^{n \times n}$, 
and let $\vv\sim \ball^n$ and $\uv \sim \sphere^n$.
Define the smoothed version of $f$ with respect to $A$:
\begin{align*} 
\hat{f}(\x) = \E[ f(\x+A \vv)  ].
\end{align*}
Then the following holds:
\begin{align*}
\nabla \hat{f}(\x) = n  \E[ f(\x+A \uv) A^{-1} \uv ].
\end{align*}
\end{corollary}
\begin{proof}
Let $g(\x) = f(A \x)$, and $\hat{g}(\x) = \E_{\vv \in \ball} [g(\x + \vv)]$. 
\begin{eqnarray*}
n  \E[ f(\x+A \uv) A^{-1} \uv ] & = n A^{-1}  \E[ f(\x+A \uv)  \uv ] \\
& = n A^{-1}  \E[ g ( A^{-1} \x+ \uv)  \uv ] \\
& = A^{-1}   \nabla \hat{g}(A^{-1} \x)  & \mbox { Lemma \ref{lem_stokes} } \\
& = A^{-1}  A   \nabla \hat{f}( \x)  = \nabla \hat{f}(\x). 
\end{eqnarray*}
\end{proof}

\sectionmark{OGD without a gradient} 
\section{Online Gradient Descent without a Gradient} 
\sectionmark{OGD without a Gradient} 

The simplest and historically earliest application of the BCO-to-OCO reduction outlined before is the application of the online gradient descent algorithm to the bandit setting. The FKM algorithm (named after its inventors, see bibliographic section) is outlined in algorithm \ref{FKM_alg}.

For simplicity, we assume that the set $\K$ contains the unit ball centered at the zero vector, denoted $\bzero$. Denote  $\K_\delta = \{ \x \ | \  \frac{1}{1-\delta} \x  \in \K  \} $.  It is left as an exercise to show that $\K_\delta$ is convex for any $0 <  \delta < 1 $ and that all  balls of radius $\delta$ around points in $\K_\delta$ are contained in $\K$.  

We also assume for simplicity that the adversarially chosen cost functions are bounded by one over $\K$, i.e., that $| \fv(\x) | \leq 1$ for all $\x \in \K$.

\begin{figure}[ht]
\begin{center}
\includegraphics[width=3.5in]{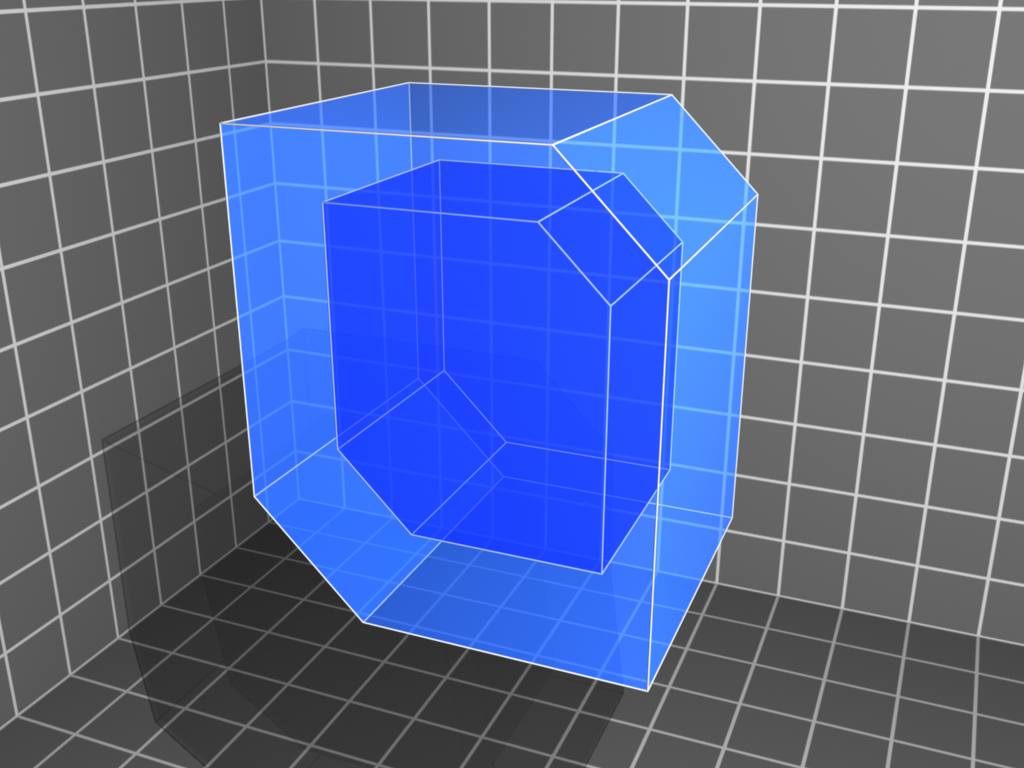}
\end{center}
\caption{The Minkowski set $\K_\delta$ \label{fig:Minkowski}}
\end{figure}

\begin{algorithm}[ht]
\caption{FKM Algorithm \label{FKM_alg}}
\label{algFKM}
\begin{algorithmic}[1]
\State Input: decision set $\K$ containing $ \bzero $, set $\x_1 = \bzero$, parameters $\delta,\eta$. 
\For{$t=1$ to $T$}	
\State Draw $\uv_t \in \sphere_1$ uniformly at random, set $\y_t  = \x_t + \delta \uv_t $.
\State Play $\y_t$, observe and incur loss $f_t \left( \y_t \right)$. Let $\gv   =  \frac{n}{\delta} f_{t}\left(\y_{t}\right)\uv_{t}$.
\State Update $\x_{t+1}=  \underset{\K_\delta}{\proj}\left[\x_{t}- \eta \gv \right] $.
\EndFor
\end{algorithmic}
\end{algorithm}

The FKM algorithm is an instantiation of the generic reduction from bandit convex optimization  to online convex optimization with spherical gradient estimators over the set $\K_\delta$.  It iteratively projects onto $\K_\delta$, in order to have enough space for spherical gradient estimation. This degrades its performance by a controlled quantity. Its regret is bounded as follows.

\begin{theorem}
\label{FKM_prop}
Algorithm \ref{algFKM} with parameters $\ \eta = \frac{D}{n T^{3/4} } , \delta = \frac{1}{T^{1/4}}  $ guarantees the following expected regret bound
\[
\sum_{t=1}^T \E [ f_{t}(\y_{t}) ]-\min_{\x \in\mathcal{K}} \sum_{t=1}^T  f_{t} (\x)  \leq 9 n D G T^{3/4}  = O(T^{3/4}) .
\]
\end{theorem}

\begin{proof}
Recall our notation of $\x^\star = \argmin_{\x \in \K} \sum_{t=1}^T f_t(\x)$. Denote 
\[
\x_{\delta}^{\star}= \proj_{\K_\delta} (\x^\star ) .
\]
Then by properties of projections we have $\|\x_\delta^\star - \x^\star\| \leq \delta D$, where $D$ is the diameter of $\K$. Thus, 
assuming that the cost functions $\{f_t\}$ are $G$-Lipschitz, we have 
\begin{equation} \label{Lip_step}
\sum_{t=1}^T \E [ f_{t}(\y_{t}) ]-  \sum_{t=1}^T  f_{t} (\x^\star)  
 \leq  \sum_{t=1}^T \E [ f_{t}(\y_{t}) ]-  \sum_{t=1}^T  f_{t} (\x_\delta^\star)  +  \delta T G D .
 \end{equation}

Denote  $\hat{f}_t = \hat{f}_{\delta,t} = \E_{\uv \sim \ball} [f(\x + \delta \uv) ] $ for shorthand. We can now bound the regret by
\begin{align*}
& \sum_{t=1}^T \E [ f_{t}(\y_{t}) ]-  \sum_{t=1}^T  f_{t} (\x^\star) \\
& \leq \sum_{t=1}^T \E [ f_{t}(\x_{t}) ]-  \sum_{t=1}^T  f_{t} (\x^\star) + \delta D GT & \mbox{$f_t$ is $G$-Lipschitz } \\
& \leq \sum_{t=1}^T \E [ {f}_{t}(\x_{t}) ]- \sum_{t=1}^T  {f}_{t} (\x^\star_\delta) +  2 \delta D G  T & \mbox{Inequality (\ref{Lip_step}})  \\
& \leq \sum_{t=1}^T \E [ \hat{f}_{t}(\x_{t}) ]- \sum_{t=1}^T  \hat{f}_{t} (\x^\star_\delta) + 4 \delta D G T  & \mbox{ Lemma \ref{lem:SmoothingLemma} } \\
& \leq \regret_{OGD}( \gv[1] , ..., \gv[T] ) + 4 \delta D G T & \mbox{ Lemma \ref{Lemma:Flaxman_FirstOrderAlgos} } \\
& \leq  \eta \sum_{t=1}^T \| \gv\|^2 + \frac{D^2}{\eta}  +  4 \delta D G T & \mbox{ OGD regret, Theorem \ref{thm:gradient} } \\
& \leq  \eta \frac{n^2}{\delta^2} T  + \frac{D^2}{\eta}  +  4 \delta D G T & |\fv(\x)| \leq 1  \\
& \leq  9 n D G T^{3/4}  . & \eta = \frac{D}{n T^{3/4} } , \delta = \frac{1}{T^{1/4}} 
\end{align*}

\end{proof}

\sectionmark{Optimal Regret for BLO}
\section{* Optimal Regret Algorithms for Bandit Linear Optimization}
\sectionmark{Optimal Regret for BLO}

A special case of BCO that is of considerable interest is BLO---Bandit Linear Optimization. This setting has linear cost functions, and captures the network routing and ad placement examples discussed in the beginning of this chapter, as well as the non-stochastic MAB problem. 

In this section we give near-optimal regret bounds for BLO using techniques from interior point methods for convex optimization. 

The generic OGD method of the previous section suffers from three pitfalls: 
\begin{enumerate}
\item
The gradient estimators are biased, and estimate the gradient of a smoothed version of the real cost function. 
\item
The gradient estimators require enough ``wiggle room'' and are thus ill-defined on the boundary of the decision set. 
\item
The gradient estimates have potentially large magnitude, proportional to the distance from the boundary. 
\end{enumerate}

Fortunately, the first issue is non-existent for linear functions - the gradient estimators turn out to be unbiased for linear functions. In the notation of the previous chapters, we have for linear functions:
$$ \hat{f}_\delta(\x)  = \E_{\vv \sim \ball} [f(\x + \delta \vv) ] = f(\x) .$$ 
Thus,  Lemma \ref{lem_stokes} gives us a stronger guarantee:
\[
\E_{\uv\in \sphere}\left[f\left(\x+\delta \uv \right) \uv\right]=\frac{\delta}{n}\nabla\hat{f}_\delta \left( \x\right) = \frac{\delta}{n} \nabla f(\x) .
\]

To resolve the second and third issues we use self-concordant barrier functions, a rather advanced technique from interior point methods for convex optimization. 

\subsection{Self-concordant barriers} 

Self-concordant barrier functions were devised in the context of interior point methods for optimization as a way of ensuring that the Newton method converges in polynomial time over bounded convex sets. In this brief introduction we survey some of their beautiful properties that will allow us to derive an optimal regret algorithm for BLO.

\begin{definition}
Let $\K\in \reals^n$ be a convex set with a nonempty interior $\text{int}(\K)$.
 A function $\R:\text{int}(\K)\to \reals$ is called $\nu$-self-concordant if:
\begin{enumerate}
\item $\R$ is three times continuously differentiable and convex, and 
approaches infinity along any sequence of points approaching the boundary of $\K$.
\item For every $\h \in \reals^n$ and $\x\in \text{int}(\K)$ the following holds:
\begin{align*}
&|\nabla^3\R(\x)[\h,\h,\h] |\leq 2( \nabla^2\R(\x)[\h,\h])^{3/2} ,\\
&|\nabla\R(\x)[\h] |\leq \nu^{1/2}( \nabla^2\R(\x)[\h,\h])^{1/2}
\end{align*}
\end{enumerate}
\end{definition}
where the third order differential is defined as:
\begin{align*}
\nabla^3\R(\x)[\h,\h,\h] \equaltri   \left.  \frac{\partial^3}{\partial t_1 \partial t_2 \partial t_3}  \R(\x+t_1 \h+t_2 \h+t_3 \h)\right \vert_{t_1=t_2=t_3=0}
\end{align*}
The Hessian of a self-concordant barrier induces a local norm at every $\x \in \text{int}(\K)$, 
we denote this norm by $||\cdot||_\x$ and its dual by $||\cdot||_\x^{*},$ which are defined $\forall \h \in \reals^n$ by
\begin{align*}
\|\h\|_\x = \sqrt{\h^\top \nabla^2\R(\x) \h}, \qquad \|\h\|_\x^{*} = \sqrt{\h^\top (\nabla^2\R(\x))^{-1} \h}. 
\end{align*}
We assume that $\nabla^2\R(\x)$ always has full rank.  In BCO applications this is easy to ensure by adding a fictitious quadratic function to the barrier, which does not affect the overall regret by more than a constant.

Let $\R$ be a self-concordant barrier and $\x \in \text{int}(\K)$. The \emph{Dikin ellipsoid} is
\begin{align*} %\label{Definition:Dikin_ellipsoid}
\ellipsoid_1(\x) :=\{\y\in\reals^n : \|\y-\x \|_\x\leq 1\}, 
\end{align*}
i.e., the $\|\cdot\|_\x$-unit ball centered around $\x$, is completely contained in $\K$.  
%In addition, the local norms defined by the Hessians of all points within the Dikin ellipsoid are very similar. This is captured in the following Lemma. 
%\begin{lemma} \label{Lemma:HessiansInsideDikin}
%Let $\y \in \ellipsoid_{1/2} (\x)$ be strictly inside the Dikin ellipsoid of $\x$, such that $\|\y-\x\|_\x \leq \frac{1}{2} $. Then 
%$$ \frac{1}{4}  \nabla^2  \R(\x)  \preceq \nabla^2 \R(\y) \preceq 4 \nabla^2 \R(\x)  . $$
%\end{lemma}

In our next analysis we will need to bound $\R(\y) - \R(\x)$ for $\x,\y\in \text{int}(\K)$, for which the following lemma is useful:
%In the analysis of our regret bound we will require to  bound $\R(y) - \R(x)$ for $x,y\in \text{int}(\K)$, the following Lemma supplies us with such a bound:
\begin{lemma} \label{Lemma:MinkowskiBarrier}
Let $\R$ be a $\nu$-self concordant function over $\K$, then for all $\x,\y \in \text{int}(\K) $:
$$\R(\y)-\R(\x)\leq \nu \log \frac{1}{1-\pi_{\x}(\y)},  $$
where  $\pi_\x(\y) = \inf\{t\geq 0: \x+t^{-1}(\y-\x)\in\K \} .$
\end{lemma}
The function  $\pi_\x(\y)$ is called the Minkowski function for $\K$, and  its output is always in the interval  $[0,1]$. Moreover, 
as $y$ approaches the boundary of $\K$ then $\pi_\x(\y)\to 1$.

Another important property of self-concordant functions is the relationship between a point and the optimum, and the norm of the gradient at the point, according to the local norm, as given by the following lemma.
\begin{lemma} \label{Lemma:DistanceAndGradients}
Let $\x \in \text{int}(\K)$  be such that  $\|\nabla \R(\x) \| _\x^*  \leq \frac{1}{4} $, and let $\x^\star = \argmin_{\x \in \K} \R(\x) $. Then 
$$ \| \x - \x^\star\|_x \leq  2  \|\nabla \R(\x) \| _\x^* . $$
\end{lemma}

\subsection{A near-optimal algorithm} 

We have now set up all the necessary tools to derive a near-optimal BLO algorithm, presented in algorithm \ref{alg:scrible}.

\begin{algorithm}[H]
\label{alg:egmincut}
\begin{algorithmic}[1]
\caption{SCRIBLE} \label{alg:scrible}
\State Input: decision set $\K$ with self concordant barrier $\R$, set $\x_1 \in \text{int}(\K)$ such that $\nabla \R(\x_1) = 0$, parameters $\eta,\delta$. 
\For{$t =1$ to $T$}	
\State Let $\A = \left[\nabla^2 \R(\x_t) \right]^{-1/2}$ .
\State Pick $\uv_t \in \sphere$ uniformly, and set $\y_t  = \x_t +  \A \uv_t $.
\State Play $\y_t$, observe and suffer loss $f_t \left( \y_t \right)$. let $\gv   =  n f_{t}\left(\y_{t}\right) \A^{-1 }\uv_{t}$.
\State \label{line:rftl} Update 
$$\x_{t+1}= \argmin_{\x \in \K_\delta} \left\{ \eta \sum_{\tau =1 }^t \gv[\tau]^\top \x +  \R(\x)  \right\} .$$
\EndFor
\end{algorithmic}
\end{algorithm}

\begin{theorem}
For appropriate choice of $\eta,\delta$, the SCRIBLE algorithm guarantees
\[
\sum_{t=1}^T \E [ f_{t}(\y_{t}) ]-\min_{\x \in\mathcal{K}} \sum_{t=1}^T  f_{t} (\x)  \leq O\left(\sqrt{T} \log T \right).
\]
\end{theorem}

\begin{proof}
First, we note that $\y_t \in \K$ never steps outside of the decision set. The reason is that $\x_t \in \K$ and $\y_t$ lies in the Dikin ellipsoid centered at $\x_t$. 

Further, by Corollary \ref{Corollary:Gradient_Estimate_SinglePoint}, we have that 
$$ \E [ \gv] = \nabla \hat{f}_t (\x_t)  = \nabla f_t(\x_t), $$
where the latter equality follows since $f_t$ is linear, and thus its smoothed version is identical to itself. 

A final observation is that line \ref{line:rftl}  in the algorithm is an invocation of the RFTL algorithm with the self-concordant barrier $\R$ serving as a regularisation function. The RFTL algorithm for linear functions is a first order OCO algorithm and thus Lemma \ref{Lemma:Flaxman_FirstOrderAlgos} applies. 

We can now bound the regret by
\begin{align*}
& \sum_{t=1}^T \E [ f_{t}(\y_{t}) ]-  \sum_{t=1}^T  f_{t} (\x^\star) \\
& \leq \sum_{t=1}^T \E [ \hat{f}_{t}(\x_{t}) ]- \sum_{t=1}^T  \hat{f}_{t} (\x^\star)  & \mbox{ $\hat{f}_t = f_t$, $\E[\y_t] = \x_t $  } \\
& \leq \regret_{RFTL}( \gv[1] , ..., \gv[T])   & \mbox{ Lemma \ref{Lemma:Flaxman_FirstOrderAlgos} } \\
& \leq   \sum_{t=1}^T  \gv^\top (\xv - \xv[t+1])  + \frac{\R(\x^\star) - \R( \x_1)}{\eta}   & \mbox{ Lemma \ref{lem:FTL-BTL}} \\
& \leq   \sum_{t=1}^T  \| \gv\|_{t}^* \| \xv - \xv[t+1] \|_{t}   + \frac{\R(\x^\star) - \R( \x_1)}{\eta}  . & \mbox{ Cauchy-Schwarz} 
%& \leq  2 \eta \sum_{t=1}^T \| \gv\|_t^{*  \ 2} + \frac{\R(\x^\star) - \R( \x_1)}{\eta}   & \mbox{ Theorem \ref{thm:RFTLmain1}} \\
%& \leq  2 \eta n^2 T  + \frac{\R(\x^\star) - \R( \x_1) }{\eta}   & \mbox {$\|\gv\|^{* \ 2}_t \leq  n^2 $} .
\end{align*}
Here we use our notation from the previous chapter for the local norm $\| \h \|_t = \| \h \|_{\x_t} = \sqrt{\h^\top \nabla^2 \R(\x_t) \h}$.

To bound the last expression, we use Lemma \ref{Lemma:DistanceAndGradients}, and the definition of $\xv[t+1] = \argmin_{\x \in \K} \Phi_t(\x) $ where $\Phi_t(\x) =  \eta \sum_{\tau =1 }^t \gv[\tau]^\top \x +  \R(\x) $ is a self-concordant barrier. Thus, 
$$ \| \xv - \xv[t+1] \|_{t} \leq  2 \| \nabla \Phi_t(\xv) \|_{t}^{*} = 2 \|  \nabla \Phi_{t-1}(\xv) +  \eta \gv \| _{t}^* = 2 \eta \| \gv \|_{t}^* , $$
since $\nabla \Phi_{t-1}(\xv) = 0$ by definition of $\xv$. Recall that to use Lemma \ref{Lemma:DistanceAndGradients}, we need $\| \nabla \Phi_t(\xv)\|_{t}^* = \eta \| \gv\|_{t} ^* \leq \frac{1}{4}$, which is true by choice of $\eta$ and since 
$$\|\gv\|^{* \ 2}_{t} \leq n^2 \uv_t^T \A^{-T} \nabla^{-2} \R(\x_t) \A^{-1} \uv_t \leq n^2 . $$
Thus,
\begin{align*}
& \sum_{t=1}^T \E [ f_{t}(\y_{t}) ]-  \sum_{t=1}^T  f_{t} (\x^\star)  \leq  2 \eta \sum_{t=1}^T \| \gv\|_{t}^{*  \ 2} + \frac{\R(\x^\star) - \R( \x_1)}{\eta}   \\
& \leq 2 \eta n^2 T  + \frac{  \R(\x^\star) - \R(\x_1) }{\eta}   .
\end{align*}
It remains to bound the Bregman divergence with respect to $\x^\star$, for which we use a similar technique as in the analysis of algorithm \ref{algFKM}, and bound the regret with respect to $\x^\star_\delta$, which is the projection of $\x^\star$ onto $\K_\delta$.  Using equation \eqref{Lip_step}, we can bound the overall regret by:
\begin{align*}
& \sum_{t=1}^T \E [ f_{t}(\y_{t}) ]-  \sum_{t=1}^T  f_{t} (\x^\star) \\
& \leq  \sum_{t=1}^T \E [ f_{t}(\y_{t}) ]-  \sum_{t=1}^T  f_{t} (\x_\delta^*)  +  \delta T G D & \mbox{ equation \eqref{Lip_step} } \\ 
%& \leq  \eta n T  + \frac{B_\R(\x_1,\x^\star_\delta) }{\eta}   +  \delta T G D  & \mbox { above derivation} \\
%& = 2 \eta n T  + \frac{  \R(\x^\star_\delta) - \R(\x_1) }{\eta}   +  \delta T G D  & \mbox { Since $\nabla(\x_1) = 0$ } \\
& = 2 \eta n^2 T  + \frac{  \R(\x^\star_\delta) - \R(\x_1) }{\eta}   +  \delta T G D  &\mbox { above derivation}  \\
& \leq 2 \eta n^2 T  + \frac{\nu \log \frac{1}{1-\pi_{\x_1}(\x^\star_\delta)} }{\eta}   +  \delta T G D  & \mbox { Lemma \ref{Lemma:MinkowskiBarrier} } \\
& \leq 2 \eta n^2 T  + \frac{\nu \log \frac{1}{\delta }}{\eta}   +  \delta T G D  &  \x^\star_\delta \in \K_\delta  .
\end{align*}
Taking $\eta = O(\frac{1}{\sqrt{T}})$ and $\delta = O(\frac{1}{T})$, the above bound implies our theorem. %becomes $O(\sqrt{n \nu T \log T})$. 

\end{proof}

\newpage
\section{Bibliographic Remarks}

The Multi-Armed Bandit problem has history going back more than fifty years to the work of \citet{Robbins52}, see the survey of \citet{BubeckC12} for a much more detailed history. The non-stochastic MAB problem and the EXP3 algorithm, as well as tight lower bounds were given in the seminal paper of \citet{AueCesFreSch03nonstochastic}. The logarithmic gap in attainable regret for non-stochastic MAB was resolved in \citep{AudibertB09}.

Bandit Convex Optimization for the special case of linear cost functions  and the flow polytope, was introduced and studied by \citet{AweKle08} in the context of online routing. The full generality BCO setting was introduced by \citet{FlaxmanKM05}, who gave the first efficient and low-regret algorithm for BCO. Tight bounds for BCO were obtained by  \citet{bubeck2015bandit} for the one dimensional case, via an inefficient algorithm by \citet{hazan2016optimal}, and finally with a polynomial time algorithm in \citet{bubeck2017kernel}.  

The special case in which the cost functions are linear, called Bandit Linear Optimization, received significant attention. \citet{DanHayKak07price} gave an optimal regret algorithm up to constants depending on the dimension. \citet{AbernethyHR08}   gave an efficient algorithm and introduced self-concordant barriers to the bandit setting. Self-concordant barrier functions were devised in the context of polynomial-time algorithms for convex optimization in the seminal work of \citet{NesterovNemirovskii94siam}. Lower bounds for regret in the bandit linear optimization setting were studied by \cite{shamir2015complexity}.

In this chapter we have considered the expected regret as a performance metric. Significant literature is devoted to high probability guarantees on the regret. High probability bounds for the MAB problem were given in \citep{AueCesFreSch03nonstochastic}, and for bandit linear optimization in \citep{AbernethyR09}.  Other more refined metrics have been recently explored in \citep{DekelTA12} and in the context of adaptive adversaries in \citep{NeuGSA14,YuMa09,EvenDarKaMa09,MannorSh03,YuMaSh09}.

For a recent comprehensive text on bandit algorithms see \citep{lattimore2020bandit}.

\newpage
%\section{Exercises}
\begin{exercises}
%\begin{enumerate}
	
%\item
\exer{Prove a lower bound on the regret of any algorithm for BCO: show that for the special case of BCO over the unit sphere, any online algorithm must incur a regret of $\Omega(\sqrt{T})$. }

\exer{$^*$ % \item $^*$
Strengthen the above bound: show that for the special case of BLO over the  $d$-dimensional  unit simplex, with cost functions bounded in $\ell_\infty$ norm by one, any online algorithm must incur a regret of ${\Omega}(\sqrt{dT})$ as $T \rightarrow \infty$, up to logarithmic terms in $T$.
}

\exer{%\item
Let $\K$ be convex. Show that the set $\K_\delta$ is convex. 
}

\exer{%\item
Let $\K$ be convex and contain the unit ball centered at zero. Show that for any point $\x \in \K_\delta$, the ball of radius $\delta$ centered at $\x$ is contained in $\K$. 
}

\exer{%\item 
Consider the BCO setting with $H$-strongly convex functions, $H$ is known a-priori to the online learner. Show that in this case we can attain a regret bound of $\tilde{O}(T^{2/3})$.\\ 
{Hint:} recall that we can attain a regret bound of $O(\log T)$ in the full-information OCO with $H$-strongly convex functions, and recall that the notation $\tilde{O}( \cdot )$ hides constant and poly-logarithmic terms. 
}	

\exer{%\item
Consider the BCO setting with the following twist: at every iteration, the player is allowed to observe {\bf two} evaluations of the function, as opposed to just one. That is, the player gives $x_t,y_t$, and observes $f_t(x_t), f_t(y_t)$. Regret is measured w.r.t. $x_t$, as usual:
$$ \sum_t f_t(\x_t) - \min_{\x^\star \in \K} \sum_t f_t(\x^\star) .$$
%\begin{enumerate}
    \subexer{% \item 
    Show how to construct a biased gradient estimator for $f_t$ with arbitrary small bias and {\it constant} variance, that is independent of the bias. 
    }
    \subexer{%\item
    Show how to use the gradient estimator from the previous part to give an efficient algorithm for this setting that attains $O(\sqrt{T})$ regret.
    }
%\end{enumerate}
}

\end{exercises}
	
%\end{enumerate}

%!TEX root = OCObook.tex

%%%%%%%%%%%%%%%%%%%%%%%%%%%%%%%%%%%%%%%%%%%%%%%%%%%%%%%%%%%%
%%%%%%%%%%%%%%%%%%%%%%%%%%%%%%%%%%%%%%%%%%%%%%%%%%%%%%%%%%%%
%  Games
%%%%%%%%%%%%%%%%%%%%%%%%%%%%%%%%%%%%%%%%%%%%%%%%%%%%%%%%%%%%
%%%%%%%%%%%%%%%%%%%%%%%%%%%%%%%%%%%%%%%%%%%%%%%%%%%%%%%%%%%%
\chapter{Projection-Free Algorithms} \label{chap:FW}
%\begin{chapquote}{Author's name, \textit{Source of this quote}}
%``This is a quote and I don't know who said this.''
%\end{chapquote}

In many computational and learning scenarios the main bottleneck of optimization, both online and offline, is the computation of projections onto the underlying decision set (see \S \ref{sec:projections}). In this chapter we introduce projection-free methods for online convex optimization, that yield more efficient algorithms in these scenarios.

The motivating example throughout this chapter is the problem of matrix completion, which is a widely used and accepted model in the construction of recommendation systems. For matrix completion and related problems, projections amount to expensive linear algebraic operations and avoiding them is crucial in big data applications. 

We start with a detour into classical offline convex optimization and describe the conditional gradient algorithm, also known as the Frank-Wolfe algorithm. Afterwards, we describe problems for which linear optimization can be carried out much more efficiently than projections. We conclude with an OCO algorithm that eschews projections in favor of linear optimization, in much the same flavor as its offline counterpart.  

\section{Review: Relevant Concepts from Linear Algebra}

This chapter addresses rectangular matrices, which model  applications  such as recommendation systems naturally.  Consider a matrix $X \in \reals^{n \times m}$. A non-negative number $\sigma \in \reals_+$ is said to be a singular value for $X$ if there are two vectors $\uv \in \reals^n, \vv \in \reals^m$ such that 
$$  X^\top \uv  = \sigma \vv   , \quad X \vv = \sigma \uv. $$
The vectors $\uv,\vv$ are called the left and right singular vectors respectively. The non-zero singular values are the square roots of the eigenvalues of the matrix $X X^\top$ (and $X^\top X$).  The matrix $X$ can be written as 
$$ X = U \Sigma V^\top  \ , \ U \in \reals ^{n \times \rho} \ , \ V^\top  \in \reals^{ \rho \times m} ,$$
where $\rho = \min\{n,m\}$, the matrix $U$ is an orthogonal basis of the left singular vectors of $X$, the matrix $V$ is an orthogonal basis of right singular vectors, and $\Sigma$ is a diagonal matrix of singular values. This form is called the singular value decomposition for $X$. 

The number of non-zero singular  values for $X$ is called its rank, which we denote by $k \leq \rho$. 
The nuclear norm of $X$ is defined as the $\ell_1$ norm of its singular values, and denoted by
$$ \|X \|_* = \sum_{i=1}^\rho \sigma_i . $$
It can be shown (see exercises) that the nuclear norm is equal to the trace of the square root of the matrix times its transpose, i.e., 
$$ \|X\|_* = \trace( \sqrt{ X^\top X}  ) $$
We denote by $A \bullet B$ the inner product of two matrices as vectors in $\reals^{n \times m}$, that is
$$A \bullet B = \sum_{i = 1}^n \sum_{j=1}^m A_{ij} B_{ij} = \trace(AB^\top) . $$

\section{Motivation: Recommender Systems} \label{sec:recommendation_systems}
\sectionmark{Recommendation systems}

Media recommendations have changed significantly with the advent of the Internet and rise of online media stores. The large amounts of data collected allow for efficient clustering and accurate prediction of users' preferences for a variety of media. A well-known example is the so called ``Netflix challenge''---a competition of automated tools for recommendation from a large dataset of users' motion picture preferences.

One of the most successful approaches for automated recommendation systems, as proven in the Netflix competition, is matrix completion. Perhaps the simplest version of the problem can be described as follows.  

The entire dataset of user-media preference pairs is thought of as a partially-observed matrix. Thus, every person is represented by a row in the matrix, and every column represents a media item (movie). For simplicity, let us think of the observations as binary---a person either likes or dislikes a particular movie. Thus, we have a matrix $M \in \{0,1,*\}^{n \times m}$  where $n$ is the number of persons considered, $m$ is the number of movies at our library, and $0/1$ and $*$ signify ``dislike'', ``like'' and ``unknown'' respectively:
$$ M_{ij} = \mythreecases {0}{\mbox{person $i$ dislikes movie $j$}}{1}{\mbox{person $i$ likes movie $j$}}{*}{\mbox{preference unknown}} .$$ 

The natural goal is to complete the matrix, i.e., correctly assign $0$ or $1$ to the unknown entries. As defined so far, the problem is ill-posed, since any completion would be equally good (or bad), and no restrictions have been placed on the completions.  

The common restriction on completions is that the ``true'' matrix has low rank. Recall that a matrix $X \in \reals^{n \times m}$ has rank $k < \rho = \min \{n,m\} $ if and only if it can be written as 
$$ X = U V \ , \ U \in \reals^{n \times k} , V \in \reals^{k \times m}.  $$

The intuitive interpretation of this property is that each entry in $M$ can be explained by only $k$ numbers. In matrix completion this means, intuitively, that there are only $k$ factors that determine a persons preference over movies, such as genre, director, actors and so on. 

Now the simplistic matrix completion problem can be well-formulated as in the following mathematical program. Denote by $\| \cdot \|_{ob}$ the Euclidean norm only on the observed (non starred) entries of $M$, i.e., 
$$\|X\|_{ob}^2 = \sum_{M_{ij} \neq *} X_{ij}^2.$$ 
The mathematical program for matrix completion is given by
\begin{align*}
& \min_{X \in \reals^{n \times m} } \frac{1}{2} \| X - M \|_{ob}^2 \\
& \text{s.t.} \quad \rank(X) \leq k. 
\end{align*}  

Since the constraint over the rank of a matrix is non-convex, it is standard to consider a relaxation that replaces the rank constraint by the nuclear norm. %Recall that the nuclear norm of a matrix, denoted $\|M\|_*$, is the sum of its singular vectors, also equal to the trace of the square root of the matrix times its transpose, 
%$$ \| X\|_*^2 =  \left(\sum_{i=1}^{\rho} \sigma_i \right)^2 $$
It is known that the nuclear norm is a lower bound on the matrix rank if the singular values are bounded by one (see exercises). 
Thus, we arrive at the following convex program for matrix completion: 
\begin{align} \label{eqn:matrix-completion}
& \min_{X \in \reals^{n \times m} }  \frac{1}{2} \| X - M \|_{ob}^2 \\
& \text{s.t.} \quad \|X\|_* \leq k. \notag 
\end{align}  

We consider algorithms to solve this convex optimization problem next.

\section{The Conditional Gradient Method}\label{subsec:cond_grad_intro}

In this section we return to the basics of convex optimization---minimization of a convex function over a convex domain as studied in chapter \ref{chap:opt}. 

The conditional gradient (CG) method, or Frank-Wolfe algorithm, is a simple algorithm for minimizing a smooth convex function $f$ over a convex set $\K \subseteq \reals^n$. The appeal of the method is that it is a first order interior point method - the iterates always lie inside the convex set, and thus no projections are needed, and the update step on each iteration simply requires to minimize a linear objective over the set. The basic method is given in algorithm \ref{alg:condgrad}.
\begin{algorithm}[H]
	\caption{Conditional gradient}
	\label{alg:condgrad}
	\begin{algorithmic}[1]
		\State Input: step sizes $\{ \eta_t \in (0,1] , \ t \in [T]\}$, initial point $\x_1 \in \K$. 
%		\State Let $\x_1$ be an arbitrary point in $\K$.
		\For{$t = 1$ to $T$}
		\State $\vv_{t} \gets \arg \min_{\x \in \K} \left\{\x^\top \nabla{}f(\x_t) \right\} $. \label{algstep:linearopt}
		\State $\x_{t+1} \gets \x_t + \eta_t(\vv_t - \x_t)$.
		\EndFor
	\end{algorithmic}
\end{algorithm}

\begin{figure}[ht]
\begin{center}
\includegraphics[width=3.5in]{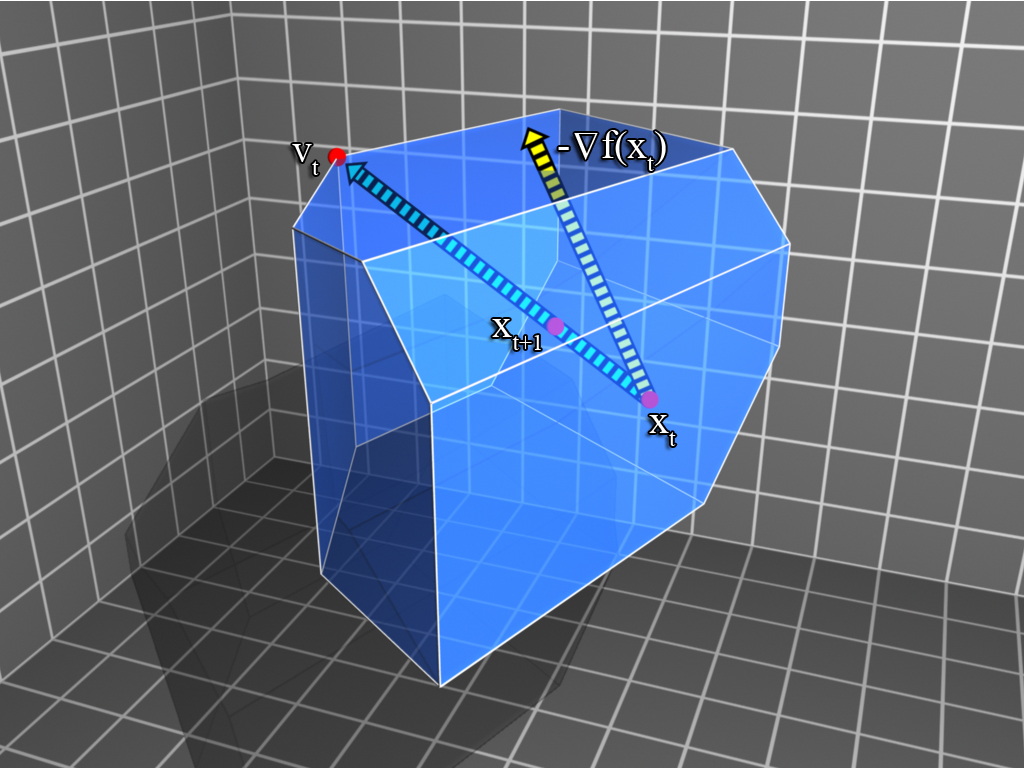}
\end{center}
\caption{Direction of progression of the CG algorithm \label{fig:OFW}}
\end{figure}

Note that in the CG method, the update to the iterate $\x_t$ may be not be in the direction of the gradient, as $\vv_t$ is the result of a linear optimization procedure in the direction of the negative gradient. This is depicted in figure \ref{fig:OFW}. 

The following theorem gives an essentially tight performance guarantee of this algorithm over smooth functions. Recall our notation from chapter \ref{chap:opt}: $\x^\star$ denotes the global minimizer of $f$ over $\K$, $D$ denotes the diameter of the set $\K$, and $h_t = f(\x_t) - f(\x^\star)$  denotes the suboptimality of the objective value in iteration $t$. 
\begin{theorem} \label{thm:offlineFW}
The CG algorithm applied to $\beta$-smooth functions with step sizes $\eta_t =  \min\{1,\frac{2}{t}\}$  attains the following convergence guarantee
$$ h_t \leq \frac{2 \beta  D^2 }{t} $$
\end{theorem}
\begin{proof}
As done before in this manuscript, we denote $\nabla_t = \nabla f(\x_t)$.   For any set of step sizes, we have
\begin{align}\label{old_fw_anal}
	&  f(\x_{t+1}) - f(\x^\star) 
	= f(\x_t + \eta_t(\vv_t - \x_t)) - f(\x^\star) \notag \\
	&\leq  f(\x_t) - f(\x^\star) + \eta_t(\vv_t-\x_t)^{\top}\nabla_t + \eta_t^2 \frac{\beta}{2}\Vert{\vv_t-\x_t}\Vert^2 &  \textrm{smoothness } \nonumber \\
	&\leq  f(\x_t) - f(\x^\star) + \eta_t(\x^\star-\x_t)^{\top}\nabla_t + \eta_t^2 \frac{\beta}{2}\Vert{\vv_t-\x_t}\Vert^2 &  \textrm{$\vv_t$ choice} \nonumber \\
	&\leq  f(\x_t) - f(\x^\star) + \eta_t(f(\x^\star)-f(\x_t)) + \eta_t^2 \frac{\beta}{2}\Vert{\vv_t-\x_t}\Vert^2 & \textrm{convexity} \nonumber \\
	&\leq  (1-\eta_t)(f(\x_t)-f(\x^\star)) + \frac{\eta_t^2\beta}{2} D^2. 
\end{align}
We reached the recursion $ h_{t+1} \leq (1- \eta_t) h_t + \eta_t^2\frac{  \beta D^2}{2} $, and by Lemma \ref{lemma:FW-recursion} we obtain,
$$ h_{t} \leq \frac{2 \beta D^2 }{t} . $$ 
\ignore{
\begin{align*}
 h_{t+1} & \leq (1- \eta_t) h_t + \eta_t^2 \frac{\beta D^2}{2}  \\
 & \leq (1- \eta_t) \frac{2  \beta H D ^2 }{ t} + \eta_t^2 \frac{ \beta  D^2}{2} & \mbox{induction hypothesis}\\
 & \leq (1- \frac{2 H}{ t}) \frac{2  \beta H D^2}{t} + \frac{4 H^2}{  t^2} \frac{\beta D^2 }{2}& \mbox{value of $\eta_t$}\\
 & = \frac{2  \beta H D^2 }{t} -  \frac{2 H^2 \beta D^2 }{ t^2 } \\
 & \leq \frac{2  \beta H D^2 }{t} (1 - \frac{1}{t} ) & \mbox{since $H \geq 1$} \\
 & \leq \frac{2  \beta H D^2 }{t+1 }.  & \mbox{$\frac{t-1}{t} \leq \frac{t}{t+1} $ } \\
 \end{align*}
}
\end{proof}

\begin{lemma} \label{lemma:FW-recursion}
Let $\{ h_t \} $ be a sequence that satisfies the recurence 
$$ h_{t+1} \leq h_t (1 - \eta_t) + \eta_t^2 c . $$
Then taking $\eta_t = \min\{1,\frac{2}{t}\} $ implies
$$ h_t \leq \frac{4c}{t} . $$ 
\end{lemma}
\begin{proof}
This is proved by induction on $t$. 

{\bf Induction base.}  For $t=1$, we have 
$$ h_2 \leq h_1 (1-\eta_1 ) + \eta_1^2 c = c \leq 4c . $$

{\bf Induction step.} 
\begin{align*}
 h_{t+1} & \leq (1- \eta_t) h_t + \eta_t^2 c  \\
 & \leq \left(1- \frac{2}{t} \right) \frac{4c}{ t} + \frac{4c}{t^2}  & \mbox{induction hypothesis}\\
 & = \frac{4c}{t} \left( 1 - \frac{1}{t} \right) \\
 & \leq  \frac{4c}{t} \cdot  \frac{t}{t+1} & \mbox{$\frac{t-1}{t} \leq \frac{t}{t+1} $ }   \\
 & = \frac{4c}{t+1} .
 \end{align*}
\end{proof}

%%%%%%%%%%%%%%%
% REMINDER THAT THIS CANNOT BE IMPROVED 
%%%%%%%%%%%%%%%
%\begin{lemma} \label{lemma:FW-recursion}
%Let $\{ h_t \} $ be a sequence that satisfies the recurence 
%$$ h_{t+1} \leq h_t (1 - \eta_t) + \eta_t^2 c . $$
%Then taking $\eta_t = \min\left \{1,\frac{2c+h_1}{2c \  t} \right\} $ implies
%$$ h_T \leq \frac{2c+h_1}{T + 1} $$ 
%\end{lemma}
%\begin{proof}
%This is proved by induction on $t$. To simplify the derivation below, let $a = 2c+h_1$, such that 
% the hypothesis is $h_t \leq \frac{a}{t}$, and set $\eta_t = \frac{b}{t}$, for $b = \frac{a}{2c}$. 

%{\bf Induction base.}  For $t=1$, and for as long as $\eta_t = 1$, we have 
%$$ h_2 \leq h_1 (1-\eta_1 ) + \eta_1^2 c \leq h_1 + c $$

%{\bf Induction step.} 
%\begin{align*}
% h_{t+1} & \leq (1- \eta_t) h_t + \eta_t^2 c  \\
% & \leq \left(1- \frac{b}{t} \right) \frac{a }{ t} + \frac{b^2 c}{t^2}  & \mbox{induction hypothesis}\\
% & = \frac{a}{t} - \frac{ab}{t^2} + \frac{b^2 c}{t^2 }\\
% & = \frac{a}{t} - \frac{a^2}{2c t^2} & b = \frac{a}{2c} \\
% & \leq  \frac{a}{t} - \frac{a}{t^2} & a \geq 2c  \\
% & = \frac{a (t-1) }{t^2} \leq \frac{a}{t+1} & \mbox{$\frac{t-1}{t} \leq \frac{t}{t+1} $ } \\
% \end{align*}
%\end{proof}

\subsection{Example: matrix completion via CG} 

As an example of an application for the conditional gradient algorithm, recall the mathematical program given by \eqref{eqn:matrix-completion}.  The gradient of the objective function at point $X^t$ is
\begin{equation} \label{eqn:matrix-gradient}
\nabla f(X^t) = (X^t - M)_{ob}  = \mycases { X_{ij}^t - M_{ij} } {(i,j) \in OB} {0}{\text{otherwise}} .
\end{equation}
Over the set of bounded-nuclear norm matrices, the linear optimization  of line \ref{algstep:linearopt} in algorithm \ref{alg:condgrad}  becomes, 
\begin{align*} 
&   \min     X \bullet \nabla_t \ , \quad \nabla_t = \nabla f(X_t)    \\
& \mbox{s.t.  }  \|X\|_* \leq k. 
\end{align*}  
For simplicity, let's consider square symmetric matrices, for which the nuclear norm is equivalent to the trace norm, and the above optimization problem becomes  
\begin{align*} 
& \min  X \bullet \nabla_t  \\
& \mbox{s.t.  }   \trace(X)  \leq k. 
\end{align*}  
It can be shown that this program is equivalent to the following (see exercises):
\begin{align*} 
& \min_{\x \in \reals^n}  \x^\top \nabla_t \x  \\
& \mbox{s.t.  }   \|\x\|_2^2  \leq k. 
\end{align*}  
Hence, this is an eigenvector computation in disguise! Computing the largest eigenvector of a matrix takes linear time via the power method, which also  applies more generally to computing the largest singular value of rectangular matrices. With this, step \ref{algstep:linearopt} of algorithm \ref{alg:condgrad}, which amounts to mathematical program  \eqref{eqn:matrix-completion}, becomes computing $v_{\max}(- \nabla f(X^t))$, the largest eigenvector of $- \nabla f(X^t)$. Algorithm \ref{alg:condgrad} takes on the modified form described in Algorithm \ref{alg:condgrad4matrixcompletion}.
\begin{algorithm}[H]
	\caption{Conditional gradient for matrix completion}
	\label{alg:condgrad4matrixcompletion}
	\begin{algorithmic}[1]
		\State Let $X^1$ be an arbitrary matrix  of trace $k$ in $\K$.
		\For{$t = 1$ to $T$}
		\State $\vv_{t} = \sqrt{k} \cdot v_{\max}(-\nabla_t )  $.
		\State $X^{t+1} = X^t + \eta_t(\vv_t  \vv_t^\top - X^t)$ for $\eta_t\in(0,1)$.
		\EndFor
	\end{algorithmic}
\end{algorithm}

\paragraph*{Comparison to other gradient-based methods.}
How does this compare to previous convex optimization methods for solving the same matrix completion problem? 
As a convex program, we can apply gradient descent, or even more advantageously in this setting, stochastic gradient descent as in \S \ref{sec:sgd}. Recall that the gradient of the objective function at point $X^t$ takes the simple form \eqref{eqn:matrix-gradient}. 
A stochastic estimate for the gradient can be attained by observing just  a single entry of the matrix $M$, and the update itself takes constant time as the gradient estimator is sparse.  However, the projection step is significantly more difficult.

In this setting, the  convex set $\K$ is the set of bounded nuclear norm matrices. Projecting a matrix onto this set amounts to calculating the SVD of the matrix, which is similar in computational complexity to algorithms for matrix diagonalization or inversion. The best known algorithms for matrix diagonalization are superlinear in the matrices' size, and thus impractical for large datasets that are common in applications.  

In contrast, the CG method does not require projections at all, and replaces them with linear optimization steps over the convex set, which we have observed to amount to singular vector computations. The latter can be implemented to take linear time via the power method or the Lanczos algorithm (see bibliography).

Thus, the Conditional Gradient method allows for optimization of the mathematical program \eqref{eqn:matrix-completion} with a linear-time operation (eigenvector using power method)  per iteration, rather than a significantly more expensive computation (SVD) needed for gradient descent.

\section{Projections versus Linear Optimization}

The conditional gradient (Frank-Wolfe) algorithm described before does not resort to projections, but rather computes a linear optimization problem of the form
\begin{equation} \label{eqn:linopt}
 \arg \min_{\x \in \K} \left\{\x^\top \uv \right\}. 
\end{equation}
When is the CG method computationally preferable?  The overall computational complexity of an iterative optimization algorithm is the product of the number of iterations and the computational cost per iteration. The CG method does not converge as well as the most efficient gradient descent algorithms, meaning it requires more iterations to produce a solution of a comparable level of accuracy. However, for many interesting scenarios the computational cost of a linear optimization step \eqref{eqn:linopt} is {\em significantly} lower than that of a projection step. 

%The latter argument is especially relevant in online convex optimization: where as in offline optimization the CG algorithm behaves polynomially in $\eps$ as compared to logarithmically for well-conditioned problems, for OCO such differences are impossible - we have seen in previous chapters that the regret is bounded by square root in the number of iterations. 

Let us  point out several examples of problems for which we have very efficient linear optimization algorithms, whereas our state-of-the-art algorithms for computing projections  are significantly slower.

\paragraph*{Recommendation systems and matrix prediction.}

In the example pointed out in the preceding section of matrix completion, known methods for projection onto the spectahedron, or more generally the bounded  nuclear-norm ball, require singular value decompositions, which  take superlinear time via our best known methods. In contrast, the CG method requires maximal eigenvector computations which can be carried out in linear time via the power method (or the more sophisticated Lanczos algorithm).

\paragraph*{Network routing and convex graph problems.}

Various routing and graph problems can be modeled as convex optimization problems over a convex set called the flow polytope. 

Consider  a directed acyclic graph with $m$ edges, a source node marked $s$ and a target node marked $t$. Every path from $s$  to $t$ in the graph can be represented by its identifying vector, that is a vector in $\lbrace{0,1}\rbrace^m$ in which the entries that are set to 1 correspond to edges of the path. The flow polytope of the graph is the convex hull of all such identifying vectors  of the simple paths from $s$ to $t$. This polytope is also exactly the set of all unit $s$--$t$ flows in the graph if we assume that each edge has a unit flow capacity (a flow is represented here as a vector in $\mathbb{R}^m$ in which each entry is the amount of flow through the corresponding edge). 

Since the flow polytope is just the convex hull of $s$--$t$ paths in the graph, minimizing a linear objective over it amounts to finding a minimum weight path given weights for the edges. For the shortest path problem we have  very efficient combinatorial optimization algorithms, namely Dijkstra's algorithm. 

Thus, applying the CG algorithm to solve {\bf any} convex optimization problem over the flow polytope will only require iterative shortest path computations.

\paragraph*{Ranking and permutations. }

A common way to represent a permutation or ordering is by a permutation matrix. Such are square matrices over $\{0,1\}^{n \times n}$ that contain exactly one $1$ entry in each row and column.

Doubly-stochastic matrices are square, real-valued matrices with non-negative entries, in which the sum of entries of each row and each column amounts to 1. The polytope that defines all doubly-stochastic matrices is called the Birkhoff-von Neumann polytope. 
The Birkhoff-von Neumann theorem states that this polytope is the convex hull of exactly all $n\times{n}$ permutation matrices. 

Since a permutation matrix corresponds to a perfect matching in a fully connected bipartite graph, linear minimization over this polytope corresponds to finding a minimum weight perfect matching in a bipartite graph.

Consider a convex optimization problem over the Birkhoff-von Neumann polytope. The CG algorithm will iteratively solve a linear optimization problem over the BVN polytope, thus iteratively solving a minimum weight perfect matching in a bipartite graph problem, which is a well-studied combinatorial optimization problem for which we know of efficient algorithms. In contrast, other gradient based methods will require projections, which are quadratic optimization problems over the BVN polytope.

\paragraph*{Matroid polytopes.}

A matroid is pair $(E,I)$ where $E$ is a set of elements and $I$ is a set of subsets of $E$ called the independent sets which satisfy various interesting proprieties that resemble the concept of linear independence in vector spaces. 
Matroids have been studied extensively in combinatorial optimization and a key example of a matroid is the graphical matroid in which the set $E$ is the set of edges of a given graph and the set $I$ is the set of all subsets of $E$ which are cycle-free. In this case, $I$ contains all the spanning trees of the graph. A subset $S\in{I}$ could be represented by its identifying vector which lies in $\lbrace{0,1}\rbrace^{\vert{E}\vert}$ which also gives rise to the matroid polytope which is just the convex hull of all identifying vectors of sets in $I$. It can be shown that some matroid polytopes are defined by exponentially many linear inequalities (exponential in $\vert{E}\vert$), which makes optimization over them difficult. 

On the other hand, linear optimization over matroid polytopes is easy using a simple greedy procedure which runs in nearly linear time. Thus, the CG method serves as an efficient algorithm to solve any convex optimization problem over matroids iteratively using only a simple greedy procedure.

\section{The Online Conditional Gradient Algorithm}
\sectionmark{Online Conditional Gradient}

In this section we give a projection-free algorithm for OCO based on the conditional gradient method, which is projection-free and thus carries the  computational advantages of the CG method to the online setting. 

It is tempting to apply the CG method straightforwardly to the online appearance of functions in the OCO setting, such as the OGD algorithm in \S \ref{section:ogd}. However, it can be shown that an approach that only takes into account the last cost function is doomed to fail. The reason is that the conditional gradient method takes into account the {\it direction} of the gradient, and is insensitive to its {\it magnitude}. 

Instead, we apply the CG algorithm step to the aggregate sum of all previous cost functions with added Euclidean regularization. The resulting algorithm is given formally in Algorithm \ref{alg:ocg}.

\begin{algorithm}[ht]
		\caption{\label{alg:ocg} Online conditional gradient }
		\begin{algorithmic}[1]
\State Input: convex set $\K$, $T$, $\x_1 \in \mathcal{K}$, parameters $\eta  , \ \{\sigma_t\} $.
\For{$t=1, 2, \ldots, T$}
\State Play $\bx_t$ and observe $f_t$.
%\State Set $\ftil_t(\x)  = \nabla_t^\top x + \sigma \|\x - \x_1\|^2 $
\State\label{eq:F_t-def} Let $F_t(\x) = \eta \tsum_{\tau=1}^{t-1} \nabla_\tau^\top \x +  \|\x - \x_1\|^2  $. 
\State Compute $ \bv_t = \arg \min_{\bx \in \K} \{\nabla F_t(\bx_t) \cdot \bx \} $.
%\State Set $\eta_{t} = t^{-a}$.
\State Set $\bx_{t+1} = (1 - \sigma_t)\bx_{t} + \sigma_t \bv_t$.
\EndFor
		\end{algorithmic}
\end{algorithm}

We can prove the following regret bound for this algorithm. While this regret bound is suboptimal in light of the previous upper bounds we have seen, its suboptimality is compensated by the algorithm's lower computational cost. 

\begin{theorem} \label{thm:FWonline-main} 
Online conditional gradient (Algorithm \ref{alg:ocg})  with parameters $\eta = \frac{D}{2 G T^{3/4} }, \sigma_t = \min\{1,\frac{2}{t^{1/2}}\}$, attains the following
guarantee
$$ \regret_T = \sum_{t=1}^{T} f_t(\bx_t) -\min_{\bx^\star \in \K}\sum_{t=1}^{T}
f_t(\bx^\star)\ \leq  8 D G T^{3/4} $$
\end{theorem}

As a first step in analyzing Algorithm \ref{alg:ocg}, consider the points 
$$ \xv^\star = \arg \min _{\x \in \K } F_t(\x) .$$
These are exactly the iterates of the RFTL algorithm from chapter \ref{chap:regularization}, namely Algorithm \ref{alg:RFTLmain} with the regularization being $R(\x) = \| \x - \xv[1]\|^2 $, applied to cost functions with a shift, namely:
$$ \ftil_t = f_t( \x + (\x_t^\star - \x_t) ) .$$
The reason is that $\nabla_t$ in Algorithm \ref{alg:ocg} refers to $\nabla f_t(\x_t)$, whereas in the RFTL algorithm we have $\nabla_t = \nabla f_t(\x_t^\star)$. 
Notice that for any point $\x \in \K$ we have $| f_t(\x) - \ftil_t(\x) | \leq G \|\x_t - \x_t^\star\|$.  Thus, according to Theorem  \ref{thm:RFTLmain1}, we have that 
\begin{eqnarray} \label{eqn:FW1}
 & \sum_{t=1}^{T} f_t(\x_t^\star) - \sum_{t=1}^{T} f_t(\bx^\star) \notag \\
 & \leq 2 G\sum_t  \|\x_t - \x_t^\star\| +    \sum_{t=1}^{T} \ftil_t(\x_t^\star) - \sum_{t=1}^{T} \ftil_t(\bx^\star )  \notag \\
 & \leq  2 G\sum_t  \|\x_t - \x_t^\star\| +   2 \eta G T + \frac{1}{\eta} D .
\end{eqnarray}

Using our previous notation,  denote by $h_t(\x) = {F_t(\x) - F_t(\x^\star_t)} $, and by $ h_t = h_t(\x_t)$.
The main lemma we require to proceed is the following, which relates the iterates $\x_t$ to the optimal point according to the aggregate function $F_t$.

\begin{lemma} \label{lem:mainFW}
%Assume that the parameters $\eta,\sigma_t $ are chosen such that  $\eta G \sqrt{h_{t+1}}  \leq \frac{D^2}{2} \sigma_t^2$.
The iterates $\x_t$ of Algorithm~\ref{alg:ocg} satisfy for all $t \ge 1$
$$ h_t \leq  {  2 D^2} \sigma_t.   $$
\end{lemma}

\begin{proof}

%First, we claim that 
%\begin{equation}\label{eqn:shalom-fw}
%\eta G \sqrt{h_t} \leq \frac{1}{2} h_t . 
%\end{equation}
%If that is not the case, then we have that 
%$ h_t \leq 2 \eta G \sqrt{h_t}$, which implies
%$$  h_t \leq 4 G^2 \eta^2 \leq 4 G^2 \cdot \frac{D^2}{4 G^2 T^{3/2} } = \frac{D^2 }{T^{3/2} } \leq D^2 \sigma_t ,$$
%which is what we wanted to prove. 

As the functions $F_t$ are $1$-smooth, applying the offline Frank-Wolfe analysis technique, and in particular Equation \eqref{old_fw_anal} to the function $F_t$ we obtain:
\begin{align*}
 h_{t}(\x_{t+1}) & = F_{t} (\x_{t+1}) -  F_{t} (\x^\star_{t}) \\
&\leq  (1-\sigma_t)( F_t (\x_t)- F_t (\x^\star_t)) + \frac{D^2}{2} \sigma_t^2 & \mbox { Equation \eqref{old_fw_anal}} \\
& =  (1-\sigma_t)  h_t + \frac{D^2}{2} \sigma_t^2.
\end{align*}

\noindent In addition, by definition of $F_t$ and $h_t$ we have
\begin{align*}
& h_{t+1}  \\
& = F_t(\x_{t+1})  - F_t(\x_{t+1}^\star) + \eta \nabla_{t+1}(\x_{t+1} - \x^\star_{t+1} ) \\
& \leq h_t(\x_{t+1})  + \eta \nabla_{t+1}(\x_{t+1} - \x^\star_{t+1}) & \mbox{$F_t(\x_t^\star) \leq F_t(\x_{t+1}^\star)$} \\
 & \leq\ h_t(\x_{t+1}) + \eta G \| \x_{t+1} - \x_{t+1}^\star\| .  & \mbox{Cauchy-Schwarz}
\end{align*}
Since $F_t$ is $1$-strongly convex, we have
$$ \| \x - \xv^\star \|^2 \leq F_t(\x) - F_t(\xv^\star)  .$$
Thus, 
\begin{align*}
h_{t+1} & \leq\ h_t(\x_{t+1}) + \eta G \| \x_{t+1} - \x_{t+1}^\star\| \\
& \leq h_t(\x_{t+1}) + \eta G \sqrt{h_{t+1}}  \\
%& \leq h_t(\x_{t+1}) + \eta^{4/3} G^2 + \eta^{2/3} \|\x_{t+1} - \x_{t+1}^\star\|^2 \\
%& \leq h_t(\x_{t+1}) + \eta^{4/3} G^2 + \eta^{2/3} h_{t+1}  \\
%& \leq h_t(\x_{t+1}) +  G^2 \sigma_t^2 + \sigma_t h_{t+1} 
 &  \leq h_t (1 - \sigma_t)   + \frac{1}{2} {D^2 } \sigma_t^2 + \eta G \sqrt{h_{t+1}}  & \mbox{above derivation} \\
& \leq h_t (1 - \frac{5}{6} \sigma_t)   + \frac{5}{8}{D^2 } \sigma_t^2. & \mbox{ equation \eqref{prop:fwhelperprop} below} 
\end{align*}

Above we used the following derivation, that holds by choice of  parameters  $\eta = \frac{D}{2 G T^{3/4} }$ and $\sigma_t = \min\{1,\frac{2}{t^{1/2}}\}$:
since $\eta,G,h_t$ are all non-negative, we have
\begin{eqnarray}
\eta G \sqrt{h_{t+1}} &  = \left(  \sqrt{D} {G \eta} \right)^{2/3} \left( \frac{G  \eta}{D}  \right)^{1/3} \sqrt{h_{t+1}} \notag \\
& \leq \frac{1}{2} \left( { \sqrt{D} G \eta}{} \right)^{4/3} + \frac{1}{2}  \left( \frac{G \eta}{D}  \right)^{2/3} h_{t+1}  \notag \\
& \leq \frac{1}{8} D^2  \sigma_t^2 + \frac{1}{6}  \sigma_t h_{t+1} \label{prop:fwhelperprop}
\end{eqnarray}

%satisfy  $\eta G \sqrt{h_{t+1}}  \leq \frac{1}{4} D^2 \sigma_t^2 + \frac{1}{4} \sigma_t h_{t+1}$.

We now claim that the theorem follows inductively. 
%Recall our assumption that $\eta G \sqrt{h_{t+1}}  \leq \frac{D^2}{2} \sigma_t^2$, which in particular implies, since  $\eta < 1$ and $h_t = F_t(\x_t) - F_t(\x_t^\star) \geq \| \x_t - \x_t^\star\|^2$  
%$$ G \leq \frac{D \sigma_1^2}{2 \eta} \leq 2D$$
The base of the induction holds since, for $t = 1$, the definition of $F_1$ implies
$$h_1 = F_1(\x_1) - F_1(\x^\star) = \| \x_1 - \x^\star\|^2 \leq  D^2 \leq  2 D^2 \sigma_1  .$$ 

Assuming the bound is true for $t$, we now show it holds for $t+1$ as well:
\begin{eqnarray*}
h_{t+1} & \leq & h_t (1 - \frac{5}{6} \sigma_t)   + \frac{5}{8} {D^2 } \sigma_t^2 \\
& \leq &   2 D^2 \sigma_t  \left(1 -  \frac{5}{6} \sigma_t \right) + \frac{5}{8} { D^2}\sigma_t^2  \\
& \leq &  { 2 D^2 }\sigma_t  \left(1 -  \frac{\sigma_t}{2} \right) \\
& \le & {  2 D^2} \sigma_{t+1},
\end{eqnarray*}
as required. The last inequality follows by the definition of $\sigma_t$ (see exercises). 

\end{proof}
\ignore{
Before continuing with the proof of the theorem, we give the proof of the intermediate proposition. 
\begin{proof}[Proof of Proposition \ref{prop:fwhelperprop}]
Since $\eta,G,h_t$ are all non-negative, we have
\begin{eqnarray*}
\eta G \sqrt{h_{t+1}} &  = \left(  \sqrt{D} {G \eta} \right)^{2/3} \left( \frac{G  \eta}{D}  \right)^{1/3} \sqrt{h_{t+1}} \\
& \leq \frac{1}{2} \left( { \sqrt{D} G \eta}{} \right)^{4/3} + \frac{1}{2}  \left( \frac{G \eta}{D}  \right)^{2/3} h_{t+1}  \\
& \leq \frac{1}{4} D^2  \sigma_t^2 + \frac{1}{4}  \sigma_t h_{t+1}
\end{eqnarray*}
\end{proof}
}

We proceed to use this lemma in order to prove our theorem:
\begin{proof}[Proof of Theorem \ref{thm:FWonline-main}]
	
By definition, the functions $F_t$ are $1$-strongly convex. Thus, we have for $\xv^\star = \arg \min_{\x \in \K} F_t(\x)$:
$$ \| \x - \xv^\star \|^2 \leq F_t(\x) - F_t(\xv^\star)  .$$
Let $\eta = \frac{D}{2G T^{3/4}} $, and notice that this satisfies the constraint of Lemma \ref{lem:mainFW}, which requires $\eta G \sqrt{h_{t+1}}  \leq \frac{D^2}{2} \sigma_t^2$. In addition, $\eta < 1$ for $T$ large enough. Hence,
\begin{align}
f_t(\xv) - f_t(\x^\star_t) & \leq G \| \xv - \x^\star_t \|  \notag \\
& \leq {G} \sqrt{  F_t(\xv) - F_t(\xv^\star) } \notag \\
& \leq { 2 G D } \sqrt{\sigma_t} .  & \mbox{ Lemma \ref{lem:mainFW} }  \label{eqn:FW2} 
\end{align}	
Putting everything together we obtain:

\begin{align*}
& \text{\em Regret}_T(\text{\em OCG})  = \sum_{t=1}^{T} f_t(\bx_t) - \sum_{t=1}^{T}
f_t(\bx^\star) \\
& = \sum_{t=1}^T \left[ f_t(\bx_t) - f_t(\xv^\star) + f_t(\xv^\star) - f_t(\x^\star) \right] \\
& \leq  \sum_{t=1}^{T} 2 G {D} \sqrt{\sigma_t}  + \sum_t \left[  f_t(\xv^\star) - f_t(\x^\star) \right] & \mbox{by \eqref{eqn:FW2}} \\ 
& \leq 4 G D {T}^{3/4} + \sum_t \left[  f_t(\xv^\star) - f_t(\x^\star) \right]   \\
& \leq 4 G  D {T}^{3/4} + 2 G\sum_t  \|\x_t - \x_t^\star \| +   2 \eta G T + \frac{1}{\eta} D .  & \mbox{by \eqref{eqn:FW1}}   \\
 %2 \eta G^2 T  + \frac{D^2}{\eta}.  & \mbox { \eqref{eqn:FW1}} 
\end{align*}

We thus obtain:
\begin{align*}
	\regret_T(\text{\em OCG}) 
	& \leq 4 G D {T}^{3/4} + 2 \eta G^2 T  + \frac{D^2}{\eta}  \\
	& \leq 4 G D T^{2/3} + DG T^{1/4} + 2 DG T^{3/4} \leq 8 D G T^{3/4}.  
\end{align*}
\end{proof}

\newpage
\section{Bibliographic Remarks}

The matrix completion model has been extremely popular since its inception in the context of recommendation systems \citep{SrebroThesis,Rennie:2005,salakhutdinov:collaborative,lee:practical,CandesR09,ShamirS11}.

The conditional gradient algorithm was devised in the seminal paper by \citet{FrankWolfe}. Due to the applicability of the FW algorithm to large-scale constrained problems, it has been a method of choice in recent machine learning applications, to name a few: 
\citep{Hazan08,Jaggi10, Jaggi13a, Jaggi13b, Dudik12a, Dudik12b, Hazan12, ShalevShwartz11, Bach12, Tewari11, Garber11, Garber13, Florina14}. In the context of matrix completion and recommendation systems, several faster variants of the Frank-Wolfe method were proposed  \citep{garber2016faster,allen2017linear}

The online conditional gradient algorithm is due to \citet{Hazan12}. An optimal regret algorithm, attaining the $O(\sqrt{T})$ bound, for the special case of polyhedral sets was devised in \citep{Garber13}. 

Recent works consider accelerating projection-free optimization using variance reduction \citep{lan2016conditional,hazan2016variance}, and the case of projection-free algorithms with stochastic gradient oracles \citep{mokhtari2018stochastic, chen2018projection, xie2019stochastic}.

For an analysis of the running time of the power and Lanczos methods for computing eigenvectors see \citep{kuczynski1992estimating}. For modern algorithms for fast computation of the singular value decomposition see \citep{allen2016lazysvd,musco2015randomized}.

\newpage
\begin{exercises}
%\section{Exercises}

%\begin{enumerate}

\exer{ %\item
Prove that if the singular values are smaller than or equal to one, then the nuclear norm is a lower bound on the rank, i.e., show
$$ \rank(X) \geq \|X\|_* .$$ 
}

\exer{ %\item
Prove that the trace  is related to the nuclear norm via
$$ \| X \|_* = \trace( \sqrt{X X^\top} ) = \trace( \sqrt{ X^\top X} ) .$$
}

\exer{ %\item
In this question we show that maximizing a linear function over the spectahedron can be reduced to a  maximal eigenvector computation. 
%\begin{enumerate}
\subexer{ %    \item 
Consider the the following mathematical program for a symmetric $C \in \reals^{d \times d}$:
\begin{align*} 
& \max  X \bullet C  \\
& X \in S_d = \{ X \in \reals^{d \times d} \ , \ X \succcurlyeq 0 \ , \ \trace(X)  \leq 1 \} .
\end{align*}  
Prove that it has the same solution as the mathematical program: 
\begin{align*} 
& \max_{\x \in \reals^d}  \x^\top C \x  \\
& \mbox{s.t.  }   \|\x\|_2  \leq 1 .
\end{align*}  
}
\subexer{ %\item
Show how to use eigenvector computations to maximize a general (a-symmetric) linear functions over the spectahedron. 
%\end{enumerate}
}}

\exer{ %\item
Prove that for positive integers $t>0$, any $ c \in [0,1]$ and $\sigma_t = \frac{2}{t^c}$ it holds that
$$ \sigma_t (1- \frac{\sigma_t}{2} ) \leq \sigma_{t+1}  . $$ 
}

\exer{ %\item
Download the MovieLens dataset from the web. Implement an online recommendation system based on the matrix completion model: implement the OCG and OGD algorithms for matrix completion. Benchmark your results. 
}

\end{exercises}
%\end{enumerate}

%!TEX root = OCObook.tex

%%%%%%%%%%%%%%%%%%%%%%%%%%%%%%%%%%%%%%%%%%%%%%%%%%%%%%%%%%%%
%%%%%%%%%%%%%%%%%%%%%%%%%%%%%%%%%%%%%%%%%%%%%%%%%%%%%%%%%%%%
%  Games
%%%%%%%%%%%%%%%%%%%%%%%%%%%%%%%%%%%%%%%%%%%%%%%%%%%%%%%%%%%%
%%%%%%%%%%%%%%%%%%%%%%%%%%%%%%%%%%%%%%%%%%%%%%%%%%%%%%%%%%%%
\chapter{Games, Duality, and Regret} \label{chap:games}

In this chapter we tie the material covered thus far to some of the  most intriguing concepts in optimization and game theory. We shall use the existence of online convex optimization algorithms with sublinear regret  to prove two fundamental properties: convex duality in mathematical optimization, and von Neumann's minimax theorem in game theory. 

Historically, the theory of games was developed by von Neumann in the early 1930's. In an entirely different scientific thread, the theory of linear programming (LP) was  advanced by Dantzig a decade later. Dantzig describes in his memoir a notable meeting between himself and von Neumann at Princeton in 1947. In this meeting, according to Dantzig, after describing the geometric and algebraic versions of linear programming,  von Neumann essentially formulated and proved linear programming duality:
\begin{quote}
``I don't want you to think I am pulling all this
out of my sleeve at the spur of the moment like a magician. I have just recently completed a book with
Oscar Morgenstern on the theory of games. What I am doing is conjecturing that the two problems
are equivalent. The theory that I am outlining for your problem is an analogue to the one we have
developed for games." \endnote{Taken from \citep{von2006ifors}}
\end{quote}

At that time, the topic of discussion was not the existence and  uniqueness of equilibrium in zero-sum games, which is captured by the minimax theorem. Both concepts were originally captured and proved using very different mathematical techniques: the minimax theorem was originally proved using machinery from mathematical topology, whereas linear programming duality was shown using convexity and geometric tools.  
   
More than half a century later, Freund and Schapire tied both concepts, which were by then known to be strongly related, to regret minimization. We shall follow their lead in this chapter, introduce the relevant concepts and give concise proofs using the machinery developed earlier in this manuscript.  
  
The chapter can be read with basic familiarity with linear programming and little or no background in game theory. We define linear programming and zero-sum games succinctly, barely enough to prove the duality theorem and the minimax theorem. The reader is referred to the numerous wonderful texts available on linear programming and game theory for a much more thorough introduction and definitions.

\section{Linear Programming and Duality}

Linear programming is a widely successful and practical convex optimization framework. Amongst its numerous successes is the Nobel prize award given on account of its application to economics. It is a special case of the convex optimization problem from chapter \ref{chap:opt} in which $\K$ is a polyhedron (i.e., an intersection of a finite set of halfspaces) and the objective function is a linear function.   Thus, a linear program can be described as follows, where  $(A \in \mathbb{R}^{n\times m})$:
\begin{eqnarray*}
 \min  \quad &  c^{\top} \x &      \\
 \text{s.t.} \quad &  A \x \geq b  & .  
\end{eqnarray*}
The above formulation can be transformed into several different forms via basic manipulations. For example, any LP can be transformed to an equivalent LP with the variables taking only non-negative values. This can be accomplished by writing every variable $x$ as $x = x^{+}  - x^{-}$, with $x^{+}, x^{-} \geq 0$. It can be verified that this transformation leaves us with another LP, whose variables are non-negative, and contains at most twice as many variables (see exercises section for more details).  

We are now ready to define a central notion in LP and state the duality theorem: 
\begin{theorem}[The duality theorem]
Given a linear program:
\begin{eqnarray*}
 \min \quad & & c^{\top} \x      \\
 \text{s.t.} \quad & & A \x \geq b , \\
 & & \x \geq 0 ,
\end{eqnarray*}
its dual program is given by:
\begin{eqnarray*}
 \max \quad & & b^\top \y      \\
 \text{s.t.} \quad & & A^{\top} \y \leq c, \\
   & & \y \geq 0 .
\end{eqnarray*}
and the objectives of both problems are either equal or unbounded.

\end{theorem}

Instead of studying duality directly, we proceed to define zero-sum games and an analogous concept to duality.

\section{Zero-sum Games and Equilibria}

The theory of games is an established research field in economic theory. We give here brief definitions of the main concepts studied in this chapter. 

Let us start with an example of a zero-sum game we all know: the rock-paper-scissors game. In this game each of the two players chooses a strategy: either rock, scissors or paper. The winner is determined according to the following table, where $0$ denotes a draw, $-1$ denotes that the row player wins, and $1$ denotes a column player victory. 
%\begin{example}
\begin{table}[ht]
  \centering
    \makebox[\linewidth]{\begin{tabular}{|c|c|c|c|}
		\hline
		- & \textbf{scissors} & \textbf{paper} & \textbf{rock} \\
		\hline
		\textbf{rock} & $-1$ & $1$ & $0$ \\
		\hline
		\textbf{paper} & $1$ & $0$ & $-1$ \\
		\hline
		\textbf{scissors} & $0$ & $-1$ & $1$ \\
		\hline
    \end{tabular}}%
    \caption{Example of a zero-sum game in matrix representation.}
\end{table}%
%\end{example}

The rock-paper-scissors game is called a ``zero-sum'' game since one can think of the numbers as losses for the row player (loss of $-1$ resembles victory, $1$ loss and $0$ draw), in which case the column player receives a loss  which is exactly the negation of the loss of the row player. Thus the sum of losses which both players suffer is zero in every outcome of the game. 

Noticed that we termed one player as the ``row player'' and the other as the ``column player'' corresponding to the matrix losses. 
Such a matrix representation is far more general: 

\begin{definition} \label{defn:zsg}
A two-player zero-sum-game in normal form is given by  a matrix $A \in [-1,1]^{n \times m}$. The loss for the row player playing strategy $i \in [n]$ is equal to the negative loss (reward) of the column player playing strategy $j \in [m]$ and equal to $A_{ij}$. 
\end{definition}

The fact that the losses were defined in the range $[-1,1]$ is arbitrary, as the concept of main importance we define next is invariant to scaling and shifting by a constant. 

A central concept in game theory is equilibrium. There are many different notions of equilibria. In two-player zero-sum games, a pure equilibrium is a pair of strategies $(i,j) \in [n] \times [m]$ with the following property: given that the column player plays $j$, there is no strategy that dominates $i$ - i.e., every other strategy $k \in [n] $ gives higher or equal loss to the row player. Equilibrium also requires that a symmetric property for strategy $j$ holds - it is not dominated by any other strategy given that the row player plays $i$.

It can be shown that some games do not have a pure equilibrium as defined above, e.g., the rock-paper-scissors game. However, we can extend the notion of a strategy to a {\it mixed} strategy - a distribution over pure strategies. The loss of a mixed strategy is the expected loss according to the distribution over pure strategies. 
More formally, if the row player chooses $\x \in \Delta_n$ and column player chooses $\y \in \Delta_m,$
then the expected loss of the row player (which is the negative reward to the column player) is given by:

\[
    \textbf{E}[\text{loss}] = \sum_{i \in [n]}{\x_i \sum_{j \in [m]}{\y_j A_{ij}}} = \x^{\top} A \y.
\]

We can now generalize the notion of equilibrium to mixed strategies. Given a row strategy $\x$, it is dominated by $\tilde{\x}$ with respect to a column strategy $\y$ if and only if 
$$ \x^\top A \y > \tilde{\x}^\top A \y.$$ 
We say that $\x$ is dominant with respect to  $\y$ if and only if it is not dominated by any other mixed strategy. A pair $(\x,\y)$ is an equilibrium for game $A$ if and only if both $\x$ and $\y$ are dominant with respect to each other. It is a good exercise for the reader at this point to find an equilibrium for the rock-paper-scissors game.

At this point, some natural questions arise: Is there always an equilibrium in a given zero-sum game? Can it be computed efficiently? Are there natural repeated-game-playing strategies that reach it? 

As we shall see, the answer to all questions above is affirmative. Let us rephrase these questions in a different way. Consider the optimal row strategy, i.e., a  mixed strategy $\x$, such that the expected loss  is minimized, no matter what the column player does.
The optimal strategy for the row player would be: 
$$ \x^\star \in \argmin_{\x \in \Delta_n} {\max_{\y \in \Delta_m} \x^{\top}A \y}.$$
Notice that we use the notation $\x^\star \in$ rather than $\x^\star = $, since in general the set of strategies attaining the minimal loss over worst-case column strategies can contain more than a single strategy. 
Similarly, the optimal strategy for the column player would be: 
$$ \y^\star \in \argmax_{\y \in \Delta_m} {\min_{\x \in \Delta_n} \x^{\top}A \y}.$$

Playing these strategies, no matter what the column player does, the row player would pay no more than
\[
    \lambda_R = \min_{\x \in \Delta_n} \max_{\y \in \Delta_m} \x^{\top} A \y = \max_{\y \in \Delta_m} {\x^{\star}}^{\top} A \y ,
\]
and column player would earn at least
\[
    \lambda_C =  \max_{\y \in \Delta_m} \min_{\x \in \Delta_n} \x^{\top} A \y = \min_{\x \in \Delta_n} {\x^{\top}} A \y^\star .
\]
 
With these definitions we can state von Neumann's famous minimax theorem:

\begin{theorem}[von Neumann minimax theorem]
In any zero-sum game, it holds that $\lambda_R = \lambda_C$.
\end{theorem}

This theorem answers all our above questions on the affirmative. The value $\lambda^\star = \lambda_C = \lambda_R$ is called the {\bf value of the game}, and its existence and uniqueness imply that any $\x^\star$ and $\y^\star$ in the appropriate optimality sets are an equilibrium.

We proceed to give a constructive proof of von Neumann's theorem which also yields an efficient algorithm as well as natural repeated-game playing strategies that converge to it.

\subsection{Equivalence of von Neumann Theorem and LP duality}

The von Neumann theorem is equivalent to the duality theorem of linear programming in a very strong sense, and either implies the other via simple reduction. Thus, it suffices to prove only von Neumann's theorem to prove the duality theorem.

The first part of this equivalence is shown by representing a zero-sum game as a primal-dual linear program instance, as we do now. 

Observe that the definition of an optimal row strategy and value is equivalent to the following  LP:
\begin{eqnarray*}
 \min \quad & & \lambda \\
\text{s.t.} \quad & & \sum{\x_i}=1 \\
   & & \forall i \in [m] \ . \ \x^{\top}A e_i \leq \lambda  \\
   & & \forall i \in [n] \ . \ \x_i \geq 0  .
\end{eqnarray*}
To see that the optimum of the above $LP$ is attained at $\lambda_R$, note that the constraint $\x^{\top}A e_i \leq \lambda \quad \forall i \in [m] $ is equivalent to the constraint $\forall \y \in \Delta_m \ . \ \x^\top A \y \leq \lambda$, since: 
\begin{eqnarray*}
\forall \y \in \Delta_m \ . \quad \x^\top A \y = \sum_{j=1}^m {\x^{\top}A e_j} \cdot \y_j   \leq  \lambda \sum_{j=1}^m {\y_j} = \lambda     
\end{eqnarray*}

The dual program to the above LP is given by
\begin{eqnarray*}
	\max  \quad  & & \mu \\
	\text{s.t.}  \quad  & & \sum{\y_i}=1 \\
	& & \forall i \in [n] \ . \ e_i^\top A \y \geq \mu  \\
	& & \forall i \in [m] \ . \ \y_i\geq0  .
\end{eqnarray*}

By similar arguments, the dual program precisely defines $\lambda_C$ and $\y^\star$. The duality theorem asserts that $\lambda_R = \lambda_C = \lambda^\star$, which gives von Neumann's theorem. 

The other direction, i.e., showing that von Neumann's theorem implies LP duality, is slightly more involved. Basically, one can convert any LP into the format of a zero-sum game. Special care is needed to ensure that the original LP is indeed feasible, as zero-sum games are always feasible and linear programs need not be. The details are left as an exercise at the end of this chapter.

\section{Proof of von Neumann Theorem}

In this section we give a proof of von Neumann's theorem using online convex optimization algorithms with sublinear regret. 

The first part of the theorem, which is also known as weak duality in the LP context, is rather straightforward: 

\textbf{Direction 1 ($\lambda_R \geq \lambda_C$):}

\begin{proof}
\begin{align*}
\lambda_R &  =   \min_{\x \in \Delta_n} \max_{\y \in \Delta_m} \x^{\top} A \y \\
&  = \max_{\y \in \Delta_m} {\x^{\star}}^{\top} A \y  & \mbox{ definition of $\x^\star$} \\
& \geq  \max_{\y \in \Delta_m} \min_{\x \in \Delta_n} \x^\top A \y  \\
& =  \lambda_C.
\end{align*}
\end{proof}
The second and main direction, known as strong duality in the LP context, requires the technology of online convex optimization we have proved thus far:

\textbf{Direction 2 ($\lambda_R \leq \lambda_C$):}

\begin{proof}
We consider a repeated game defined by the $n \times m$ matrix $A$.  For $t=1,2,...,T$, the row player provides a mixed strategy $\x_t \in \Delta_n$, column player plays mixed strategy $\y_t \in \Delta_m$, and the loss of the row player, which equals to the reward of the column player, equals $\x_t^\top A \y_t$. 

The row player generates the mixed strategies $\x_t$ according to an OCO algorithm --- specifically using the Exponentiated Gradient algorithm \ref{alg:eg} from chapter \ref{chap:regularization}.
The convex decision set is taken to be the $n$ dimensional simplex $\mathcal{K} =  \Delta_n = \{ \x \in \mathbb{R}^n \; | \; \x(i) \geq 0, \sum{\x(i)}=1 \}$. The loss function at time $t$ is given by  
$$ f_t(\x) =  \x^{\top}A\y_t \mbox{\ \ \ ($f_t$ is linear with respect to $\x$) } . $$

Spelling out the EG strategy for this particular instance, we have 
\[
    \x_{t+1}(i)  \gets \frac{  \x_t(i) e^{ -\eta A_i \y_t } } { \sum_j \x_{t}(i) e^{ -\eta A_j \y_t} }  \;.
\]
Then, by appropriate choice of $\eta$ and Corollary  \ref{cor:eg}, we have 
\begin{eqnarray} \label{eq:shalom5}
 \sum_t{f_t (\x_t)} & \leq  &  \min_{\x^\star \in \mathcal{K}}{\sum_t{f_t (\x^\star)}} + {\sqrt{2 T \log n }}  \;.  
\end{eqnarray}

The column player plays her best response to the row player's strategy, that is:
\begin{align} \label{shalom2}
\y_t = \arg \max_{\y \in \Delta_m} \x_t^\top A \y .
\end{align}

Denote the average mixed strategies by:
\[
\bar{\x} = \frac{1}{t} \sum_{\tau=1}^t {\x_\tau} \quad,\quad  \bar{\y} = \frac{1}{t} \sum_{\tau=1}^t {\y_\tau} \;.
\]

Then, we have
\begin{align*}
\lambda_R & = \min_\x \max_\y \ \x^\top A \y \\
& \leq \max_\y \bar{\x}^\top A \y & \mbox{special case}\\
& = \frac{1}{T} \sum_t \x_t A \y^\star \\
& \leq \frac{1}{T} \sum_t \x_t A \y_t & \mbox{ by \eqref{shalom2} }\\
& \leq \frac{1}{T} \min_\x \sum_t \x^\top A \y_t + \sqrt{2 \log n /T} & \mbox{ by \eqref{eq:shalom5} } \\
& = \min_\x \x^\top A \bar{\y} + \sqrt{2 \log n /T} \\
& \leq \max_\y \min_\x \x^\top A \y + \sqrt{2 \log n /T} & \mbox{special case}\\
& = \lambda_C	+ \sqrt{2 \log n /T}.
\end{align*}
Thus $\lambda_R \leq \lambda_C + \sqrt{2 \log n /T}$. As $T \rightarrow \infty$, we obtain part 2 of the theorem. 
\end{proof}

Notice that besides the basic definitions, the only tool used in the proof is the existence of sublinear regret algorithms for online convex optimization. The fact that the regret bounds for OCO algorithms were defined without restricting the cost functions, and that they can be adversarially chosen, is crucial for the proof. The functions $f_t$ are defined according to $\y_t$, which is chosen based on $\x_t$. Thus, the cost functions we constructed are adversarially chosen after the decision $\x_t$ was made by the row player.

\section{Approximating Linear Programs}

The technique in the preceding section not only proves the minimax theorem, and thus linear programming duality, but also entails an efficient algorithm. Using the equivalence of zero-sum games and  linear programs,  this efficient algorithm can be used to solve linear programming. We now spell out the details of this algorithm in the context of zero-sum games. 

Consider the following  algorithm: 
\begin{algorithm}
	[H] \caption{Simple LP } \label{alg:simpleLP}
	\begin{algorithmic}
            [1] \State Input: linear program in zero-sum game format,  by matrix $A \in \reals^{n \times m}$. 
            \State Let $\x_1  =  [ 1/n ,1/n,...,1/n] $
            \For{$t=1$ to $T$}
            \State Compute $\y_t  =  \max_{\y \in \Delta_m} {\x_t^\top A \y}  $
            \State Update $\forall i \ . \  \x_{t+1}(i)  \gets \frac{  \x_t(i) e^{ -\eta A_i \y_t } } { \sum_j \x_{t}(j) e^{ -\eta A_j \y_t} }  $
            \EndFor
            \State \Return $\bar{\x} = \frac{1}{T} \sum_{t=1}^T \x_t $
	\end{algorithmic}
\end{algorithm}

Almost immediately we obtain from the previous derivation the following:
\begin{lemma}
The returned vector $\bar{\x}$ of Algorithm \ref{alg:simpleLP} is a $\frac{\sqrt{2 \log n}}{\sqrt{T}}$-approximate solution to the zero-sum game and linear program it describes. 
\end{lemma}
\begin{proof}
Following the exact same steps of the previous derivation, we have
\begin{align*}
\max_\y \bar{\x}^\top A \y & = \frac{1}{T} \sum_t \x_t A \y^\star \\
& \leq \frac{1}{T} \sum_t \x_t A \y_t & \mbox{ by \eqref{shalom2} }\\
& \leq \frac{1}{T} \min_\x \sum_t \x^\top A \y_t + \sqrt{2 \log n /T} & \mbox{ by \eqref{eq:shalom5} } \\
& = \min_\x \x^\top A \bar{\y} + \sqrt{2 \log n /T} \\
& \leq \max_\y \min_\x \x^\top A \y + \sqrt{2 \log n /T} & \mbox{special case}\\
& = \lambda^\star	+ \sqrt{2 \log n /T} .
\end{align*}
Therefore, for each $i \in [m]$:
\[
    \bar{\x}^\top A e_i  \leq \lambda^\star + \frac{\sqrt{2 \log n}}{\sqrt{T}} 
\]
\end{proof}

Thus, to obtain an $\eps$-approximate solution, one would need $\frac{2 \log n}{\eps^2}$ iterations, each involving a  simple update procedure.

\newpage
\section{Bibliographic Remarks}

Game theory was founded in the late 1920's-early '30s, whose cornerstone was laid  in the classic text ``Theory of Games and Economic Behavior'' by \citet{neumann44a}. 

Linear programming is a fundamental mathematical optimization and modeling tool, dating back to the 1940's and the work of \citet{kantorovich40} and \citet{dantzig51}.  Duality for linear programming was conceived by von Neumann, as described by Dantzig in an interview \citep{dantzig}. For in depth treatment of the theory of linear programming there are numerous comprehensive texts, e.g., \citep{BertsimasLP,matousek2007understanding}. 

The beautiful connection between low-regret algorithms and solving zero-sum games was discovered by \citet{Freund199979}. More general connections of convergence of low-regret algorithms to equilibria in games were studied by \citet{hart2000simple}, and more recently in \citep{Even-dar:2009,Roughgarden:2015}.

Approximation algorithms that arise via simple Lagrangian relaxation techniques were pioneered by  \citet{PST}. See also the survey \citep{AHK-MW} and more recent developments that give rise to sublinear time algorithms \citep{CHW,hazan2011beating}.

\newpage
\begin{exercises}

%\section{Exercises}

%\begin{enumerate}
\exer{
Prove that equilibrium strategy pairs in zero-sum games are not unique. That is, construct a zero-sum game for which there is more than one equilibrium.
}

\exer{
In this question we prove a special case of Sion's generalization to the minimax theorem. Let $f: X \times Y \mapsto \reals$ be a real valued function on $X \times Y$, where $X,Y$ are bounded, closed and convex sets in Euclidean space $\reals^d$. Let $f$ be convex-concave, i.e., 
%\begin{enumerate}
	\subexer{
	For every $\y \in Y$, the function $f(\cdot,\y):X \mapsto \reals$ is convex. }
	\subexer{
	For every $\x \in X$, the function $f(\x,\cdot):Y \mapsto \reals$ is concave. }
%\end{enumerate}
Prove that
$$ \min _{\x \in X} \max_{\y \in Y} f(\x,\y) = \max_{\y \in Y} \min_{\x \in X} f(\x,\y) $$ 
}

\exer{
Read Adler's exposition on the equivalence of linear programming and zero sum games \citep{adler}. Explain how to convert a linear program to a zero-sum game. 
}

\exer{
Consider a repeated zero-sum game over a matrix $A$ in which both players change their mixed strategies according to a low-regret algorithm over the linear cost/reward functions of the game. Prove that the average value of the game approaches that of an equilibrium of the game given by $A$. 
}

\exer{ $^*$
Write a semidefinite program as a zero-sum game. Write down an algorithm for approximating the solution of a semidefinite program using OCO algorithms, and sketch an analysis of its correctness and performance bound. 
}

\end{exercises}
%\end{enumerate}

%!TEX root = OCObook.tex

%%%%%%%%%%%%%%%%%%%%%%%%%%%%%%%%%%%%%%%%%%%%%%%%%%%%%%%%%%%%
%%%%%%%%%%%%%%%%%%%%%%%%%%%%%%%%%%%%%%%%%%%%%%%%%%%%%%%%%%%%
%  PAC learning
%%%%%%%%%%%%%%%%%%%%%%%%%%%%%%%%%%%%%%%%%%%%%%%%%%%%%%%%%%%%
%%%%%%%%%%%%%%%%%%%%%%%%%%%%%%%%%%%%%%%%%%%%%%%%%%%%%%%%%%%%
\chapter{Learning Theory, Generalization, and Online Convex Optimization} \label{chap:online2batch}
\chaptermark{Learning Theory and OCO} 

In our treatment of online convex optimization so far we have only implicitly discussed learning theory.  The framework of OCO was shown to capture applications such as learning classifiers online, prediction with expert advice, online portfolio selection and matrix completion,  all of which have a learning aspect. We have introduced the metric of regret and gave efficient algorithms to minimize regret in various settings. We have also argued that minimizing regret is a meaningful approach for many online prediction problems.  However, the relation to other theories of learning was not discussed thus far. 

In this section we draw a formal and strong connection between OCO and the theory of statistical learning. We begin by giving the basic definitions of statistical learning theory, and proceed to describe how the applications studied in this manuscript relate to this model. We then continue to show how regret minimization in the setting of online convex optimization gives rise to computationally efficient statistical learning algorithms.

\section{Statistical Learning Theory}
\sectionmark{Statistical Learning Theory}

The theory of statistical learning addresses the problem of learning a concept from examples. A concept is a mapping from domain $\X$ to labels $\Y$, denoted $C : \X \mapsto \Y$.  

As an example, consider the problem of optical character recognition. In this setting,  the domain $\X$ can be  all $n \times n$ bitmap images,  the label set $\Y$ is  the Latin (or other) alphabet, and the concept $C$ maps a bitmap into the character depicted in the image. 

Statistical theory models the problem of learning a concept by allowing access to labelled examples from the target distribution. The learning algorithm has access to  pairs, or samples, from an unknown distribution
$$ (\bx,y) \sim \D  \quad , \quad \bx \in \X \ , \ y \in \Y. $$ 
%For example, one can think of email spam detection, in which $\bx$ are emails (usually represented in the bag of words representation as discussed in chapter \ref{chap:intro}), and $y$ is either one or zero, indicating that the message is spam or not spam. 

The goal is to be able to predict $y$ as a function of $\bx$, i.e., to {\bf learn} a hypothesis, or a mapping from $\X$ to $\Y$, denoted $ h: \X \mapsto \Y $, with small error with respect to the distribution $\D$. In the case that the label set is binary $\Y = \{0,1\}$, or discrete such as in optical character recognition, the {\it generalization error} of an hypothesis $h$ with respect to distribution $\D$ is given by
\[ \err(h) \equaltri \E_{(\bx,y)\sim \D} [ h(\bx) \neq y ] .\]

More generally, the goal is to learn a hypothesis that minimizes the loss according to a (usually convex) loss function $\ell: \Y \times \Y \mapsto \reals$.  In this case the generalization error of a hypothesis is defined as: 
\[  \err(h) \equaltri \E_{(\bx,y)\sim \D} [ \ell(h(\bx), y) ] .\]

We henceforth consider learning algorithms $\mA$ that observe a sample from the distribution $\D$ , denoted $S \sim \D^m$ for a sample of $m$ examples, $S = \{(\x_1,y_1),...,(\x_m,y_m)\}$, and produce a hypothesis $\mA(S) : \X \mapsto \Y $  based on this sample.  

\noindent The goal of statistical learning can thus be summarised as follows:

\begin{center}
\begin{minipage}{.81\textwidth}
\it
Given access to i.i.d. samples from an arbitrary distribution over $\X \times \Y$ corresponding to a certain concept, learn a hypothesis $h : \X \mapsto \Y$ which has arbitrarily small generalization error with respect to a given loss function. 
\end{minipage}
\end{center}

\subsection{Overfitting}

In the problem of optical character recognition  the task is to recognize a character from a given image in bitmap format. To model it in the statistical learning setting,   the domain $\X$ is the set  of  all $n \times n$ bitmap images for some integer $n$. The label set $\Y$ is  the latin alphabet, and the concept $C$ maps a bitmap into the character depicted in the image.  

Consider the naive algorithm which fits the perfect hypothesis for a given sample, in this case set of bitmaps. Namely, $\mA(S)$ is the hypothesis which correctly maps any given bitmap input $\x_i$ to its correct label $y_i$, and maps all unseen bitmaps to the character $``1."$

Clearly, this hypothesis does a very poor job of generalizing from experience - all images that have not been observed yet will be classified without regard to their properties, surely an erroneous classification most times.   However - the training set, or observed examples, are perfectly classified by this hypothesis! 

This disturbing  phenomenon is called ``overfitting,'' a central concern in machine learning.  Before continuing to add the necessary components in learning theory to prevent overfitting,  we turn our attention to a formal statement of when overfitting can appear.

\subsection{No free lunch?}

The following theorem shows that learning, as stated in the goal of statistical learning theory,  is impossible without restricting the hypothesis class being considered. For simplicity, we consider the zero-one loss in this section.

\begin{theorem} [No Free Lunch Theorem] \label{thm:nfl}
Consider  any  domain $\mathcal{X}$ of size $|\mathcal{X}| = 2m > 4$, and any algorithm $\mA$ which outputs a hypothesis $\mA(S) $ given a sample $S$ of size $m$. Then there exists a concept $C: \mathcal{X} \rightarrow \{0,1\}$ and a distribution $\mathcal{D}$ such that:
\begin{itemize}
\item The generalization error of the concept $C$ is zero.
\item With probability at least $\frac{1}{10}$, the error of the hypothesis 
generated by $\mA$ is at least $\err(A(S)) \geq \frac{1}{10}$. 
\end{itemize} 
\end{theorem}

The proof of this theorem  is based on the probabilistic method, a useful technique for showing the existence of combinatorial objects by showing that the probability they exist in some distributional setting is bounded away from zero. In our setting, instead of explicitly constructing a concept $C$ with the required properties, we show it exists by a probabilistic argument.

\begin{proof}
We  show that for any learning algorithm, there is some learning task (i.e., ``hard'' concept) that it will not learn well. Formally, take $\mathcal{D}$ to be the uniform distribution over $\X$. Our proof strategy will be to show the following inequality, where we take a uniform distribution over all concepts $\X \mapsto \{0,1\}$

$$Q \overset{def}{=} \E_{C:\X\rightarrow\{0,1\}} [\E_{S\sim \mathcal{D}^m} [\err(\mA(S))]] \geq \frac{1}{4} . $$
After showing this step, we will use Markov's Inequality to conclude the theorem. 

We proceed by using the linearity property of expectations, which allows us to swap the order of expectations, and then conditioning on the event that $\x \in S$.

\begin{equation*}
\begin{aligned}
Q & = \E_{S} [\E_{C} [\E_{\x \in \mathcal{X}} [\mA(S)(\x) \neq C(\x)]]]  \\
& = \E_{S,\x} [ \E_{C} [\mA(S)(\x) \neq C(\x)|\x \in S] \Pr [\x \in S] ]  \\ & + \E_{S,\x} [ \E_{C} [\mA(S)(\x) \neq C(\x)|\x \not \in S]  \Pr[ \x \not \in S] ].
\end{aligned}
\end{equation*}

All terms in the above expression, and in particular the first term, are non-negative and at least $0$. Also note that since the domain size is $2m$ and the sample is of size $|S| \leq m$, we have $\Pr(\x \not \in S ) \geq \frac{1}{2}$. Finally, observe that $\Pr[ \mA(S)(\x) \neq C(\x)] = \frac{1}{2}$ for all $\x \not\in S$ since we are given that the ``true'' concept $C$ is chosen uniformly at random over all possible concepts. Hence, we get that:

$$ Q \geq 0 + \frac{1}{2} \cdot \frac{1}{2} = \frac{1}{4}, $$
which is the intermediate step we wanted to show. The random variable $\E_{S\sim \mathcal{D}^m} [\err(\mA(S))] $ attains values in the range $[0,1]$. Since its expectation is at least $\frac{1}{4}$, the event that it attains a value of at least $\frac{1}{4}$ is non-empty. Thus, there exists a concept such that 
$$ \E_{S\sim \mathcal{D}^m} [\err(\mA(S))] \geq \frac{1}{4} $$
where, as assumed beforehand, $\mathcal{D}$ is the uniform distribution over $\X$.

We now conclude with Markov's Inequality: since the expectation above over the error is at least one-fourth, the probability over examples such that the error of $\mA$ over a random sample is at least  one-tenth is at least
$$ \Pr_{S \sim \D^m}  \left( \err(\mA(S) )  \geq \frac{1}{10}\right) \geq \frac{\frac{1}{4}-\frac{1}{10}}{1-\frac{1}{10}} > \frac{1}{10}.$$
\end{proof}

\subsection{Examples of learning problems}

The conclusion of the previous theorem is that the space of possible concepts  being considered in a learning problem needs to be restricted for any meaningful guarantee.   
Thus, learning theory concerns itself with concept classes, also called hypothesis classes, which are sets of possible hypotheses from which one would like to learn. We denote the concept (hypothesis) class by $\H = \{h : \X \mapsto \Y\}$. 

%\subsection{Examples of learning problems: domains, concepts and loss functions} 

Common examples of learning problems that can be formalized in this model and the corresponding definitions include: 
\begin{itemize}
\item
Optimal character recognition: In the problem of optical character recognition  the domain $\X$ consists of  all $n \times n$ bitmap images for some integer $n$,  the label set $\Y$ is a certain alphabet, and the concept $C$ maps a bitmap image into the character depicted in it.  A common (finite) hypothesis class for this problem is the set of all decision trees with bounded depth.

\item
Text classification:  In the problem of text classification the domain is a subset of Euclidean space, i.e., $\X \subseteq \reals^d$. Each document is represented in its bag-of-words representation, and $d$ is the size of the dictionary. The label set $\Y$ is binary, where one indicates a certain classification or topic, e.g.,``Economics'', and zero others. 

A commonly used hypothesis class for this problem is the set of  all bounded-norm vectors in Euclidean space  $\H = \{ h_\w \ , \ \w \in \reals^d \ , \ \|\w\|_2^2 \leq \omega \}$ such that $h_\w(\x) = \w^\top \x$. The loss function is chosen to be the hinge loss, i.e., $\ell(\hat{y},y) = \max\{0 , 1 - \hat{y} y \}  $.

\item
Recommendation systems: recall the online convex optimization formulation of this problem in section \ref{sec:recommendation_systems}. A  statistical learning formulation for this problem is very similar. The domain  is a direct sum of two sets $\X = \X_1 \oplus \X_2$. Here $\x_1 \in \X_1$ is a certain media item, and every person is an item $\x_2 \in \X_2$.  The label set $\Y$ is binary, where one indicates a positive sentiment for the person to the particular media item, and zero a negative sentiment. 

A commonly considered hypothesis class for this problem is the set of all mappings $\X_1 \times \X_2 \mapsto \Y$ that, when viewed as a matrix in $\reals^{|\X_1| \times |\X_2|}$, have bounded algebraic rank. 

\end{itemize}

\subsection{Defining generalization and learnability} 

We are now ready to give the fundamental definition of statistical learning theory, called Probably Approximately Correct (PAC) learning: 

\begin{definition}[PAC learnability] \label{def:learnability}
A hypothesis class $\H$  is PAC learnable with respect to loss function $\ell : \Y \times \Y \mapsto \reals$ if the following holds. There exists an algorithm $\mA$ that accepts $S_T = \{(\bx_t,y_t), \ t \in [T]\}$ and returns hypothesis  $\mA(S_T) \in \H$ that satisfies: 
 for any $\eps,\delta > 0$ there exists a sufficiently large natural number $T  = T(\eps,\delta)$,  such that for any distribution $\D$ over pairs $(\bx,y)$ and $T$ samples from this distribution, it holds that with probability at least $1-\delta$ 
$$ \err( \mA(S_T) ) \leq \eps. $$
\end{definition}

\noindent A few remarks regarding this  definition:
\begin{itemize}
\item
The set $S_T$ of samples from the underlying distribution is called the training set. The error in the above definition is called the {\bf generalization error}, as it describes the overall error of the concept  as generalized from the observed training set.  
The behavior of the number of samples $T$ as a function of the parameters $\eps,\delta$ and the concept class is called the {\bf sample complexity} of $\H$. 

\item
The definition of PAC learning says nothing about computational efficiency.  
Computational learning theory usually requires, in addition to the definition above, that the algorithm $\mA$ is efficient, i.e., polynomial running time with respect to $\eps,\log \frac{1}{\delta}$ and the representation of the hypothesis class. The representation size for a discrete set of concepts is taken to be the logarithm of the number of hypotheses in $\H$, denoted $\log |\H|$. 

\item
If the hypothesis $\mA(S_T)$ returned by the learning algorithm  belongs to the hypothesis class  $\H$, as in the definition above, we say that $\H$ is {\bf properly learnable}. More generally, $\mA$ may return hypothesis from a different hypothesis class, in which case we say that $\H$ is {\bf improperly learnable}.

\end{itemize}

The fact that the learning algorithm can learn up to {\it any} desired accuracy $\varepsilon > 0$ is called the {\bf realizability assumption} and greatly reduces the generality of the definition. It amounts to requiring that a hypothesis with near-zero error belongs to the hypothesis class. In many cases, concepts are only approximately learnable by a given hypothesis class, or inherent noise in the problem prohibits realizability (see exercises). 

This issue is addressed in the definition of a more general learning concept, called {\bf agnostic learning}:

\begin{definition}[agnostic PAC learnability] \label{def:agnosticlearnability}
The hypothesis class $\H $  is agnostically PAC learnable with respect to loss function $\ell : \Y \times \Y \mapsto \reals$ if the following holds. There exists an algorithm $\mA$ that accepts $S_T = \{(\bx_t,y_t), \ t \in [T]\}$ and returns hypothesis  $\mA(S_T) $ that satisfies: 
 for any $\eps,\delta > 0$ there exists a sufficiently large natural number $T  = T(\eps,\delta)$ such that for any distribution $\D$ over pairs $(\bx,y)$ and $T$ samples from this distribution, it holds that with probability at least $1-\delta$ 
$$ \err( \mA(S_T) ) \leq \min_{h \in \H} \{ \err(h) \} + \eps. $$
\end{definition}

With these definitions, we can state the fundamental theorem of statistical learning theory for finite hypothesis classes:
\begin{theorem}[PAC learnability of finite hypothesis classes] 
Every finite concept class $\H$ is agnostically PAC learnable with sample complexity that is $\poly(\eps,\delta, \log |\H|)$.
\end{theorem}

In the following sections we prove this theorem, and in fact a more general statement that holds also for certain infinite hypothesis classes.
The complete characterization of which infinite hypothesis classes are learnable is a deep and fundamental question, whose complete answer was  given by Vapnik and Chervonenkis (see bibliography).  
The question of which (finite or infinite) hypothesis classes are {\bf efficiently} PAC learnable, especially in the improper sense,  is still at the forefront of learning theory today.

\section{Agnostic Learning using Online Convex Optimization}

In this section we show how to use online convex optimization for agnostic PAC learning. Following the paradigm of this manuscript, we describe and analyze a reduction from agnostic learning to online convex optimization. The reduction is formally described in Algorithm \ref{alg:reductionOCO2LRN}.

\begin{algorithm}[H]
	\caption{Reduction: Learning $\Rightarrow$ OCO }
	\label{alg:reductionOCO2LRN}
	\begin{algorithmic}[1]
		\State Input: OCO algorithm $\mA$, convex hypothesis class $\H \subseteq \reals^d$, convex loss function $\ell$. 
		\State Let $h_1 \leftarrow \mA(\emptyset) $. 
		\For{$t = 1$ to $T$}
		\State Draw labeled example $(\bx_t,y_t) \sim \D$. 
		\State Let $f_t(h) = \ell( h(\bx_t) , y_t) $. 
		\State Update 
			$$ h_{t+1} = \mA( f_1,...,f_t) .$$
		\EndFor
		\State  Return $\bar{h} = \frac{1}{T} \sum_{t=1}^T h_t $.
	\end{algorithmic}
\end{algorithm}

For this reduction we assumed that the concept (hypothesis) class is a convex subset of Euclidean space. A similar reduction can be carried out for discrete hypothesis classes (see exercises).  In fact, the technique we explore below will work for any hypothesis set $\H$ that admits a low regret algorithm, and can be generalized to infinite hypothesis classes that are known to be learnable.

%Before we present the reduction algorithm, let's recall the regret bound we've seen in previous lectures for OCO:
%\begin{equation} \label{eq1}
%\sum_{t=1}^{\top}{f_{(x_{t},y_{t})}(h_{t})-\underset{h\in H}{\min}\sum_{t=1}^{\top}{f_{(x_{t},y_{t})}(h) = O\left(\sqrt{T}\right)}}
%\end{equation}
%

Let $h^\star = \arg\min_{h \in \H} \{ \err(h) \}$ be the hypothesis in the class $\H$ that minimizes the generalization error.   
Using the assumption that $\mA$ guarantees sublinear regret, our simple reduction implies PAC learning, as given in the following theorem. 

\begin{theorem} \label{thm:OCO2LRN}
Let $\mA$ be an OCO algorithm whose regret after $T$ iterations is guaranteed to be bounded by $\regret_T(\mA)$. 
%For $T = \frac{G^2D^2}{\epsilon^2}8\log(\frac{1}{\delta}) $ (using our standard notation of $D$ being the diameter of $\H$ and $G$ an upper bound on gradient size),  it holds that,
Then for any $\delta > 0$, with probability at least $1-\delta$, it holds that 
\begin{equation*}
\err(\bar{h})\le \err (h^\star) + \frac{ \regret_T(\mA) }{T} +\sqrt{\frac{8\log (\frac{2}{\delta})}{T}}. 
\end{equation*}
In particular, for $T = O( \frac{1}{\eps^2} \log \frac{1}{\delta}  + T_\eps(\mA) )  $, where $T_\eps(\mA)$ is the integer $T$ such that $\frac{\regret_T(\mA)}{T} \leq \eps$, we have 
\begin{equation*}
\err(\bar{h})\le \err (h^*) + \eps. 
\end{equation*}
\end{theorem}

How general is the theorem above? In the previous chapters we have described and analyzed OCO algorithms with regret guarantees that behave asymptotically as $O(\sqrt{T})$ or better. This translates to sample complexity of  $O(\frac{1}{\eps^2} \log \frac{1}{\delta})$ (see exercises), which is known to be tight for certain scenarios.

To prove this theorem we need some tools from probability theory, such as the concentration inequalities that we survey next.

\subsection{Reminder: measure concentration and martingales}

\begin{figure}[ht]
\begin{center}
\includegraphics[width=3.0in]{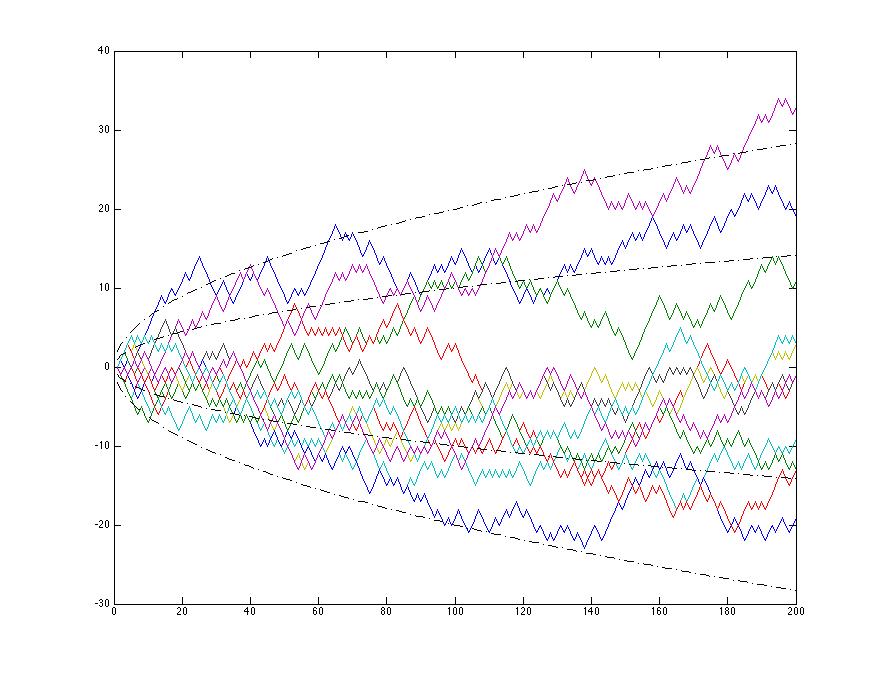}
\end{center}
\caption{Symmetric random walk: 12 trials of 200 steps. The black dotted lines show the functions $\pm \sqrt{x}$ and $\pm 2 \sqrt{x}$, respectively.  \label{fig:randomwalk}}
\end{figure}

Let us briefly discuss the notion of a martingale in probability theory. For intuition, it is useful to recall the simple random walk. Let $X_i$ be a Rademacher random variable which takes values
$$ X_i = \mycases {1} {\text{with probability } \quad \frac{1}{2} } {-1} {\text{with probability } \quad \frac{1}{2} } . $$ 
A simple symmetric random walk is described by the sum of such random variables, depicted in figure \ref{fig:randomwalk}.  Let $X = \sum_{i=1}^T X_i$ be the position after $T$ steps of this random walk. 
The  expectation and variance of this random variable  are $\E[ X] = 0 \ , \ \var(X) = T$.

The phenomenon of measure concentration addresses the probability of a random variable to attain values within range of its standard deviation. For the random variable $X$, this probability is much higher than one would expect using only the first and second moments. Using only the variance, it follows from Chebychev's inequality that
$$ \Pr\left[ |X| \geq c \sqrt{T} \right] \leq \frac{1}{c^2}. $$
However, the event that $|X|$ is centred around $O(\sqrt{T})$ is in fact much tighter, and can be bounded by the Hoeffding-Chernoff lemma
as follows 
\begin{eqnarray} \label{lem:chernoff}
\Pr[|X|  \ge c \sqrt{T}] \le 2  e^{\frac{-c^2}{2}} & \mbox  { Hoeffding-Chernoff lemma.}
\end{eqnarray}

Thus, deviating by a constant from the standard deviation decreases the probability exponentially, rather than polynomially.  This well-studied phenomenon generalizes to  sums of weakly dependent random variables and martingales, which are important for our application. 

\begin{definition}
A sequence of random variables $X_1, X_2,...$  is called a \emph {martingale} if it satisfies: 
\[
	\E [X_{t+1}|X_{t}, X_{t-1}...X_{1}] = X_{t} \quad \forall \; t>0.
\]
\end{definition}

A similar concentration phenomenon to the random walk sequence occurs in martingales. This is captured in the following theorem by Azuma.

\begin{theorem}[Azuma's inequality]
Let $ \big \lbrace X_{i} \big \rbrace _{i=1}^{T}$ be a  martingale of $T$ random variables  that satisfy  $|X_{i} - X_{i+1}|  \leq {1}$. Then:
\begin{equation*}
\Pr \left[ |X_{T} - X_{0}|>c \right]  \le 2 e^{\frac{-c^2}{2T}}.
\end{equation*}
\end{theorem}

By symmetry, Azuma's inequality implies,  
\begin{equation} \label{eqn:azuma2}
\Pr \left[ X_{T} - X_{0}> c \right]  = \Pr \left[X_{0} - X_{T}> c \right] \le e^{  \frac{-c^2}{2T}}. 
\end{equation}

\subsection{Analysis of the reduction}

We are ready to prove the performance guarantee for the reduction in Algorithm \ref{alg:reductionOCO2LRN}. Assume for simplicity that the loss function $\ell$ is bounded in the interval $[0,1]$, i.e., 
$$ \forall \hat{y},y \in \Y \ , \ \ell(\hat{y}, y) \in [0,1]. $$

\begin{proof}[Proof of Theorem \ref{thm:OCO2LRN}]
We start by defining a sequence of random variables that form a martingale. Let 
$$ 
Z_{t} \equaltri  \err(h_{t})- \ell(h_{t}(\x_{t}),y_{t})   , \quad  X_{t} \equaltri \sum_{i=1}^t Z_{i}. 
$$
Let us verify that $\{X_t\}$ is indeed a bounded martingale. Notice that by definition of $\err(h)$, we have that
$$ \E_{(\bx,y)\sim \D}[Z_{t} | X_{t-1}]  = \err(h_t) - \E_{(\bx,y)\sim \D} [\ell(h_t(\x) , y)] =  0  . $$   
Thus, by the definition of $Z_t$, 
\begin{eqnarray*}
\E [ X_{t+1}|X_{t},...X_{1} ] & = \E[Z_{t+1} | X_t]+ X_{t} 
 = X_t.
\end{eqnarray*}
In addition, by our assumption that the loss is bounded,
we have that (see exercises)
\begin{align} \label{eqn:martingale-bound}
|X_{t} - X_{t-1}| = |Z_{t}|  \le 1.
\end{align}
Therefore we can apply Azuma's theorem to the martingale $\{X_t\}$, or rather its consequence \eqref{eqn:azuma2}, and get 

\begin{equation*}
\Pr[X_{T}  > c] \le  e^{\frac{-c^2}{2T}}. 
\end{equation*}
Plugging in the definition of $X_{T}$, dividing by $T$ and using $c = \sqrt{2T\log (\frac{2}{\delta})}$:
\begin{equation} \label{eqn:oco2lrn1}
\Pr\left[ 
\frac{1}{T}\sum_{t=1}^{T}\err(h_{t})- \frac{1}{T}\sum_{t=1}^{T}{\ell(h_{t}(\x_{t}),y_{t})}  > \sqrt{\frac{2\log (\frac{2}{\delta})}{T}} 
\right] \le \frac{\delta}{2}.
\end{equation}

A similar martingale can be defined for $h^\star$ rather than $h_t$, and repeating the analogous definitions and applying Azuma's inequality we get:
\begin{equation} \label{eqn:oco2lrn2}
\Pr\left [  \frac{1}{T}\sum_{t=1}^{T}\err(h^\star)- \frac{1}{T}\sum_{t=1}^{T}{l(h^\star(\x_{t}),y_{t})}  < - \sqrt{\frac{2\log (\frac{2}{\delta})}{T}} 
\right] \le \frac{\delta}{2}.
\end{equation}
For notational convenience, let us use the following notation:
$$ \Gamma_1  = \frac{1}{T}\sum_{t=1}^{T}\err(h_{t})- \frac{1}{T}\sum_{t=1}^{T}{\ell(h_{t}(\x_{t}),y_{t})}, $$
$$ \Gamma_2  =  \frac{1}{T}\sum_{t=1}^{T}\err(h^\star)- \frac{1}{T}\sum_{t=1}^{T}{l(h^\star(\x_{t}),y_{t})}. $$
Next, observe that
\begin{align*} 
& \frac{1}{T}\sum_{t=1}^{T}\err(h_t)- \err(h^\star) \\
&  =   \Gamma_1 - \Gamma_2 
+ \frac{1}{T}\sum_{t=1}^{T}{\ell(h_{t}(\x_{t}),y_{t})} -  
\frac{1}{T}\sum_{t=1}^{T}{\ell(h^\star (\x_{t}),y_{t})}  \\
& \leq \frac{\regret_T(\mA )}{T} + \Gamma_1 - \Gamma_2, % & \mbox { $f_t(h) = \ell(h(\bx_t),y_t) $  } % \\
% & \leq |\Gamma_1 | + | \Gamma_2 | +  \regret_T(\mA) & \mbox { $\triangle$-inequality } \\
\end{align*}
where in the last inequality we have used the definition $f_t(h) = \ell(h(\bx_t),y_t) $.  From the above and Inequalities \eqref{eqn:oco2lrn1}, \eqref{eqn:oco2lrn2}  we get 
\begin{align*}
& \Pr \left [ \frac{1}{T}\sum_{t=1}^{T}\err(h_{t})- \err(h^\star) >  \frac{\regret_T(\mA)}{T} +2\sqrt{\frac{2\log (\frac{2}{\delta})}{T}} \right]  \\
& 
\le \Pr \left [   \Gamma_1 - \Gamma_2  >  2\sqrt{\frac{2\log (\frac{1}{\delta})}{T}} \right] \\
& \leq   \Pr \left [ \Gamma_1    >  \sqrt{\frac{2\log (\frac{1}{\delta})}{T}} \right]  + \Pr \left [   \Gamma_2  \le - \sqrt{\frac{2\log (\frac{1}{\delta})}{T}} \right] \\ 
& \leq \delta. \quad\quad\quad \mbox{Inequalities \eqref{eqn:oco2lrn1}, \eqref{eqn:oco2lrn2}}
\end{align*}
By convexity we have that $\err(\bar{h}) \le \frac{1}{T}\sum_{t=1}^{T}\err (h_{t})$. Thus, with probability at least $1-\delta$,
\begin{equation*}
\err(\bar{h}) \le \dfrac{1}{T}\sum_{t=1}^{T}\err (h_{t}) \le \err (h^\star) + \frac{ \regret_T(\mA) }{T} +\sqrt{\frac{8\log (\frac{2}{\delta})}{T}}. 
\end{equation*}
\end{proof}

\section{Learning and Compression}

Thus far we have considered finite and certain infinite hypothesis classes, and shown that they are efficiently learnable if there exists an efficient regret-minimization algorithm for a corresponding OCO setting. 

In this section we describe yet another property which is sufficient for PAC learnability: the ability to compress the training set. This property is particularly easy to state and use, especially for infinite hypothesis classes. It does not, however, imply efficient algorithms. 

Intuitively, if a learning algorithm is capable to express an hypothesis using a small fraction of the training set, we will show that it generalizes well to unseen data. For simplicity, we only consider learning problems that satisfy a variant of the realizablility assumption, i.e., the compression scheme generates a hypothesis that attains zero error. 

More formally, we define the notion of a compression scheme for a given learning problem as follows. The definition and theorem henceforth can be generalized to allow for loss functions, but for simplicity, consider only the zero-one loss function for this section. 
\begin{definition}(Compression Scheme) 
A distribution $\D$ over $\X \times \Y$ admits a compression scheme of size $k$, realized by an algorithm $\mA$, if the following holds.  For any $T > k$, let $S_T = \{(\bx_t,y_t), \ t \in [T]\}$ be a sample from $\D$. There exists an $S' \subseteq S_T \ , \ |S'| = k $, such that the algorithm $\mA$ accepts the set of $k$ examples $S'$, and returns a hypothesis  $\mA(S') \in \{ \X \mapsto \Y \}$, which satsifies: 
$$ \err_{S_T}( \mA(S') ) = 0 . $$
\end{definition}

The main conclusion of this section is that a learning problem that admits a compression scheme of size $k$ is PAC learnable with sample complexity proportional to $k$. 
This is formally given the following theorem. 

\begin{theorem} \label{thm:compression2generalization}
Let $\D$ be a data distribution that admits a compression scheme of size $k$ realized by algorithm $\mA$. 
Then with probability at least $1-\delta$ over the choice of a training set $|S_T|=T$, it holds that 
$$ \err ( \mA(S_T)) \leq \frac{8 k \log \frac{T}{\delta}}{T} . $$
\end{theorem}
\begin{proof}

Denote by $\err_S(h)$ the error of an hypothesis $h$ on a sample $S$ of i.i.d.\ examples, where the sample is taken independently of $h$. Since the examples are chosen independently, the probability that a hypothesis with $\err(h) > \eps$ has $\err_S(h) = 0$ is at most  $(1-\eps)^{|S|}$. Denote the event of $h$ satisfying these two conditions by ${h \in {\mbox{bad}}}$.

Consider a compression scheme for distribution $\D$ of size $k$, realized by $\mA$, and a sample of size $|S_T|=T \gg  k$. By definition of a compression scheme, the hypothesis returned by $\mA$ is based on $k$ examples chosen from the  set $S' \subseteq S_T$. We can bounds the  probability of the event that $\err_{S_T}(\mA(S')) = 0$ and $\err(\mA(S')) > \eps$, denoted by ${\mA(S') \in \mbox{bad}}$, as follows,
\begin{eqnarray*}
& \Pr[ {\mA(S') \in \mbox{bad}} ] \\
& = \sum_{S' \subseteq S_T, |S'|=k} \Pr[ {\mA(S') \in \mbox{bad}}   ] \cdot  \Pr[ S'] & \mbox{law of total probability} \\
& \leq \binom{T}{k}  (1-\eps)^T . 
\end{eqnarray*}

For $\eps = \frac{8k \log \frac{T}{\delta} }{T} $, we have that 
$$  \binom{T}{k}  (1-\eps)^T \leq T^k e^{-\eps T} \leq \delta .$$  
Since the compression scheme is guaranteed to return a hypothesis such that $\err_{S_T}(\mA(S')) = 0$, this implies that with probability at least $1-\delta$, the hypothesis $\mA(S')$ has $\err(\mA(S')) \leq \eps$.
\end{proof}

An important example of the use of compression schemes to bound the generalization error is for the hypothesis class of hyperplanes in $\reals^d$. It is left as an  exercise to show that this hypothesis class admits a compression scheme of size $d$.

\newpage
\section{Bibliographic Remarks}

The foundations of statistical and computational learning theory were put forth in the seminal works of \citet{Vapnik1998} and \citet{Valiant1984} respectively. There are numerous comprehensive texts on statistical and computational learning theory, see e.g.,  \citep{Kearns1994}, and the recent text \citep{shalev-shwartz_ben-david_2014}. 

Reductions from the online to the statistical (a.k.a. ``batch'') setting were initiated by Littlestone \citep{Littlestone89}. Tighter and more general bounds were explored in \citep{Cesa06,CesaGen08,Zhang05}.

The probabilistic method is attributed to Paul Erdos,  see the illuminating text of Alon and Spencer  \citep{AlonS92}. 

The relationship between compression and PAC learning was studied in the seminal work of \citet{LittlestoneW86}. For more on the relationship and historical connections between statistical learning and compression see the inspiring chapter in \citep{avibook}. More recently \citet{moran2016sample,david2016statistical} show that compression is equivalent to learnability in general supervised learning tasks and give quantitative bounds for this relationship.  

The use of compression for proving generalization error bounds has been applied in \citep{hanneke2019sample} for regression and in \citep{gottlieb2018near,kontorovich2017nearest} for nearest neighbor classification.  Another application is the recent work of \citet{bousquet2020proper} which gives optimal generalization error bounds for support vector machines using compression.

\newpage
\begin{exercises}

\exer{
Strengthen the no free lunch Theorem \ref{thm:nfl} to show  the following:  For any   $\varepsilon > 0$, there exists a finite domain $\X$, such that for any learning algorithm $\mA$ which given a sample $S$ produces hypothesis $\mA(S)$, there exists a distribution $D$ and a concept $C: X \mapsto \{0,1\}$ such that  
\subexer{
$\err(C) = 0$}
\subexer{
$ \E_{S \sim \D^m} [ \err (\mA(S)) ] \geq \frac{1}{2} - \varepsilon $.
}
}

\exer{
Let $\mA$ be an agnostic learning algorithm for the finite hypothesis class $\H : \X \mapsto \Y$ and the zero-one loss. Consider any concept $C: X \mapsto Y$ which is realized by $\H$, and the concept $\hat{C}$ which is obtained by  replacing the label associated with each domain entry $x \in X$ randomly with probability $\eps_0 > 0$ every time $x$ is sampled independently. That is:
$$ \hat{C}(x)  = \mythreecases {1} { \text{with probability }   \frac{ \eps_0}{2}} {0}{\text{with probability }  \frac{\eps_0}{2}} { C(x) } {\text{otherwise}} $$  

Prove that $\mA$ can $\eps$-approximate the concept $\hat{C}$: that is, show that  $\mA$ can be used to produce a hypothesis $h_{\mA} $ that has error
$$ \err_D (h_{\mA}) \leq  \frac{1}{2} \eps_0 + \eps  $$
with probability at least $1-\delta$ for every $\eps,\delta$ with sample complexity polynomial in $\frac{1}{\eps}, \log \frac{1}{\delta}, \log |H|$.  }

\exer{
Prove inequality \ref{eqn:martingale-bound}. 
}

\exer{ (Sample complexity of SVM) \\
Consider the class of hypothesis given by hyperplanes in Euclidean space with bounded norm 
$$ \H = \{ \x \in \reals^d \ , \ \| \x \|_2 \leq \lambda \} .$$  
Give an algorithm to PAC-learn this class with respect to the hinge loss function using reduction \eqref{alg:reductionOCO2LRN}. Analyze the resulting computational and sample complexity.  
}

\exer{
Show how to use a modification of reduction \ref{alg:reductionOCO2LRN} to learn a finite (non-convex) hypothesis class efficiently, i.e., without enumerating over all hypothesis. For this question, success probability of $\frac{1}{2}$ is sufficient. \\
Hint: instead of returning $\bar{h}$, consider returning a hypothesis at random.
}

\exer{
Consider the hypothesis class of all axis-aligned rectangles in the plane. That is, consider all  hypothesis parametrized by four real numbers, $a_x,b_x,a_y,b_y$, such that 
$$ h_{a_x,b_x,a_y,b_y}(\x) = \mycases {1} {\x_1 \in [a_x,b_x], \x_2 \in [a_y,b_y]} {0} {o/w} . $$
Prove that this hypothesis class admits a compression scheme of size $4$. 
}

\exer{$^*$
Consider the hypothesis class of all hyperplanes in $\reals^d$. This class is parameterized by all vectors in the unit sphere, such that 
$$ \forall \y \in \sphere_d \ , \  h_{\y} (\x) = \sign (\x^\top \y) .$$
Prove that this class has a compression scheme of size $d$. 
}

\end{exercises}

%%%%%%%%%%%%%%%%%%%%%%%%%%%%%%%%%%%%%%%%%%%%%%%%%%%%%%%%%%%%
%%%%%%%%%%%%%%%%%%%%%%%%%%%%%%%%%%%%%%%%%%%%%%%%%%%%%%%%%%%%
%  Adaptivity
%%%%%%%%%%%%%%%%%%%%%%%%%%%%%%%%%%%%%%%%%%%%%%%%%%%%%%%%%%%%
%%%%%%%%%%%%%%%%%%%%%%%%%%%%%%%%%%%%%%%%%%%%%%%%%%%%%%%%%%%%
\chapter{Learning in Changing Environments} \label{chap:adaptive}
\chaptermark{Adaptive Regret}

In online convex optimization the decision maker iteratively makes a decision without knowledge of the future, and pays a cost based on her decision and the observed outcome. The algorithms that we have studied thus far are designed to perform nearly as well as the best single decision in hindsight. The performance metric we have advocated for, average {\it regret} of the online player, approaches zero as the number of game iterations grows.

In scenarios in which the outcomes are sampled from some (unknown) distribution, regret minimization algorithms effectively ``learn" the environment and approach the optimal strategy. This was formalized in chapter \ref{chap:online2batch}. However, if the underlying distribution changes, no such claim can be made.

Consider for example the online shortest path problem we have studied in the first chapter. It is a well observed fact that traffic in networks exhibits changing cyclic patterns. A commuter may choose one path from home to work on a weekday, but a completely different path on the weekend when traffic patterns are different. Another example is the stock market: in a bull market the investor may want to purchase technology stocks, but in a bear market perhaps they would shift their investments to gold or government bonds.

When the environment undergoes many changes, standard regret may not be the best measure of performance. In changing environments, the online convex optimization algorithms we have studied thus far for strongly convex or exp-concave loss functions exhibit undesirable ``static" behavior, and converge to a fixed solution. 

In this chapter we introduce and study a generalization of the concept of regret called {\it adaptive} regret, to allow for a changing prediction strategy. 
We start with examining the notion of adapting in the problem of prediction from expert advice. We then continue to the more challenging setting of  online convex optimization, and derive efficient algorithms for minimizing this more refined regret metric.

\section{A Simple Start: Dynamic Regret}

Before giving the main performance metric studied in this chapter, we consider the first natural approach: measuring regret w.r.t. any sequence of decisions. Clearly,  in general it is impossible to compete with an arbitrary changing benchmark. However, it is possible to give a refined analysis that shows what happens to the regret  of an online convex optimization algorithm vs. changing decisions. 

More precisely, define the {\it dynamic regret} of an OCO algorithm with respect to a sequence $\uv_1,\ldots,\uv_T$ as:
\begin{eqnarray*}
\dregreteqn_T(\mA,\uv_1,\ldots,\uv_T) & \equaltri &  \sum_{t=1}^T f_t(\x_t) - \sum_{t=1}^T f_t(\uv_t)  
\end{eqnarray*}

To analyze the dynamic regret, some measure of the complexity of the sequence $\uv_1,\ldots,\uv_T$ is necessary. Let $\mP(\uv_1,\ldots,\uv_T)$ be the path length of the comparison sequence defined as
$$ \mP(\uv_1,\ldots,\uv_T) = \sum_{t=1}^{T-1} \|\uv_t - \uv_{t+1}\|  + 1. $$

It is natural to expect the regret to scale with the path length, as indeed the following theorem shows. For a fixed comparator $\uv_t = \x^\star $, the path length is one, and thus Theorem \ref{thm:dynamic-regret} recovers the $O(\sqrt{T})$ standard regret bound. For simplicity, we assume that the time horizon $T$ is known ahead of time, and so is the path length of the comparator sequence, although these limitations can be removed (see bibliographic section). 
\begin{theorem}\label{thm:dynamic-regret}
Online Gradient Descent (algorithm \ref{alg:ogd}) with step size $\eta \approx \sqrt{\frac{\mP(\uv_1,...,\uv_T) }{T}} $ guarantees the following dynamic regret bound:
$$ \dregreteqn_T(\mA,\uv_1,\ldots,\uv_T) = O( \sqrt{T  \mP(\uv_1,\ldots,\uv_T) } ) $$ 
\end{theorem}

\begin{proof}
Using our notation, and following the steps of the proof of Theorem \ref{thm:gradient},
\begin{eqnarray*}
\|\bx_{t+1}-\uv_t\|^2\ \leq  \|\by_{t+1}-\uv_t\|^2 = \|\bx_t- \uv_t\|^2 + \eta^2
\|\nabla_t\|^2 -2 \eta \nabla_t^\top (\bx_t -\uv_t) .
\end{eqnarray*}
Thus, 
\begin{eqnarray*}
2 \nabla_t^\top (\bx_t-\uv_t)\ &\leq  \frac{ \|\bx_t-
\uv_t\|^2-\|\bx_{t+1}-\uv_t\|^2}{\eta} + \eta G^2 %\\
%& \leq \frac{1}{\eta} \left( \|\x_t\|^2 - \|\x_{t+1}\|^2 + 
\end{eqnarray*}
Using convexity and summing this inequality across time we get
\begin{align*}
& 2 \left( \sum_{t=1}^T f_t(\bx_t)-f_t(\uv_t) \right ) \leq 2\sum_{t=1}^T \nabla_t^\top (\xv - \x^\star) \\
&\leq  \sum_{t=1}^T \frac{ \|\bx_t-
	\uv_t\|^2-\|\bx_{t+1}-\uv_t\|^2}{\eta} + \eta G^2 T     \\
& =   \frac{1}{\eta}  \sum_{t=1}^T  \left( \|\x_t\|^2 - \|\x_{t+1}\|^2 + 2 \uv_t^\top (\x_{t+1} - \x_{t})  \right) 
 + \eta  G^2 T  \\
&\leq \frac{2}{\eta} \left( D^2  +  \sum_{t=2}^{T}  \x_t^\top ( \uv_{t-1} - \uv_{t})   + \uv_T^\top \x_{T+1} - \uv_1^\top \x_1 \right)  + \eta G^2 T   \\
&\leq \frac{3}{\eta} \left(  D^2   + D \sum_{t=2}^{T}  \| \uv_{t-1} - \uv_{t} \|  \right)  + \eta G^2 T &  \uv_t \in \K   \\
& \leq \frac{3D^2 }{\eta} \mP(\uv_1,...,\uv_{T} ) + \eta G^2 T . %\leq 3 DG \sqrt{T \mP(\uv_1,\ldots,\uv_T) }.
\end{align*}
The theorem now follows by choice of $\eta$. 
\end{proof}

This simple modification to the analysis of online gradient descent   naturally extends to online mirror descent, as well as to other notions of path distance of the comparison sequence. 

We now turn to another metric of performance that requires more advanced   methods than we have seen thus far. This metric can be shown to be more general than dynamic regret, in the sense that the bounds we prove also imply low dynamic regret.

\section{The Notion of Adaptive Regret}

The main performance metric we consider in this chapter is designed to measure the performance of a decision maker in a changing environment. It is formally given in the following definition.
\begin{definition} \label{def:adaptiveregret}
The adaptive regret of an online convex optimization algorithm $\mA$ is defined as the maximum regret it achieves over any contiguous time interval. Formally,
\begin{eqnarray*}
\newregreteqn_T(\mA) & \equaltri & \sup_{I = [r,s] \subseteq [T]} \left\{   \sum_{t=r}^s f_t(\x_t) - \min_{x^*_I \in \K} \sum_{t=r}^s f_t(\x^*_I) \right\} \\
& =  & \sup_{I = [r,s] \subseteq [T]} \left\{ \regret_{[r,s]}(\mA) \right\} .
\end{eqnarray*}
\end{definition}

As opposed to standard regret, the power of this definition stems from the fact that the comparator is allowed to change. In fact, it is allowed to change indefinitely with every interval of time. 

For an algorithm with low adaptive regret, as opposed to standard regret, how would its performance guarantee differ in a changing environment? Consider the problem of portfolio selection, for which time can be divided into disjoint segments with different characteristics: bear market in the first $T/2$ iterations and bull market in the last $T/2$ iterations. 
A (standard) sublinear regret algorithm is only required to converge to the average of both optimal portfolios, clearly an undesirable outcome. However, an algorithm with sublinear \newregret bounds would \emph{necessarily} converge to the optimal portfolio in both intervals.

Not only does this definition make intuitive sense, but it generalizes other natural notions. For example, consider an OCO setting that can be divided into $k$ intervals, such that in each a different comparator is optimal. Then an \newregret guarantee of $\newregreteqn_T = o(T)$ would translate to overall regret of $k \times \newregreteqn_{T/k}$ compared to the best $k$-shifting comparator\endnote{In certain cases this can be $k \times \newregreteqn_T$, depending on the particular algorithm used.}.

\subsection{Weakly and strongly adaptive algorithms}
%\subsection{Convex vs. exp-concave loss functions}

The Online Gradient Descent algorithm over general convex losses, with step sizes $O(\frac{1}{\sqrt{t}})$, attains an adaptive regret guarantee of 
$$ \newregreteqn_T(OGD) = O(\sqrt{T}) ,$$
and  this bound is tight. 
This is a simple consequence of the analysis we have already seen in chapter \ref{chap:first order}, and left as an exercise.  Unfortunately this guarantee is meaningless for intervals of length $o(\sqrt{T})$.

Recall that for strongly convex loss functions, the OGD algorithm with the optimal learning rate schedule attains $O(\log T)$ regret. However, it  does {\bf not} attain any non-trivial \newregret guarantee: its adaptive regret can be as large as $\Omega(T)$, and this is also left as an exercise.

%However, it is possible to obtain more refined bounds in terms of the interval length. 

An OCO algorithm $\mA$ is said to be {\it strongly adaptive} if its adaptive regret can be bounded by its regret over the interval up to logarithmic terms in $T$, i.e. 
\begin{eqnarray*}
%& \sup_{I = [r,s] \subseteq [T]} \left\{   \sum_{t=r}^s f_t(\x_t) - \min_{x^*_I \in \K} \sum_{t=r}^s f_t(\x^*_I) \right\} \\
&  \newregreteqn_T(\mA)   = O(\regret_I(\mA)  \cdot \log^{O(1)} T). 
\end{eqnarray*}

%Our goal henceforth is  to design strongly adaptive algorithms. % As a consequence of the definitions, logarithmic regret algorithms that are strongly adaptive have poly-logarithmic adaptive regret.

The natural question is thus: are there algorithms that attain the optimal regret guarantee, and simultaneously the optimal adaptive regret guarantee?  As we shall see, the answer is affirmative in a strong sense: we shall describe and analyze algorithms that are optimal in both metrics.  Furthermore, these algorithms can be implemented with small computational overhead over the non-adaptive methods we have already studied.

\section{Tracking the Best Expert}

Consider the fundamental problem studied in the first chapter of this text, prediction from expert advice, but with a small twist. Instead of a static best expert, consider the setting in which different experts are the ``best expert" in different time intervals. 
More precisely, consider the situation in which time $[T]$ can be divided into $k$ disjoint intervals such that each admits a different ``locally best" expert. Can we learn to track the best expert? 

This tracking problem was historically the first motivation to study adaptivity in online learning. Indeed, as shown by Herbster and Warmuth (see bibliographic section), there is a natural algorithm that attains optimal regret bounds. 

The Fixed Share algorithm, describe in Algorithm \ref{alg:fixed-share}, is a variant of the Hedge Algorithm \ref{alg:Hedge}. On top of the familiar multiplicative updates,  it adds a uniform exploration term whose purpose is to avoid the weight of any expert from becoming too small. This provably allows a regret bound that tracks the best expert in any interval. 

\begin{algorithm}[ht]
\caption{Fixed Share}  \label{alg:fixed-share}
\begin{algorithmic}[1]
\State Input: parameter $\delta < \frac{1}{2}$. Initialize $\forall i \in [N] , p_i^1 = \frac{1}{N}$.
\For{$t=1$ to $T$}
\State Play $\x_t = \sum_{i=1}^N p_t^i \x_t^{i}$.
\State  After receiving $f_t$,  update for $1 \leq i \leq N$  \\
$$ \hat{p}^{i}_{t+1} = \frac{p^{i}_t e^{-\alpha f_t(\x^{i}_t)}}{\sum_{j=1}^N p^{j}_t e^{-\alpha f_t(\x^{j}_t)}} $$
\State Fixed-share step: 
$$ p_{t+1}^{i}  = (1 - \delta )\hat{p}^{i}_{t+1} + \frac{\delta}{N} $$
\EndFor
\end{algorithmic}
\end{algorithm}

In line with the notation we have used throughout this manuscript, we denote decisions in a convex decision set by $\x \in \K$. An expert $i$ suggests decision $\x_t^i$, and suffers loss according to a convex loss function denoted $f_t(\x_t^i)$. 
The main performance guarantee for the Fixed Share algorithm is given in the theorem below. 

\begin{theorem} \label{thm:fixed-share} 
Given a sequence of $\alpha$-exp-concave loss functions, the Fixed-Share algorithm with $\delta = \frac{1}{2 T}$ guarantees 
$$ \sup_{I = [r,s] \subseteq [T]} \left\{   \sum_{t=r}^s f_t(\x_t) - \min_{i^* \in [N]} \sum_{t=r}^s f_t(\x^{i^*}_t) \right\} =  O\left(\frac{1}{\alpha} \log N T \right) .$$
\end{theorem}

Notice that this is a different guarantee than adaptive regret as per Definition \ref{def:adaptiveregret}, as the decision set is discrete. However, it is a crucial component in the adaptive algorithms we will explore in the next section. 

As a direct conclusion from this theorem, it can be shown (see exercises) that if the best expert changes $k$ times in a sequence of length $T$, the overall regret compared to the best expert in every interval is bounded by
$$ O \left( k \log \frac{NT}{k} \right) . $$

To prove this theorem, we start with the following lemma, which is a fine-grained analysis of the multiplicative weights properties: 
\begin{lemma} \label{lem:round-reg}
For all $1 \leq i < N$, 
$$ f_t(\x_t) - f_t(\x^{i}_t) \leq \alpha^{-1} (\log \hat{p}^{i}_{t+1} - \log \hat{p}^{i}_{t} - \log (1 - \delta) ) .$$
\end{lemma}

\begin{proof} Using the $\alpha$-exp concavity of $f_t$, 
\begin{eqnarray*}
e^{-\alpha f_t(\x_t)} & = & e^{-\alpha f_t(\sum_{j=1}^N p^{j}_t \x^{j}_t)}  \geq  \sum_{j=1}^N p^{j}_t e^{-\alpha f_t(\x^{j}_t)}.
\end{eqnarray*}
Taking the natural logarithm,
$$f_t(\x_t) \leq  -\alpha^{-1} \log \sum_{j=1}^N p^{j}_t e^{-\alpha f_t(\x^{j}_t)} \nonumber $$
Hence,
\begin{eqnarray*}
  f_t(\x_t) - f_t(\x^{i}_t)  & \leq  \alpha^{-1}(\log e^{-\alpha f_t(\x^{i}_t)} - \log \sum_{j=1}^N p^{j}_t e^{-\alpha f_t(\x^{j}_t)})  \\
& =  \alpha^{-1} \log \frac{e^{-\alpha f_t(\x^{i}_t)}}{\sum_{j=1}^N p^{j}_t e^{-\alpha f_t(\x^{j}_t)}}  \\
& =  \alpha^{-1} \log \left(\frac{1}{p^{i}_t} \cdot
\frac{p^{i}_te^{-\alpha f_t(\x^{i}_t)}}{\sum_{j=1}^N p^{j}_t e^{-\alpha f_t(\x^{j}_t)}} \right)  \\
& =  \alpha^{-1} \log \frac{\hat{p}^{i}_{t+1}}{p^{i}_t} =  \alpha^{-1} (\log \hat{p}^{i}_{t+1} - \log {p}^{i}_{t} ) 
\end{eqnarray*}
The proof is completed observing that:
\begin{eqnarray*}
\log p^{i}_t & = \log \left( (1 - \delta)\hat{p}^{i}_{t} + \frac{\delta}{N}  \right) \\
& \geq  \log \hat{p}^{i}_t + \log (1 - \delta ) . %\geq \log \hat{p}^{i}_t  \log(1- \delta)   . 
\end{eqnarray*}

\end{proof}

Theorem \ref{thm:fixed-share} can now be derived as a corollary:
\begin{proof}[Theorem \ref{thm:fixed-share}]
By summing up over the interval $I = [r,s]$, and using the lower bound on $p_t^i$, we have
\begin{eqnarray*}
& \sum_{t \in I } f_t(\x_t) - \sum_{t \in I} f_t(\x^{i}_t ) \\
& \leq \sum_{t \in I} \alpha^{-1} (\log \hat{p}^{i}_{t+1} - \log \hat{p}^{i}_{t} - \log (1 - \delta) ) \\
& \leq \frac{1}{\alpha} \left[ \log \frac{1}{\hat{p}^i_r} - |I| \log (1-\delta) \right] \\
& \leq \frac{1}{\alpha} \left[ \log \frac{N}{\delta} + 2 \delta |I| \right] & \hat{p}^i_r \geq \frac{\delta}{N}, \delta < \frac{1}{2} \\
& \leq \frac{1}{\alpha} \log 2 N T + \frac{1}{\alpha}  & \delta = \frac{1}{2T} 
\end{eqnarray*}
\end{proof}

\section{Efficient Adaptive Regret for Online Convex Optimization} \label{sec:basic}

The Fixed-Share algorithm described in the previous section is extremely practical and efficient for discrete sets of experts. However, to exploit the full power of OCO we require an efficient algorithm for continuous decision sets. 

Consider for example the problems of online portfolio selection and online shortest paths: na\"ively applying the Fixed-Share algorithm is computationally inefficient.  Instead, we seek an algorithm which takes advantage of the efficient representation of these problems in the language of convex programming. 

We present such a method called FLH, or Follow the Leading History. The basic idea is to think of different online convex optimization algorithms starting at different time points as experts, and apply a version of Fixed Share to these experts.

\begin{algorithm}
\caption{Follow the Leading History}  \label{alg:flh1}
\begin{algorithmic}[1]
\State Let $\mA$ be an OCO algorithm. Initialize $p_1^1 = 1$
\For{$t=1$ to $T$}
\State Set $\forall j \leq t \ , \ \x^{j}_t \leftarrow \mA(f_j,...,f_{t-1})$  \label{eqn:shalom12} 
\State Play $\x_t = \sum_{j=1}^t p^{j}_t \x^{j}_t$.
\State After receiving $f_t$,  update for $1 \leq i \leq t$  \\
$$ \hat{p}^{i}_{t+1} = \frac{p^{i}_t e^{-\alpha f_t(\x^{i}_t)}}{\sum_{j=1}^t p^{j}_t e^{-\alpha f_t(\x^{j}_t)}} $$
\State Mixing step: set $ p^{t+1}_{t+1} = \frac{1}{t+1}$ and
$$ \forall i \neq t+1 \ , \ p^{i}_{t+1} = \left(1 - \frac{1}{t+1} \right)\hat{p}^{i}_{t+1} . $$
\EndFor
\end{algorithmic}
\end{algorithm}

The main performance guarantee is given in the following theorem. 
\begin{theorem} \label{thm:main-flh1} 
Let  $\mA$ be an OCO algorithm for $\alpha$-exp-concave loss function with 
$\regret_T(\mA)$. Then, 
$$\mbox{\newregreteqn}_T(FLH) \leq \regret_T(\mA)  + O(\frac{1}{\alpha} \log T) . $$
In particular, taking  $\mA\equiv ONS$ guarantees 
$$ \mbox{\newregreteqn}_T =  O(\frac{1}{\alpha} \log
T) .$$
\end{theorem}
Notice that FLH invokes $\mA$ at iteration $t$ at most $T$ times. Hence its running time is bounded by $T$ times that of $\mA$. This can still be prohibitive as the number of iterations grows large. In the next section, we show how the ideas from this algorithm can give rise to an efficient adaptive algorithm with only $O(\log T)$ computational overhead and slightly worse regret bounds.

The analysis of FLH is very similar to that of Fixed Share, with the main subtleties due to the fact that the time horizon $T$ is not assumed to be known ahead of time, and thus the number of experts varies with time. 

Instead of giving the full analysis, which is deferred to the exercises, we give a simplified version of FLH which does assume a-priory knowledge of $T$, and whose analysis can be directly reduced to that of Theorem \ref{thm:fixed-share}. 

\begin{algorithm}
\caption{Simple-FLH}  \label{alg:simple-flh}
\begin{algorithmic}[1]
\State Let $\mA$ be an OCO algorithm. Set $N=T, \delta= \frac{1}{2T}$. 
\For{$t=1$ to $T$}
\State For all $i \leq t$, set $ \x^{i}_t \leftarrow \mA(f_j,...,f_{t-1})$. Otherwise, set $\x_t^i = \bzero$. 
\State Apply the Fixed Share algorithm with expert predictions $\x_t^i$. 
\EndFor
\end{algorithmic}
\end{algorithm}

The simplified version of FLH is given in Algorithm \ref{alg:simple-flh}, and it guarantees the following adaptive regret bound. 
\begin{theorem} \label{thm:simple-flh}
Algorithm \ref{alg:simple-flh} guarantees:
$$\mbox{\newregreteqn}_T(\mbox{Simple-FLH}) \leq \regret_T(\mA)  + O(\frac{1}{\alpha} \log T) . $$
\end{theorem}
\begin{proof}
Applying Theorem \ref{thm:fixed-share} to the experts defined in Simple FLH, guarantees for every interval in time $I = [r,s]$, and by choice of $N$, for every $i \leq s$, 
\begin{eqnarray*}
\sum_{t \in I } f_t(\x_t) - \sum_{t \in I} f_t(\x^{i}_t ) \leq  \frac{1}{\alpha} \log 2 N T + 1  = O(\frac{1}{\alpha} \log T) .
\end{eqnarray*}
In particular, consider the sequence of predictions given by the $r$'th expert, for which we have
\begin{eqnarray*}
\sum_{t \in I }  f_t(\x^{r}_t ) = \regret_{s-r+1}(\mA) \leq \regret_T(\mA).
\end{eqnarray*}
The theorem now follows since this holds for every iterval $I \subseteq [T]$. 
\end{proof}

\section{* \ Computationally Efficient Methods } \label{sec:pruning}

In the previous section we studied adaptive regret, introduced and analyzed an algorithm that attains near optimal \newregret bounds. However, FLH suffers from a significant computational and memory overhead: it requires maintaining $O(T)$ copies of an online convex optimization algorithm. This computational overhead, which is proportional to the number of iterations, can be prohibitive in many applications.  
In this section our goal is to implement the algorithmic template of FLH efficiently and using little space.

To be more precise, henceforth denote the running time per iteration of algorithm $\mA$ as $V_t(\mA)$. 
Recall that at time $t$,  FLH stores all predictions $\{\x_t^i  \ | \  i \in [t]\}$  and has to compute weights for
all of them. This requires running time of at least $O(V_t(\mA) \cdot t)$.

The FLH2 algorithm, described in Algorithm \ref{alg:flh2}, significantly cuts down this running time to being only logarithmic in the current time iteration parameter $t$. To achieve this, FLH2 applies a pruning method to cut down the number of active online algorithms from $t$ to $O(\log t)$. However, its adaptive regret guarantee is slightly worse, and suffers a multiplicative factor of $O(\log T)$ as compared to FLH.

\begin{algorithm}
\caption{FLH2}\label{alg:flh2}
\begin{algorithmic}[1]
\State Let $\mA$ be an OCO algorithm. Initialize $p_1^1 = 1, S_1 = \{1\}$

\For{$t=1$ to $T$}
\State Set $\forall j \in S_t \ , \ \x^{j}_t \leftarrow \mA(f_j,...,f_{t-1})$ 
\State Play $\x_t = \sum_{j \in S_t} p^{j}_t \x^{j}_t$.
\State \label{algstep:mw} After receiving $f_t$, perform update for $i \in S_t$:  \\    
$$ \hat{p}^{i}_{t+1} = \frac{p^{i}_t e^{-\alpha f_t(\x^{i}_t)}}{\sum_{j \in S_t} p^{j}_t e^{-\alpha f_t(\x^{j}_t)}} $$
\State \label{algstep:mixing} Pruning: set $S_{t+1} \leftarrow \mbox{Prune}(S_t) \cup \{t+1\}$.  Set $\hat{p}^{t+1}_{t+1}$ to $\frac{1}{t}$, and update:
	$$ \forall i \in S_{t+1} \ . \   p^{i}_{t+1} = \frac{ \hat{p}^{i}_{t+1}} {\sum_{j \in S_{t+1}} \hat{p}^j_{t+1} } $$ 
\EndFor
\end{algorithmic} 
\end{algorithm}

 Before giving the exact details of this pruning method, we state the performance guarantee for FLH2.
\begin{theorem} \label{thm:flh2} 
Given an OCO algorithm $\mA$ with regret $\regret_T(\mA)$ and running time 
$V_T(A)$,  algorithm  $FLH2$  guarantees:
 $V_T(FLH2) \leq V_T(\mA)  \log T $ and 
$$\mbox{\newregreteqn}_T(FLH2) \leq \regret_T(\mA) \log T + O(\frac{1}{\alpha} \log^2 T) . $$
\end{theorem}

The main conclusion from this theorem is obtained by using FLH2 with $\mA$ being the ONS algorithm from chapter \ref{chap:second order-methods}. This gives adaptive regret of $O(\frac{1}{\alpha} \log^2 T)$ and  running time which is polynomial in natural parameters of the problem and poly-logarithmic in the number of iterations.

Before diving into the analysis, we explain the main new ingredient. 
At the heart of this algorithm is a new method for incorporating history. We will show that it suffices to store only $O(\log t)$
experts at time $t$, rather than all $t$ experts as in FLH. 

At time $t$, there is a working set
$S_t$ of experts. In FLH, this set
can be thought of to contain
$E^1,\cdots,E^t$, where each $E^i$ is the algorithm $\mA$ starting from iteration $i$.  For the next round, a new expert $E^{t+1}$
is added to get $S_{t+1}$. The complexity and regret of FLH is directly related to the cardinality of these sets. 

The key to decreasing the sizes
of the sets $S_t$ is to  also \emph{remove} (or {\it prune})
some experts. Once an expert
is removed, it is never used again. The algorithm
will perform the multiplicative update and mixing steps (steps \ref{algstep:mw} and \ref{algstep:mixing}  in algorithm \ref{alg:flh2})
only on the working set of experts.

The problem of maintaining the set of active experts can be thought of as the following abstract data streaming problem. Suppose the integers $1,2,\cdots$ are being ``processed"
in a streaming fashion. At time $t$, we have ``read"
the positive integers up to $t$ and maintain a very
small subset
of them in $S_t$.  At time $t$ we create $S_{t+1}$ from $S_t$: we are allowed to add to $S_t$ only the integer $t+1$, and remove some integers already in $S_t$.
Our aim is to maintain a set $S_t$ which satisfies:
\begin{enumerate}
\item
 For every positive $s \leq t$, $[s,(s+t)/2] \cap S_t \neq \emptyset$.
 \item
  For all $t$, $|S_t| = O(\log T)$.
  \item For all $t$, $S_{t+1}\backslash S_t = \{t+1\}$.
\end{enumerate}

\begin{figure}[ht]
\begin{center}
\includegraphics[width=3in]{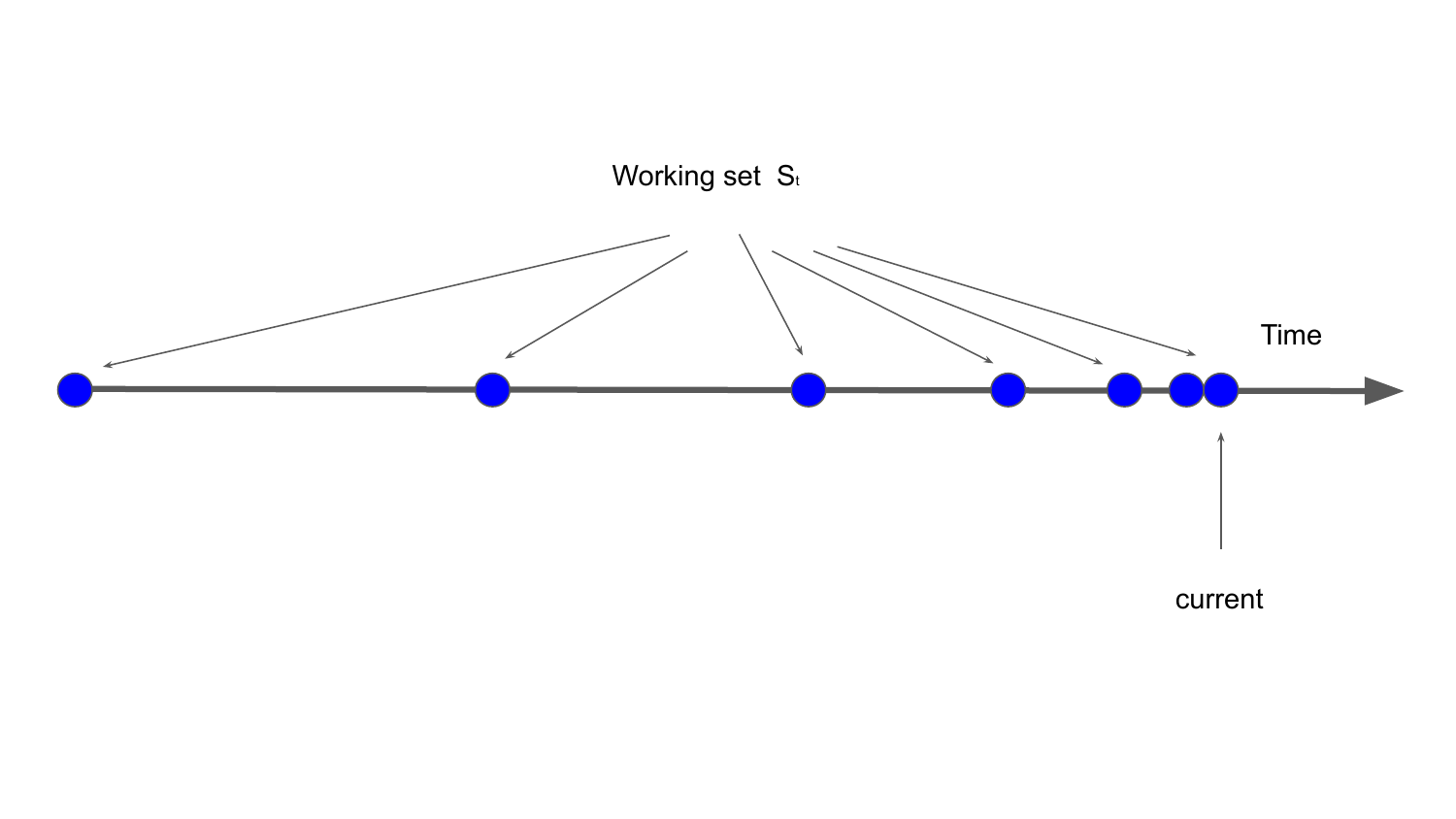}%projection}
\end{center}
\caption{Illustration of the working set $S_t$}\label{fig-st}
\end{figure}

The first property of the sets $S_t$ intuitively means that $S_t$ is ``well spread out" in a logarithmic scale. This is depicted in Figure~\ref{fig-st}. The second property ensures computational efficiency.

Indeed, the procedure ``Prune" maintains $S_t$ with these exact properties, and is detailed after we prove Theorem \ref{thm:flh2}.

We proceed to prove the main theorem. We start with an analogue of Lemma \ref{lem:round-reg}. 
\begin{proposition} \label{round-reg2}
The following holds for all $i \in S_t$, 
\begin{enumerate}
	\item 
	$f_t(\x_t) - f_t(\x^{i}_t) \leq \alpha^{-1} (\log \hat{p}^{i}_{t+1} - \log \hat{p}^{i}_{t} + \log \frac{t-1}{t} ) $
	\item $f_t(\x_t) - f_t(\x^{t}_t)\leq  \alpha^{-1} (\log \hat{p}^{t}_{t+1} + \log t) $
\end{enumerate}
\end{proposition}

\begin{proof} Using the $\alpha$-exp concavity of $f_t$ -
\begin{eqnarray*}
e^{-\alpha f_t(\x_t)} & = & e^{-\alpha f_t(\sum_{j \in S_t} p^{j}_t \x^{j}_t)}  \geq  \sum_{j \in S_t} p^{j}_t e^{-\alpha f_t(\x^{j}_t)}
\end{eqnarray*}
Taking the natural logarithm,
$$f_t(\x_t) \leq  -\alpha^{-1} \log \sum_{j \in S_t} p^{j}_t e^{-\alpha f_t(\x^{j}_t)} \nonumber $$
Hence,
\begin{eqnarray*}
  f_t(\x_t) - f_t(\x^{i}_t)  & \leq  \alpha^{-1}(\log e^{-\alpha f_t(\x^{i}_t)} - \log \sum_{j \in S_t} p^{j}_t e^{-\alpha f_t(\x^{j}_t)})  \\
& =  \alpha^{-1} \log \frac{e^{-\alpha f_t(\x^{i}_t)}}{\sum_{j \in S_t} p^{j}_t e^{-\alpha f_t(\x^{j}_t)}}  \\
& =  \alpha^{-1} \log \left( \frac{1}{p^{i}_t} \cdot
\frac{p^{i}_te^{-\alpha f_t(\x^{i}_t)}}{\sum_{j \in S_t} p^{j}_t e^{-\alpha f_t(\x^{j}_t)}}\right)  \\
& =  \alpha^{-1} \log \frac{\hat{p}^{i}_{t+1}}{p^{i}_t} 
\end{eqnarray*}
To complete the proof, we note the following two facts that are analogous to the ones used in Claim \ref{lem:round-reg}:
\begin{enumerate}
	\item For $1 \leq i < t$, $\log p^{i}_t \geq \log \hat{p}^{i}_t + \log \frac{t-1}{t}$
	\item $\log p^{t}_t \geq -\log t$
\end{enumerate}
Proving these facts is left as an exercise. 

%The second fact holds since $p^{t}_t \ge \hat{p}_t^t \cdot \frac{1}{1+\frac{1}{t-1}} =  \frac{1}{t-1} \frac{t-1}{t} = \frac{1}{t} $, and by taking the natural logarithm. The first fact follows since even without pruning we would have ${p}^{i}_t \geq \hat{p}^{i}_t \cdot \frac{1}{1 + \frac{1}{t-1} } = \hat{p}^i_t \cdot \frac{t-1}{t} $. Pruning only strengthens this inequality. 
\end{proof}

Using this we can prove the following Lemma.
\begin{lemma} \label{eff-int-reg} Consider some
time interval $I = [r,s]$. Suppose that
$E^r$  was in the working set $S_t$, for all $t \in I$.
Then the regret incurred in $I$ is at most $\frac{1}{\alpha} \log (s) + \regret_{T}(\mA)$.
\end{lemma}
\begin{proof}  
Consider the regret in  $I$
with respect to expert $E^r$,
\begin{eqnarray*}
& \sum_{t = r}^{s} (f_t(\x_t) - f_t(\x^{r}_t)) \\
& =  (f_r(\x_r) - f_r(\x^{r}_r)) + \sum_{t = r+1}^{s} (f_t(\x_t) - f_t(\x^{r}_t)) \nonumber \\
& \leq  \alpha^{-1} \bigl(\log \hat{p}^{r}_{r+1} + \log r + \sum_{t = r+1}^{s} (\log \hat{p}^{r}_{t+1} - \log \hat{p}^{r}_{t} + \log \frac{t}{t-1} )\bigr) & \mbox{ Claim \ref{round-reg2}} \\
& =  \alpha^{-1} (\log r + \log \hat{p}^{r}_{s+1}	+ \sum_{t = r+1}^{s} \log \frac{t}{t-1}) \nonumber \\
& =  \alpha^{-1} (\log (s) + \log \hat{p}^{r}_{s+1}	) \nonumber 
\end{eqnarray*}

Since $\hat{p}^{r}_{s+1} \leq 1$, $\log \hat{p}^{r}_{s+1} \leq 0$. This
implies that the regret w.r.t. expert $E^r$ is bounded by $ \alpha^{-1} \log (s) $. Since $E^r$ has regret bounded by $\regret_I(\mA) \leq \regret_T(\mA)$ over $I$, the conclusion follows.
\end{proof}

%Finally, we reach the main proof of this section.
Given
the properties of $S_t$, we can
show that in any interval the regret incurred
is small.

\begin{lemma} \label{aflh-reg} For any interval $I$ the
regret incurred by the FLH2 is at most \\ $(\frac{1}{\alpha} \log(s) + \regret_{T}(\mA)) (\log_2 |I|+1)$. \end{lemma}

\begin{proof} Let $|I| \in [2^q, 2^{q+1})$, and denote for simplicity $R_T = \frac{1}{\alpha} \log(s) + \regret_{T}(\mA)$.
We will prove by induction on $q$.

{\bf base case:} For $q=0$ the regret is bounded by 
$$f_r(\x_r) \leq \regret_T(\mA) \leq R_T $$

{\bf induction step:}
By the properties of the $S_t$'s, there is an expert $E^i$ in the pool such that
$i \in [r,(r+s)/2]$. This expert $E^i$ entered the pool
at time $i$ and stayed throughout $[i,s]$. By
Lemma~\ref{eff-int-reg}, the algorithm
incurs regret at most $R_T = \frac{1}{\alpha} \log (s) + \regret_{T}(\mA)$ in $[i,s]$.

The interval
$[r,i-1]$ has size at most $\frac{|I|}{2} \in [2^{q-1},2^q)$, and by induction
the algorithm has regret of at most $R_T \cdot q$ on this interval.
This gives a total of $R_T(q+1)$ regret on $I$.
\end{proof}

We can now prove Theorem \ref{thm:flh2}:
\begin{proof}[Theorem \ref{thm:flh2}]
The running time of FLH2 is bounded by $|S_t| \cdot V_T(\mA)$.
Since $|S_t| = O(\log t)$, we can bound the running time
by $O(V_T(\mA) \log T)$. This fact, together with Lemma~\ref{aflh-reg},
completes the proof.
\end{proof}

\subsection{The pruning method} \label{section:streamsoln}

We now explain the pruning procedure used to maintain the set $S_t \subseteq \{1,2,...,t\}$.

We specify the \emph{lifetime} of integer $i$ -
if $i = r2^k$, where $r$ is odd, then the lifetime
of $i$ is  $2^{k+2}+1$. Suppose
the lifetime of $i$ is $m$.
Then for any time $t \in [i,i+m]$, integer $i$ is \emph{alive} at $t$.
The set $S_t$ is simply
the set of all integers that are alive at time $t$.
Obviously, at time $t$, the only integer
added to $S_t$ is $t$ - this immediately
proves Property (3). We now
prove the other properties.

\begin{proof} (Property (1)) We need
to show that some integer in $[s,(s+t)/2]$ is alive
at time $t$. This is trivially true when $t-s < 2$,
since $t-1, t \in S_t$. Let $2^\ell$ be the
largest power of $2$ such that $2^\ell \leq (t-s)/2$.
There is some integer $x \in [s,(s+t)/2]$
such that $2^\ell | x$. The lifetime of $x$ is
larger than $2^\ell \times 2 + 1 > t-s$, so $x$
is alive at $t$.
\end{proof}

\begin{proof} (Property (2)) For each
$0 \leq k \leq \lfloor \log t \rfloor$, let us
count the number of integers of the form $r2^k$ ($r$ odd)
alive at $t$. The lifetimes of these integers
are $2^{k+2}+1$. The only integers alive lie
in the interval $[t-2^{k+2}-1,t]$. Since all
of these integers of this form are separated
by gaps of size at least $2^k$, there are at most a constant
number of such integers alive at $t$. In total,
the size of $S_t$ is $O(\log t)$.
\end{proof}

\newpage
\section{Bibliographic Remarks}

Dynamic regret bounds for online gradient descent were proposed by \citet{Zinkevich03}, and further studied in \citep{besbes2015non}.  It was shown in \citep{zhang2018dynamic} that  adaptive regret bounds imply dynamic regret bounds. 

The study of learning in changing environments can be traced to the seminal work of  \citet{HW} in the context of tracking for the problem of prediction from expert advice. Their technique was later extended to tracking of experts from a small pool \citep{BW}.

The problem of tracking a large set of experts efficiently was studied using the Fixed-Share technique in \citep{singer-portfolios,asinger-portfolios,Gyorgy05trackingthe}. 

The deviation from Fixed-Share to the FLH technique and the notion of adaptive regret were introduced in \citet{hazan2007adaptive}. These techniques were subject of later study and extensions \citep{adamskiy2016closer,zhang2019adaptive}.
\citet{daniely2015strongly} study adaptive regret for weakly convex loss functions and introduced the term ``strongly adaptive", which differentiates the weakly and strongly convex settings. They note that FLH is a strongly adaptive algorithm.

The use of an exponential look-back for prediction has roots in information theory \citep{WillemsK97,ShamirM06}. Efficient methods for streaming, that were used in this chapter to maintaining a small set of active experts, were studied in the steaming algorithms literature \citep{gopalan2007estimating}. 

Adaptive regret algorithms were motivated by applications involving changing environments, such as the portfolio selection problem. More recently they were applied for time series prediction \citep{anava2013online} and the control of dynamical systems \citep{gradu2020adaptive}.

\newpage
\begin{exercises}

%\section{Exercises}
%\begin{enumerate}[I]
\exer{
Consider an OCO setting that can be divided into $k$ intervals, such that in each a different comparator is optimal. Let $\mA$ be an algorithm that has an  \newregret guarantee of $\newregreteqn_T(\mA) = o(T)$. Prove that the regret of $\mA$ vs. the best $k$-shifting comparator is bounded by $k \times \newregreteqn_{T}$. 
}

\exer{
Prove that the OGD algorithm for convex functions, with step sizes $O(\frac{1}{\sqrt{t}})$, has an \newregret guarantee of $O(\sqrt{T})$, and that this is tight. Prove that the lazy version of OGD, from chapter \ref{chap:regularization}, behaves differently and has an adaptive regret bound of $\Omega(T)$. 
}
\exer{
Prove that the OGD algorithm with step size $O(\frac{1}{t})$ for strongly convex functions, has adaptive regret which is lower bounded by $\Omega(T)$. 
}

\exer{
Consider the problem of prediction from expert advice with $\alpha$-exp-concave loss functions, where the best expert switches $k$ times. That is, time can be divided into $k$ segments $I_1,...,I_k$, such that the best expert in each segment is different. 
\\
Show that the regret of the  Fixed Share algorithm vs. the best $k$-switching expert (a strategy that is allowed to change experts $k$ times) is bounded by 
$$ O\left( k \log \frac{NT}{k} \right) . $$
}

\exer{
Spell out a choice of $\delta$ parameter for the Fixed Share algorithm that does not require knowledge of the number of iterations $T$ in advance. Prove an analogue of Theorem \ref{thm:fixed-share} with your choice of $\delta$. 
}

\exer{
Complete the proof of Theorem \ref{thm:main-flh1}. 
}

\exer{
Prove the following facts used in the proof of Proposition \ref{round-reg2}:
%\begin{enumerate}
	\subexer{ For $1 \leq i < t$, $\log p^{i}_t \geq \log \hat{p}^{i}_t + \log \frac{t-1}{t}$}
	\subexer{ $\log p^{t}_t \geq -\log t$}
%\end{enumerate}
}

\exer{
Implement meta-algorithms FLH and FLH2 with the ONS algorithm, and apply the resulting method on the portfolio selection problem. Benchmark your results and compare to ONS and OGD. 
}
%\end{enumerate}
\end{exercises}

%!TEX root = OCObook.tex

%%%%%%%%%%%%%%%%%%%%%%%%%%%%%%%%%%%%%%%%%%%%%%%%%%%%%%%%%%%%
%%%%%%%%%%%%%%%%%%%%%%%%%%%%%%%%%%%%%%%%%%%%%%%%%%%%%%%%%%%%
%  Boosting
%%%%%%%%%%%%%%%%%%%%%%%%%%%%%%%%%%%%%%%%%%%%%%%%%%%%%%%%%%%%
%%%%%%%%%%%%%%%%%%%%%%%%%%%%%%%%%%%%%%%%%%%%%%%%%%%%%%%%%%%%
\chapter{Boosting and Regret} \label{chap:boosting}

In this chapter we consider a fundamental methodology of machine learning:  {\it boosting}. In the statistical learning setting, roughly speaking, boosting refers to the process of taking a set of rough ``rules of thumb'' and combining them into a more accurate predictor. 

Consider for example the problem of Optical Character Recognition (OCR) in its simplest form: given a set of bitmap images depicting hand-written postal-code digits, classify those that contain the digit ``1'' from those of ``0''.  

\begin{figure}[h!]
	\begin{center}
		\includegraphics[width=3.3in]{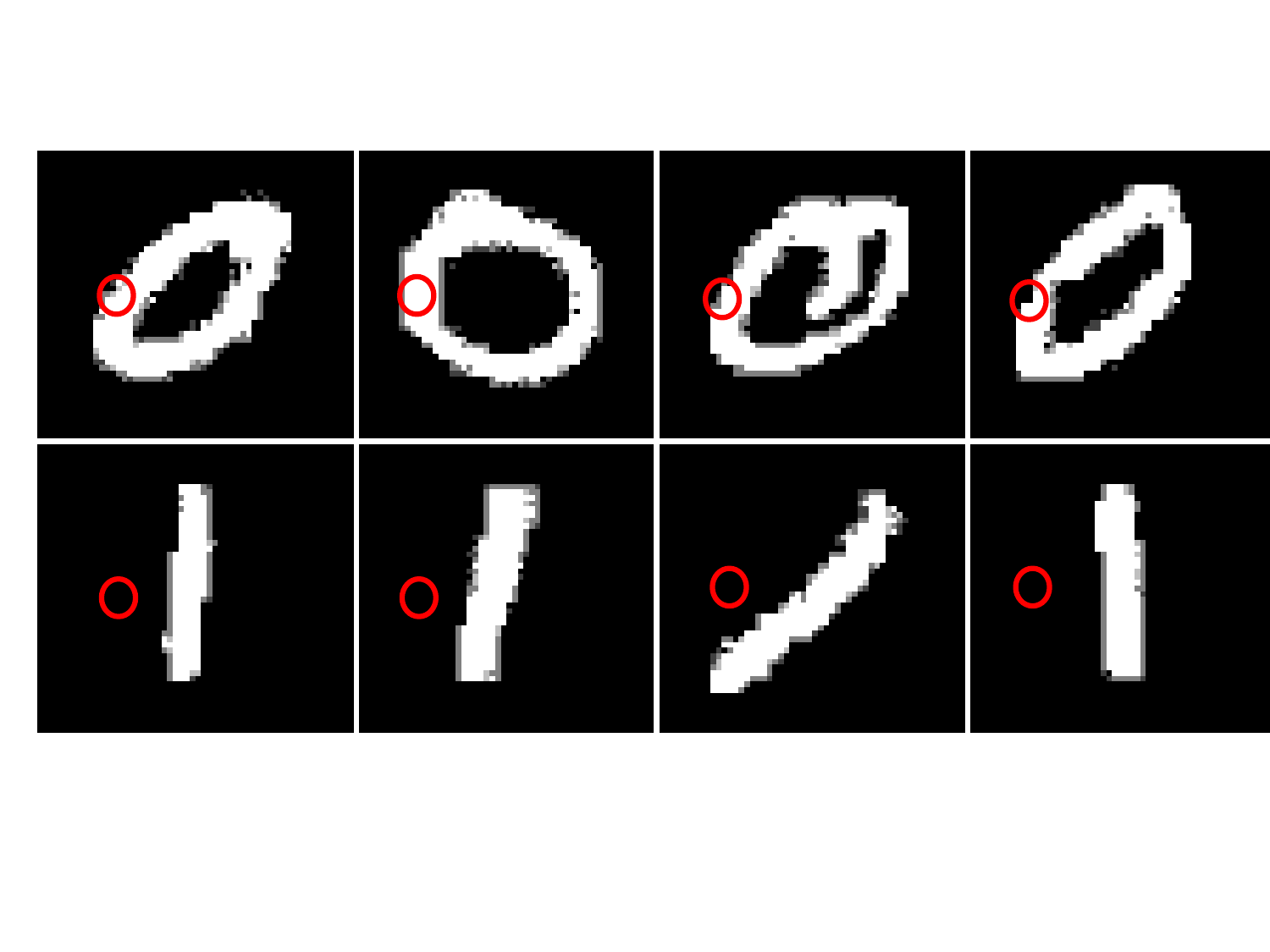}
	\end{center}
	\caption{Distinguishing zero versus one from a single pixel}
\end{figure}

Seemingly, discerning the two digits seems a difficult task taking into account the different styles of handwriting, inconsistent styles even for the same person, label errors in the training data, etc. However, an inaccurate rule of thumb is rather easy to produce: in the bottom-left area of the picture we'd expect many more dark bits for ``1''s than if the image depicts a ``0''. This is, of course, a rather inaccurate classifier. It does not consider the alignment of the digit, thickness of the handwriting, and numerous other factors. Nevertheless, as a rule of thumb - we'd expect better-than-random performance and some correlation with the ground truth. 

The inaccuracy of the crude single-bit predictor is compensated by its simplicity. It is not hard to implement  a classifier based upon this rule of thumb,  which is very efficient indeed. The natural and fundamental question which now arises is: can several such rules of thumb be combined into a single, accurate and efficient classifier? 

In the rest of this chapter we shall formalize this question in the statistical learning theory framework. We then proceed to use the technology developed in this manuscript, namely regret minimization algorithms for online convex optimization, to answer this question in the affirmative. Our development will be somewhat non-standard: we'll describe a black-box reduction from regret-minimization to boosting. This allows any of the  OCO methods previously discussed in this text to be used as the main component of  a boosting algorithm.  

%In the next chapter we move away from the classical boosting results for the statistical learning setting, and study the notion of online boosting. 

\section{The Problem of Boosting}

Throughout this chapter we use the notation and definitions of chapter \ref{chap:online2batch} on learning theory, and focus on statistical learnability rather than agnostic learnability. More formally, we assume the so called  ``realizability assumption'', which states that for a learning problem over hypothesis class $\H$ there exists  some  $ h^\star \in \mathcal{H}$ such that its generalization error is zero, or formally $\err(h^\star)=0.$

Using the notations of the previous chapter, we can define the following seemingly weaker notion than statistical learnability.  

\begin{definition}[Weak learnability] 
The concept class $\H : \X\mapsto \Y $  is said to be $\gamma$-weakly-learnable if the following holds. There exists an algorithm $\mA$ that accepts $S_m = \{(\bx,y)\}$ and returns an hypothesis in $\mA(S_m)  \in \H$ that satisfies: \\
 for any $\delta > 0$ there exists $m  = m(\delta)$ large enough such that for any distribution $\D$ over pairs $(\bx,y)$, for $y = h^\star(\x)$, and $m$ samples from this distribution, it holds that with probability  $1 - \delta$, 
$$ \err( \mA(S_m) ) \leq \frac{1}{2} - \gamma  $$
\end{definition}

This is an apparent weakening of the definition of statistical learnability that we have described in chapter \ref{chap:online2batch}:  the error is not required to approach zero. The standard case of statistical learning in the context of boosting is called ``strong learnability''.  An algorithm that achieves weak learning is referred to as a weak learner, and respectively we can refer to a strong learner as an algorithm that attains statistical learning, i.e., allows for generalization error arbitrarily close to zero, for a certain concept class. 

The central question of boosting can now be formalized:  are weak learning and  strong learning equivalent? In other words, is there an (efficient?) procedure that has access to a weak oracle for a concept class, and returns a strong learner for the class?

Miraculously, the answer is affirmative, and gives rise to one of the most effective paradigms in machine learning. 

\section{Boosting by Online Convex Optimization}

In this section we describe a {\it reduction} from OCO to boosting. The template  is similar to the one we have used in chapter \ref{chap:online2batch}: using one of the numerous algorithms for online convex optimization we have explored in this manuscript, as well as access to a weak learner, we create a procedure for strong learning.

\subsection{Simplification of the setting} 

Our derivation focuses on simplicity rather than generality. As such, we make the following assumptions:
\begin{enumerate}
\item
We restrict ourselves to the classical setting of binary classification. Boosting to real-valued losses is also possible, but outside our scope.
Thus, we assume the loss function to be the zero-one loss, that is:
$$ \ell(\hat{y}, y) = \mycases {0}{y = \hat{y}} {1}{0/w} $$
\item
We assume that the concept class is realizable, i.e., there exists an $h^\star \in \H$ such that $\err(h^\star) = 0$. There are results on boosting in the agnostic learning setting, these are surveyed in the bibliographic section. 

\item
We  denote the distribution over examples $\X \times \Y = \{(x,y)\} $, where $y = h^\star(\x)$, as a point in $\Delta_{\X}$. That is, a point $\bp \in \Delta_{\X}$ is a non-negative vector that integrates to one over all examples. %$ \int_{(x,y)  \in \X \times \Y} p_{(x,y)} d (x,y) = 1  $.
For simplicity, we think of $\X,\Y$ as a finite, and therefore 
$\bp \in \Delta_{m} $ belongs to the $m$ dimensional simplex, i.e., is a discrete distribution over $m$ elements. 

\item
We henceforth denote the weak learning algorithm by $\W$, and denote by $\W(\bp , \delta )$ a call to the weak learning algorithm over distribution $\bp$ that satisfies
$$ \Pr[  \err_{\bp} (\W(\bp,\delta))   \geq \frac{1}{2}- \gamma ]  \leq \delta . $$

%\item
%For a sample $S \sim \D$, let $\err_S(h)$ be the empirical error of an hypothesis $h \in \H$ on the sample, i.e., 
%$$  \err_S(h) \equaldef \E_{(\bx,y)\sim S} [ h(\bx) \neq y ]. $$
%We assume that any hypothesis $h \in \H$ that attains zero error on the sample $S_m$ for $m = m(\epsilon)$ is guaranteed at most $\epsilon$ generalization error. 

%This assumption is well  justified in statistical learning theory, and its central theorems address exactly this scenario: taking a large enough sample and finding a hypothesis which is consistent with the sample (zero error on it), implies $\epsilon$ generalization error. 
%However, the conditions in which the above holds are beyond our scope. 
\end{enumerate}

With these assumptions and definitions we are ready to prove the main result: a reduction from weak learning to strong learning using an online convex optimization algorithm with a sublinear regret bound. Essentially, our task would be to find a hypothesis which attains zero error on a given sample.

\subsection{Algorithm and analysis}
 
Pseudocode for the boosting algorithm is given in Algorithm \ref{alg:boost1}. This reduction  accepts as input a $\gamma$-weak learner and treats it as a black box, returning a function which we'll prove is a strong learner. 

The reduction also accepts as input an online convex optimization algorithm denoted $\mA^{OCO}$. The underlying decision set for the OCO algorithm is the $m$-dimensional simplex, where $m$ is the sample size. Thus, its decisions are distributions over examples. The cost functions are linear, and assign a value of zero or one, depending on whether the current hypothesis errs on a particular example. Hence, the cost at a certain iteration is the expected error of the current hypothesis (chosen by the weak learner) over the distribution chosen by the low-regret algorithm. 

\begin{algorithm}[h]
\caption{Reduction from  Boosting to OCO}
\label{alg:boost1}
\begin{algorithmic}
\State \textbf{Input}: $\H,\delta$,  OCO algorithm $\mA^{OCO}$, $\gamma$-weak learning algorithm $\W$, sample $S_m \sim \D$. 
%\State Draw $m$ labeled example $\{ (\bx_i,y_i) \sim \D \ , \ i \in [m] \} $.\\ 
%	 Take sample $S=\{(x_i,y_i) \}_{i=1}^m$ of size $m =\mathcal{O}( \frac{4d}{\gamma^2\epsilon^2} \log(4d/\epsilon\delta))$ from $\mathcal{D}$, $d=\text{VC}_{dim}(\mathcal{H})$\\
\State Set $T$ such that $\frac{1}{T} {\regret_T(A^{OCO})} \leq \frac{\gamma}{2} $ %=\frac{4}{\gamma^2} \log(m)$ 
%Set $\delta' = \delta/T$\\
\State Set distribution $\bp_1 = \frac{1}{m} \bone \in \Delta_m $ to be the uniform distribution.
\For{$t=1,2 \ldots T$ }
\State Find hypothesis $h_t \leftarrow \W(\bp_t ,\frac{\delta}{2T} )$   
\State Define the loss function $f_t( \bp) = \br_t^\top \bp $, where the vector  $\br_t \in \reals^m$ is defined as
$$ \br_t( i) = \mycases {1} { h_t(\x_i)=y_i } {0}{o/w} $$
\State 	Update $\bp_{t+1} \leftarrow \mA^{OCO} (f_1,...,f_t) $ %  = \frac{P_t(i)exp(-\eta r_t(i))}{\sum_{j=1}^m P_t(j)exp(-\eta r_t(j))}$
\EndFor
\State \Return  $\bar{h}(\x) =\text{sign}(\sum_{t=1}^T h_t(\x))$ %  $\bar{h}(\x) = \mycases {1} {  \frac{1}{T} \sum_{t=1}^T h_t(\x)  \geq \frac{1}{2} } {0 } {0/w}  $.
\end{algorithmic}
\end{algorithm}

It is important to note that the final hypothesis $\bar{h}$ which the algorithm outputs does not necessarily belong to $\H$ - the initial hypothesis class we started off with.

\begin{theorem} \label{thm:boosting-basic}
Algorithm \ref{alg:boost1} returns a hypothesis  $\bar{h}$ such that with probability at least $1-\delta$,
$$ \err_S(\bar{h}) =0 . $$
\end{theorem}
\begin{proof}
Given $h\in \mathcal{H}$, we denote its empirical error on the sample $S$, weighted by the distribution $\bp \in \delta_m$, by:
 \begin{align} \nonumber
\err_{S,\bp}(h) = \sum_{i=1}^m \bp(i) \cdot \bone_{ h(\x_i) \neq y_i } .
 \end{align}
 Notice that by definition of $\br_t$ we have  $ \br_t^\top  \bp_t  = 1 - \err_{S , \bp_t} (h_t) $. 
Since $h_t$ is the output of a $\gamma$-weak-learner on the distribution $\bp_t$, we have for all $t \in [T]$,
 \begin{align*} \nonumber
\Pr[ \br_t^\top  \bp_t  \leq \frac{1}{2} + \gamma ]  & = \Pr[ 1 - \err_{S , \bp_t} (h_t) \leq \frac{1}{2}+\gamma ] \\
& = \Pr[  \err_{S , \bp_t} (h_t) \geq \frac{1}{2}- \gamma ] \\
& \leq \frac{\delta}{2T} .
 \end{align*}
This applies for each $t$ separately, and by the union bound we have 
$$ \Pr[ \frac{1}{T} \sum_{t=1}^T \br_t^\top  \bp_t  \geq \frac{1}{2} + \gamma ]  \geq 1- \delta $$ 

Denote by  $S_\phi \subseteq S$ be the set of all missclassified examples by $\bar{h}$. Let $\bp^*$ the uniform distribution over $S_\phi$. 
 \begin{align*} \nonumber
  \sum_{t=1}^T \br_t^\top \bp^* & = \sum_{t=1}^T \frac{1}{|S_\phi|}\sum_{(\x,y) \in S_\phi} \bone_{h_t(\x) = y}  \\
  & =       \frac{1}{|S_\phi|} \sum_{(\x,y) \in S_\phi} \sum_{t=1}^T \bone_{h_t(\x_j) = y_j} \\
  &  \leq  \frac{1}{|S_\phi|} \sum_{(\x,y) \in S_\phi}  \frac{T}{2} & \mbox{ $\bar{h}(\x_j) \neq y_j $} \\
  & =\frac{T}{2} .
  \end{align*}  
Combining the previous two observations, we have with probability at least $1-\delta$ that
\begin{eqnarray*}
\frac{1}{2} +  \gamma & \leq \frac{1}{T}  \sum_{t=1}^T \br_t^\top  \bp_t \\
& \leq  \frac{1}{T}  \sum_{t=1}^T \br_t^\top \bp^* + \frac{1}{T}  \regret_T(\mA^{OCO}) & \mbox { low regret of $\mA^{OCO}$} \\
& \leq \frac{1}{2} + \frac{1}{T} \regret_T(\mA^{OCO} ) \\
& \leq \frac{1}{2} + \frac{\gamma}{2}   .
\end{eqnarray*}
This is a contradiction. We conclude that a distribution $\bp^*$ cannot exist, and thus all examples in $S$ are classified correctly. 
 
\end{proof}

\subsection{AdaBoost}

A special case of the template reduction we have described is obtained when the OCO algorithm is taken to be the Multiplicative Updates method we have come to know in the manuscript. 

Corollary \ref{cor:eg} gives a bound of $O(\sqrt{T\log m})$ on the regret of the EG algorithm in our context. This bounds $T$ in Algorithm \ref{alg:boost1} by $O(\frac{1}{\gamma^2} \log m)$.

Closely related is the AdaBoost algorithm, which is one of the most useful and successful algorithms in Machine Learning at large (see bibliography). 
Unlike the Boosting algorithm that we have analyzed, AdaBoost doesn't have to know  in advance the parameter  $\gamma$ of the weak learners. 
Pseudo code for the AdaBoost algorithm is given in \ref{alg:adaboost}.
\begin{algorithm}[h]
\caption{AdaBoost}
\label{alg:adaboost}
\begin{algorithmic}
\State \textbf{Input}: $\H,\delta$,  $\gamma$-weak-learner  $\W$, sample $S_m \sim\D$. 
\State  Set $\bp_1 \in \Delta_m$ be the uniform distribution over $S_m$. 
\For{$t=1,2 \ldots T$ }
\State Find hypothesis $h_t \leftarrow \W(\bp_t,\frac{\delta}{T})$   
\State Calculate $\epsilon_t = \err_{S,\bp_t}(h_t)$, $\alpha_t = \frac{1}{2}\log(\frac{1-\epsilon_t}{\epsilon_t})$
\State 	Update, 
$$ \bp_{t+1}(i) = \frac{\bp_t(i) e^{ -\alpha_t y_i h_t(i)} } {\sum_{j=1}^m \bp_t(j)e^{-\alpha_t y_j h_t(j)}} $$
\EndFor
\State \Return $\bar{h}(\x) =\text{sign}(\sum_{t=1}^T \alpha_t h_t(\x))$
\end{algorithmic}
\end{algorithm}

\subsection{Completing the picture}

In our discussion so far we have focused only on the empirical error over a sample. To show generalization and complete the Boosting theorem, one must show that zero empirical error on a large enough sample implies $\epsilon$ generalization error on the underlying distribution. 

Notice that the hypothesis returned by the Boosting algorithms does not belong to the original concept class. This presents a challenge for certain methods of proving generalization error bounds that are based on measure concentration over a fixed hypothesis class. 

Both issues are resolved using the implication that compression implies generalization, as given in Theorem \ref{thm:compression2generalization}. We sketch the argument below, and the precise derivation is left as an exercise. 

Roughly speaking, boosting algorithm \ref{alg:boost1} runs on $m$ examples for $T = O(\frac{\log m}{\gamma^2})$ rounds,  returns a final hypothesis $\bar{h}$ that is the majority vote of $T$ hypothesis, and classifies correctly all $m$ examples of the training set. 

Suppose that the weak learning algorithm has sample complexity of size $k(\gamma,\delta)$:  given $k = k(\gamma,\delta)$ examples, it returns a hypothesis with generalization error at most $\frac{1}{2} - \gamma$ with probability at least $1-\delta$. Further, suppose the original training set of $m$ examples was sampled from distribution $\D$. 

Since $\bar{h}$ classifies correctly the entire training set,  it follows that the distribution $\D$ has a compression scheme of size 
$$Tk  = O\left( \frac{k(\gamma,\frac{\delta}{T}) \log m}{\gamma^2} \right) . $$ 

Therefore, using Theorem \ref{thm:compression2generalization}, we have that, %for $S'$ being the set of $T k$ examples that are given to the weak learning algorithm throughout the boosting procedure, 
$$ \err_{\D}(\bar{h}) \leq O\left( \frac{k \log^2 \frac{m}{\delta}} {\gamma^2 m} \right) .$$

Now one can obtain an arbitrary small generalization error by choosing $m$ as a function of $k,\delta,\gamma$. 
Notice that this argument makes an assumption only about the sample complexity of the weak learning algorithm, rather than the hypothesis class $\H$.

\newpage
\section{Bibliographic Remarks}

The theoretical question of Boosting and posed and addressed in the  work of \citet{Schapire90,freund1995boosting}. The AdaBoost algorithm was proposed in the seminal paper of  \citet{FreundSch1997}. The latter paper also contains the essential ingredients for the reduction from general low-regret algorithms to boosting. 

Boosting has had significant impact on theoretical and practical data analysis as described by the statistician Leo Breiman  \citep{Breiman01}. 
For a much more comprehensive survey of Boosting theory and applications see the recent book \citep{schapire2012boosting}. 

The theory for agnostic boosting is more recent, and several different definitions and settings exist, see ~\citep{kalai2008agnostic, KalaiS05, kanade2009potential, feldman2009distribution,bendavid2001agnostic}, the most general of which is perhaps by \citet{kanade2009potential}. %, which we revisit in chapter \ref{chap:ocoboost}. 

A unified framework for realizable and agnostic boosting, for both the statistical and online settings, is given in \citep{brukhim2020online}.

The theory of boosting has been extended to real valued learning  via the theory of gradient boosting \citep{friedman2002stochastic}. More recently it was extended to online learning \citep{leistner2009robustness, chen2012online, chen2014boosting, beygelzimer2015optimal, beygelzimer2015online, agarwal2019boosting, jung2017online, jung2018online,brukhim2020online2}.

\newpage
\begin{exercises}
\exer{
Describe a boosting algorithm based on the \ogd algorithm. Give a bound on its running time. 
}

\exer{
Download the MNIST dataset from the web. Implement weak learners to differentiate a pair of digits of your choice according to a single bit. 
Implement AdaBoost and apply it to your weak learners. Summarize your results and conclusions. 
}

\exer{$^*$
Consider the problem of {\it agnostic boosting}, in which the existence of a zero-error hypothesis is not assumed. 

\subexer{
Write down an alternative definition of a weak learning algorithm for the agnostic setting. 
}
\subexer{
Write down a reasonable goal for a boosting algorithm. 
}
\subexer{
Write down an analogue of Theorem \ref{thm:boosting-basic} for agnostic boosting (without proof). 
}
}

\exer{
Compute the number of samples required to achieve a generalization error of $\eps$, using boosting algorithm \ref{alg:boost1} and a  weak learning algorithm with sample complexity bound $k(\gamma,\delta)$. 
}
\end{exercises}

\chapter{Online Boosting} \label{chap:ocoboost}

This text considers online optimization and learning, and it is a natural question to ask whether the technique of boosting has an analogue in the online world? What is a ``weak learner" in online convex optimization, and how can one strengthen it? This is the subject of this chapter, and we shall see that boosting can be extremely powerful and useful in the setting of online convex optimization.

\section{Motivation: Learning from a Huge Set of Experts}

Recall the classical problem of prediction from expert advice from the first chapter of this text. A learner iteratively makes decisions and receives loss according to an arbitrarily chosen loss function. For its decision making, the learner is assisted by a pool of experts. Classical algorithms such as the Hedge algorithm \ref{alg:Hedge}, guarantees  a regret bound of  $O(\sqrt{T \log N} )$, where $N$ is the number of experts, and this is known to be tight. 

However, in many problems of interest, the class of  experts is too large to efficiently manipulate. This is particularly evident in contextual learning, as formally defined below, where the experts are {\it policies} -- functions mapping contexts to action. In such instances, even if a regret bound of $O(\sqrt{T \log N})$ is meaningful, the algorithms achieving this bound are computationally inefficient; their running time is linear in $N$. 
This linear dependence is many times unacceptable: the effective number of policies  mapping contexts to actions is exponential in the number of contexts.

The boosting approach to address this computational intractability is motivated by the observation that it is often possible to design simple {\it rules-of-thumb} that perform slightly better than random guesses. Analogously to the weak learning oracles from chapter \ref{chap:boosting}, We propose that the learner has access to an ``online weak learner" - a computationally cheap mechanism capable of guaranteeing multiplicatively approximate regret against a base hypotheses class.  

In the rest of this chapter we describe efficient algorithms that when provided weak learners, compete with the convex hull of the base hypotheses class with near-optimal regret.

\subsection{Example: boosting online binary classification}

As a more precise example to the motivation we just surveyed, we formalize online boosting for binary prediction from expert advice. At iteration $t$, a set of experts denoted $h \in \H$,  observe a context $\ba_t$, and predict a binary outcome $h(\ba_t) \in \{-1 ,1 \}$. The loss of each expert is taken to be the binary loss, $- h(\ba_t) \cdot y_t$ for a true label $y_t \in \{-1,1\}$. 
The Hedge algorithm from the first chapter applies to this problem, and guarantees a regret of $O(\sqrt{T \log |\H|})$ for a finite $\H$. However, the case in which $\H$ is extremely large, maintaining the weights is prohibitive computationally. 

A weak online learner $\W$ in this setting is an algorithm which is guaranteed to attain at most a factor $\gamma$ loss from the best expert in class, for some $\gamma \in [0,1]$, up to an additive sublinear regret term. Formally, for any sequence of contexts and labels $\{\ba_t,y_t\}$,
\begin{equation*}
 \sum_{t=1}^T y_t \cdot \W(\ba_t)  \le \gamma \cdot \underset{h \in \H}{\min} \sum_{t=1}^T y_t \cdot h(\ba_t)  + \regret_{T}(\W ).
\end{equation*}

%Notice that since the loss can be negative, for this binary classification problem we inverted the signs and wrote the guarantee in terms of gains. 

The online boosting question can now be phrased as follows: given access to a weak online learning algorithm $\W$, can we design an efficient online algorithm $\mA$ that guarantees vanishing regret over $\H$? More formally, let
\begin{equation*}
\regret_T(\mA) =  \sum_{t=1}^T y_t \cdot \W(\ba_t)  -  \underset{h \in \H}{\min} \sum_{t=1}^T y_t \cdot h(\ba_t)  . 
\end{equation*}

Can we design an algorithm $\mA$ that has $\frac{\regret_T(\mA)}{T} \mapsto 0$, without explicit access to $\H$? 
As we will see, the answer to this question is affirmative in a strong sense: boosting does have an online analogue which is a powerful technique in online learning. In the next section we describe a more powerful notion of boosting that applies to the full generality of online convex optimization. This in turn implies an affirmative answer to this question for online binary classification.

\subsection{Example: personalized article placement} 

In the problem of matching articles to visitors of a web-page on the Internet, a number of articles are available to be placed in a given web-page for a particular visitor. The goal of the decision maker, in this case the article placer, is to find the most relevant article that will maximize the probability of a visitor click. 

It is usually the case that context is available, in the form of user profile, preferences surfing history and so forth. This context is invaluable in terms of placing the most relevant article. Thus, the decision of the article-placer is to choose from a {\it policy}: a mapping from context to article. 

The space of policies is significantly larger than the space of articles and space of contexts: its size is the power of articles to the cardinality of contexts. This motivates the use of online learning algorithms whose computational complexity is independent of the number of experts.

The natural formulation of this problem is not binary prediction, but rather multi-class prediction. Formulating this problem in the language of online convex optimization is left as an exercise.

\section{The Contextual Learning Model}

Boosting in the context of online convex optimization is most useful for the contextual learning problem which we now describe.

Let us consider the familiar OCO setting over a general convex decision set $\K \subseteq \reals^d$, and adversarially chosen convex loss functions $f_1,...,f_t : \K \mapsto \reals$.  
Boosting is particularly important in settings that we have a very large number of possible experts that makes running one of the algorithms we have considered thus far infeasible.  Concretely, suppose we have access to a hypothesis class $\H \subseteq \{\ba\} \mapsto \K$, that given a sequence of contexts $\ba_1, ...,\ba_t$, produces a new point $h( \ba_{t+1} ) \in \K$.

We have studied numerous methods capable of minimizing regret for this setting  in this text, all assumed that we have access to the set $\H$, and depend on its diameter in some way. 

To avoid this dependence, we consider an alternative access model to $\H$. A weak learner for the OCO setting is defined as follows. 

\begin{definition}%[$\gamma$-weak OLO and OCO algrithms]
\label{online_agnostic_wl}
    An online learning algorithm $\W$ is a $\gamma$-\textbf{weak OCO learner (WOCL)} for~$\H$ and $\gamma \in (0,1)$,
    if for any sequence of contexts $ \{ \ba_t \} $ and {\bf linear} loss functions $f_1,...,f_T$, for which $\max_{\x\in \K} f_t(\x) - \min_{\y\in \K} f_t(\y) \leq 1$ \endnote{The results henceforth hold with a different constant if we replace one by a different scaling.}, we have 
\begin{equation}
    \label{eq:wl-stepwise}
 \sum_{t=1}^T f_t(\W(\ba_t))  \le \gamma \cdot \underset{h \in \H}{\min} \sum_{t=1}^T f_t(h(\ba_t))  + (1-\gamma) \sum_{t=1}^T f_t(\bar{\x}) + \regret_{T}(\W ),
\end{equation}
where $\bar{\x} = \int_{\x \in \K } \x $ is the center of mass of $\K$.
\end{definition}

This definition differs in two aspects from the types of regret minimization guarantees we have seen thus far. For one, the algorithm competes with a $\gamma$-multiple of the best comparator in hindsight, and is ``weak" in this precise manner. 

Secondly, a multiplicative guarantee is not invariant for a constant shift. This is the reason for the existence of an  additional component, $\sum_t f_t(\bar{\x})$, in the regret bound. This can be thought of as the cost of a random, or naive predictor. A weak learner must, at the very least, perform better than this naive and non-anticipating predictor! 

It is convenient to henceforth assume that the loss functions are shifted such that $f_t(\bar{\x}) = 0$. Under this assumption, we can rephrase $\gamma$-WOCL as 
\begin{equation} \label{eqn:simpleWOLL}
 \sum_{t=1}^T f_t(\W(\ba_t))  \le \gamma \cdot \underset{h \in \H}{\min} \sum_{t=1}^T f_t(h(\ba_t))  + \regret_{T}(\W).
 \end{equation}
%where we denote by $G$ the upper bound on the gradients of $f_t$. 

\section{The Extension Operator}

The main difficulty is coping with the approximate guarantee that the WOCL provides. Therefore the algorithm we describe henceforth scales the predictions returned by the weak learner by a factor of $\frac{1}{\gamma}$. This means that the scaled decisions do not belong to the original decision set anymore, and need to be projected back. 

Here lies the main challenge. First, we assume that the loss functions $f \in \F$ are defined over all of $\reals^d$ to enable valid decisions outside of $\K$. 
%We then define the projection operator to return points inside of $\K$ that always have lesser value w.r.t. any function $f \in \F$, 
%\begin{definition}[$\F$-projection]
%A projection operator $\prod_\K^\F: \reals^d \mapsto \K$ w.r.t. family of functions $\F$ is defined to be a point insider $\K$ that is smaller than the given point for all $f \in \F$, i.e.
%$$ \forall \x \in \reals^d , f \in \F \ , \ f\left(\prod_\K^\F(\x) \right) \leq f(\x) .$$ 
%\end{definition}
Next, we need to be able to project onto $\K$ without increasing the cost. It can be seen that some natural families of functions, i.e., linear functions,  do not admit any such projection.  To remedy this situation, we define the extension operator of a function over a convex domain $\K$ as follows. 

First, denote the Euclidean distance function to a set $\K$ as (see also section \ref{sec:approach-dist}),
$$ \dist(\cdot, \K) \ , \  \dist(\x,\K) = \min_{\y \in \K} \|\y-\x\| .  $$

\begin{definition}[$(\K,\kappa,\delta)$-extension] \label{defn:extension}
The extension operator over $\K \subseteq \reals^d$ is defined as:
$$ X_{\K,\kappa,\delta}[f] : \reals^d \mapsto \reals \ \ , \ \ X[f ]  = S_\delta[  f  + \kappa \cdot \dist(\cdot,\K)  ] , $$ 
where the smoothing operator $S_\delta$ was defined as per Lemma \ref{lem:SmoothingLemma}. 
\end{definition}

The important take-away from these operators is the following lemma, whose importance is crucial in the OCO boosting algorithm  \ref{alg:ocoboost}, as it projects infeasible points that are obtained from the weak learners to the feasible domain.

\begin{lemma} \label{lem:extensionX}
The $(\K,\kappa,\delta)$-extension of a function $\fhat = X[f]$ satisfies the following:
\begin{enumerate}
    \item For every point $\x \in \K$, we have $ \| \fhat(\x) - f(\x) \|_2 \leq \delta G$.
    
    \item The projection of a point, whose gradient is bounded by $G$,  onto $\K$ improves the $(\K,\kappa,\delta)$-extension function, for $\kappa = G$, value up to a small term,
    $$ \fhat \left( \prod_\K (\x)\right) \leq \fhat(\x) + \delta G . $$
\end{enumerate}
\end{lemma}
\begin{proof}

\begin{enumerate}
    \item
    Since $\dist(\x,\K) = 0$ for all $x \in \K$, this follows immediately from Lemma \ref{lem:SmoothingLemma}. 

    \item 
    Denote $\x_\pi =  \prod_\K (\x)  $ for brevity. Then 
    \begin{eqnarray*}
    & \fhat(\x_\pi) - \fhat(\x) \\
    & \leq f(\x_\pi) - f(\x) - \kappa  \dist(\x,\K)  + \delta G & \mbox{part 1} \\
    & \leq f(\x_\pi) - f(\x) - \kappa \| \x - \x_\pi\| + \delta G \\
    & = \nabla f(\x) (\x - \x_\pi) - \kappa \| \x - \x_\pi\| + \delta G  \\
    & \leq \| \nabla f(\x) \| \| \x - \x_\pi\| - \kappa \| \x - \x_\pi\| + \delta G & \mbox{Cauchy-Schwartz} \\
    & \leq G  \| \x - \x_\pi\| - \kappa \| \x - \x_\pi\| + \delta G \\
    & \leq \delta  G & \mbox{choice of $\kappa$} .
    \end{eqnarray*}

\end{enumerate}

\end{proof}

\section{The Online Boosting Method}

The online boosting algorithm we describe in this section is closely related to the online Frank-Wolfe algorithm from chapter \ref{chap:FW}. Not only does it deliver in boosting WOCL to strong learning, but it gives an even stronger guarantee: low regret over the convex hull of the hypothesis class.

Algorithm \ref{alg:ocoboost} efficiently converts a weak online learning algorithm into an OCO algorithm with vanishing regret in a black-box manner. The idea is to apply the weak learning algorithm on linear functions that are gradients of the loss. The algorithm then recursively applies another weak learner on the gradients of the residual loss, and so forth.

\begin{algorithm}[h]
\caption{Boosting for Online Convex Optimization}
\label{alg:ocoboost}
\begin{algorithmic}[1]
\State Input:  $N$ copies of the $\gamma$-WOCL  $\W^1, \W^2, \ldots, \W^N$, parameters $\eta_1,...,\eta_T$,$\delta,\kappa=G$.
\For{$t=1$ {\bfseries to} $T$}
    \State Receive context $\ba_t$, choose $\x_t^0 = \bzero$ arbitrarily. 
    \For{$i=1$ {\bfseries to} $N$}
    	\State Define $\x_t^i = (1 - \eta_i) \x_t^{i-1} + \eta_i \frac{1}{\gamma} \W^i(\ba_t)$.
    \EndFor
    \State Predict $\x_t = \prod_{\K}[ \x_t^N ] $, suffer loss $f_t(\x_t)$.
    \State Obtain loss function $f_t$,  create $\fhat_t = X_{\K,\kappa,\delta}[f_t]$.
    \For{$i=1$ {\bfseries to} $N$}
        \State Define and pass to $\W^i$ the linear loss function $f_t^i$,
        $$f_t^i(\x) = \nabla \fhat_t(\x_t^{i-1}) \cdot  \x  . $$
    \EndFor
\EndFor
\end{algorithmic}
\end{algorithm}

However, the Frank-Wolfe method is not applied directly to the loss functions, but rather to a proxy loss which defined using the extension operation in \ref{defn:extension}. 
Importantly, algorithm \ref{alg:ocoboost} has a running time that is independent of $|\H|$. 

Notice that if $\gamma=1$, the algorithm stills gives a significant advantage as compared to the weak learner: the regret guarantee is vs. the convex hull of $\H$, as compared to the best single hypothesis. 

\begin{theorem}[Main] \label{thm:oco-boost-convex}
The predictions $\x_t$ generated by Algorithm~\ref{alg:ocoboost} with $\delta = \sqrt{ \frac{ D^2}{\gamma N}  }, \eta_i = \min \{\frac{2}{i}, 1\}$ satisfy
\begin{align*}
\sum_{t=1}^T f_t(\x_t)\ - \min_{h^\star \in \ch(\H)} \sum_{t=1}^T f_t( h^\star(\ba_t) ) \leq   \frac{ 5 d G  D T}{\gamma\sqrt{  N}}  +  \frac{2GD}{\gamma } \regret_T(\W)  .
\end{align*}
\end{theorem}

\paragraph{Remark 1:} It is possible to obtain tighter bounds by a factor of the dimension, and other constant terms, using a more sophisticated smoothing operator. References for these tighter results are given in the bibliographic section at the end of this chapter.

\paragraph{Remark 2:} The regret bound of Theorem \ref{thm:oco-boost-convex} is nearly as good as we could hope for. The first term approaches zero as the number of weak learners $N$ grows. The second term is sublinear as the regret of the weak learner. It is scaled by a factor of $\frac{1}{\gamma}$, which we can expect due to the approximate guarantee of the weak learner. 

\medskip

Before proving the theorem, let us define some notations we use. The algorithm defines the extension of the loss functions as 
$$ \fhat_t = X[f_t] = S_\delta [ f_t  +  G\cdot \dist(\x,\K) ] . $$
We apply the setting of $\kappa=G$, as required by Lemma \ref{lem:extensionX}, and by Lemma \ref{lem:SmoothingLemma}, $\fhat_t$ is $\frac{dG }{\delta}$-smooth.
Also, denote by $\ch(\H) = \{ \sum_{h \in \H} \bp_h  h | \bp \in \Delta_\H \}$  the convex hull of the set $\H$, and let
$$ h^\star = \argmin_{h^\star \in \ch(\H)}\sum_{t=1}^T f_t(h^\star(\ba_t)) $$ 
to be the best hypothesis in the convex hull of $\H$ in hindsight, i.e., the best convex combination of hypothesis from $\H$.  Notice that since the loss functions are generally convex and non-linear, this convex combination is not necessarily a singleton. 
We define $\x_t^\star = h^\star(\ba_t) $ as the decisions of this hypothesis.

The main crux of the proof is given by the following lemma.
\begin{lemma} \label{lem:main-analysis}
For smoothed loss functions $\{ \fhat_t \}$ that are ${\beta}$-smooth and $\hat{G}$ Lipschitz, it holds that 
\begin{eqnarray*}
\sum_{t=1}^T \fhat_t(\x_t^N)\ -  \sum_{t=1}^T \fhat_t( \x^\star_t )  \leq  \frac{2 {\beta} D^2 T}{\gamma^2 N} +  \frac{\hat{G}D}{\gamma} \regret_T(\W) .
\end{eqnarray*}
\end{lemma}
\begin{proof}
Define for all $i = 0, 1, 2, \ldots, N$, 
$$\Delta_i = \sum_{t=1}^T \left(\fhat_t(\x_t^i) - \fhat_t(\x^\star_t)\right) . $$ 
Recall that $\fhat_t$ is ${\beta}$ smooth by our assumption. Therefore:
	\begin{align*}
	  & \Delta_i  =  \sum_{t=1}^T \left[ \fhat_t(\x_t^{i-1} + \eta_i ( \frac{1}{\gamma} \W^i(\ba_t) - \x_t^{i-1})) - \fhat_t(\x^\star_t) \right]\\
	\leq & \sum_{t=1}^T \Bigl[ \fhat_t(\x_t^{i-1}) - \fhat_t(\x^\star_t)  + \eta_i \nabla \fhat_t(\x_t^{i-1}) \cdot ( \frac{1}{\gamma} \W^i(\ba_t) - \x_t^{i-1}) \\
	& + \frac{\eta_i^2{\beta}}{2} \| \frac{1}{\gamma} \W^i(\ba_t) - \x_t^{i-1} \|^2  \Bigr] .
	\end{align*}
By using the definition and linearity of $f_t^i$, we have
\begin{align*}
    \Delta_i \leq&  \sum_{t=1}^T \left[\fhat_t(\x_t^{i-1}) - \fhat_t(\x^\star_t)  +\eta_i ( f_t^i( \frac{1}{\gamma} \W^i(\ba_t))  - f_t^i(\x_t^{i-1}))  + \frac{\eta_i^2{\beta} D^2}{2 \gamma^2} \right]  \\
	 =& \Delta_{i-1}  + \sum_{t=1}^T \eta_i ( \frac{1}{\gamma} f_t^i(  \W^i(\ba_t))  - f_t^i(\x_t^{i-1})) + \sum_{t=1}^T\frac{\eta_i^2{\beta} D^2}{2 \gamma^2} .
\end{align*}
Now, note the following equivalent restatement of the WOCL guarantee, which again utilizes linearity of $f_t^i$ to conclude: linear loss on a convex combination of a set is equal to the same convex combination of the linear loss applied to individual elements.
\begin{align*}
    \frac{1}{\gamma} \sum_{t=1}^T f_t^i (\W^i(\ba_t)) \leq &\min_{h^\star \in \H} \sum_{t=1}^T f_t^i (h^\star (\ba_t)) + \frac{\hat{G}D \regret_T(\W)}{\gamma}\\
    =& \min_{h^\star \in \ch(\H)} \sum_{t=1}^T f_t^i (h^\star(\ba_t)) + \frac{\hat{G}D \regret_T(\W)}{\gamma} .
\end{align*}
Using the above and that $h^\star\in \ch(\H)$, we have
\begin{align*}
     & \Delta_i  \\
     & \leq  \Delta_{i-1}  + \sum_{t=1}^T [ \eta_i \nabla \fhat_t(\x_t^{i-1}) \cdot (\x_t^\star - \x_t^{i-1})   + \frac{\eta_i^2{\beta} D^2}{2 \gamma^2 }     ] + \eta_i \frac{\hat{G}D}{\gamma} \regret_T(\W ) \\
    & \leq  \Delta_{i-1}  (1 -  \eta_i )  + \frac{\eta_i^2{\beta} D^2 T }{2 \gamma^2 }  + \eta_i  {R_T} .
\end{align*}
where the last inequality uses the convexity of $\hat{f}_t$ and we denote $R_T = \frac{\hat{G}D}{\gamma} \regret_T(\W )$.
We thus have the recurrence
$$ \Delta_i \leq \Delta_{i-1} (1 - \eta_i) +  \eta_i^2 \frac{{\beta} D^2 T }{2 \gamma^2 }  + \eta_i  {R_T} . $$
Denoting $\hat{\Delta}_i = \Delta_i - {R_T}  $,	we are left with
$$ \hat{\Delta}_i \leq \hat{\Delta}_{i-1} (1 - \eta_i) +  \eta_i^2 \frac{{\beta} D^2 T }{2 \gamma^2 }   . $$

This is a recursive relation that can be simplified by applying Lemma \ref{lemma:FW-recursion} from chapter \ref{chap:FW}. We obtain that $\hat{\Delta}_N \leq \frac{2 {\beta} D^2 T}{\gamma^2 N}$.
\end{proof}

We are ready to prove the main guarantee of Algorithm \ref{alg:ocoboost}.

\begin{proof}[Proof of Theorem \ref{thm:oco-boost-convex}]
Using both parts of Lemma \ref{lem:extensionX} in succession, we have
\begin{align*}
 \sum_{t=1}^T f_t(\x_t)\ -  \sum_{t=1}^T f_t( \x^\star_t ) 
& \leq \sum_{t=1}^T \fhat_t(\x_t)\ - \sum_{t=1}^T \fhat_t( \x^\star_t ) + 2 \delta G T \\
& \leq  \sum_{t=1}^T \fhat_t(\x_t^N)\ -  \sum_{t=1}^T \fhat_t( \x^\star_t ) + 3 \delta G T. \label{eqn:shalom4}
\end{align*}
Next, recall by Lemma \ref{lem:SmoothingLemma}, that $\fhat_t$ is $\frac{d G}{\delta}$-smooth. By applying Lemma \ref{lem:main-analysis}, and optimizing $\delta$, we have \begin{align*}
 \sum_{t=1}^T f_t(\x_t)\ -  \sum_{t=1}^T f_t( \x^\star_t ) 
& \leq  3 \delta G T +  \frac{2  d G  D^2 T}{\delta \gamma^2 N} +  \frac{\hat{G}D}{\gamma} \regret_T(\W) \\
& =  \frac{ 5 \sqrt{d} G  D T}{\gamma \sqrt{N}}  +  \frac{\hat{G}D}{\gamma} \regret_T(\W)  \\
& \leq \frac{ 5 d G  D T}{\gamma \sqrt{N}}  +  \frac{\hat{G}D}{\gamma} \regret_T(\W) ,
\end{align*}
where the last inequality is only to obtain a nicer expression. 

It remains to bound $\hat{G}$, and we claim that $\hat{G} \leq 2G$. To see this, notice that the function $\dist(\x,\K)$ is $1$-Lipschitz, since 
\begin{align*}
&  \dist (\x, \K) - \dist(\y, \K) \\
& = \|\x-\Pi_\K(\x)\|  - \|\y-\Pi_\K(\y)\| \\
& \leq \|\x-\Pi_\K(\y)\|  - \|\y-\Pi_\K(\y)\| & \mbox{ $\Pi_\K(\y)\in\K$}\\
& \leq \|\x-\y\| .  & \mbox{ $\Delta$-inequality}
\end{align*}
Thus, by the definition of the extension operator and the functions $f_t^i$,  we have that 
$\|\nabla f_t^i(\x_t^i)\|=\|\nabla \hat{f}_t(\x_t^i)\| \leq 2G$.
\end{proof}

\newpage
\section{Bibliographic Remarks}

The theory of boosting, which we have surveyed in chapter \ref{chap:boosting}, originally applied to binary classification problems. Boosting for real-valued regression was studied in the theory of   gradient boosting by \citet{friedman2002stochastic}. 

Online boosting, for both the classification and regression settings was studied much later  \citep{leistner2009robustness, chen2012online, chen2014boosting, beygelzimer2015optimal, beygelzimer2015online, agarwal2019boosting, jung2017online, jung2018online,brukhim2020online2}. The relationship to the Frank-Wolfe method was explicit in these works, and also studied in \citep{10.1214/16-AOS1505,wang2015functional}. A framework which encapsulates both agnostic and realizable boosting, for both offline and online settings, is given in \citep{brukhim2020online}. 

Boosting for the full online convex optimization setting, with a multiplicative approximation and general convex decision set, was  obtained in \citep{hazan2021boosting}. The latter also give tighter bounds by a factor of the dimension than those presented in this text using a  more sophisticated smoothing technique known as the Moreau-Yoshida regularization \citep{beck2017first}.

The contextual experts and bandits problems have been proposed by \citet{langford2008epoch} as a decision making framework with large number of policies. In the online setting, several works study the problem with emphasis on efficient algorithms given access to an optimization oracle \citep{rakhlin2016bistro,syrgkanis2016improved,syrgkanis2016efficient,rakhlin2016bistro}. For surveys on contextual bandit algorithms and applications of this model see \citep{zhou2015survey,bouneffouf2019survey}.

%Another related line of work in online learning studies the possibility of $\alpha$-regret minimization given access to approximation oracles with an approximate weak learner \citet{kakade2009playing, hazan2018online}. This weaker notion of $\alpha$-regret only generates approximately optimal decisions. In contrast, our result gives a stronger guarantee of standard regret, and the regret is compared to a stronger benchmark of convex combination of experts. 

\newpage
\begin{exercises}

\exer{
Derive, from Algorithm \ref{alg:ocoboost}, an algorithm for online binary classification. Spell out its regret guarantee. 
}

\exer{ 
Formulate the online personalized article placement problem in the language of online convex optimization (what is the decision set?).  Define a weak learner in this context, and the final boosting guarantee from Theorem \ref{thm:oco-boost-convex}. 
}

\exer{ $^*$
Define the Moreau-Yoshida regularization (or smoothing operator) as: 
\[ M_\delta[f](\x) = \inf_{\y \in \reals^d} \left\{f(\y) + \frac{1}{2\delta} \|\x-\y\|^2 \right\}.\]
Prove that given any $G$-Lipschitz $f$, the smoothed function $\fhat_\delta = M_\delta[f] $ satisfies: 
        \subexer{ $\hat{f}_\delta$ is $\frac{1}{\delta}$-smooth, and $G$-Lispchitz.}
        \subexer{ $ \left|\hat{f} _\delta (\x) - f(\x) \right| \le \frac{\delta G^2}{2}$ for all $ \x \in \reals^d$.}
}

\exer{ $^*$
Use the Moreau-Yoshida regularization to improve the bounds of Theorem \ref{thm:oco-boost-convex} by a factor of the dimension. 
}

\end{exercises}

\chapter{Blackwell Approachability and Online Convex Optimization} \label{chap:approach}
\chaptermark{Blackwell Approachability}

The history of adversarial prediction started with the seminal works of mathematicians David Blackwell and James Hannan.  In most of the text thus far, we have presented the viewpoint of sequential prediction and loss minimization, taken by  Hannan. This was especially true in chapter \ref{chap:regularization}, as the FPL algorithm dates back to his work. In this chapter we turn to a dual view of regret minimization, called ``Blackwell approachability". Approachability  theory  originated in the work of Blackwell, and was discovered simultaneously to that of Hannan. A short historical account is surveyed in the bibliographic materials at the end of this chapter. 

For decades the relationship between regret minimization in general convex games and Blackwell approachability was not fully understood. The common thought was, in fact, that Blackwell approachability is a stronger notion. In this chapter we show that approachability and online convex optimization are equivalent in a strong sense: an algorithm for one task implies an algorithm for the other with no loss of computational efficiency. 

As a side benefit to this equivalence, we deduce a proof of Blackwell's approachability theorem using the existence of online convex optimization algorithms. This proof applies to a more general version of approachabilty, over general vector games, and comes with rates of convergence  that are borrowed from the OCO algorithms we have already studied. 

While previous chapters had a practical motivation and introduced methods for online learning, this chapter is purely theoretical, and devoted to give an alternate viewpoint of online convex optimization from a game theoretic perspective.

\section{Vector-Valued Games and Approachability}
\sectionmark{Vector-Valued Games}

Von Neumann's minimax theorem, that we have studied in chapter \ref{chap:games}, establishes a central result in the theory of two-player zero-sum games by providing a prescription to both players. This prescription is in the form of a pair of optimal mixed strategies: each strategy attains the optimal worst-case value of the game without knowledge of the opponent's strategy. However, the theorem fundamentally requires that both players have a utility function that can be expressed as a \emph{scalar}.

In 1956, in response to von Neumann's result, David Blackwell posed an intriguing question: what guarantee can we hope to achieve when playing a two-player game with a \emph{vector-valued payoff}?

A vector-valued game is defined similarly to zero-sum games as we have defined in Definition \ref{defn:zsg}, with reward/loss vectors replacing the scalar rewards/losses.  
\begin{definition} \label{defn:vectorgame}
A two-player vector game is given by a set of $n \times m$ vectors $\{ \uv(i,j) \in \reals^d \}$. 
The reward vector for the row player playing strategy $i \in [n]$, and column player playing strategy $j \in [m]$, is given by the vector $ \uv(i,j) \in \reals^d $.
\end{definition}

Similar to scalar games, we can define mixed strategies as distributions over pure strategies, and denote the expected reward vector for playing mixed strategies by 
$$ \forall \x \in \Delta_n , \y \in \Delta_m \ . \ \uv(\x,\y) = \E_{i \sim \x, j \sim \y} \left[ \uv(i,j) \right] .$$

We henceforth consider more general vector games than originally considered in the literature. The additional generality allows for uncountably  many strategies for both players, and allows the strategies to originate from bounded convex and closed sets in Euclidean space. 
\begin{definition} \label{defn:generalizedvectorgame}
A generalized two-player vector game is given by a set of  vectors $\{ \uv \in \reals^d \}$, and two bounded convex and closed decision sets $\K_1,\K_2 $. The reward vector for the row player playing strategy $\x \in \K_1$, and column player playing strategy $\y \in \K_2$, is given by the vector $ \uv(\x,\y) \in \reals^d $.
\end{definition}

The goal of a zero-sum game is clear: to guarantee a certain loss/reward. What should be the vector game generalization? Blackwell proposed to ask ``can we guarantee that our vector payoff lies in some closed convex set $S$?''   

It is left as an exercise at the end of this chapter to show that an immediate analogue of Von Neumann's theorem does not exist: there is no single mixed strategy that ensures the vector payoff lies in a given set. 
However, this does not rule out an asymptotic notion, if we allow the game to repeat indefinitely, and ask whether there exists a strategy to ensure that the {\it average} reward vector lies in a certain set, or at least approaches it in terms of Euclidean distance. 
This is exactly the solution concept that Blackwell proposed as defined formally below. 

Using the notation we have used throughout this text, we denote the (Euclidean) distance to a bounded, closed and convex set $S$ as
$$ \dist(\w,S) = \min_{\x \in S} \|\w-\x\| . $$ 

\begin{definition}\label{def:approach}
Given a generalized vector game  $\K_1,\K_2, \{\uv(\cdot,\cdot)\}$, we say that a set $S \subseteq \reals^d$ is {\bf approachable} if there exists some algorithm $\mA$, called an {\bf approachability algorithm},  which iteratively selects points $\K_1 \ni \x _{t} \leftarrow \mA(\y _{1}, \y _{2}, \ldots, \y _{t-1})$, such that, for any sequence \\ $\y _1, \y _2, \ldots , \y_T \in \K_2$, we have
	\[
		\dist \textstyle{\left(\frac 1 T \sum_{t=1}^T \uv(\x _t, \y _t), S \right) }\to 0
		\quad \text{ as } \quad
		T \to \infty . 
	\]
\end{definition}

Under this notion, we can now allow the player to implement an {adaptive strategy} for a repeated version of the game, and we require that the average reward vector comes arbitrarily close to $S$. Blackwell's theorem characterizes which sets in Euclidean space are approachable. We give it below in generalized form, 
\begin{theorem}[Blackwell's Approachability Theorem]  \label{thm:blackwell}
For any vector game $\K_1,\K_2,\{\uv(\cdot,\cdot)\}$, the closed, bounded and convex set $S \subseteq \reals^d$ is approachable if and only if the following condition holds:
$$ \forall \y \in \K_2 \ , \ \exists \x \in \K_1 \ , \mbox{s.t. } \uv(\x,\y) \in S . $$
\end{theorem}

The approachability condition spelled out in the equation above is both necessary and sufficient. The necessity of this condition is left as an exercise, and the more interesting implication is that any set that satisfies this condition is, in fact, approachable. Our reductions henceforth give an explicit proof of Blackwell's theorem, and we leave it as an exercise to draw the explicit conclusion of this theorem from the first efficient reduction.  

The relationship between Blackwell approachability in vector games  and OCO may not be evident at this point. However, we proceed to show that the two notions are in fact algorithmically equivalent. 

In the next section we show that any algorithm for OCO can be efficiently converted to an approachability algorithm for vector games. Following this, we show the other direction as well: an approachability algorithm for  vector games gives an OCO algorithm with no loss of efficiency!

\section{From Online Convex Optimization to Approachability} \label{sec:approach-dist}
\sectionmark{OCO to Approachability}

In this section we give an efficient reduction from OCO to approachability. Namely, assume that we have an OCO algorithm denoted $\mA$, that attains sublinear regret. Our goal is to design a Blackwell approachability algorithm for a given vector game and closed, bounded convex set $S$.  Thus, the reduction in this section shows that OCO is a stronger notion than approachabiliy. This direction is perhaps the more surprising one, and was discovered more recently, see bibliographic section for an historical account of this development.

Since we are looking to approach a given set, it is natural to  consider minimizing the distance of our reward vector to the set. Recall we denote the (Euclidean) distance to a set as $ \dist(\w,S) = \min_{\x \in S} \|\w-\x\| $. 
The support function of closed convex set $S$ is given by
$$ h_S(\w) = \max_{\x\in S} \{ \w^\top \x \}. $$
Notice that this function is convex, since it is a maximum over linear functions. 
\begin{lemma} \label{lem:dist-equivalence}
The distance to a set can be written as
\begin{equation} 
\dist(\uv,S) = \max_{\|\w\|\leq 1} \left\{ \w^\top \uv  - h_S(\w) \right\} .
\end{equation}
\end{lemma}
\begin{proof}
Using the definition of the support function,
\begin{eqnarray*}
& \max_{\|\w\|\leq 1} \left\{  \w^\top \uv  - h_S(\w) \right\} \\
 & =  
 \max_{\|\w\|\leq 1} \left\{ \w^\top \uv - \max_{\x \in S} \w^\top \x  \right\} \\
 & = \max_{\|\w\|\leq 1} \min_{\x \in S} \left\{ \w^\top \uv  -   \w^\top \x  \right\} & \mbox{negation}\\ 
 & =  \min_{\x \in S} \max_{\|\w\|\leq 1} \left\{ \w^\top \uv  -   \w^\top \x  \right\}  & \mbox{minimax theorem} \\
 & = \min_{\x \in S} \| \x - \uv\| \\
& =   \dist(\uv,S) .
\end{eqnarray*}
\end{proof}

Blackwell's theorem characterizes approachable sets: it is necessary and sufficient to be able to find a best response $\x \in \K_1$, to any $\y \in \K_2$, such that $\uv(\x,\y) \in S$.  To proceed with the reduction, we need an equivalent condition stated formally as follows. 
\begin{lemma} \label{lem:blackwell-1}
For a generalized vector game $\K_1,\K_2,\{\uv\}$, the following to conditions are equivalent:
\begin{enumerate}
    \item There exists a feasible best response,
    $$ \forall \y \in \K_2 \ , \ \exists \x \in \K_1 \ , \mbox{s.t. } \uv(\x,\y) \in S . $$

    \item  For all  $\w \in \reals^d, \|\w\|\leq 1$, there exists $\x \in \K_1$ such that 
    $$ \forall \y \in \K_2 \ \ , \ \  \w^\top \uv(\x,\y) - h_S(\w) \leq 0 .$$ 
\end{enumerate}
\end{lemma}
\begin{proof}
Consider the scalar zero sum game
$$ \min_{\x} \max_{\y}  \dist(\uv(\x,\y) , S) = \lambda . $$ 
Blackwell's theorem asserts that $\lambda = 0$ if and only if $S$ is approachable. Using Sion's generalization to the Von Neumann minimax theorem from chapter \ref{chap:games}, 
\begin{eqnarray*}
& \lambda  = \min_{\x} \max_\y \dist(\uv(\x,\y) , S) \\
& = \min_\x \max_\y \max_{\| {\w} \|\leq 1} \left\{ {\w}^\top \uv(\x,\y) - h_S ({\w}) \right\} & \mbox{Lemma \ref{lem:dist-equivalence}} \\
& =  \max_{\|{\w}\|\leq 1} \min_\x \max_\y \left\{ {\w}^\top \uv(\x,\y) - h_S ({\w}) \right\} & \mbox{minimax} 
%& \geq   \min_\x \max_\y \left\{ \w^\top \uv(\x,\y) - h_S (\w) \right\}.  & \mbox{particular $\w$ } 
\end{eqnarray*}
Thus, the second statement of the lemma is satisfied if and only if $\lambda = 0$. 

\end{proof}

As mentioned previously, the necessity of Blackwell's condition is left as an exercise. To prove sufficiency, we assume that form (2) of Blackwell's condition as in Lemma \ref{lem:blackwell-1} is satisfied. Formally, we henceforth assume that the vector game and set $S$ are equipped with a best response oracle $\mO$, such that
\begin{equation} \label{eqn:blackwell-oracle}
\forall \y \in \K_2 \ , \ \  \w^\top \uv( \mO(\w),\y) - h_S(\w) \leq 0  .  
\end{equation}

We proceed with the formal proof of sufficiency, constructively specified in Algorithm \ref{alg:oco2approach}. Notice that in this reduction, the functions $f_t$ are concave, and the OCO algorithm is used for maximization.

\begin{algorithm}[h]
	\caption{OCO to Approachability reduction} \label{alg:oco2approach}
	\begin{algorithmic}[1]
		\State Input: generalized vector game $\K_1,\K_2,\{ \uv(\cdot,\cdot)\}$, set $S$, best response oracle $\mO$, OCO algorithm $\mA$
		\State Set $\K =  \ball \in \reals^d$ to be the unit Euclidean ball, as decision set for $\mA$. 
		\For{$t=1, \ldots, T$}
		\State set $f_t(\w) =  \w^\top \uv_{t-1}  - h_S(\w) $
		\State Query $\mA$: $\w_t \leftarrow \mA(f_1, \ldots, f_{t-1})$
		\State Query $\mO$: $\x _t \leftarrow {\mO}(\w_t)$  
		\State Observe $\y_t$ and let $\uv_t = \uv(\x_t,\y_t)$
		\EndFor
        \State 	\Return $\bar{\uv}_T = \frac{1}{T} \sum_{t=1}^T \uv(\x_t,\y_t) $
	\end{algorithmic}
\end{algorithm}

\begin{theorem}
Algorithm \ref{alg:oco2approach}, with input OCO algorithm $\mA$, returns the vector $\bar{\uv}_T = \frac{1}{T} \sum_{t=1}^T \uv(\x_t,\y_t) $ that approaches the set $S$ at a rate of 
$$ \dist( \bar{\uv}_T , S ) \leq \frac{\regret_T(\mA)}{T} $$ 
\end{theorem}

\begin{proof}
Notice that equation \eqref{eqn:blackwell-oracle} implies that for $\w_t$ as defined in the algorithm, we have
$$ \forall \y \in \K_2 \ \ , \ \      \w_t^\top \uv(\mO(\w_t),\y)  - h_S(\w_t)   \leq 0  .  $$
This implies that for any $t$, 
\begin{equation}\label{eqn:blackwell-e1}
f_t(\w_t) =    \w_t^\top  \uv(\mO(\w_t),\y_t)  - h_S(\w_t)  \leq 0 .  
\end{equation}

Therefore we have, using Lemma \ref{lem:dist-equivalence},
\begin{eqnarray*}
\dist(\bar{\uv}_T, S) & = \max_{\|\w\|\leq 1} \left\{  \w^\top \bar{\uv}_T  - h_S(\w) \right\}  \\
& = \max_{\w^\star \in \K}  \frac{1}{T} \sum_t f_t(\w^\star) & \mbox{definition of $f_t$} \\
& \leq \frac{1}{T} \sum_t f_t(\w_t) + \frac{\regret_T(\mA)}{T} & \mbox{OCO guarantee of $\mA$} \\
& \leq \frac{\regret_T(\mA)}{T} & \mbox{equation \eqref{eqn:blackwell-e1}}
\end{eqnarray*}

\end{proof}

This theorem explicitly relates OCO with approachability, and since we have already proved the existence of efficient OCO algorithms in this text, it can be used to formally prove Blackwell's theorem. Completing the details is left as an exercise.

\section{From Approachability to Online Convex Optimization}

In this subsection we show the converse reduction: given an approachability algorithm, we design an OCO algorithm with no loss of computational efficiency. This direction was essentially shown by Blackwell for discrete decision problems, as described in more detail in the bibliographic section. We prove it here in the full generality of OCO.  

Formally, given an approachability algorithm $\mA$, denote by $\dist_T(\mA)$ an upper bound on its rate of convergence to the set $S$ as a function of the number of iterations $T$. That is, for a given vector game, denote by $\bar{\uv}_T = \frac{1}{T} \sum_{t=1}^T \uv(\x_t,\y_t) $ the average reward vector. Then $\mA$  guarantees 
$$ \dist(\bar{\uv}_T,S) \leq \dist_T(\mA) \ , \ \lim_{T \mapsto \infty} \dist_T(\mA) = 0 . $$
Given an approachability algorithm $\mA$, we henceforth create an OCO algorithm with vanishing regret.

\subsection{Cones and polar cones}

Approachability is in a certain geometric sense a dual to OCO. To see this, we require several geometric notions, that are explicitly required for the reduction from approachability to OCO. 

For a given convex set $\K \subseteq \reals^d$, we define its cone as the set of all vectors in $\K$ multiplied by a non-negative scalar,  
$$ \cone(\K) = \{ c \cdot \x \ | \ \x \in \K , 0 \leq c \in \reals \} .$$

The notion of a convex cone is not strictly required for the proofs below, but they are commonly used in the context of approachability. 
The polar set to a given set $\K \subseteq \reals^d$  is defined to be
$$ \K^0 \equaldef \{ \y  \in \reals^{d} \ \mbox{ s.t. } \  \forall \x \in \K \ , \ \x^\top  \y \leq 0 \} .$$
It is left as an exercise to prove that $\K^0$ is a convex set, and that for cones, the polar to the polar is the original set.

Henceforth we need the extension of a convex set defined as follows. Denote by $1 \oplus \K$ as the direct sum of the scalar one and the set $\K$, i.e., all vectors of the form $\tilde{\x} = 1 \oplus \x$ for $\x \in \K$.  
Denote the bounded polar extension of a set $\K$ by
%$$ Q(\K) = Q_\kappa(\K) = \cone(1 \oplus \K)^0 \cap B_\kappa .$$
$$ Q(\K) = (1 \oplus \K)^0 .$$
That is, we take all points in the polar set to the direct sum $1 \oplus \K$. %That is, we take all points in the polar cone to $\cone(1 \oplus \K)$, intersection with the Euclidean ball of radius $\kappa$. We omit the subscript $\kappa$ if clear from the context. 

%Henceforth we assume that the diameter of $\K$ is at least one for simplicity of presentation.  

This definition of the polar set gives rise to the following quantitative characterization.
%Henceforth we denote the diameter of the set $Q(\K)$ by $D^0$, and assume it is at least one for simplicity of presentation.  
%This definition of the polar set gives rise to the following quantitative characterization.

\begin{lemma} \label{lem:dual-dist}
Let $\y \in \reals^{d+1}$ be such that $\dist(\y,Q(\K)) \leq \eps$. Then, denoting by $D$ the diameter of $\K$, 
$$ \forall \tilde{\x} \in 1 \oplus \K \ ,  \ \y^\top \tilde{\x} \leq \eps (D+1) . $$
\end{lemma}
\begin{proof}
By definition of distance to a set, we have that $\dist(\y,Q(\K)) \leq \eps$, implies the existence of a point $\z \in Q(\K)$ such that $\| \y - \z \| \leq \eps$. Thus, for all $\tilde{\x} \in 1 \oplus \K$, we have
\begin{eqnarray*}
\y^\top \tilde{\x} & = (\y - \z + \z)^\top \tilde{\x} \\
& \leq \| \y - \z\| \|\tilde{\x} \| + \z^\top \tilde{\x} & \mbox{Cauchy-Schwartz} \\
& \leq \eps \|\tilde{\x} \| + \z^\top \tilde{\x} & \|\y-\z\|\leq \eps \\
& \leq \eps \|\tilde{\x}\| + 0 & \tilde{\x} \in 1 \oplus \K , \z \in (1\oplus \K)^0 \\
& \leq \eps (1+D) .
\end{eqnarray*}
\end{proof}

\subsection{The reduction}

Algorithm \ref{alg:bwa_to_lra} takes as an input a Blackwell approachability algorithm that guarantees, under the necessary and sufficient condition, convergence to a given set. It also takes as an input a set $\K$ for OCO. 

The reduction considers a vector game with decision sets $\K,\F$ and approachability set $S =  Q(\K)$, and generates a sequence of decisions that guarantee low regret as we prove next.  

Since this reduction creates the approachability set $S$ as a function of $\K$, we need to prove that indeed the set $S$ is approachable. We show this in the next subsection.

\begin{algorithm}[h]
	\caption{Conversion of Approachability Alg. $\mA$ to Online Convex Optimization Alg. $\L$} \label{alg:bwa_to_lra}
	\begin{algorithmic}[1]
		\State Input: closed, bounded and convex decision set $\K \subset \reals^d$, approachability oracle $\mA$. 
		\State Let: vector game w. $\K_1 = \K$,  $\K_2 = \F$,
		and set  $S := Q(\K) $.

        \For{$t=1, \ldots, T$}
			\State Query $\mA$: $\x _t \leftarrow \mA(f_1, \ldots, f_{t-1})$
			\State Let: $\L(f_1, \ldots, f_{t-1}) := \x _t$
			\State Receive: cost function $f_t$
			\State Construct reward vector $\uv(\x_t ,f_t) := \nabla_t^\top  \x_t  \oplus (-\nabla_t)$ 
		\EndFor
	\end{algorithmic}
\end{algorithm}

\begin{theorem}\label{thm:blackwell_to_olo}
The reduction defined in Algorithm~\ref{alg:bwa_to_lra}, for any input algorithm $\mA$, produces an OLO algorithm $\L$ such that $$\regret(\L) \leq  T (D +1) \cdot {\dist_T(\mA)} . $$
\end{theorem}

% Proof in the appendix

\begin{proof}
The approachability algorithm guarantees $\dist( \bar{\uv_T} , S ) \leq  \dist_T(\mA)$. Using the definition of $S$ and Lemma \ref{lem:dual-dist} we have  

\begin{eqnarray*}
& \forall \xtil[] \in Q(\K) \ . \ (D+1) \cdot \dist_T(\mA) \\
& \geq  ( \frac{1}{T} \sum_{t=1}^T \uv(\x _t, f_t)) ^\top \xtil[] \\
& \geq ( \frac{1}{T} \sum_{t=1}^T \uv(\x _t, f_t)) ^\top (1 \oplus \x^\star )  \\
& =   \frac{1}{T} \sum_{t=1}^T   \nabla_t^\top \x_t  -  \frac{1}{T} \sum_{t=1}^T \nabla_t^\top \x^\star   \\
		& \geq  \frac{1}{T}   \regret_T(\L),
\end{eqnarray*}
where the second inequality holds since the first inequality holds for every $\xtil[]$, in particular for the vector $1 \oplus \x^\star$.

\end{proof}

\subsection{Existence of a best response oracle}
Notice that the reduction of this section from approachability to OCO does not require the best response oracle. However, Blackwell's approachability theorem does require this oracle as sufficient and necessary, and thus for the set $S$ we constructed to be approachable at all, such an oracle needs to exist. This is what we show next. 

Consider the vectors $\uv_t$ constructed in the reduction. A best response oracle finds, for every vector $\y$, a vector $\x$ that guarantees 
$   \uv(\x,\y) \in S $. In our case, this translates to the condition
$$  \forall f \in \F \ , \ \exists \x \in  \K \ ,  \ \nabla f(\x)^\top \x \oplus (- \nabla f(\x)) \in  (1 \oplus \K)^0 . $$ 
By definition of the polar set, this implies that for all $\tilde{\x} \in \K$, we have 
$$ \nabla f(\x)^\top \x  - \nabla f(\x)^\top \tilde{\x}  \leq 0 . $$
In other words, the best response oracle corresponds to a procedure that given $f$, finds a vector $\x^\star$ such that 
$$ \forall \x \in \K \ . \ f(\x^\star) - f(\x) \leq \nabla f(\x^\star)^\top (\x^\star - \x) \leq 0 . $$ 
This is an optimization oracle for the set $\K$!

\newpage
\section{Bibliographic Remarks}

David Blackwell's celebrated Approachability Theorem was published in  \citep{blackwell_analog_1956}.  The first no-regret algorithm for a discrete action setting was given in a seminal paper by James Hannan in \citep{Hannan57} the next year.  That same year,  Blackwell pointed out \citep{blackwell1954controlled} that his approachability result leads, as a special case, to an algorithm with essentially the same low-regret guarantee proven by Hannan. For Hannan's account of events see \citep{gilliland2010conversation}.

Over the years several other problems have been reduced to Blackwell approachability, including asymptotic calibration \citep{foster_asymptotic_1998}, online learning with  global cost functions \citep{even-dar_online_2009} and more \citep{mannor2008regret}. Indeed, it has been presumed that approachability, while establishing the existence of a no-regret algorithm, is strictly more powerful than regret-minimization; hence its utility in such a wide range of problems.

However, this was recently shown not to be the case. \citet{abernethy2011blackwell} showed that approachability is in fact equivalent to OCO. This result is the basis of the material presented in this chapter. One side of their reduction was simplified and generalized in  \citep{shimkin2016online}.

\newpage
\begin{exercises}

\exer{
Prove that  Von Neumann's theorem does not hold for vector games, i.e., give an example of a vector game such that there is no single mixed strategy that ensures the vector payoff lies in a given set. Show that this is true even for approachable sets. 
}

\exer{
Prove that the polar set to a given convex set $\K \in \reals^d$ is convex. 
}
%\exer{
%Prove that the polar to a polar of a cone is itself again, i.e., 
%$$ (\cone(\K)^0)^0 = \cone(\K) . $$

\exer{
Prove that Blackwell's condition is necessary, i.e., prove that for a set $S$ to be approachable in a given vector game, there must exist an oracle $\mO$ such that 
$$ \forall \y \in \Delta_m \ , \ \uv(\mO(\y),\y) \in S . $$
}
\exer{
Complete the proof of Blackwell's theorem using the first reduction of this chapter and the existence of OCO algorithms. That is, prove that for a given vector game, the existence of an oracle as per the previous question, is sufficient for a set $S \subseteq \reals^d$ to be approachable. 
}

%\exer{ 
%Prove that
%$$ \dist(\uv,S) = \max_{\|\w\|\leq 1} \left\{  \w^\top \uv  - h_S(\w) \right\} . $$

\end{exercises}

\backmatter

\theendnotes

\bibliographystyle{unsrtnat}
\bibliography{bookbib}

\end{document}